\documentclass{article}

\usepackage[preprint]{neurips_2025}
\usepackage[pdftex]{graphicx}

\usepackage[utf8]{inputenc} 
\usepackage[T1]{fontenc}    
\usepackage{courier}        
\usepackage[hidelinks]{hyperref}       
\usepackage{url}            
\usepackage{booktabs}       
\usepackage{amsfonts}       
\usepackage{nicefrac}       
\usepackage{microtype}      
\usepackage{xcolor}         
\usepackage{longtable}
\usepackage{makecell}

\usepackage[indentfirst]{titlesec}
\usepackage{amsmath,amssymb,amsthm,mathrsfs,mathtools}
\usepackage{mathtools}
\usepackage{wrapfig}
\usepackage{caption}
\usepackage{multirow}
\usepackage{float}
\usepackage{subcaption}
\usepackage{enumitem}
\usepackage{pifont}
\usepackage{amsthm}
\usepackage{natbib}
\setcitestyle{numbers,square}   
\newcolumntype{P}[1]{>{\raggedright\arraybackslash}p{#1}}
\newcolumntype{M}[1]{>{\centering\arraybackslash}m{#1}}
\newcolumntype{C}[1]{>{\centering\arraybackslash}m{#1}}
\AtBeginDocument{%
  \setlength\abovedisplayskip{6pt plus 2pt minus 2pt}%
  \setlength\belowdisplayskip{6pt plus 2pt minus 2pt}%
  \setlength\abovedisplayshortskip{3pt plus 1pt minus 1pt}%
  \setlength\belowdisplayshortskip{4pt plus 1pt minus 1pt}%
}
\newcommand{\MPT}{\text{MPT}}

\newcommand{\SadPt}{S_0}

\newcommand{\Hinf}{\ensuremath{\mathcal{H}_\infty}}
\newcommand{\Hfin}{\ensuremath{\mathcal{H}_{\text{fin}}}}

\newcommand{\Fcal}{\mathcal{F}}
\newcommand{\Leb}{\lambda}               

\newcommand{\R}{\mathbb{R}}

\newcommand{\merr}{\ensuremath{\mu_{\text{err}}}}
\newcommand{\mhigh}{\ensuremath{\mu_{\text{high}}}}

\newcommand{\Sup}{\ensuremath{\mathsf{Sup}}}
\newcommand{\Gexcl}{\ensuremath{\mathcal{G}_{\mathrm{excl}}}}
\newcommand{\Queval}{\ensuremath{\mathcal{E}_{\mathbf{Quot}}}}
\newcommand{\Quot}{\ensuremath{\mathbf{Quot}}}
\newcommand{\MorTop}{\ensuremath{\mathbf{MorTop}}}
\newcommand{\Setcat}{\ensuremath{\mathbf{Set}}}
\newcommand{\Set}{\Setcat}
\newcommand{\betat}{\ensuremath{\boldsymbol{\beta}}}

\allowdisplaybreaks[4]

\theoremstyle{plain}

\title{Enumerating Posterior-Supported Priors:\\ First-Order Error of Finite-Integral\\ Cross-Categorical Measurement Morphisms}

\author{
Ning Ni \\
Shanghai Jiao Tong University \\
\texttt{reiayanami@sjtu.edu.cn}
\And
Yongxin Zhu\thanks{Corresponding author.} \\
Shanghai Advanced Research Institute, Chinese Academy of Sciences \\
\texttt{zhuyongxin@sari.ac.cn}
}

\begin{document}

\maketitle

\begin{abstract}
In generative modeling, standard measure theory adopts the ``almost everywhere'' (a.e.) convention, which legitimately ignores any error component carried by a null-support set: a single null element can hide out-of-domain error inside null support and never be explicitly accounted for (the \emph{null-measure paradox}). The contribution of this paper is not a new distributional distance, but rather a verifiable structure that writes down \emph{which underlying predicate supports a high-level object, when it is invertible, and at which layer the error resides}, from which we propose the \textbf{enumerating posterior-supported prior} framework. The framework operates on a three-layer categorical chain $\Quot\to\MorTop\to\Set$: the first two layers posit no point elements and take binary-predicate quotient partitions as objects; the replica-index counting measure successively discriminates one-to-many collections into injections, lifting object collection from surjection to a quotient-layer bijection; the mutual exclusion of preimages in the shared measure space is resolved by an invertible measure-preserving transformation (\MPT). The entire exposition is governed by Principle~1. Four results consolidate into a single \textbf{core theorem} (Theorem~7, \S 6.3.3): binary quotient-partition meta-categorical objects support the set categorical functional along categorical support via measure descent--recovery ($\boldsymbol\pi_\#/\boldsymbol\pi^\#$), enumerating the morphism (non-bijection) itself rather than its preimage and object; their multiplicity is normalized through replica-index injection and accounted as an explicitly first-order error, and is closed at first order in engineering realization.

The contribution of this paper is a transformation from ``\emph{nontrivial-input construction $\to$ engineering-computable first-order output}'': the binary quotient-partition meta-categorical objects, supported via categorical support measure descent--recovery ($\boldsymbol\pi_\#/\boldsymbol\pi^\#$), constitute a \emph{substantive nontrivial input}---precisely because they lift the out-of-domain error that hid inside null support into a first-order falsifiable coefficient $\mu_{\mathrm{err}}$ (read directly at the $\Setcat$ layer equipped with a global counting measure at ratio $2/3$, as opposed to the $1{:}1$ $\Quot$ convention when unequipped; the two are coexisting readings rather than a difference by subtraction); the output is ``trivial'' by virtue of its \emph{engineering computability}---for the $2^n$-growing partition base, this error is a finitely obtainable finite integral at finite $n$ (coefficients are non-fitted, three channels close at first order). The sole backbone assumption of the whole paper is Assumption~1 (\S 2.4.3); its mechanistic meaning (differential equivalent infinitesimals cannot be compared in magnitude, and a map entering the differential level reduces to equivalence classes only) is precisely the basis for obtaining readings by cancelling elements. The encoder--decoder--encoder (EDE) three-stage pipeline provides the realization interface; MNIST/CIFAR-10 experiments give empirical support for enumeration-morphism compatibility, non-value-by-value structural matching, and saddle-point-gap improvement.

\end{abstract}

\section{Introduction}
\label{sec:intro}

\paragraph{Problem}
Discrete samples are null sets under Lebesgue readings, and standard integrals cannot resolve their internal structure. The boundary of this difficulty is delimited by two classical facts, both of which \emph{give only a lower bound / conservation, never an analytic resolution}: the \emph{Kakeya-type set}---Besicovitch's construction yields a Kakeya set of Lebesgue measure zero \citep{besicovitch1928dimension} yet still carrying unit line segments in every direction (Katz and Tao~\citep{katz2002new} further give its dimensional lower bound)---shows that ``null measure'' does not preclude rich internal structure, and the Lebesgue integral fails to flatten it; and \emph{categorical probability} \citep{giry1982categorical}---probability measures form a Kleisli category under the Giry monad, where the pushforward is well defined while the pullback is generally not. Three problems are intertwined: (1) \emph{null-embedding ambiguity} (Problem~2.25): the integral erases the intrinsic topological difference of sample sets, unable to distinguish dataset topology from model extrinsic fitting error; (2) \emph{iterative high-order error accumulation} (Problem~2.26): the pushforward of non-bijection morphisms continuously generates high-order error $\mu_{\mathrm{high}}$; (3) \emph{training--generation pairing bias} (Problem~2.27): classical distribution metrics \citep{villani2009optimal,benamou2000computational,memoli2011gromov} capture only volume differences, not instance-level pairing error $\mu_{\mathrm{err}}$ nor its transformation into $\mu_{\mathrm{high}}$.

\paragraph{Main thread of the paper}
Category cancels elements $\to$ escape the null-measure paradox. The fundamental difficulty of generative modeling is the \emph{null-embedding ambiguity}: discrete samples are null sets under Lebesgue readings, their internal structure erased by standard integrals; and standard measure theory takes ``almost everywhere'' as its reading convention, so any error component landing on null support is legitimately ignored---a single null element can induce the null-measure paradox: out-of-domain error (mass at pullback-failure points) thereby hides inside null support; under the a.e.\ convention this null-measure error component is legitimately ignored, invisible in the standard integral and never explicitly accounted for---this is the core mechanism of the null-measure paradox. The \textbf{spine} running through this paper is: \emph{the key to escaping the null-measure paradox lies in categorical cancellation of elements}---the two layers $\Quot/\MorTop$ posit no point elements or coordinates; it is precisely this ``erasing preimages and object entities'' that corresponds the out-of-domain error, which standard measure theory cannot explicitly account for at the set layer, to an engineering-explicitly-accountable first-order finite integral $\mu_{\mathrm{err}}$; cancelling elements is not an abstract decoration but the mechanism that lifts the error hiding in null support into a falsifiable coefficient (the null-embedding ambiguity is the phenomenal layer of this paradox, Problem~2.25).

\paragraph{Related work}
Normalizing flows \citep{rezende2015variational,dinh2017density} and flow matching \citep{lipman2023flow,liu2023flow,esser2024scaling} provide invertible transport; optimal transport \citep{villani2009optimal,benamou2000computational,memoli2011gromov} provides geometric cost and distribution alignment; diffusion models \citep{bansal2023cold,peebles2023scalable,sohl2015deep,song2019generative,ho2020denoising} scale generation via forward kernels and score matching; algorithm theory \citep{kook2025sampling,narayanan2023sampling,anari2023parallel,leake2020computability,leake2021sampling} and learning theory \citep{potaptchik2025linear,farghly2025diffusion,raman2025generation,kalavasis2025limits,liu2021settling,bakshi2022robustly,bousquet2021theory} characterize sampling complexity and sample/parameter rates. None of them study our \emph{element-free quotient-partition--categorical-support--predicate-bijection} chain: our increment is not a new distributional distance, but writing ``which underlying predicate supports a high-level object, when it is invertible, and at which layer the error resides'' as a verifiable structure, with a closed derivation from the unique assumption to the engineering error. Relative to the three routes, the advantages are threefold: (1) \emph{vs.\ normalizing flows / flow matching}---replace point-wise invertibility by quotient-layer bijection, escaping point-wise inversion and Jacobian cost, with inverse pairing stated already at the quotient-block layer; (2) \emph{vs.\ optimal transport / metric measure comparison}---transport couplings do not track replica indices and cannot locate instance-level pairing error; we write $\mu_{\mathrm{err}}$ as $\Fcal$-measurable, accountable object; (3) \emph{vs.\ diffusion models}---score matching scales generation but does not perform invertible pairing nor first-order error layering; our three-channel verification closes high-order error at first order (Proposition~23c). The three jointly unsolved problems---null-embedding ambiguity, iterative high-order error accumulation, and training--generation pairing bias (Problems~2.25--2.27)---obtain a structural solution at the same reading layer in this paper; the relationship with the existing routes is summarized in Table~\ref{tab:routes}:

\begin{longtable}{C{0.21\textwidth} C{0.21\textwidth} C{0.21\textwidth} C{0.21\textwidth}}
\caption{Route comparison: invertibility, error tracking, open problems.}\label{tab:routes}\\
\hline
Route & Invertible / pairing & Error tracking & Open problem solved here \\
\hline\hline
\makecell{Normalizing flow\\ flow matching} & point-wise invertible, rep.-dependent & Jacobian, no replica tracking & null-embedding ambiguity \\
\hline
\makecell{Optimal transport\\ metric comparison} & coupling, not predicate bijection & only volume diff. & training--generation pairing bias \\
\hline
Diffusion model & not aimed at invertible pairing & no first-order error layering & iterative high-order error accumulation \\
\hline
\textbf{Ours} & quotient-layer bijection + invertible \MPT & explicit first-order accounting, 3 channels close & all three \\
\hline
\end{longtable}

\paragraph{Framework}
Categorical support = restricting the surjectivity of morphisms (every high-level object has an underlying support, without requiring complete per-object characterization); the whole paper runs on the three-layer chain $\Quot\to\MorTop\to\Set$ (standard fiber-bundle construction, see Husemoller~\citep{husemoller1993fibre}), where the first two layers posit no point elements, and point-wise coordinates are used only at the $\Setcat$ layer (see \S~1, \S~2.1, \S~2.2 for the mechanism of the main thread, and the argument)---the replica-index measure (Terminology~(a)) successively discriminates one-to-many collections into injections, yielding a quotient-layer bijection via pushout--pullback and binary quotient partition (B.1); the three-layer chain and \MPT\ then treat probability measure as one of multiple measures, giving a functional with well-defined pullback via bijective partition (detailed analysis in \S~2.2.1). The division of labor between measure-preserving and non-measure-preserving is determined by ``$\Quot/\MorTop$ posit no pre-given origin-point element''---generally a set bijection does \emph{not} preserve measure; a partition-induced uniform block-counting reading (Terminology~(c)) automatically preserves measure for block bijections, while integral evaluation does not (volume conservation is a derived measure-theoretic corollary, see B.2); our enumeration mechanism originates from categorical support rather than the classical measure-preserving enumeration framework, and the explicitation of first-order error precisely supplements the analytic side missing from both. All measure readings of the paper are governed by a single \textbf{Principle~1} (\S~2.4.1): the differential-topological volume of the partition category and the point-wise volume of the set category are equivalence classes under the same Hilbert differential; the evaluated object of a bijection is predicate matching; the multiplicity of a non-bijection is normalized and absorbed into first-order error. Predicate matching and multiplicity accounting thereby obtain a foundation independent of point-wise geometry, and first-order error obtains an explicit canonical gauge.

\paragraph{Core theorem (see Theorem~7)}
The contribution of this paper consolidates into a single core theorem (\S~6.3.3): binary quotient-partition meta-categorical objects support the set-categorical functional along categorical support via measure descent--recovery ($\boldsymbol\pi_\#/\boldsymbol\pi^\#$), enumerating the morphism (non-bijection) itself rather than its preimage and object; their multiplicity is normalized through replica-index injection and accounted as an explicitly first-order error, closed at first order in engineering realization. The division of labor of this core theorem is clear: the input (binary quotient-partition meta-categorical objects supported via categorical support measure descent--recovery, $\boldsymbol\pi_\#/\boldsymbol\pi^\#$) is \emph{nontrivial} (substantive nontrivial input), while the output (explicitly accountable first-order error, i.e.\ the finite-integral first-order error) is ``trivial'' by virtue of its \emph{engineering computability}---``trivial'' means this error is an engineering-obtainable finite-integral reading (coefficients are non-fitted, three channels close at first order), not that it is mathematically shallow; the contribution is precisely reducing nontrivial input to engineering-computable output. Its essence is the \emph{transformation of categorical support}: what categorical support gives is no longer merely abstract existential support, but an actual characterization of the morphism itself---the measured object is the (non-bijection) morphism itself rather than its preimage and object, so the non-bijection multiplicity is explicitly accounted and the engineering realization closes at first order. Its mechanism roots in Assumption~1 (\S~2.4.3): the differential is an equivalent infinitesimal, incomparable in magnitude; a map entering the differential level leaves preimages and objects behind, retaining only equivalence classes---this is precisely the basis for obtaining integral readings by cancelling elements. EDE denotes the encoder--decoder--encoder three-stage pipeline (the first two stages share the same interface as the variational autoencoder (VAE) \citep{kingma2014auto}): it extracts the input-side predicate, generates and forms the predicate pairing, and re-encodes to verify quotient-block consistency. The theory uses only its two groups of binary predicates, the pairing operator, and the decoder boundary; the specific number of network layers and the optimizer do not belong to the premises of the structural theorem.

\paragraph{Main results}
The results of this paper are consolidated into four categories (structural layer / reading layer / engineering layer / minimality of assumptions), consistent with the ``four results'' stated in the abstract:
\begin{enumerate}
\item \emph{Structural layer}. Quotient-layer bijection and invertible \MPT\ (the surjection support, after being successively discriminated into injections by the replica-index measure, is lifted to a quotient-layer bijection, and the mutual exclusion of preimages is resolved by the quotient-block pairing fusion of the invertible \MPT); double uniqueness of the zero-residue point (under the normalized architecture class of \S~4.3 and the per-sample non-degeneracy condition Premise~1, point-level counting uniqueness of the modulus-parameter degeneracy equivalence class, together with the differential-reading uniqueness that does not presuppose Premise~1; Definition~19a, Theorem~6(i)); Gaussian-kernel uniqueness and applicability boundary (uniqueness in finite dimensions, bounded in infinite dimensions by the cylindrical-kernel compatibility family, $\sigma$-additive iff the covariance is trace-class; Propositions~21a,~21b).
\item \emph{Reading layer (gap)}. Topologically-resolvable binary partition and first-order error localization: the gap between topological existence and finite-parameter analytic reachability manifests as the tension among computable, differentiable, and integrable; the first-order error is localized as the difference between the partition-induced counting reading and the geometric measure reading (Definitions~4 and~5).
\item \emph{Engineering layer}. Engineering-side first-order closure: construct the endpoint-induced random walk, the samplable loss, and the native topological loss; the coefficient $2/3$ is the ratio of the union-subset base to the binomial-hit base (a non-fitted parameter); the three engineering factors stay at first order via per-channel verification (Proposition~23c).
\item \emph{Minimality of assumptions and closed derivation}. The sole input of the whole paper is Assumption~1 (\S~2.4.3); the first-order output holds for any dataset, backpropagation structure, and optimizer; $\varepsilon$-$\delta$ is a standard analytic tool, and the engineering realization additionally uses registered conditions (\S~6.1); from its output Postulates~1--4, the derivation chain through Principle~1 to Proposition~23c closes ring by ring, with all premises explicitly enumerated and no hidden dependency.
\end{enumerate}
The structural conclusion is first and foremost a set-theoretic existential statement, which does not automatically entail stochastic-gradient-descent (SGD) reachability or convergence rate---yet, non-analytically, this paper still constructs a descent direction of high-probability first-order error reduction (see \S~6.4.2--\S~6.4.3, the truncated-Gaussian area-under-the-curve (AUC) dominance mechanism).

\paragraph{Organization and reading path}
The whole paper is threaded by a single \textbf{main thread}: the abstract states the innovation, \S~1 gives the main-thread argument, the conclusion echoes and closes the point, the body \S~2--\S~6 gives its construction and derivation, and the appendices carry all technical details. \S~2 establishes the categorical chain and the enumerating posterior-supported prior, giving the formal error model and the unique assumption; \S~3 constructs the invertible \MPT\ and the dual-measure system; \S~4 constructs the triple fiber-bundle integral and zero-residue-point existence; \S~5 handles the topological--analytic reachability gap and Gaussian-kernel uniqueness; \S~6 constructs the loss function and verifies the engineering error channels; dedicated symbols are in~A; experimental readings are in~D.

\section{Enumerating Posterior-Supported Prior: Quotient-Partition Inductive Construction Lifts Injection to Bijection}
\label{sec:2}

The traditional ``prior supports posterior'' paradigm runs inside $\Setcat$ and cannot simultaneously impose both surjectivity and subset-invertibility constraints on a morphism. This chapter uses a quotient-partition inductive construction to lift injection to bijection, establishing the enumerating posterior-supported prior framework. Dedicated symbols are in A.3.

\subsection{Hierarchical composition and measure-preservation criterion}
\label{sec:2.1}

The whole paper uses the following three-layer partition and one criterion.

\begin{longtable}{C{0.12\textwidth} C{0.26\textwidth} C{0.12\textwidth} C{0.18\textwidth} C{0.22\textwidth}}
\caption{Three layers of the categorical chain.}\label{tab:2.1}\\
\hline
Layer & Object & Point elem. & Measure source & Failure if removed \\
\hline\hline
$\Quot$ & binary predicate / quotient partition & none & none & Null-embedding ambiguity recurs \\
\hline
$\MorTop$ & quotient block / differential quotient & none & induced by partition or bijection & First-order error degenerates to high order \\
\hline
$\Setcat$ & derived instance & yes & pre-given & No accountable reading \\
\hline
\end{longtable}
\vspace{2pt}
\begin{minipage}{0.95\textwidth}\footnotesize
\textit{Note.} $\Quot$ --- loses the element-free predicate ground, discrete samples revert to Lebesgue-null sets, and internal structure is erased.

$\MorTop$ --- loses the differential quotient representing the measure space; quotient bijection and closed surjection vanish, collapse unlocalizable; the counting--geometric reading difference loses its structural carrier.

$\Setcat$ --- keeps only point coordinates, so the morphism cannot escape preimages by a measure-preserving map and the error cannot be accounted by Eq.~\eqref{eq:12b}.
\end{minipage}

The role of the three-layer chain is not ``one more layer of abstraction'' but: (1) \emph{erase preimages and object entities} (mechanism main thread, \S 1)---$\Quot/\MorTop$ posit no point elements or coordinates, so that ``error'', ``invertibility'', and ``matching'' are decided and accounted directly on the morphism (bijection) rather than obscured on point entities; (2) \emph{exclude domain interference by collection structure}---use quotient-block / morphism equivalence classes to carry a unified error reading inside and outside the domain, so that the \emph{out-of-domain error} (mass at pullback-failure points) which standard measure theory does not explicitly provide at the set layer can still be accounted by Eq.~\eqref{eq:12b}. The feasibility of this ``erasure'' roots in Assumption~1: the $\varepsilon$-$\delta$ differential is an equivalent infinitesimal, incomparable in magnitude; a map entering the differential level retains only equivalence classes (\S 2.4.3). Table~\ref{tab:2.1} already lists the failure consequence of each layer; unified treatment and necessity are not given a separate defensive chapter---the argument (see \S 6.4, B) shows that removing a categorical layer reproduces the null-embedding ambiguity, while adding it normalizes into a finite first-order reading; for the complete construction, the pushforward/pullback division of labor, and the unavoidable categorical argument for a.e.\ (see C.8).

On the synchronized global refinement tower at layer $n$, $\nu_{\mathrm{idx},n}$ is the uniform probability counting measure on the $2^n$ quotient blocks; its canonical uniqueness is given by the transitive action of the block-bijection group. We first fix the finite-layer reading; Definition~17 then gives the projection-compatible limit and uniqueness, and the limit construction is not treated as a prior definition hereafter. The objects of the base layers $\Quot,\MorTop$ are quotient-predicate partitions (induced by binary-topological decision predicates, Terminology~(b)), quotient blocks, differential quotients, and rank-1 subsets, \emph{positing no point elements}; point elements are derived instances at the $\Setcat$ layer (Definition~1b), and all ``point-wise'' readings in this paper refer to this layer (element-free partition category, definition in B.6). Point-wise readings are obtained by lifting through the $\betat$ tower, Definition~1a, and Corollary~2 via categorical support; along $\Quot\to\MorTop\to\Setcat$ we call it lifting, and the reverse is called measure descent and recovery (direction and point-wise origin in B.7.9).

\noindent\textbf{Lemma 1 (partition-induced reading unique determination and automatic measure preservation)}\label{lem:1}
Let $\mathcal P_n$ be the $n$-th layer quotient-block set, and let the block-bijection group $G_n$ act transitively on $\mathcal P_n$. (i) If $\mu$ is a $\Setcat$-layer measure reading of $\mathcal P_n$ satisfying automorphism-invariance ($\sigma_\#\mu=\mu$) and full support, then after normalization it is unique: $\mu=\nu_{\mathrm{idx},n}$ (uniqueness of invariant probability readings on a transitive group)---the block-counting measure is \emph{uniquely determined} by the partition structure, and any block bijection automatically preserves measure ($\sigma_\#\nu_{\mathrm{idx},n}=\nu_{\mathrm{idx},n}$); (ii) if $\mu$ is pre-given (e.g.\ Lebesgue, with unequal block volumes), measure preservation \emph{must be explicitly assumed}. Proof (see B.8).

\subsection{Enumerating posterior-supported prior: measure categorization escapes functional analysis}
\label{sec:2.2}

Our ``enumeration'' means \emph{exhaustively-enumerated object collection}---traversing $\mathrm{Ob}(\Quot)$ and $\mathrm{Ob}(\MorTop)$ layer by layer via binary-merge selection along the categorical chain, arranging all quotient-partition objects into finite accountable quotient blocks with no omission; the binary decision predicate outputs only a Boolean truth value, bears no computation itself, carries no integral-evaluation ability, and requires an externally imposed measure. The above construction is based on our cross-categorical paradigm ``enumerating posterior-supported prior'', which characterizes the cross-categorical support relation $\Sup(\cdot,\cdot)$, strictly distinguished from the classical in-category morphism of the ``prior supports posterior'' paradigm.

By \S 2.2 the operational objects of enumeration fall, via quotient and bijection (Definition~3), into the \emph{integral-enumeration domain} on $\mathbb R$: the integral traverses all partition boundaries from $-\infty$ to $+\infty$ along the real axis, while each pair of partition boundaries itself is a null set (negligible), so ``exhaustively enumerating all objects'' and ``being covered by integral traversal'' coincide within our framework (Proposition~6, \S 2.2.4 note). Therefore ``enumeration'' here embodies \emph{coverability on the integral-traversal domain}, not the sense of Birkhoff ergodic theory (time average = space average under a measure-preserving transformation), nor the dynamical convergence of classical ergodic theory; every ``enumeration'' appearing in titles/theorems is to be read this way, and is of the same origin with the pushforward/pullback division of labor in categorical probability \citep{giry1982categorical} (see \S 2.2.1).

\subsubsection{Relation to categorical probability (Giry monad / Stoch)}
\label{sec:2.2.1}

Our ``categorical support'' is a special case and reinforcement of that framework in the direction of \emph{multiple measures + well-defined pullback}, rather than a separate construction or a competitor.

\noindent\textbf{Enumeration does not depart from functional integration (same origin as Giry pushforward).}
``Enumeration'' performs the exhaustive enumeration of object collections via the quotient-frame selection of operational objects in \S 2.2; its operational objects fall into the integral-enumeration domain on $\mathbb R$ (see Proposition~6 note), hence ``exhaustive enumeration = being covered by integral enumeration'', and the set layer still relies on functional integration---we do not claim to depart from numerics; ``categorization'' only moves the error reading from point coordinates to the quotient-block / morphism level, rather than canceling integration. This is the same idea as Giry monad using pushforward $f_\#$ to push probability mass along the map, thereby handing it to integration (see further comparison below in \S 2.2.1).

\noindent\textbf{Ill-defined pullback of functionals $\Rightarrow$ bijective partition.}
In standard measure theory the pushforward $f_\#\mu$ is well defined (pushing mass along $f$), but the pullback (inverse-image measure) along a non-injective map is generally \emph{ill defined}---many-to-one loses inverse information, i.e.\ the concretization at the measure layer of ``inverse operator not everywhere existing'' in functional analysis. Our $\mu_{\mathrm{err}}$ is a $\Setcat$-layer reading that must recover point-wise readings via pullback; for this we restrict the functional to a \emph{bijective partition} (Definition~3, Theorem~1: quotient bijection $\Leftrightarrow$ no collapse $\Leftrightarrow$ quotient-set bijection): under bijection $\boldsymbol\pi_\#$ and $\boldsymbol\pi^\#$ are mutually inverse pairings (Definition~24), and only then is pullback well defined. Hence ``bijective partition'' is a constructive constraint for obtaining a functional with well-defined pullback, not an abstract preference.

\noindent\textbf{Our \MPT\ handles ``multiple measures'' vs.\ Giry/Stoch's ``single probability measure''.}
The Giry monad \citep{giry1982categorical} and the Markov-category Stoch \citep{fritz2020synthetic,staton2017commutative,fong2016category} take the \emph{probability measure} (total mass $1$) as the only object layer: Giry gives pushforward via kernel composition, Stoch gives random-map composition via string diagrams; neither equips a multiple-measure structure. Our \MPT\ (Definition~12, invertible measure-preserving transformation) instead unifies Lebesgue (geometric volume, infinite mass on $\mathbb R^d$, not itself a probability measure), replica-index counting measure (finite quotient-block counting), and binomial measure (Gaussian-vertex partition shared space) into the same shared measure space, without relying on the random-kernel-composition semantics of any single probability measure. Our increment is to write the first-order error $\mu_{\mathrm{err}}$ (the difference between block-counting and geometric volume) as an explicitly accountable, falsifiable reading under ``multiple measures + well-defined pullback (bijective partition)''---an object with no place and no definition in the single-measure Kleisli/Markov framework (the single-measure framework does not distinguish this difference). Our replica-index measure is precisely the engineering-reading baseline of the Giry pushforward on quotient blocks: the two are two notations of the same counting measure, and under the infinite-refinement limit (the $\betat$ tower $\to$ Bernoulli$(1/2)^{\mathbb N}$ $\to$ binary-expansion pushforward) they are normalization-equivalent to the Lebesgue probability measure---the only difference is that we additionally model the out-of-domain / inter-block error $\mu_{\mathrm{err}}$ explicitly. Details (see C.9).

Standard categorical probability (Giry monad, Stoch Markov category) gives reusable conclusions: probability measures form a monad, random maps form a Markov category, pushforward is well defined while pullback is generally not---we \emph{reuse} these and make no claim of new contribution. But its single-measure framework does not equip multiple measures nor block-collapse structure, hence has \emph{no apparatus to recover ``mass coming from outside the domain'' via pullback}: only pushforward gives the folded image, unable to explicitly turn the out-of-domain error legitimately ignored by a.e.\ on null support into a first-order falsifiable coefficient (the $\mathbb R^-$ positive-measure case of argument C.8; the ``category cancels elements $\to$ force out $\mu_{\mathrm{err}}$'' transformation, see main thread \S 1). Our \emph{contribution} fills the gap: the composition of bijective partition and \MPT\ multiple measures---using bijective partition (Theorem~1) to give a functional with well-defined pullback, and on the \MPT\ multiple-measure shared space reading the ``block-counting $\leftrightarrow$ geometric volume'' difference explicitly as $\mu_{\mathrm{err}}$, together with the three clauses of Theorem~7 in \S 6.3.3---this increment has no corresponding entry in Giry/Stoch. By $\Quot/\MorTop$ erasing preimages and object entities (mechanism in \S 2.1 categorical-layer essence, rooting in Assumption~1), the error reading is purified into a morphism (bijective-partition) reading, of the same origin with ``pullback must be well defined via bijective partition'', rather than competing with the standard framework.

\subsubsection{Antecedent axiom, system invariant, and generalized meta-categorical support}
\label{sec:2.2.2}

\noindent\textbf{Axiom 2 (binary-decidability axiom)}
The \emph{finite binary quotient-partition predicate} (complementary equivalence-class pair) used in this paper possesses binary-decision semantics (true / false); this universal claim covers only the predicate classes actually used in this paper, and does not claim that arbitrary formalizable topological propositions are binary-decidable (the decidability of general topological propositions is limited by Church--Turing, G\"odel incompleteness, and descriptive-set-theory hierarchies, beyond the scope of this axiom). \textbf{Corollary 1 (binary-predicate well-foundedness)} (see B.9), \textbf{Corollary 1a (no-derived-instance satisfiability)} (see B.11). Given a binary quotient-partition object $Q$ (equivalent binary predicate), at the meta level take the negation of this predicate, extract the complementary equivalence-class pair $(X_0,X_1)$, and use it to establish the correspondence between quotient blocks (Terminology~(f)(a)), providing support objects for morphism construction. The predicate itself is not computation and does not belong to the internal morphisms of the category; set intersection is realized by the topological-lattice intersection operation induced by the quotient partition, and the base category introduces no algebraic multiplication operator symbol.

\noindent\textbf{Definition 1 (generalized meta-categorical support object $\Sup(f,B)$)}\label{def:1}
Let $\mathcal C$ be a meta-category, and let $Src,Tgt$ be meta-collections taken from $\mathrm{Ob}(\mathcal C)$ or $\mathrm{Mor}(\mathcal C)$. The categorical-support object $\Sup(f,B)\subseteq Src$ is a source-side sub-meta-collection satisfying that the restricted morphism $f\big|_{\Sup(f,B)}:\Sup(f,B)\to B$ is a surjective morphism.

Categorical support is a morphism between sub-collections of two meta-collections, whose morphism form carries a pairing of inclusion and restriction morphisms (Proposition~3(i)); ``inter-categorical support'' is uniformly read as ``support between sub-collections''. Categorical support is one-to-many and irreversible (\S 1): surjectivity is guaranteed by Definition~1 and Proposition~3(ii), the missing injectivity is successively injected by the replica-index measure (Definition~17, \S 4.2), and lifted to a bijection via pushout--pullback and binary quotient partition; taking $B=Tgt$ degenerates to the initial form. The footing and narrowing motivation of this definition (see B.15).

\noindent\textbf{Definition 1a (differential-quotient relation $\mathrm dx\sim_\beta\mathcal D$)}\label{def:1a}
Fix the differential--predicate bijection tower $\betat$ (Lemma~4). For a differential-topological element $\mathrm dx$ in $\Hinf$ and a rank-1 minimal non-empty topological-differential subset $\mathcal D$, define
\begin{equation}
\mathrm dx\sim_\beta\mathcal D\quad\Longleftrightarrow\quad \mathrm dx\ \text{is carried within the quotient block marked by}\ \beta_n(\mathcal D)
\end{equation}
\textbf{Definition 1b (derived object)} (see B.16).

The object of $\MorTop$ is the quotient under this relation, called the \emph{differential quotient} (Terminology~(f)(b)). Any map's image on the minimal quotient is either completely paired with the source block (bijection) or falls into the same target quotient block (collapse), with binary decision and no intermediate state---this is the minimal-granularity entry of the quotient-collapse / quotient-bijection decision (Definition~8, Theorem~1).

\noindent\textbf{Corollary 2 (see Postulate~2 first-order equality; output)}\label{cor:2}
Under Assumption~1 (\S 2.4.3), let $\mathrm dx\sim_\beta\mathcal D$. Equip the differential value with any measure reading of $\mathcal D$ (block-counting or replica-index measure $\nu_{\mathrm{idx}}$, collectively denoted $\mathrm dx\odot\nu_{\mathrm{idx}}$); let the counting differential be $\mathrm d\#$. Then
\begin{equation}
\mathrm dx\odot\nu_{\mathrm{idx}}\ \sim_1\ \mathrm d\#
\end{equation}
That is, the difference between the two readings is borne exactly by \emph{one block collapse} (see the quotient-collapse image set of Definition~8), readable via measure descent in the partition category; $\mu_{\mathrm{err}}$ (Definition~5) is the measure reading of this carrier set. Thereby $\sim_1$ defines the \emph{relation between first-order error of the partition category and the set-category reading}: when the set layer preserves the binary quotient partition, the set-layer reading \emph{inherits} this first-order error---the criterion is ``difference is one block collapse'', not referring back to $\mu_{\mathrm{err}}$ itself. Proof (see B.17). Boundary: when the two readings are induced by mutually incompatible partition structures, first-order equality fails (see the metric-theoretic symptom of Definition~21 categorical leakage).

\noindent\textbf{Proposition 3 (morphism-ization of categorical support and surjectivity)}\label{prop:3}
Support carries an inclusion morphism and a restriction morphism (composable as morphism composition); for quotient bijections, the restriction morphism is surjective; first-order equality guarantees lifting without losing the binary quotient-partition constraint (no categorical leakage). Surjectivity only asserts ``every upper-layer object has a lower-layer support'', not a per-object reverse surjectivity. Statement and proof (see B.18).

\paragraph{Paradigm convention}
\begin{enumerate}
\item \emph{Prior supports posterior (classical, inside $\Setcat$)}: the prior pre-constructs the forward-propagation morphism and measure space inside the set category, unable to enumerate all forward-propagation outputs, and cannot directly construct $\Sup(f,B)$, hence reparameterization is introduced---the \emph{core contradiction}: lacking an analytic tool that simultaneously imposes surjectivity and subset-invertibility constraints on the morphism (comparison in D.1).
\item \emph{Enumerating posterior-supported prior (ours, cross-categorical)}: enumerating the base-object collection is equivalent to enumerating the lifted later-category object collection. If all forward-propagation outputs can be enumerated within the base-object collection, then a reversible forward propagation can be constructed to obtain a bijection, and backpropagation becomes the categorical-support object $\Sup(\cdot,\cdot)$ of forward propagation.
\end{enumerate}

\subsubsection{Three-layer categorical inductive construction}
\label{sec:2.2.3}

\noindent\textbf{Definition 2a--2c (three-layer categorical inductive construction)}\label{def:2ac}
Inductive base $\Quot$ (Definition~2, B.19), inductive step 1 $\MorTop$ (Definition~2a, B.20), canonical lifting $L':\MorTop\to\Setcat$ (Definition~2b), canonical descent functor $\Gamma$ (Definition~2c, B.21). The inductive base has only primitive object collections (quotient-predicate partitions, induced by binary-topological decision predicates), no morphisms, maps, or computation; the inductive step generates morphisms and operators layer by layer, producing the prior categorical structure from posterior primitive objects. \emph{Inductive base}: each primitive object of $\mathrm{Ob}(\Quot)$ corresponds to a binary decision predicate, encoding the posterior (backward correction \citep{bousquet2021theory}). \emph{Inductive step 1}: binary-merge selection picks a complementary equivalence-class pair $(X_0,X_1)$; the fiber bundle divides into self-map and cross-set two-set classes; enumerating $\mathrm{Ob}(\Quot)$ is equivalent to enumerating $\mathrm{Ob}(\MorTop)$, encoding the prior (forward propagation). \emph{Inductive step 2}: $L'$ lifts and assembles the fiber-nesting structure to $\Setcat$ (derived instance is a maximal compatible section). \emph{Convention}: only after lifting to $\Setcat$ do topological-lattice intersection and measure product unify into a formal multiplication operator.

\noindent\textbf{Definition 3 (quotient bijection)}\label{def:3}
The lifting morphism $\tau:\Quot\to\MorTop$ realizes a bijection between the quotient-block set and $\mathrm{Ob}(\MorTop)$ at the object-collection layer, called a \emph{quotient bijection}---the legitimate lifting channel of our categorical support, with no derivation relation to the quotient-block-set permutation itself nor to Assumption~1 (distinction in B.22).

\subsubsection{Measure categorization: meta-categorical object supports multiple categorical objects}
\label{sec:2.2.4}

The set category $\Setcat$ naturally accommodates multiple measures and their joint measure; two measures at the $\Setcat$ layer are a \emph{pushout surjection} of the fiber bundle (pushforward well defined, restriction morphism surjective), but the \emph{pullback is well defined only on bijective partitions} (Definition~1)---the pullback along a non-injective map is generally ill defined, which is precisely the motivation for introducing bijective partitions; whereas $\Setcat$ is \emph{difficult to construct a measure-preserving invertible transformation (\MPT)} (Definition~12). This section induces a chain \emph{surjection $\to$ bijection $\to$ \MPT}: use the analytically resolvable counting measure to decouple the (unresolvable) Lebesgue geometric side of \MPT; staying at the measure level forces confronting the dimensionality curse---no global translation-invariant measure in infinite dimensions, and translation invariance cannot be functionalized, three difficulties. For this, \emph{degenerate to the element-free support-partition category} (see \S 2.1): this degeneration preserves surjectivity (Proposition~3), obtains binary bijections (Proposition~4, Theorem~1), escapes the difficult-to-solve functional (Lemma~2), and the meta-categorical object collection supports the entire structure of the multi-category (Proposition~5).

\noindent\textbf{Proposition 4 (simplest binary-merge bijection)}\label{prop:4}
The binary-merge relation has exchange symmetry; the selection orders $(X_0,X_1)$ and $(X_1,X_0)$ induce isomorphic morphism constructions. Therefore the simplest cross-set fiber bundle providing a support object $\Sup(\cdot,\cdot)$ is a bijection (this bijection belongs to the inter-category $\Quot\to\MorTop$ lifting construction of Definition~3, \S 2.2.3). This proposition gives the simplest form of the bijection of inductive step 1 (inter-category lifting), is a special case of Definition~3 and the direct premise of Proposition~3(ii). \emph{Proof (sketch)}: bijection holds at the object-collection layer: binary-merge / split pairing gives an $\mathrm{Ob}$-layer bijection (merge takes quotient, split takes preimage, pairing reversible); exchange symmetry gives selection-order independence; this proof does not promise a point-wise bijection at the $\Setcat$ layer (see Corollary~4). Full proof (see B.27).

\noindent\textbf{Proposition 5 (existence of cross-category object-collection lift)}\label{prop:5}
There exists an \emph{inter-category} meta-level object-collection lifting process (the inter-category lifting functor $\Quot\to\MorTop$ realized by categorical support $\Sup(\cdot,\cdot)$, Definition~1), which from the primitive object collection $\mathrm{Ob}(\Quot)$ generates the complete category $\Quot$ and the category $\MorTop$, so that all objects, morphisms, and fiber-bundle structures of $\MorTop$ are uniquely determined by $\mathrm{Ob}(\Quot)$ (this lifting is an inter-category construction, presupposing no $\Setcat$-layer point elements). \emph{Necessary and sufficient conditions}: (1) the reverse path of the lifting process has topological invertibility; (2) the construction satisfies micro-enumerability; where (1) is given by the exchange symmetry of Proposition~4, and (2) is given by the object-collection bijection of inductive step 1 (Definition~3). \emph{Proof (sketch)}: categorical induction in five steps: inductive base (Axiom~2, Corollary~1) $\to$ necessary condition $\to$ fiber-bundle operator derivation $\to$ enumerability argument $\to$ top-level closure. Full proof (see B.27).

\noindent\textbf{Corollary 3--8 (structural readings at the object-collection layer)}
Object-collection bijection does not equate to categorical equivalence; our construction does not guarantee a global object-set bijection between the $\Setcat$ derived-instance layer and $\Quot$. Legal bijections are allowed between binary quotient blocks, and the prior fiber topology can be reconstructed from the posterior quotient partition; classical stochastic reparameterization does not preserve invertibility. The six conclusions are stated respectively in B.28, with proofs corresponding item by item in B.29.

\noindent\textbf{Definition 4 (topological-resolvable binary-partition criterion)}\label{def:4}
\emph{First existence} (topological measure-theoretic categorical existence): the object stays in $\Quot/\MorTop$, not lifted to $\Setcat$, and a functional cannot be written; \emph{second existence} (finite-dimensional analytic / parameter-family reachability): the object is lifted to $\Setcat$ and can be approximated by a finite parameter family under a specified measure.

The deviation of generative models cannot be reduced to either single side: topological objects are reachable by binary-bijection collection within the partition category (Proposition~6), but after lifting they are not necessarily approximable by a finite parameter family---this gap is the \emph{core source of first-order error} (the two existences are called first and second existence in order). Integral reading executes by quotient blocks (finite truncation: per-block finite average on $Z_m$), not point-wise sampling; inner compactness (Proposition~18) guarantees that the lifted-integral output is a finite scalar.

\noindent\textbf{Proposition 6 (meta-level enumeration existence)}\label{prop:6}
Enumerating all quotient-partition objects within $\mathrm{Ob}(\Quot)$ is equivalent to enumerating all objects of $\mathrm{Ob}(\MorTop)$. This is a \emph{broader} existence than the two existences of Definition~4: (i) enumeration equivalence holds; (ii) it does not promise finite-parameter analytic reachability at the $\Setcat$ layer; (iii) it is not implied by separate topological existence---the bijection thereby falls into the enumerable domain of object collection (finite layer $2^n$ blocks; the integral boundary on $\mathbb R$ searchable to $\pm\infty$, the boundary itself null and negligible), and is the operational object of search, rather than a pure existential assertion or a guarantee of effective construction (see \S 5.1). It is precisely this third existence that frees categorical support from categorical leakage (Proposition~3(iii); Proposition~19, \S 5). \emph{Proof (sketch)}: (i) enumeration equivalence is directly given by the binary-merge selection--quotient-equivalence construction of inductive step 1; (ii) enumeration equivalence only asserts an object-collection-layer bijection, containing no parameter-family component, hence is mutually non-implicative with the two existences of Definition~4. Full proof (see B.31).

Our ``searchable'' is only read as a surjection-cover reading, not promising that one can effectively enumerate term by term or find every bijection in the infinite predicate tower (constructive search, explicitly not promised, belonging to \S 5.1). Details (see E.1).

\subsection{Iterative morphism first-order error: injection lifted to bijection}
\label{sec:2.3}

After categorical support, via the dual-identity transformation of the replica-index measure, is lifted to a bijection (see \S 1), the error consequence of morphisms other than bijections---morphisms still staying at injection---becomes an accountable object: replica-index measurability (non-null measure) is the price of this transformation, and the conversion of non-null measure into first-order measurable error is precisely the realization form of this price at the set-measure level (see \S 1, ``price: \MPT-ization and first-order error''). This section thereby turns this price into a formal error model.

\subsubsection{Formal error modeling}
\label{sec:2.3.1}

This section uses Definition~4 to distinguish the categorical layer of objects: all errors are defined as set-measure-layer objects; distinguishable conditions are placed in \S 2.4 as fiber-bundle constraints, and the primitive construction introduces no functional; the primitive category uses differential-partition enumeration, not a point-wise enumeration concept (point-wise belongs to the upper $\Setcat$ reading, \S 2.1). Point-wise enumerability is not a corollary of Assumption~1 (\S 2.4.3)---even without imposing any point-wise-layer assumption, differential enumerability (Proposition~6, Definition~3, Theorem~1) is fully preserved.

\noindent\textbf{Definition 5 (first-order measurable error $\mu_{\mathrm{err}}$)}\label{def:5}
Let $T^\ast$ be the ideal truth quotient self-map endogenously given by the quotient-partition category (Proposition~7, $\mathcal T^\ast$), and let $T$ be the model-fitting morphism. The canonical definition of the deviation set is given in the \emph{partition category} (point-wise comparison is ill defined at the $\Setcat$ layer): fix $\betat$ (Lemma~4):
\begin{equation}\label{eq:def5-E}
\mathcal E_{\Quot}\triangleq\big\{B_w\in\mathcal P_n:\ \text{under predicate word }w,\ T\text{ and }\mathcal T^\ast\text{ disagree at the block level}\big\}
\end{equation}
Its $\Setcat$-layer preimage reading is $\mathcal E\triangleq\bigcup_{B_w\in\mathcal E_{\Quot}}B_w^{(X)}$ ($B_w^{(X)}$ is the preimage carrier of quotient block $B_w$). Taking $\mu_\otimes$ as the partition-induced measure product ($\nu_{\mathrm{idx},n}\times\nu_{\mathrm{idx},n}$), with $\Delta$ the diagonal embedding, the first-order error measure is derived from the fiber-bundle measure-invariance constraint (\S 2.4) as
\begin{equation}\label{eq:def5-err}
\mu_{\mathrm{err}}(A)\triangleq\Sup^{\Setcat}_{\Quot}\Big[\nu_{\mathrm{idx},n}\big(\mathcal E_{\Quot}\cap\Sup^{-1}_{\Quot}(A)\big)\Big],\qquad A\in\sigma(\mathcal P_n)
\end{equation}
where $\Sup^{\Setcat}_{\Quot}[\cdot]$ is the \emph{measure descent--re-ascent symbol of categorical support} ($\Sup^{-1}_{\Quot}$ is its quotient-block-layer pullback): all definitions and derivations written in the partition category $\Quot$ (quotient-block set $\mathcal P_n$, normalized counting measure $\nu_{\mathrm{idx},n}$, scale $2^{-n}$) have a \emph{unique bijective characterization} at the $\Setcat$ layer (preimage carrier $X$) (excluding trivial characterizations). Hence $\mu_{\mathrm{err}}$ does not directly intersect the diagonal embedding $\Delta(\mathcal E_{\Quot})\subset\mathcal P_n\times\mathcal P_n$ with $\Setcat$-layer $A\times X$ (that would simultaneously cause a scale-difference factor $2^{-n}$ and a type mismatch), but first descends $A$ to the quotient-block layer, takes the intersection, then re-ascends the reading---$\nu_{\mathrm{idx},n}$'s normalization appears only once at the quotient-block layer, and the $\Setcat$-layer area $u/2^n$ is exactly its re-ascended image. $\mu_{\mathrm{err}}$ is $\Fcal$-measurable, quantifiable, and characterizes instance-level pairing deviation; its source is the gap between partition-induced block-counting measure and geometric measure (B.7.6).

Volume-error \MPT\ sharing and morphism escape from preimages: first order. Our ``first order'' is a \emph{structural order}: taking the quotient bijection (Definition~3, \S 2.2.3) as the baseline, any deviation that collapses a quotient bijection into many-to-one, or separates block-counting reading from geometric-measure reading, is uniformly read as first order---rather than an asymptotic Taylor remainder with respect to an external small parameter ($\varepsilon$, partition fineness, step size, dimension). The binary quotient-partition bijection is induced into a differential bijection via the Hilbert differential (Lemma~4, Corollary~2, $\betat$ tower); its \emph{collapse} (some quotient block repeatedly covered or mis-paired at the $\Setcat$ layer, so that the block-counting multiplicity and the Lebesgue-volume reading no longer agree) is precisely the accounting object of $\mu_{\mathrm{err}}$; Postulate~2's $\sim_1$ (first-order equality) is this reading: the differential reading and the counting reading differ by exactly one collapse, absorbable within the same order. On the triple fiber-bundle integral (Definition~16, Eq.~\eqref{eq:12b} factor~\ding{172}), $\mu_{\mathrm{err}}$ obtains measure-theoretic content: \emph{measure-space volume error}---the difference between block-counting multiplicity and geometric-measure volume (see \S 6.2.2, the area-difference of zero / non-zero saddle-point division is its manifestation), and is \emph{shared by the invertible \MPT}: the invertible measure-preserving transformation conserves multiplicity weight (Proposition~11), introducing and changing no volume error; $\mu_{\mathrm{err}}$ is the residue of ``bijection collapse destroys multiplicity'' outside the measure-preserving channel---i.e.\ the precise meaning of the ``price'' in \S 1. Thereby the morphism \emph{escapes preimage objects}: the measure space is a construction space, not a real space, so ``measure preservation'' can be stated as a characterization of the morphism itself, detached from point-wise preimages; but measure preservation must first be \emph{invertibilized} (see Definition~10, Definition~12's measure-preserving invertible pairing, i.e.\ invertible \MPT), as mere measure preservation is insufficient to escape preimages. \emph{Measure-preserving / non-measure-preserving division of labor} (Proposition~8) (bijection does not preserve measure, quotient-translation-invariance preserves measure, integral does not preserve measure, three-way division): differential-topological equivalence gives a closed surjection (Corollary~2, Proposition~3) without implying measure preservation (Lemma~1(ii)); measure preservation is automatically satisfied by partition-induced readings (Lemma~1(i)) or explicitly assumed when taking Lebesgue. Operability closes at two points: the differential measure space makes the morphism \emph{measurable} ($\mu_{\mathrm{err}}$ is $\Fcal$-measurable, quantifiable), and the bijection makes the morphism \emph{functionally measurable} (Theorem~1, Proposition~20: quotient bijection is the unique channel turning predicate matching into an integrable reading). Above this lies the pushforward accumulation that destroys measure preservation / bijection, classified into high order $\mu_{\mathrm{high}}$ (two sources, Definition~6); first order and high order thereby obtain a decidable boundary at the measure layer.

\noindent\textbf{Definition 6 (high-order error $\mu_{\mathrm{high}}$)}\label{def:6}
Let $T^\dagger$ be the ideal measurable-bijection reference morphism, $T$ any measurable morphism, and $\boldsymbol T_{\otimes\#}$ the product-space pushforward operator. Restrict $\mu_{\mathrm{err}}$ to the sample support set $S$ and apply pushforward on the product space, defining the signed measure
\begin{equation}
\mu_{\mathrm{high}}\triangleq\boldsymbol T_{\otimes\#}\big(\mu_{\mathrm{err}}\vert_S\big)-\boldsymbol T_{\otimes\#}^\dagger\big(\mu_{\mathrm{err}}\vert_S\big)
\end{equation}
$\mu_{\mathrm{high}}$ characterizes the structural deviation accumulated by iterating non-bijection morphisms; its generation has two sources: \emph{categorical leakage} and \emph{topological collapse} (the binary quotient-partition category is broken, degenerating into multi-partition, with minimal cross-partition quotient blocks / straddling quotient blocks). \emph{Pure dimension-lifting does not endogenously generate high-order error, independently given by Theorem~4 (\S 4.2)}; this paper does not use the exogenous-operation concept of ``dimension-reduction projection''; the endogenous reading of dimension reduction is in the declaration after this definition. \emph{Order criterion (distinguished from Definition~5's ``first order = single quotient-bijection collapse'')}: the ``topological collapse'' of $\mu_{\mathrm{high}}$ specifically refers to \emph{partition-boundary inconsistency}---different blocks' partition boundaries no longer coincide, the binary quotient partition degenerates into three-/four-fold (i.e.\ the precise meaning of ``destroying measure preservation / bijection''), which must hold after $\ge 2$ cumulative applications of morphism pushforward; whereas \emph{single in-block collapse} (some quotient block covered many-to-one, block-counting and Lebesgue readings separated) \emph{does not enter this paper's first-order error reading target}---the engineering-side first-order error reading \emph{only characterizes between quotient blocks} (inter-block) predicate mismatch; \emph{in-block} micro-level (point-level) many-to-one error \emph{objectively exists and is reading-visible}, but is not within $\mu_{\mathrm{err}}$'s measurement target, and the engineering side \emph{actively excludes it from the measurement scope (active neglect)}---neither claiming it does not exist nor claiming it is controlled at first order; the two must not be conflated. Categorical leakage has been excluded (Proposition~19); topological collapse (multi-partition / cross-partition cumulative form) is an exogenous operation, not an endogenous error of the Gaussian-\MPT\ functional family (Definition~6); the pure dimension-lifting layer does not endogenously generate high-order error (Theorem~4, \S 4.2); hence $\mu_{\mathrm{high}}$ has no generation mechanism in this paper's framework and its engineering realization.

Our ``dimension reduction'' is the decrease of quotient-block count under binary quotient-partition merging, absolutely not an exogenous projection operator; in-block micro-level error is actively neglected (not projection-erased) by the triple tensor full-contraction in the discrete-atom limit of the triple fiber-bundle integral, and pure dimension-lifting is independently handled by Theorem~4. Details (see C.10). The split of Definitions~5/6 is the necessary translation from differential characterization $\to$ quotient-block characterization: the canonical source of $\mu_{\mathrm{err}}$ is differential-quotient-block summation (multi-partition rather than bi-partition, $2/3$ is the discrete-counting ratio of subset base 3 to hit base 2), and one must distinguish no-collapse (block-layer injection) from block-pairing bijection (uniqueness modulo label permutation); in-block differential non-bijection error objectively exists but $\mu_{\mathrm{err}}$ only characterizes between quotient blocks. Details (see C.11).

\subsubsection{Non-null measure under quotient partition: support surjection induces unique pushout--pullback bijection}
\label{sec:2.3.2}

The quotient-layer bijection of this section is localized under the \emph{pushout--pullback} structure: categorical support is a surjective morphism (Definition~1), and surjectivity guarantees that every reading on the target side has a preimage to pull back, so the quotient-layer bijection induced by support surjection obtains the characterization of a \emph{pushout--pullback bijection}: the pushforward $\boldsymbol\pi_\#$ and the pullback $\boldsymbol\pi^\#$ are mutually inverse pairings on it (Definitions~26,~24, \S 5.3--\S 5.4); its form at the block-pairing level is the self-map quotient-equivalence bijection (Terminology~(f)(g)), existing under no-collapse and unique modulo target-block label permutation (Theorem~1).

\noindent\textbf{Definition 7 (binary quotient partition and binary iterative refinement)}\label{def:7}
A \emph{binary quotient partition} of $X$ is a set of finite, pairwise-disjoint non-empty measurable subsets $\mathcal P=\{X_i\}$ satisfying $\bigcup_i X_i=X$. \emph{Binary iterative refinement is executed within the partition category, operating on the collection of predicates (predicates are not computation)}: split each current block's predicate into two, or discard and keep without splitting; finite collections yield any finite measurable quotient partition, and the \emph{final quotient-block set is independent of the collection order} (synchronized and non-discarding gives the global refinement tower $\mathcal P_n$ of $2^n$ blocks).

Differential refinement does not cross any partition. The differential quotient at any refinement layer closes only within this layer's predicate word, hence differential refinement internally yields native finite quotient-block partitions ($\ne$ Hilbert continuous partition); the accounting object of first-order error $\mu_{\mathrm{err}}$ is the native finite quotient block locked by the partition boundary (rather than the Lebesgue continuous reading), which is precisely the root of why error must be accounted at the quotient-block layer rather than the Lebesgue layer. Details (see C.12).

\noindent\textbf{Definition 8 (no-collapse / quotient-translation-invariance)}\label{def:8}
Let $[F]:[X]\to[Y]$ be a \emph{quotient surjection}. \emph{(a)(b)(c) are set-theoretic / measure-theoretic criteria under a general quotient surjection, not presupposing $[X],[Y]$ finite or equal cardinality} (no-collapse = pairwise-disjoint image sets; quotient-translation-invariance = measure-preserving and inter-block-mutually-exclusive along the $\boldsymbol\pi_\#/\boldsymbol\pi^\#$ channel; $\Setcat$ functional well defined = excluding element-intersection-without-coincidence). ``Finite and equal cardinality'' is the special case of \emph{Theorem~1}'s ``no-collapse $\Leftrightarrow$ quotient-set bijection $\Leftrightarrow$ block-pairing bijection'' (finite-set injection $\Leftrightarrow$ surjection $\Leftrightarrow$ bijection; for unequal cardinality / non-surjection the failure is in the boundary statement) and the premise of \emph{Proposition~7}'s replica-index cardinality compatibility:
\begin{itemize}
\item[(a)] \emph{No-collapse (block-layer injection criterion).} $[F]$ is \emph{no-collapse} if for any distinct $X_i,X_k\in[X]$, $[F](X_i)\cap[F](X_k)=\varnothing$---intersecting image sets are \emph{quotient collapse} (the manifestation of $\mu_{\mathrm{err}}$ at the quotient level). No-collapse is only a block-layer injection condition, carrying no inherent surjectivity.
\item[(b)] \emph{Quotient-translation-invariance (the characterization target of this definition).} The quotient-block family $\{[F](X_i)\}$ is \emph{translation-invariant} at the quotient level: along the pushforward $\boldsymbol\pi_\#$/pullback $\boldsymbol\pi^\#$ channel, the quotient-block reading preserves measure under block pairing ($\sigma_\#\nu_{\mathrm{idx},n}=\nu_{\mathrm{idx},n}$, Proposition~8, \S 2.4.1), and inter-block mutual exclusion is not overlapped by translation pairing. \emph{No-collapse is the necessary premise of quotient-translation-invariance}, with two further criteria:
 \begin{itemize}
 \item[(b1)] \emph{Collapse $\Rightarrow$ no translation-invariance between quotients}: if there is a quotient collapse (two image sets intersect), then at the intersection both the left-block and right-block readings are read simultaneously, inter-block mutual-exclusion translation pairing cannot be maintained, hence the quotient-block family \emph{cannot be translation-invariant} at the quotient level---quotient-translation-invariance holds only under no-collapse (pairwise-disjoint blocks).
 \item[(b2)] \emph{Intersection bijection but no translation-invariance}: two intersecting quotient blocks can still each establish a \emph{block-level bijection} with two objects (each as source block to its respective target block is an injection--surjection pairing, i.e.\ the ``block-pairing level'' pairing of Theorem~1(c)), but it is precisely their \emph{intersection} that destroys the mutual exclusion of the whole block family, hence overall there is \emph{no translation-invariance}. That is, ``block-level bijection exists'' does not imply ``translation-invariance''---translation-invariance demands more than block-object bijection: it requires the entire quotient-block family to be pairwise-disjoint at the quotient level, with no overlapping readings, rather than only requiring single-block-to-single-target pairing.
 \end{itemize}
\item[(c)] \emph{Functional well defined after Hilbert descent to Euclidean (application conclusion of this definition).} When the Hilbert differential descends (projects) to Euclidean via finite-layer standard Euclidean embedding (see H3, \S 5.3, \S 3.2 finite truncation), if translation-invariance is lost (collapse / inter-block intersection appears), then \emph{the equivalent-class differential cannot be defined as a functional of the set category $\Setcat$}: a functional defined on $\Setcat$ must \emph{exclude the case ``elements intersect but do not coincide''}---elements (points) of $\Setcat$ are mutually exclusive and do not allow overlapping readings, while equivalent-class differentials (quotient-block readings) constitute well-defined elements of $\Setcat$ only when their family is \emph{pairwise-disjoint} (no-collapse, translation-invariance). Hence \emph{no-collapse / quotient-translation-invariance is the necessary and sufficient condition for ``equivalent-class differential can be defined as a $\Setcat$ functional''}, strictly established by Theorem~1 (\S 2.3.3, quotient equivalence class to element disjoint) and directly supporting the block-layer premise of Theorem~7 ``binary quotient-partition objects support the set-categorical functional via measure descent--recovery'' (\S 6.3.3) and Corollary~13 ``closure of set-categorical functional assembly'' (\S 5.3).
\end{itemize}

\subsubsection{Quotient equivalence class to element disjoint induces first-order error}
\label{sec:2.3.3}

\noindent\textbf{Theorem 1 (quotient equivalence class to element disjoint: no-collapse $\Leftrightarrow$ quotient-block family pairwise disjoint $\Leftrightarrow$ $\Setcat$ functional well defined)}\label{thm:1}
Let $[F]:[X]\to[Y]$ be a quotient surjection; the necessary-and-sufficient-condition statement of this theorem \emph{does not presuppose $[X],[Y]$ finite or equal cardinality}, and holds for any (including infinite, unequal-cardinality) quotient surjection. Its essence is to establish ``\emph{quotient equivalence class to element disjoint}'' as the necessary and sufficient condition for the later ``equivalent-class differential can be defined as a $\Setcat$ functional'' (Definition~8c) and ``quotient-translation-invariance'' (Definition~8b), rather than a finite-set bijection combinatorial rewrite; the finite equal-cardinality case is separately listed as the standard corollary (d). Specifically:
\begin{itemize}
\item[(a)] \emph{No-collapse (quotient equivalence class to element disjoint).} $[F]$ is no-collapse---the image family $\{[F](X_i)\}$ is pairwise disjoint (Definition~8a). This is precisely ``\emph{quotient equivalence class to element disjoint}'': each quotient block (quotient equivalence class) as a $\Setcat$ element is mutually disjoint, excluding ``elements intersect but do not coincide'' (Definition~8c's element-mutual-exclusion requirement); $\Setcat$'s elements (points) are mutually exclusive and do not allow overlapping readings, hence equivalent-class differentials constitute well-defined elements of $\Setcat$ only when the quotient-block family is pairwise disjoint.
\item[(b)] \emph{Quotient-translation-invariance (under the partition-induced block-counting reading $\nu_{\mathrm{idx},n}$).} The quotient-block family $\{[F](X_i)\}$ is translation-invariant at the quotient level (Definition~8b): no-collapse is its necessary premise (Definition~8b1); the existence of block-level bijection (single-block to single-target-block pairing) is insufficient to guarantee overall translation-invariance---the whole requires the entire quotient-block family to be pairwise disjoint with no overlapping readings (Definition~8b2). Under the counting measure $\nu_{\mathrm{idx},n}$, measure preservation is automatically given by Lemma~1(i); under the geometric (Lebesgue) measure, measure preservation must be explicitly assumed (Lemma~1(ii)). Hence (a)$\Leftrightarrow$(b).
\item[(c)] \emph{Can support $\Setcat$ functional (necessary and sufficient condition).} This quotient structure can support the set-categorical functional---the equivalent-class differential (quotient-block reading) thereby obtains a well-definedness, constituting a well-defined element of $\Setcat$; i.e.\ (a)$\Leftrightarrow$(c) is the necessary and sufficient condition for ``equivalent-class differential $\Rightarrow$ $\Setcat$ functional well defined'' (Definition~8c), directly serving the block-layer premise of Theorem~7 ``binary quotient-partition objects support the set-categorical functional via measure descent--recovery'' (\S 6.3.3) and Corollary~13 ``closure of set-categorical functional assembly'' (\S 5.3).
\item[(d)] \emph{(finite equal-cardinality special case, standard corollary, not our contribution).} When $[X],[Y]$ are \emph{finite and equal cardinality}, (a) degenerates to ``$[F]$ is a quotient-set bijection'' and the block pairing is unique modulo target-block label permutation---this is the standard result of finite-set ``injection $\Leftrightarrow$ surjection $\Leftrightarrow$ bijection'', which we \emph{reuse} rather than newly prove; it is only invoked where ``block-pairing uniqueness'' is needed (Definition~25a, Proposition~7, B.105's epoch+1 consistent-enumeration replica-index-cardinality-compatible block-pairing self-map family), and does not constitute the discriminative power of Theorem~1; outside the premise (unequal cardinality / non-surjection), (a) still holds but this special-case bijection uniqueness fails (boundary statement).
\end{itemize}
\emph{Proof} (see B.34) ((d)'s finite equal-cardinality special case and (a)$\Leftrightarrow$(c)'s general equivalence are both rigorously established therein).
\emph{Note: non-emptiness of Theorem~1.} The discriminative power of this theorem is not in finite-set injection$\Leftrightarrow$bijection (that is the standard corollary listed in (d), not our contribution), but in establishing ``quotient equivalence class to element disjoint'' as the necessary and sufficient condition for ``equivalent-class differential can be defined as a $\Setcat$ functional'' ((a)$\Leftrightarrow$(c)) and explicitly bridging Definition~8's quotient-translation-invariance criterion ((a)$\Leftrightarrow$(b)); it carries the decision condition of Definitions~8b(c), providing the block-layer premise for later Theorem~7 and Corollary~13. Boundary statement: the equivalence relation acts only on the block-pairing coarse granularity; intra-block morphisms may still be many-to-one (block-pairing bijection and intra-block derived-instance bijection are decoupled); discriminative-power boundary (see B.34). \emph{What is truly captured}: the elementary $\Setcat$-functional conclusion of this theorem is a precise characterization of \emph{predicate matching}---under the double counting induced by bi-partition, the counting Lebesgue measure $\nu_{\mathrm{idx},n}$'s reading on the self-morphism ``differential quotient $\in$ quotient block $\to$ differential quotient $\in$ quotient block'' (match is bijection, mismatch is collapse) converges in $\Setcat$ to the \emph{size and partition boundary} of relation $R$; this reading is precisely what standard single-space measure theory cannot explicitly give in the same domain due to the null-embedding ambiguity, hence its \emph{nontriviality lies not in the conclusion itself but in the nontrivial-input construction (fiber-bundle pullback and categorical support) that makes it readable and computable}.

The operational role of this theorem is to \emph{enable exhaustive enumeration of all epoch+1}: Definition~25a requires adjacent epochs' quotient-block pairings to be unique modulo label permutation (the bijection uniqueness of special case (d)), otherwise epoch+1's counting iteration cannot be consistently enumerated---i.e.\ \emph{without Theorem~1's block-pairing well-defined premise, one cannot enumerate all epoch+1} (see \S 2.2.4, enumerability under the multi-measure shared space depends on block-pairing bijection).

\noindent\textbf{Proposition 7 (cardinality compatibility)}\label{prop:7}
For a finite binary-iteratively-generated measurable quotient partition, if $|[X]|=|[Y]|$ and $[F]$ is no-collapse, then $[F]$ generates a quotient-layer bijection self-map $\mathcal T^\ast$; further, if $\mathrm{card}_\#(S)\le\mathrm{card}_\#(J)$ (Definition~9(a), B.35), then there is no forced replica collapse due to insufficient replica-index cardinality (sample--quotient-block attribution attaches no cardinality condition). Proof (see B.36).

\subsection{Fiber-bundle morphism measure-invariance constraint and formal problem: product-invariance induced by differential metric semantics}
\label{sec:2.4}

The $\Quot/\MorTop$-layer quotient partitions are only Boolean-decidable, possessing no metric semantics; to obtain the metric size of a differential-partition union one must introduce integration, yet infinite partitions necessarily generate infinitely many infinitesimal differential elements, destroying the well-founded condition of the categorical inductive base, hence a distinguishable partition lattice requires at least two sets of partition boundaries. This section first states the derived status of this constraint and the bearer of measure preservation (\S 2.4.1), then gives the structural characterization of preimage-volume mutual exclusion where the constraint fails (\S 2.4.2), and finally closes with the formal problem, the backbone extra mathematical assumption, and the realization condition (\S 2.4.3).

\subsubsection{Derivation of the measure-product-invariance constraint}
\label{sec:2.4.1}

This section gives the primitive and derived versions of this constraint. \emph{The measure-product-invariance constraint is a derived constraint for screening distinguishable partition lattices, not an ad hoc axiom; the compatible product measure $\mu_\otimes$ is a derived product of the constraint, not a logical starting point.} (Here $\mu_1,\mu_2$ at the $\Quot/\MorTop$ layer uniformly refer to the $\Setcat$-layer readings of \emph{partition-induced structures}; the gap between the two readings is the source of $\mu_{\mathrm{err}}$.)

\noindent\textbf{Proposition 8 (bearer of measure preservation)}\label{prop:8}
The measure preservation of partition-induced readings is given by the Hilbert differential (quotient-translation-invariance preserves measure, $\sigma_\#\nu_{\mathrm{idx},n}=\nu_{\mathrm{idx},n}$); when taking the geometric measure at the $\Setcat$ layer, measure preservation must be explicitly assumed; the $\Hinf$ layer has no non-trivial translation-invariant measure (Proposition~21b), hence the integral side inherits no measure preservation---bijection does not preserve measure, quotient-translation-invariance preserves measure, integral does not preserve measure, the three-way division of labor (B.2). Proof (see B.37).

\noindent\textbf{Principle 1 (differential-topological-volume equivalence class and predicate-matching accounting principle; Principle~1 of the whole paper)}\label{princ:1}
All measure readings of the whole paper follow one principle, whose four clauses respectively give the definition of volume, the equivalence relation of the two volumes, the evaluation object of bijections, and the accounting of non-bijections:
\begin{itemize}
\item[\textbf{P1.1}] \emph{Hilbert-differential definition of volume.} The \emph{differential-topological volume} of the partition category (the $\nu_\mathrm{idx}/\nu_M$ on quotient blocks / differential quotients) and the \emph{differential-measure-space volume} of the set category (Lebesgue / geometric) are both defined by the \emph{Hilbert differential}: the former is attributed to the quotient-block reading by $\mathrm dx\sim_\beta\mathcal D$ (Definition~1a), the latter is assembled and read via $L'$ (Definition~2b); the common material is the $\mathcal H^\Setcat$ of Assumption~1 and the $\betat$ tower.
\item[\textbf{P1.2}] \emph{The two volumes are mutually equivalence classes (translation-invariance; equal readings after lifting--descent).} Both sides are translation-invariant (quotient layer $\sigma_\#\nu_{\mathrm{idx},n}=\nu_{\mathrm{idx},n}$; $\Setcat$ layer by Postulate~1); after lifting--descent along $\boldsymbol\pi_\#,\boldsymbol\pi^\#$ the readings are equal (Postulate~2 first-order equality). Hence the two are equivalence classes under the same Hilbert differential, and cross-category volume comparison does not depend on either side's point-wise geometry.
\item[\textbf{P1.3}] \emph{The evaluation object of a bijection is predicate matching.} The main statement of Theorem~1 is ``\emph{quotient equivalence class to element disjoint} $\Leftrightarrow$ no-collapse $\Leftrightarrow$ can support $\Setcat$ functional'' (Definition~8c); its finite equal-cardinality special case (d) gives no-collapse, quotient-set bijection, block-pairing bijection as three-way equivalent (modulo target-block label permutation). Axiom~2 excludes half-right-half-wrong; the area of predicate mismatch is the normative scope of first-order error (Proposition~23a(ii), whose ``no half-credit discount'' is supported by the proof that straddling quotient blocks must be a null set, see \S 6.3.3 Theorem~7 proof (b) and B.104).
\item[\textbf{P1.4}] \emph{The multiplicity of a non-bijection is normalized by the replica-index measure.} The prior-and-posterior replica-index measure records how many times a non-bijection is repeatedly covered, and absorbs it into first-order error via counting-measure normalization ($2^{-n}$).
\end{itemize}
(Status and full chain of Principle~1, see B.38.) \noindent\textbf{Lemma 4 (differential--predicate bijection)} (see B.39) \noindent\textbf{Corollary 9 (differential mutual exclusion induces predicate mutual exclusion)} (see B.41)

Let $(X,\mathcal F)$ be a measurable space equipped with two $\sigma$-finite measures $\mu_1,\mu_2$; let $T:X\to X$ be a measurable morphism, $\boldsymbol T_\otimes(x,y)\triangleq(T(x),T(y))$.

\noindent\textbf{Definition 9 (product-invariance and counting-cardinality unified constraint)}\label{def:9}
(see B.35) \noindent\textbf{Proposition 9a (primitive version: product-invariance under partition-induced measure)} (see B.43), \noindent\textbf{9b (derived version: product-invariance under invertible \MPT)} (see B.45). \emph{Proof} (see B.47).

The measure preservation of Proposition~9a is induced by the partition structure, while that of Proposition~9b is given by explicit assumption; the two must not be conflated: when taking the Lebesgue measure at the $\Setcat$ layer, mere measurable bijection does not imply product-invariance. Details (see E.3).

\subsubsection{Measure pushforward dilemma in sets: preimage-volume mutual exclusion}
\label{sec:2.4.2}

When a morphism does not satisfy the measure-product-invariance constraint, there is no constraint-induced compatible product measure; but one can still \emph{formally construct} a formal product set-function $\mu_\otimes\triangleq\mu_1\times\mu_2$, at which point preimage-volume mutual exclusion appears.

\noindent\textbf{Corollary 10a (preimage mutual exclusion, core identity)}\label{cor:10a}
Fix $\betat$, let $\{\mathcal D_i\}$ be a family of rank-1 minimal non-empty topological-differential subsets that are pairwise differentially mutually exclusive ($\nu_M(\mathcal D_i\cap\mathcal D_j)=0$). Then for a measurable morphism $T$: (i) predicate mutual exclusion $\beta(\mathcal D_i)\wedge\beta(\mathcal D_j)=\mathsf{False}$; (ii) single-morphism preimage mutual exclusion $A_i\cap A_j=\varnothing$ (cross-morphism case holds under quotient-layer block pairing); (iii) for any $\sigma$-finite $\mu_1,\mu_2$, $\mu_1(A_i\cap A_j)=\mu_2(A_i\cap A_j)=0$ (mutual exclusion is derived from structural disjointness, independent of measure choice). Proof (see B.48). \emph{Note (split numbering, added condition)}: (i)--(iii) are set-theoretic identities; (ii) is the direct reading of (i) under $\betat$ for a single morphism; \emph{this corollary only registers the set-theoretic fact of preimage mutual exclusion, bearing no constructive content of ``how shared measures fuse'' (the latter is the independent Construction Fact~10a below)}.

\noindent\textbf{Construction Fact 10a (see Corollary~10a's MPT pairing fusion of shared measures; not a logical consequence, requires absolute continuity)}\label{fact:10a}
In the \MPT\ shared-measure space, \emph{there exists no non-zero shared measure carried on $A_i\cap A_j$---shared measures can only be obtained by invertible-\MPT\ pairing fusion between quotient blocks} (see Definition~13's restatement + the independent construction assertion of the \MPT\ pairing structure). This fact is \emph{not} a derivation of Corollary~10a (i)--(iii), and must be separately numbered to decouple it from ``preimage mutual exclusion''. Its establishment \emph{requires an explicit absolute-continuity condition}: one must have $\mu_1\ll\mu_2$ (Definition~13, \S 3.1, or the corresponding absolute-continuity relation under the pairing channel $p_T$; explicit definition of $p_T$), provided by the descent continuity of the parameter space (gradient descent) and the absolute continuity of the Gaussian functional on its codimension-1 support subset (Propositions~21a/b); this paper does not handle parameter kinematics (see Theorem~6's no-kinematics statement), hence only requires and registers this condition within the \MPT\ pairing-fusion construction channel. This construction fact only provides the variational underlying premise, and \emph{by itself cannot complete the bi-measure disentanglement}.

From Corollary~10a we obtain a set of basic corollaries (reused throughout, numbered 2.23.1--2.23.6): \noindent\textbf{2.23.1--2.23.6 (six enumeration readings)} (see B.49).

\subsubsection{Formal problem and cross-categorical bearing}
\label{sec:2.4.3}

\noindent\textbf{Problem 2.25 (null-embedding ambiguity)}
Discrete sample sets are Lebesgue-null subsets; standard integrals erase their intrinsic topological differences, unable to distinguish dataset intrinsic topology from model extrinsic fitting error. \noindent\textbf{Problem 2.26 (iterative high-order error accumulation)} The pushforward of non-bijection morphisms continuously generates $\mu_{\mathrm{high}}$; no-collapse only eliminates inter-block intersection fusion, not the high-order term induced by in-block many-to-one folding. \noindent\textbf{Problem 2.27 (training--generation pairing bias)} Classical distribution metrics only capture volume differences, unable to characterize instance-level pairing error $\mu_{\mathrm{err}}$ nor its transformation into $\mu_{\mathrm{high}}$. Based on the gaps between the three categories, we propose the \emph{cross-categorical space lift--project--descend paradigm}: embed the base-space objects with categorical conflicts into a higher-dimensional product-measure space to dissolve the constraint, then project back to the original domain; the constraint-release effect depends entirely on the properties of the invertible \MPT. This construction does not negate Corollary~10a, but merely migrates the computation that originally triggered mutual exclusion in the base space to a larger product joint-measure space; absent the invertible-\MPT\ premise, the system degenerates to the base-space preimage-mutual-exclusion behavior.

\noindent\textbf{Assumption 1 (the unique minimal assumption of the whole paper; registered conditions at the realization layer in \S 6.1; layered statement, B.3/B.4)}\label{ass:1}
This assumption requires only two clauses, no superfluous one: the measure $\mu$ on the configuration / quotient space \textbf{(a) atomless}---contains no atom; any single point and any refinement column tend to infinitesimal topological volume per cell; \textbf{(b) genuine partition (substantive partition)}---the chosen partition predicate makes both sides of the partition non-null (the binary partition is genuine, not degenerating to one side empty).

Under this assumption, infinite refinement (infinite partition) makes every cell's topological volume take no non-infinitesimal value, all collapsing into differential-quotient equivalence classes (infinitesimal topological volume). ($\varepsilon$-$\delta$ is a standard analytic tool, not our assumption; the engineering realization of Eq.~\eqref{eq:12b} additionally uses the registered condition of \S 6.1(c).) \emph{Orthogonal decomposition is a non-assumption solution}: Euclidean dimension-lifting / functional orthogonal decomposition is a constructive solution for realizing infinite partition (acting on $\mathcal H^\Setcat$, existence given by construction), not an assumption; the $\varepsilon$-$\delta$ language also belongs to the solution side. Our only structural assumption remains Assumption~1 (atomless $\land$ genuine partition $\Rightarrow$ every cell tends to differential-quotient equivalence class). Details (see C.13).

\emph{(regime statement of atomlessness)} Atomlessness holds in the \emph{generically non-degenerate regime}---i.e.\ gradient descent recovers to the Euclidean saddle point, leaving the $\Hinf$ zero-value saddle point $\SadPt^{\mathrm{Hil}}$ (Theorem~6) degenerate domain; at the zero-value saddle point this assumption degenerates (corresponding to the null / limit scope of \S 6.3.2's formal-derivation box: $2/3$ is the coefficient under a.e.\ or limit sense, not a finite exact constant). All $\varepsilon$-$\delta$ constructive arguments (solution side) are carried out on $\mathcal H^\Setcat$. The predicate tower's Hilbertization on the $\Quot/\MorTop$ side takes predicate words and quotient blocks as objects; that side's arguments are uniformly expressed in layer comparison and block permutation.

From this assumption four formal outputs are derived (item-by-item attribution in B.4): among them \emph{Postulate~1} is \emph{quotient-translation-invariance measure preservation with non-null reading} (the measure-preservation equality is given by the structural layer of Lemma~1(i)/Proposition~8, not a new assumption; Assumption~1 only contributes genuine-partition both-sides non-null and cross-layer-to-first-order reading, not derived from metric structure); \emph{Postulates~2,~4} are the quantified consequences of structural characterization under Assumption~1; \emph{Postulate~3 is not a consequence of Assumption~1, but the statement that the standard $\varepsilon$-$\delta$ limit / differential analysis framework is well defined on $\mathcal H^\Setcat$---it is the analytic tool \emph{must be used} when deriving Postulates~1,~2,~4 (not our assumption, nor our conclusion)}:

(\emph{Postulate~1})\qquad \begin{equation}\begin{gathered}
\sigma_\#\nu_{\mathrm{idx},n}=\nu_{\mathrm{idx},n}\neq 0\\
\nu_{\mathrm{idx},n}(\mathrm dx)\equiv\nu_{\mathrm{idx},m}(\mathrm dx)\ (\mathrm{mod}\ \mu_{\mathrm{err}})\\
n\le m
\end{gathered}\end{equation} (\emph{quotient-translation-invariance measure preservation with non-null reading}: the normalized counting measure $\nu_{\mathrm{idx},n}$ (Definition~17) has invariant reading under quotient-block permutation and layer comparison (Proposition~8's quotient-translation-invariance measure preservation), and because Assumption~1(b) genuine partition makes both sides non-null, each cell's reading is constantly non-zero; its cross-layer compatibility to first order is the \emph{same-origin normalization baseline} of discrete (block-counting) and continuous (Lebesgue) readings, i.e.\ the formalization of \S 2.1's quotient-translation-invariance reading; where the measure-preservation equality is given by the structural layer of Lemma~1(i)/Proposition~8, Assumption~1(b) only contributes each cell $\neq0$, and the cross-layer-to-first-order reading is given by Assumption~1 (the $\varepsilon$-$\delta$ analytic tool)). That is, the \emph{cross-layer reading of the normalized counting measure}: the value where a same differential lies is invariant across different partition finenesses, to first order (see \S 6.2.2, discrete--continuous reading baseline).

(\emph{Postulate~2})\qquad \begin{equation}
\mathrm dx\odot\nu_\mathrm{idx}\ \sim_1\ \mathrm d\#
\end{equation}, i.e.\ \emph{first-order equality} (Corollary~2, \S 2.2.2's conclusion itself): the differential value and any subset measure reading are equivalent infinitesimals (both first-order quantities).

(\emph{Postulate~3})\qquad \begin{equation}\begin{gathered}
x_n\to x\iff\big(\forall\epsilon>0\ \exists N\in\mathbb N:\ n\ge N\Rightarrow\lVert x_n-x\rVert<\epsilon\big)\\
\mathcal H^\Setcat\ \text{complete under}\ \lVert\cdot\rVert
\end{gathered}\end{equation}, i.e.\ \emph{real-number reading and limit argument well defined}: continuity, completeness, and point-wise neighborhoods are expressed in $\varepsilon$-$\delta$ manner; this is precisely the well-definedness of the standard $\varepsilon$-$\delta$ real analysis framework on $\mathcal H^\Setcat$, the \emph{analytic tool must be used} when deriving Postulates~1,~2,~4 (not our assumption, nor our conclusion).

(\emph{Postulate~4})\qquad \begin{equation}
\widehat{\gamma_\Sigma}(\lambda)=\int e^{i\langle\lambda,x\rangle}\,\mathrm d\gamma_\Sigma(x)=e^{-\frac12\langle\Sigma\lambda,\lambda\rangle}
\end{equation} ($K_\Sigma\in\mathcal K_\Sigma$ induces Gaussian measure $\gamma_\Sigma$ on $\mathcal H^\Setcat$ and is uniquely determined by its characteristic functional; see Proposition~21b; $\Sigma$ trace-class; general case only cylindrical measure, boundary given), i.e.\ \emph{Gaussian functional integral well defined} (see \S 5.3, Propositions~21a/21b): $\sigma$-additive iff covariance is trace-class, general case only cylindrical measure.

\emph{Output vs.\ Principle~1}: Postulates~1--4 are the four formal outputs of Assumption~1, and Principles~1--4 (\S 2.4.1) are the four clauses of Principle~1; the two are connected by a chain: Postulate~1, Postulate~2 + Lemma~1(i) + Definition~24 $\to$ Principle~1, Principle~2 (B.3 row~\ding{172}); Postulate~3 is the well-definedness premise of Principle~1/Principle~2's real-number reading and limit argument; Postulate~4 via Principle~2 + Proposition~21b gives the Gaussian cylindrical-kernel compatibility-family boundary (table row~\ding{176}).

The geometric content of Assumption~1 is: under atomless + genuine partition, Euclidean dimension-lifting iteration (constructive-side construction of \S 3.2) makes the distinguishable block-volume difference on finite truncation shrink by dimension ratio, all falling into $\mu_{\mathrm{err}}$; the item-by-item list of ``needs / does not need this assumption'' (see B.4). This assumption cannot be eliminated by our construction (predicate tower $\to$ Theorem~3b $\to$ Lemma~5g is the consequence after the assumption holds, belonging to solution-side construction, not the basis of its satisfaction); its necessity lies in $\varepsilon$-$\delta$ being key to enumerating morphisms (differential equivalence classes): without $\varepsilon$-$\delta$ one cannot obtain the discrete--continuous transform---first affecting integration (definable integral does not mean finitely computable), and uncomputable integral cannot measure the compactness induced by integration. \emph{(the $\ln$ factor as direct embodiment of Assumption~1)} The natural logarithm $\ln$ in the second term $(1-\ln\mathfrak A_\varsigma)$ of Eq.~\eqref{eq:12b} is the \emph{direct illustration of integration}: the reverse integral $x(1-\ln x)$ of surprisal $-\ln x$ is the explicit integral of the continuous Gaussian distribution reading, and this explicit integral is precisely the embodiment of ``first-order error on the continuous Gaussian can be read out by integration under Assumption~1''---hence the $\ln$ conduction chain from Assumption~1 to Eq.~\eqref{eq:12b} is directly borne by integration, closing ring by ring, and the engineering-layer first-order error source is exactly this.

\section{Binary Quotient-Partition Bijection Induces Invertible \MPT: Shared Measure-Space Volume and Replica-Index Measure Preimage}
\label{sec:3}

This chapter assembles $\Quot$ primitive objects via $\MorTop$ lifting to $\Setcat$: relying on Theorem~1, it constructs the invertible \MPT\ and the shared measure space, dissolving the base-space preimage mutual exclusion of Corollary~10a; the replica-index measure is uniquely determined by the block-bijection-group automorphism invariance (Lemma~1(i), Definition~17). The topological existence of a quotient-layer bijection does not automatically give a closed-form analytic realization, and first-order error objectively exists (B.51).

\subsection{Bijection induces non-null-measure coverage: invertible \MPT}
\label{sec:3.1}

\noindent\textbf{Definition 10 (measure-preserving transformation)}\label{def:10}
A map $T:X\to X$ is a \emph{measure-preserving transformation} under measure $\mu$ if for every $A\in\mathcal F$, $\mu(T^{-1}(A))=\mu(A)$. At the $\Quot/\MorTop$ layers, if the measure is induced by the partition / bijection structure, then any block bijection automatically preserves measure (Lemma~1(i), Proposition~9a); at the $\Setcat$ layer when taking the geometric measure, measure preservation must be explicitly assumed (Lemma~1(ii), Proposition~9b; standard background \citep{giry1982categorical}).

\noindent\textbf{Definition 11 (invertible measurable map)}
$T$ is a measurable bijection and its \emph{inverse map is also measurable}. \noindent\textbf{Definition 12 (invertible measure-preserving transformation, invertible \MPT)}\label{def:12}
At the $\Setcat$ layer, $T$ is an invertible measure-preserving transformation iff $T$ satisfies Definition~11 and is measure-preserving; at the $\Quot/\MorTop$ layers, ``invertible'' means the \emph{block-layer inverse pairing} (Terminology~(f)(a)), and \emph{does not require a point-wise inverse map to exist} (see \S 2.1, base layers posit no point elements). The ideal invertible \MPT\ is a topological-theoretic object in the limit sense; real models can only realize the quotient-layer no-collapse bijection self-map $\mathcal T^\ast$, and generally do not satisfy point-wise measurable bijection.

The single-measure framework cannot separate the local contribution inside a set: we introduce a second $\sigma$-finite measure $\mu_2$ to construct the product measure $\mu_\otimes=\mu_1\times\mu_2$ ($\mu_2$ plays the role of a discriminant screening element). Constrained by Corollary~10a, a stochastic fiber bundle lacking topological wrapping amplifies the \emph{measure-pushforward subtraction paradox} ($\sigma_\#\mu-\tau_\#\mu$ is generally neither non-negative nor normalizable); the invertible \MPT\ provides a mapping pathway to mitigate this conflict (formal distinction and degeneracy boundary in B.52).

\noindent\textbf{Definition 13 (shared-measure derivation channel $\mathcal G_T$)}\label{def:13}
Let $T$ be an ideal invertible \MPT, and $\mu_\otimes=\mu_1\otimes\mu_2$ the joint measure of the product space. The shared measure $\mu_{\mathrm{share}}$ is \emph{derived} from this joint measure via the \MPT\ pairing channel (not separately defined):
\begin{equation}
\mu_{\mathrm{share}}\triangleq(p_T)_\#\mu_\otimes=\mathcal G_T(\mu_1,\mu_2)
\end{equation}
Where the \emph{pairing channel $p_T$ is explicitly defined as the first-coordinate projection} $p_T:X\times X\to X,\ p_T(x,y)\triangleq x$ (measurable surjection; the discriminant-screening information on the $\mu_2$ side is merged via the ideal invertible \MPT\ $T$---under ideal invertibility $\mu_2=T_\#\mu_1$, so the shared reading of $x$ simultaneously carries the $\mu_2$-side information; at the quotient-block layer read as $p_T(B_i,B_j)=B_i$, and the $\mu_2$-side $B_j$ pairs with $B_i$ via $T$'s block action).

Multiple measure readings are allowed on the same object; this channel only derives the shared reading, without deleting the respective readings of $\mu_1,\mu_2$; its existence is a structural conclusion under the fiber-bundle pushout--pullback (the ideal invertible \MPT\ necessarily exists, not meaning point-wise analytic realizability). On the finite quotient-block algebra $\mathcal A_{\mathcal P}$, the restriction of $\mu_{\mathrm{share}}$ satisfies finite additivity over mutually exclusive quotient blocks: \begin{equation}
\mu_{\mathrm{share}}\big(\bigsqcup_{b=1}^r C_b\big)=\sum_{b=1}^r\mu_{\mathrm{share}}(C_b)
\end{equation}. Subsequent losses, if using additive volume, adopt this derived measure and its quotient-block-algebra restriction (see B.52). A complete cross-measure mapping pathway is established iff $T$ simultaneously satisfies the invertibility and double-measure-preservation conditions at the corresponding layer; at the $\Setcat$ layer a measurable bijection does not automatically preserve measure (Lemma~1(ii)), while at the $\MorTop$ layer block-bijection measure preservation under uniform block-counting reading is given by Lemma~1(i).

\noindent\textbf{Proposition 10 (layered characterization of absolute continuity)} (see B.53). \noindent\textbf{Proposition 11 (diagonal product measure invariance under invertible \MPT)}\label{prop:11}
Let $T$ be an ideal invertible \MPT. Then $\mu_\otimes(T(A)\times T(A))=\mu_\otimes(A\times A)$. This equality is the minimal expression, under the joint measure, of multiplicity-weight conservation (the number of times an element is counted in an integral does not change due to $T$), and the minimal instance of shared-measure-space volume fidelity; Proposition~12a--12C (B.55) successively generalizes it to the three layers of global product invariance, differential-element invariance, and diagonal-subset invariance, and Proposition~12(2)'s quotient-space volume representability also rests on this proposition. \emph{Proof} (see B.55). This equality expresses the joint-measure invariance of the diagonal set; it does not imply $T(A)=A$, nor does it mean point-wise pairing is entirely correct; volume fidelity is a derived consequence of multiplicity-weight conservation, and can coexist with the first-order error relative to $\mathcal T^\ast$. Details (see E.4).

\noindent\textbf{Proposition 12 (single-layer joint integral and single-measure barrier)} (see B.56).

\subsection{Unique dual measure under weight-volume transform: Fourier-induced replica-index measure}
\label{sec:3.2}

This section derives, from the measure integral of limit summation, a non-trivial map (the Fourier operator) satisfying the Proposition~9b product-invariance constraint, and constructs a Hilbert space within the binary quotient-partition topological category; in this framework Euclidean dimension-lifting is closed by Theorem~3 to the standard Euclidean embedding (modulo $O(n{+}1)$; the three basic premises and their division of labor with Assumption~1 are in B.57).

\noindent\textbf{Propositions 12a--12C (three-term generalization of product invariance)}
The three-term generalization of product-measure invariance (Proposition~9b)---product-measure invariance, differential-product invariance, and measurable-subset-volume invariance---together with its proposition chain, are stated and proved in (B.55); among them Proposition~12c is cited by the proof of \S 6.2.2 and Proposition~12 (B.58).

Under the dual-measure framework, the base-space geometric volume is characterized by the Lebesgue measure $\lambda$ derived from product invariance; relying on the completeness of the product measurable space, there exists a dual measure $\nu$ carrying the frequency-side information (the joint Borel measure gives $\mu,\nu$ via projection marginals).

\noindent\textbf{Theorem 2 (uniqueness of Euclidean-normalized Fourier correspondence)}\label{thm:2}
Take $G=\mathbb R^d$, fix the self-dual identification $\hat G\cong\mathbb R^d$, the character $\chi_\xi(x)=e^{2\pi i\xi\cdot x}$, and the frequency normalization. Then the convolution--pointwise-multiplication correspondence given by $\chi_\xi$ and the $L^1$--$L^2$ Fourier--Plancherel formula is uniquely determined.

Changing the character pairing or frequency scale yields another normalization, outside the uniqueness domain of this theorem; this theorem is not generalized to $G\to\hat G$ module-category automorphism uniqueness on general locally compact abelian (LCA) groups, nor does it specify the replica-index measure. Standard theorem adaptation and proof (see B.59, C.2.4). Fourier uniqueness gives the conceptual prototype of the frequency-counting measure, but its realization is borne independently by the \emph{replica-index measure} $\nu_{\mathrm{idx}}$ (Lemma~1(i); Definition~17); relying on Proposition~9b and Fubini iteration, product invariance generalizes to finite-group iterated products, yielding a finite-dimensional truncation sequence $\{G^{(n)}\}$ (standard exposition of Fourier analysis and Pontryagin duality, \citep{rudin1962fourier}).

\noindent\textbf{Theorem 3 (compatible linear dimension-raising: volume form, standardness, and limit)}\label{thm:3}
Let $G^{(n)}\cong\mathbb R^n$, and \emph{assume} each $\varphi_n:G^{(n)}\to G^{(n+1)}$ is a continuous injective group homomorphism compatible with the refinement tower; on finite-dimensional Euclidean groups this is equivalent to an injective linear embedding. Further assume the product invariance of Proposition~9b and dynamical normalization hold. (a) The $n$-dimensional Haar volume factor is $\sqrt{\det(D\varphi_n^{\mathsf T}D\varphi_n)}$, and there exists an Euclidean equipment compatible along the tower. (b) The standard Euclidean embedding is unique modulo $O(n{+}1)$; this is the standardness premise of Assumption~1, not a conclusion of product invariance. Let $\mathcal H_\infty$ be the \emph{completion} of this compatible direct limit under the inner-product norm; the direct limit itself is generally only a pre-Hilbert space. The infinite-dimensional layer has no non-trivial translation-invariant Lebesgue-type measure, and translation-invariance arguments are limited to finite truncations. Proof (see B.61).
\noindent\textbf{Proposition 13 (pairwise product invariance of multiple groups)} (see B.62).

\subsection{Invertible \MPT\ induces double nested fiber-bundle integral paradigm}
\label{sec:3.3}

Under the dual-measure joint-product framework we examine the \MPT\ and its product-induced map $\boldsymbol T_\otimes$: the two-dimensional product domain contains two independent differential directions---the original-space geometric differential and the differential-counting direction characterized by the replica-index measure---whose pairing constitutes the joint measure, serving as the underlying input of the fiber-bundle integral; this paradigm is not bound to a specific measure. The replica-index measure is simultaneously the measure basis for the price paid by collecting transforms into a bijection (see \S 1): the support morphism is thereby realized as a fiber-bundle-sense \MPT\ under another measure (e.g.\ Lebesgue), and the shared differential structure makes the definition of element points no longer well defined.

\noindent\textbf{Definition 14 (built-in fiber-bundle pullback operator $\overleftarrow T$)}\label{def:14}
Let $(X\times X,\mathcal F\otimes\mathcal F,\mu_\otimes)$, $\mu_\otimes=\mu_1\times\mu_2$; notation $\overleftarrow\mu_\otimes\triangleq\overleftarrow T\mu_\otimes$, and in the discrete-atom case denote the pulled-back atomic counting replica-index by $\overleftarrow w_{ij}$. \emph{Ideal invertible \MPT\ special case and quotient-partition relaxation}: if $T$ is an ideal invertible \MPT, then $\overleftarrow T\mu_\otimes=\boldsymbol T^{-1}_{\otimes\#}\mu_\otimes$; general model maps only guarantee the quotient-layer no-collapse bijection $\mathcal T^\ast$, requiring only matching the pairing constraint of the baseline $\mathcal T^\ast$ at the quotient-block layer. $\boldsymbol T^{-1}_{\otimes\#}$ is the standard measure preimage pullback (its existence premise is point-wise inverse-map existence), while $\overleftarrow T$ is a custom built-in pullback operator; the two coincide under ideal invertible \MPT, but generally differ under the quotient-partition category.

\noindent\textbf{Definition 15 (double nested fiber-bundle integral)}\label{def:15}
Let $X$ be equipped with two independent $\sigma$-finite measures: the original-space base measure $\mu_1$ and the replica-index measure $\mu_2$; let $T$ be a candidate map. For any $\mathcal F\otimes\mathcal F$-measurable function $f$:
\begin{equation}
\iint_{X\times X}f(x,T(x))\,\mathrm d_{\mu_\otimes}\ \overset{\overleftarrow T}{\longrightarrow}\ \iint_{X\times X}f(x,T(x))\,\mathrm d\overleftarrow\mu_\otimes
\end{equation}
The inner enumeration of the domain uses the original joint measure $\mu_\otimes$, keeping the replica-counting dimension differentiable; the measure actually used for integral evaluation is the $\overleftarrow\mu_\otimes$ transformed by the built-in operator. $\overset{\overleftarrow T}{\longrightarrow}$ is the built-in integral marker, an inherent component of the definition. This definition is the baseline form of the fiber-bundle integral in Chapters~3--4; Propositions~14--16 are its simplification family under bijective subdomains, ideal \MPT, and discrete-atom cases (see below in \S 3.3).

\noindent\textbf{Propositions 14--16 (measurable premises and integral-simplification family)} (see B.63), \noindent\textbf{Proposition 17 (finite discrete case: integral-equivalent tensor multiplication)} (see B.64).
\emph{Compatibility verification}: the counting replica-index (index measure) reading $w_{ij}$ of a discrete atom belongs to the product-space joint measure, not to the base-space $X$ measure, hence circumventing the base-space preimage mutual exclusion of Corollary~10a; only after projection back to the base space via $\mathcal G_T$ does one obtain $\mu_{\mathrm{share}}$. This treatment is guaranteed not self-contradictory by the following two compatibilities:
\begin{quote}\begin{enumerate}
\item \emph{Counting-replica-index-measure vs.\ base-space-measure compatibility}: the counting replica-index of a discrete atom only records the pairing replica-index, does not enter the base-space $X$ geometric measure, and does not conflict with the base-space preimage mutual exclusion of Corollary~10a; after projection back to the base space via $\mathcal G_T$, the index measure and $\mu_{\mathrm{share}}$ fall in the same measurable space, and the two are compatible.
\item \emph{Discrete-atom vs.\ continuous-binary compatibility}: the discrete case of ideal invertible \MPT\ corresponds to tensor-index permutation, and finite-dimensional truncation contraction only guarantees quotient-block coarse-granularity invariance; this is consistent with the dichotomy of ``\MPT\ topological existence without guaranteed analytic reachability''---the index permutation of discrete atoms is precisely the counting-replica-index reading of the continuous dichotomy on finite truncation, and the two are compatible under the same index-measure scope.
\end{enumerate}\end{quote}

\section{Construct First-Order Error: Quotient-Partition Embedded in Triple Fiber-Bundle Integral}
\label{sec:4}

This chapter constructs a triple nested fiber-bundle integral paradigm adapted to the EDE three-stage, completing the three-category transition via inner-layer measure descent; the existence of zero-residue points and the uniqueness of counting measure are given by Theorem~6. Dedicated symbols in A.4.

\subsection{Triple fiber-bundle integral definition}
\label{sec:4.1}

The triple fiber-bundle integral is the integral carrier of the price paid for collecting transforms into a bijection: after the replica-index measure becomes measurable, non-null mass needs to be accounted layer by layer in the inner replica-index fiber, the middle joint-conservation fiber, and the outer pan-measure topological fiber (see \S 1, ``price: \MPT-ization and first-order error''; general construction of fiber-bundle integral, \citep{husemoller1993fibre}).

\noindent\textbf{Definition 16 (triple nested fiber-bundle integral)}\label{def:16}
The three layers of fibers are: the inner replica-index fiber $\Omega_\nu$ (carrying $\mathrm d\nu_{\mathrm{idx}}$, supporting measure descent), the middle joint-conservation fiber $\Omega_\lambda$ (carrying $\mathrm d\lambda$ and the descended copy $\overleftarrow{\mathrm d\nu}_{\mathrm{idx}}$, where dimension-lifting iteration occurs), and the outer pan-measure topological fiber $\Omega_\mu$ (assembled and lifted to $\Setcat$ via $L'$, carrying $\mathrm d\mu_{\mathrm{gen}}$).

\emph{Explicit definition of layer measures}: the inner measure is the replica-index measure $\nu_{\mathrm{idx}}$ (Definition~17); the \emph{middle measure} $\nu_{\mathrm{comp}}\triangleq\lambda\otimes\overleftarrow{\nu}_{\mathrm{idx}}$, where the descended copy is defined as
\begin{equation}
\overleftarrow{\nu}_{\mathrm{idx}}\triangleq\boldsymbol\pi_{\#}\nu_{\mathrm{idx}}
\end{equation}
(Definition~24); the \emph{outer measure} $\nu_{\mathrm{Leb,idx}}\triangleq\mu_{\mathrm{gen}}$ is the generative-side pan-measure (Theorem~5). The slots of the integrand $F(\nu; x, T(x), t)$ correspond one-to-one to the three layers: $\nu\in\Omega_\nu$ (replica index, explicitly entering the integrand), $x\in\Omega_\lambda$ ($T(x)$ is its image via \MPT), $t\in\Omega_\mu$ (outer topological parameter, i.e.\ the marker of pan-measure $\mu_{\mathrm{gen}}$). \emph{$T(x)$ is determined by $x$, not an independent variable}: the middle-layer integral variable should be read as $(x,\overleftarrow{\nu})$, and $T(x)$ is uniquely determined by $x$ via \MPT\ and then substituted, so the integrand is actually a four-slot, three-layer-variable structure.

The triple nested fiber-bundle integral is three-layer nested, each layer being an \emph{iterated integral} that transforms the evaluation measure via the built-in pullback operator:
\begin{equation}
\Lambda=\int_{\Omega_\mu}\Biggl[\ \int_{\Omega_\lambda}\Biggl(\ \int_{\Omega_\nu}F(\nu; x, T(x), t)\ \mathrm d\nu_{\mathrm{idx}}(\nu)\ \Biggr)\ \mathrm d\nu_{\mathrm{comp}}(\lambda,\overleftarrow{\nu})\ \Biggr]\ \mathrm d\nu_{\mathrm{Leb,idx}}(\mu)
\end{equation}
Each integration eliminates one fiber dimension (see Construction~1), hence it is \emph{not equivalent to an ordinary Cartesian triple integral}: between layers this formula is connected by pullback / pushforward operators, the evaluation order is uniquely fixed by the categorical direction, and it degenerates into a Fubini rearrangement only under the ideal invertible \MPT\ premise. This definition is the integral carrier for the first-order error computation of the whole paper.

\noindent\textbf{Construction 1's properties: inter-layer transformation rules} (see B.65).

\subsection{\MPT\ dimension-lifting: discrete-Lebesgue continuous topology composite bijection}
\label{sec:4.2}

\noindent\textbf{Definition 17 (replica-layer finite-partition measure)}\label{def:17}
Let $\mathcal P_n=\{B_1,\dots,B_{2^n}\}$ be the $n$-th layer binary partition. On the algebra generated by $\mathcal P_n$ define
\begin{equation}\begin{gathered}
\nu_{\mathrm{idx},n}\triangleq\frac1{2^n}\sum_{i=1}^{2^n}\delta_{B_i}\\
\nu_{\mathrm{idx},n}(A)=\frac1{2^n}\#\{i:B_i\subseteq A\},\quad A\in\sigma(\mathcal P_n)
\end{gathered}\end{equation}
Satisfying $\nu_{\mathrm{idx},n}(X)=1$; this is the explicit representative of the ``partition-induced block-counting measure'' in Lemma~1(i), and under block permutation the pushforward measure is invariant ($\sigma_\#\nu_{\mathrm{idx},n}=\nu_{\mathrm{idx},n}$, the mathematical model of Proposition~9a).

\emph{Anchoring of the limit environment (element-free partition category reading)}: each layer's $\nu_{\mathrm{idx},n}$ is defined on a different quotient space, and the weak limit must first fix a common environment: all layers are sent via refinement embedding into the Cantor-type environment $\mathcal C_\beta\triangleq\varprojlim_n\mathcal P_n$ (the projective limit of Theorem~5 is taken on this environment); $\{\nu_{\mathrm{idx},n}\}_{n\ge1}$ weakly converges on $\mathcal C_\beta$, and the limit $\nu_{\mathrm{idx},\infty}$ is the unique cylinder-compatible probability measure (Bernoulli $(1/2)$ product measure). $\mathcal C_\beta$ is a topological environment, not a $\Setcat$ point space, and does not presuppose point elements.

The infinite-dimensional base $\mathcal H_\infty$ is induced under double constraints: the replica-index can be taken infinitely; the backpropagation constraint---no cross-replica information interaction is allowed before loss scalarization completes---makes different replica-indices correspond to mutually independent dimensions.

\noindent\textbf{Theorem 4 ($\mathcal H_\infty$'s bijective volume-preservation and first-order-error closure)}\label{thm:4}
Let the factor spaces satisfy Lemma~2, Theorem~1, and the measurable measure-preserving-bijection property. Lifting the replica-index fiber and the sample geometric space by product to dimension $\mathcal H_\infty$, then (a) $\mathcal H_\infty$ inherits the global measurable volume-preserving bijection structure; (b) product dimension-lifting does not change the first-order measurable property of the error; (c) $\mathcal H_\infty$ internally does not endogenously generate high-order error---its mechanism is \emph{partition-predicate anchoring}: in the dimension-lifting iteration the predicate word of the $\betat$ tower is statically fixed, and dimension-lifting only amplifies the carrier space without changing the predicate structure (see \S 6.4.1 boundary-statement reading identity). \emph{Qualification}: ``volume preservation'' means fiber-bundle volume conservation along the $\boldsymbol\pi_\#/\boldsymbol\pi^\#$ channel, \emph{not} the existence of a global translation-invariant (Lebesgue-type) measure on $\mathcal H_\infty$---the latter does not exist (see \S 3.2). \emph{Proof (sketch)} (see B.66)--B.67.

\noindent\textbf{Theorem 5 (replica continuous-measure mapping compatibility)}\label{thm:5}
The measure readings of the three-layer fibers and the integral variables finally fall on $\mathbb R$ (i.e.\ the reading / range after lifting-assembly to the $\Setcat$ layer; the inner layer still does not derive point elements at the $\Quot/\MorTop$ primitive layer); relying on non-infinite integral-boundary partition bijections / non-bijection sub-regions; all measure values fall in $\mathbb R$. Under global normalization, in the replica-infinite-partition limit, the \emph{substantive content} of the compatible reading between the replica-layer measure system and the pan-measure $\mu_{\mathrm{gen}}$ is: the projective limit of $\{\nu_{\mathrm{idx},n}\}$ and the Helly weak limit give the \emph{same limit} $\nu_{\mathrm{idx},\infty}$ on the Cantor-type environment $\mathcal C_\beta$ (the two routes agree); the same $[0,1]$ integral range and unit total mass are only concomitant properties, not where the compatible information amount lies (any two probability measures share range and total mass). This compatibility is established at the $\MorTop$ layer; this theorem does \emph{not} claim an order-preserving isomorphism of the integral range; the order structure is preserved only under quotient-layer readings. \emph{Proof (sketch)} (see B.68)--B.69.

\noindent\textbf{Definition 18 (dimension-raising integral)} (see B.70).

Relying on the Euclidean volume integral form, we express, via the functional of unknown measures, the volume of non-null-measure binary quotient partitions; multiple integrals eliminate a certain dimension to obtain that dimension's marginal-measure reading, and each layer's \MPT\ measure is mutually the marginal measure (Theorem~3, \S 3.2 dimension-lifting-map uniqueness).

\noindent\textbf{Proposition 18 (inner compactness of binary dimension-raising quotient partition)} (see B.71).

\subsection{Zero-residue-point analytic uniqueness induces random enumeration dynamics paradigm}
\label{sec:4.3}

This section carries out the set-theoretic existence deduction of the zero-residue-point set, \emph{without asserting} that the training dynamics orbit actually reaches this set: the zero-residue set $\SadPt=\{\theta:\mathcal L(\theta)=0\}$ provides a topological numerical anchor, and its existence and counting-measure uniqueness are given by Theorem~6---the uniqueness is divided into two readings: \emph{point-level} ((e)--(f), carrying Premise~1) and \emph{differential-level} ((i), the differential where zero lies is unique, Definition~19a), which mutually recognize each other via Definition~19a.

\paragraph{Realization conditions and posterior structure (theorem inputs, not to be conflated with the backbone mathematical assumption)}
\textbf{Convention 1 (architecture and zero-output calibration)}: samples normalized to $[0,1]$; the network has no bias, no residual skip-connection injecting non-zero constants, BatchNorm translation parameters set to zero while keeping scale parameters; the zero output must be precisely paired with the target predicate; \emph{block-level mapping}---EDE predicate pairing is executed at the block layer, and $T$ maps each source quotient block into a unique target quotient block; straddling quotient blocks (partially falling on the partition boundary) are handled by the null-measure argument of \S 6.3.3: their occurrence domain is only the partition boundary, and the Gaussian functional's (not degenerating to one-hot) partition boundary is null measure (B.104), $\mu_{\mathrm{err}}$ only accumulates non-null-support volume, hence the straddling-quotient-block family's volume contribution is constantly $0$, and no separate ``undefined'' exclusion is set.
\noindent\textbf{Convention 2 (non-negative per-sample loss)}: $\mathcal L=\sum_i r_i$, $r_i\ge0$, forbidding in-batch positive-negative cancellation.
\noindent\textbf{Convention 3 (measurable and local-gradient condition)}: $\Theta\subset\mathbb R^P$ is equipped with $\mathcal B(\Theta)$; $\mathcal L$, the forward map, and the gradient map are Borel measurable; where the theorem uses gradients, per-sample gradients are required to exist.
\noindent\textbf{Convention 4 (measure descent and finite evaluation)}: the measure-descent target is the measure space; finite-truncation integral domain and normalization factors guarantee the loss is bounded (only constraining integral evaluation, not asserting all criterion sets of the whole paper are bounded).

\noindent\textbf{Premise 1 (per-sample non-degeneracy; dedicated premise for point-level counting uniqueness)}: the definitional version is $E\subseteq[\boldsymbol0]$ ($E\triangleq\{\theta:N(\theta)=|\mathcal X|\}$, $[\boldsymbol0]$ being the parameter-degeneracy equivalence class inducing the same input-output function); the gradient reading is: if $\theta\notin[\boldsymbol0]$ and $\mathcal L(\theta)>0$, then at least one sample $x$ satisfies $\nabla_\theta\ell(\theta;x)\ne\boldsymbol0$. This classification is classification in the predicate-matching sense (Principle~3), and cannot be automatically deduced from Conventions~1--4. \emph{Premise~1's domain is the point-level counting reading}: the differential-reading uniqueness (Theorem~6(i)) does not presuppose Premise~1, and the two must not overstep each other (domain analysis in B.73).

\noindent\textbf{Theorem 6 (zero-residue-point existence, conditional counting uniqueness, and differential-reading uniqueness)}\label{thm:6}
Under Conventions~1--4, let $\SadPt=\{\theta:\mathcal L(\theta)=0\}$. Then: (a) $\boldsymbol0\in\SadPt$ and $\SadPt$ is Borel measurable; (b) $\boldsymbol0$ is a global minimum, with zero gradient wherever the gradient exists; (c) non-zero-residue-point loss is strictly positive; (d) the zero-parameter point has identically zero per-sample gradient; (e) \emph{adding Premise~1}, $N(\theta)=\#\{x:\nabla_\theta\ell(\theta;x)=0\}$ reaches $|\mathcal X|$ only on $[\boldsymbol0]$; (f) under Premise~1, all other per-sample stationary points belong to this equivalence class; (g) $\SadPt$ is the EDE predicate-precise-pairing set, not the functional-critical-point set; (h) under Premise~1, maximal-counting loss, and the \S 6.3.2 EDE matching mechanism, the zero-value saddle point is the unique stayable saddle point under \emph{infinite epoch $\leftrightarrow$ infinite predicate parameters}: when predicate parameters are infinite, Euclidean dimension-lifting exactly reaches the Hilbert layer, and the zero-residue equivalence class becomes the unique predicate-word static anchor; under \emph{finite epoch} there exist only stayable finite-predicate networks and finite partitions (finite predicate parameters, finite quotient blocks), whose stayability is borne by the finite partition, not requiring nor implying the Hilbert-layer unique predicate-word static saddle point; (i) \emph{differential-reading uniqueness (not presupposing Premise~1)}: the replica-index counting reading of the differential $\mathrm dx_0$ where zero lies is uniquely $|\mathcal X|$ (the 0-differential quotient), point-level degeneracy ambiguity does not enter, and it mutually recognizes (e)--(f) via Definition~19a; complete construction and proof (see B.73).
\emph{Proof} (see B.3).

This paper does not do the kinematics of parameter-descent reaching (does not promise SGD-reachability): the zero network is posterior, and the prior kinematics is not promised by this paper; the unique stayability of the zero-value saddle point is the uniqueness of the predicate-word static structure, not kinematic reachability. The negation of reachability does not weaken this construction (non-analytically, a high-probability first-order-error-reducing descent direction is still constructed); the zero network is not in the zero-output-calibrated loss class of Convention~1, hence is not the experimental loss optimum. Details (see E.5).
The ``zero saddle point'' in this paper means the zero-residue precise-pairing point; when involving the classical saddle-point concept we call it the zero-residue point.

\begin{figure}[h]
\centering
\includegraphics[width=211pt]{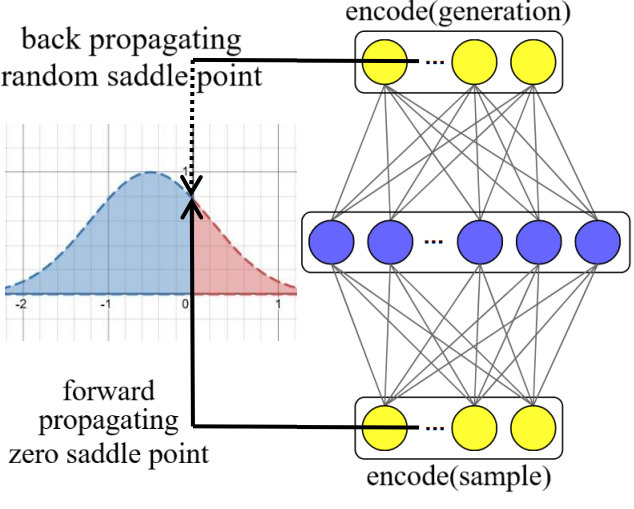}
\caption{Forward-propagated zero-value saddle point and backward-propagated random saddle point align at the zero-value saddle point.}
\label{fig:1}
\end{figure}

The propagation reading of zero-value-saddle-point uniqueness (see Figure~\ref{fig:1}): the forward-propagated zero-value saddle point and the back-propagated random saddle point align at the zero-value saddle point; among all candidate saddle points, only the zero-value saddle point does not mismatch through forward--backward bidirectional propagation---i.e.\ the image presentation of the structural feature ``same for all replica-indices'' in (h).

\noindent\textbf{Definition 19 (domain symmetric-difference offset)} (see B.74).
\noindent\textbf{Definition 19a (the differential where zero lies and equivalence-class transform)}\label{def:19a}
Let $\mathrm dx_0$ be the \emph{differential where zero lies}: the zero value (the single-point reading of the zero data-flow output given by Convention~1 and Theorem~6(a)(d)) falls, via the parameter-degeneracy equivalence class $[\boldsymbol0]$ transform, into the differential-topological element. Layer by layer: $\mathrm dx_{0,n}=B_{0^n}$ (half-open block convention, no boundary ambiguity), and $\mathrm dx_0$ is its $\betat$-tower projective limit, i.e.\ the minimal quotient block carrying the zero reading (Definition~1a, differential quotient). The transform $[\boldsymbol0]\mapsto\mathrm dx_0$ gathers the point-layer many-to-one into the single differential-quotient of the differential layer; each dimension's differential is infinite in $\Setcat$ (Lemma~4), the point-level degeneracy configuration does not change this uniqueness (Theorem~6(i)), and the maximal counting reading of $\mathrm dx_0$ is $|\mathcal X|$.

\noindent\textbf{Definition 20 (three-layer saddle-point-gap lacuna set)} (see B.75). \noindent\textbf{Corollary 11 (sample-functional zero-first-order-error set anchor)} (see B.76).

After the zero-residue point gives the topological numerical anchor (Theorem~6), the set-theoretically-legal translational enumeration is constrained by two boundaries ($\mathcal H_\infty$ has no global translation-invariant measure; assignment connotation is not a functional), and the variational descent must be borne by \emph{random enumeration}: taking the maximal replica-index measure as the variational target, using the prior-posterior dual attribute of the $\mathrm{epoch}+1$ counting measure, the unreachability of translational enumeration is transformed into the computability of random enumeration. High-order error is \emph{not} endogenous in the pure dimension-lifting layer of $\mathcal H_\infty$ (Theorem~4); its structural deviation can only originate from two types: categorical leakage and topological collapse (the binary quotient-partition category is broken, degenerating into multi-partition, with minimal cross-partition quotient blocks); both are exogenous operations, not endogenous to the Gaussian-\MPT\ functional family (Definition~6), and categorical leakage has been excluded by Proposition~19; \emph{in-block many-to-one folding is not in this category} (see \S 2.3.1, its micro-level error objectively exists and is reading-visible, but is not within this paper's first-order-error reading target, and the engineering side \emph{actively neglects} it by unified declaration), hence no $\mu_{\mathrm{high}}$ is generated.

\section{Topological--Analytic Reachability Gap and Gaussian Transform-Kernel Uniqueness}
\label{sec:5}

This chapter proceeds forward from $\Quot$: it excludes categorical leakage (see Proposition~19), opens the numerical passage across the three categories via measure descent--recovery (see Proposition~20), and gives the unique collapse of the fiber-bundle Gaussian (see Proposition~21) and the zero-residue-point variation (see Proposition~22; Corollary~13). Dedicated symbols are in A.5.

\subsection{The gap between qualitative and quantitative, and freedom from categorical leakage: the computable--differentiable--integrable gap and the constructive exclusion of categorical leakage}
\label{sec:5.1}

\noindent\textbf{Definition 21 (categorical leakage)}\label{def:21}
The loss of binary-quotient-partition constraints during cross-categorical transmission: if the categorical support is not surjective, or if the lifting is performed via a non-quotient bijection, so that the $\Setcat$ layer exhibits derived instances lacking an underlying quotient-partition support (spurious numerical-category intrusion), then categorical leakage is said to occur; Proposition~19 excludes it item by item.

\noindent\textbf{Proposition 19 (the present framework suffers no categorical leakage)}\label{prop:19}
The categorical support of this paper is induced to be surjective by enumeration (search) (see Proposition~3(ii)), and the lifting is performed only via quotient bijections---the bijectivity of the quotient bijection (object-collection layer) and the induced support surjection hold simultaneously (see Proposition~4; Proposition~5; Definition~3). Hence within the present framework there is \emph{no categorical leakage}; the classical paradigm, which indirectly constructs support via stochastic reparameterization, destroys morphism invertibility (see Corollary~7), and is a typical source of categorical leakage. Consequently, in the originating topological loss (see the abstract decomposition of Eq.~\eqref{eq:12b}, \S 6.4.1), the $\Quot$-layer reading $\tfrac12$ (i.e.\ the $1{:}1$ $\Quot$ gauge) and the $\Setcat$-layer reading $\tfrac23$ are two \emph{coexisting reading gauges} (see \S 6.3.2) for whether the morphism is equipped / not equipped with a global counting measure, and their difference is not to be read as categorical leakage.

\noindent\textbf{Proof} (see B.78).

On the finite truncation $Z_m$, objects are computable, differentiable, and integrable; the $\Hinf$ layer guarantees only topological existence: a bijection can be found by enumerable search (see Proposition~3; Theorem~1), yet cannot be computed down to a functional (see Proposition~21b)---one can \emph{search out a bijection but cannot compute down to a functional}; this is precisely the gap. The enumerating posterior-supported prior converts it into a layered structure: computability and integrability fall on $Z_m$, topological existence falls on $\Hinf$, connected by measure descent--recovery, and the first-order error is exactly the intrinsic gap along this passage.

\subsection{Dissolving the gap: set-categorical assembly of functionals after measure descent realizes computable forward and back propagation}
\label{sec:5.2}

The first-order error at the volume level originates from the differential-volume equivalence classes across Euclidean--Hilbert (see Corollary~2); the numerical equivalence classes originate from the counting measure and the rank-1 minimal non-empty sets. \emph{Categorical equivalence cannot directly assign values to functionals}---the underlying category posits no point elements (see \S 2.1), and the measure must be assembled toward the partition category via descent--recovery.

The $\Quot$ predicate itself has no morphisms and is not computable; the measure reading depends entirely on measure descent--recovery, formalized as Proposition~20---it serves as the intermediate layer between $\Quot$ and $\Setcat$.

\noindent\textbf{Proposition 20}\label{prop:20}
Proposition~20 realizes the differential-topological ascent and descent at the level of numerical expressions among $\Quot$, $\MorTop$, and $\Setcat$; the three-layer categorical composition is preserved. The two groups of EDE output predicates alleviate the computable-but-not-differentiable constraint: a single $\Quot$ predicate itself has no computational ability, yet it is fitted twice by EDE, and the equality of the two outputs yields a computable decision condition. \emph{Cardinality exchange is unconstrained}: a topological domain may be arbitrarily split into a sum of multiple differentials, and multiple differentials may be merged into the same topological domain; a topological class (differential quotient) may correspond to an equivalence class of arbitrarily many element points (B.7.7). \emph{Exchange occurs at the middle layer, not at the outer layers}: it occurs at the middle-layer joint-conservation fiber $\Omega_\lambda$ (see Definition~16); when recovering the measure, a reverse-equivalence-class sign is added, yielding the \emph{topological volume (sum of differential volumes)}; the integration object of the outer layer $\Omega_\mu$ is the \emph{differential Lebesgue}, and the integration \emph{does not commute} the differential numerical value with the topology---the Lebesgue measure \emph{cannot be recovered alone}, carrying first-order error, while the replica-index counting measure \emph{can be recovered alone without error} (see the division of labor in \S 6.2.2). The well-definedness of the integral does not take $\varepsilon$-$\delta$ as a premise (finite quotient-partition layer; $\varepsilon$-$\delta$ bears only the responsibility of real-number limit argumentation on $\mathcal H^{\Setcat}$).

\noindent\textbf{Proof} (see B.79).

At this Hilbert level of \emph{infinite predicates}, the \emph{preimage mutual-exclusion constraint of the finite partition category (see Corollary~10a)} is further lifted in dimension, constructing the \emph{mutually-exclusive additive group $\Gexcl$} under layering as the skeleton of the cross-measure-morphism support:

\noindent\textbf{Corollary 10b (multi-measure preimage mutual-exclusion constraint; propagation of Corollary~10a(i), not an independent definition; used in this paper as an argument premise)}\label{cor:10b} (see B.81).

\noindent\textbf{Definition 22 (mutual-exclusion additive group $\Gexcl$)}\label{def:22} (see B.83); the three layers and their mutual-exclusion forms are summarized in Table~\ref{tab:5.1}:

\begin{longtable}{M{0.21\textwidth} M{0.21\textwidth} M{0.21\textwidth} M{0.21\textwidth}}
\caption{Three layers of the categorical chain and their mutual-exclusion forms.}\label{tab:5.1}\\
\hline
Layer & Predicate form & Mutual-exclusion form & Compositional restriction \\
\hline\hline
$\Quot$ partition category & quotient-predicate partition (unique categorical object) & finite preimage mutual-exclusion constraint (see Corollary~10a) & no point elements; Boolean decision only \\
\hline
$\MorTop$ mapping category & infinite predicate & layered mutually-exclusive additive group $\Gexcl$ & quotient-class bijection; measure via descent \\
\hline
$\Setcat$ set category & numerical functional & measure-space topological mutual exclusion & derived instances; dual measure \\
\hline
\end{longtable}
\vspace{2pt}
\begin{minipage}{0.95\textwidth}\footnotesize
\textit{Note.} $\Quot$ --- no point elements, no numerical value, Boolean decision only, no explicit morphism numerical expression.

$\MorTop$ --- no point elements, quotient-equivalence-class bijection present, measure assembled via descent.

$\Setcat$ --- derived instances present, Lebesgue/counting dual measure.
\end{minipage}

The characterization of joint measure descent on the primitive $\Quot$ binary-quotient-partition object gives the antecedent constraint of Proposition~21 (see \S 5.3): under the nested dual-measure constraint, even if the triple fiber-bundle integral collapses to the functional level, the output is a closed \emph{functional family} (composite binomial equivalence class) rather than an isolated single functional---the elements within the family obey a natural-base-power self-consistency condition, yielding a natural-base exponential (Gaussian) functional-family form with $L^2$ as the independent variable.

\subsection{Gaussian functional: the unique collapse from fiber bundle to bijection induces the set-categorical equipped measure}
\label{sec:5.3}

\begin{figure}[h]
\centering
\includegraphics[width=283pt]{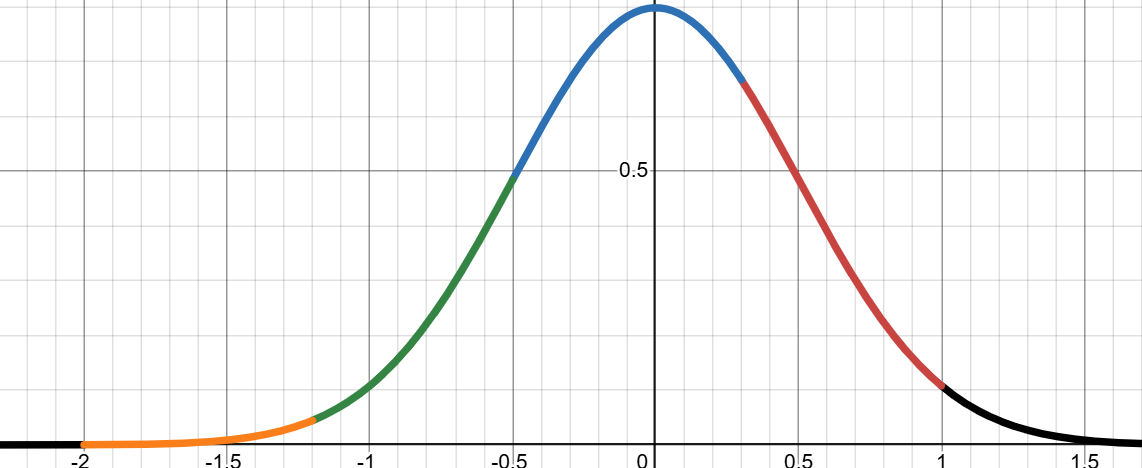}\\[2pt]
(a)\\[6pt]
\includegraphics[width=283pt]{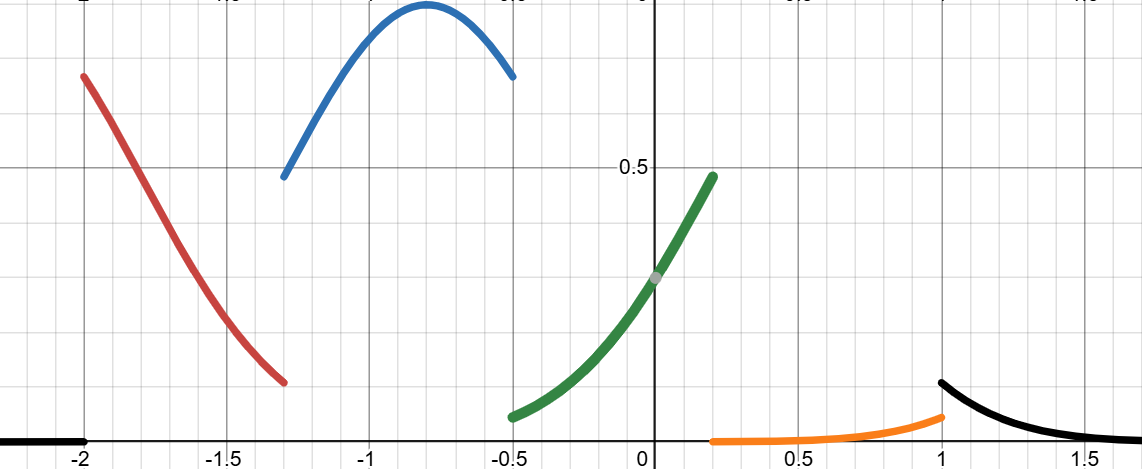}\\[2pt]
(b)
\caption{Gaussian functional and Gaussian behavior (fiber bundle): (a) Gaussian marginal distribution (colored by partition segment); (b) shuffled Gaussian---a permutation variant of the same marginal segment, topologically equivalent to (a) under the marginal-distribution reading.}
\label{fig:2}
\end{figure}

The core objects of this section act on the joint space $G\times\hat G$; the topological-predicate queue $\boldsymbol Q$ takes $\mathit S_0$ as the partition threshold; only after the replica-index fiber, the joint-conservation fiber, and the pan-measure fiber are all three complete can the assembly of the Gaussian functional family be discussed. The Fourier--Gaussian uniqueness is the manifestation of collection closure on the functional side (see \S 1): after the support uniquely collapses via bijection, the finite-output observable is determined by uniqueness (see Definition~28; \S 6.2.3 boundary constraint)---this is the uniqueness of the measure-structure layer, distinct from the problem of Gaussian-mixture identifiability in the parameter-estimation layer \citep{liu2021settling,bakshi2022robustly}. The fiber-bundle reading of the Gaussian functional and Gaussian behavior (see Figure~\ref{fig:2}).

\noindent\textbf{Definition 23 (topological-predicate queue $\boldsymbol Q$)}\label{def:23} (see B.84).

\noindent\textbf{Definition 24 (fiber-bundle pushforward $\boldsymbol\pi_\#$ and pullback $\boldsymbol\pi^\#$)}\label{def:24}
$\boldsymbol\pi_\#$ (\emph{measure descent}: the pushforward along the category chain downward, with the partition category $\Quot$ as target) and $\boldsymbol\pi^\#$ (\emph{measure recovery}, also called \emph{lifting recovery}: the pullback along the category chain upward, reading out numerical values and points, i.e.\ the back-projection operator $BP(\cdot)$) are formal set-theoretic operators that respectively encode forward and backward propagation, compensating for the absence of a reparameterization analytic inverse; the operators themselves introduce no additional set-theoretic axioms. At the $\MorTop$ layer, $\boldsymbol\pi_\#,\boldsymbol\pi^\#$ act on the quotient-block set $[Z_m]$; at the $\Hinf$ layer both remain well defined, but the Euclidean measure coordinates are erased (see the boundary conclusion of \S 3.2).

\noindent\textbf{Definition 27 (quotient self-map / quotient-misaligned fiber bundle)}\label{def:27} (see B.85).

\noindent\textbf{Axiom 1 (orthogonal-basis-coefficient uniqueness)}\label{ax:1} (see B.86).

The classical ``prior supports posterior'' paradigm, running inside $\Setcat$, relies on the standard axiom set (Borel structure, Kolmogorov probability axioms \citep{kolmogorov1933grundbegriffe}, the ``prior support pre-given'' presupposition); it contains no cross-categorical support, measure descent--recovery, or quotient-partition constraint.

\noindent\textbf{Definition 28 (fiber-bundle functional-bearing condition)}\label{def:28}
All observable outputs are given by the truncated set via the $\boldsymbol\pi_\#\circ\boldsymbol\pi^\#$ calculus: $(\boldsymbol\pi_\#\circ\boldsymbol\pi^\#)(S|_\Omega)$ is the unique observable output, and the original point-wise set object is not taken as an observable element. Definition~28 is an applicability-domain declaration that \emph{does not participate in the proof deduction and is not counted as a mathematical assumption}; it is not inside the classical axiom set $\mathbf A_0$, and Proposition~21 holds within this applicability domain.

\noindent\textbf{Lemma 5a, 5b--5d, 5e, 5f, 5g (auxiliary lemma family)}\label{lem:5}
The common tail segment is realizable under unbounded intersection (see Lemma~5a), auxiliary technical lemmas (Lemma~5b--5d), BatchNorm normalization yields doubly-bounded topologically non-zero regions (Lemma~5e), the zero-residue set is a global loss lower bound (Lemma~5f), the differential-sampling paradigm yields uniform enumeration of replica fibers (Lemma~5g); statements (see B.87--B.91).

Let the cross-space map $\mathcal F:G\to\hat G$ satisfy \emph{Premise Set~H}: (a) Theorem~2; (b) invertible \MPT, shared measure and product invariance; (c) finite-layer standard (isometric) Euclidean embedding and cylindrical-kernel construction, where this isometric embedding is a solution-side standardization convention (see the explicit assumption of Theorem~3), not belonging to Assumption~1; (d) Proposition~20; (e) the observable-output applicability domain of Definition~28; (f) the projection-compatible Gaussian cylindrical-kernel family induced by Lemma~4. The translation-invariant reading is limited to finite truncations; the attribution and complete statement of each premise (see B.92).

\noindent\textbf{Proposition 21a (finite-truncation Gaussian-kernel uniqueness)}\label{prop:21a}
Let the phase be a character, the envelope be continuous positive-definite, and satisfy any of the following unique-ification conditions: (iii-1) the envelope belongs to the extreme point of the cone of \emph{all finite-dimensional simultaneously radial positive-definite} functions; the Bessel-type extreme ray fixing a single-dimensional radial cone is not within this misuse scope. (iii-2) components are independent and the density is invariant under the full $O(m)$. Then the envelope is Gaussian; under standard coordinates it is isotropic, and in the positive-definite case $\Sigma\succ0$ is obtained by an invertible linear coordinate transformation; the singular positive-semidefinite case is handled separately only as a degenerate kernel, and the kernel family is
\begin{equation}
\mathcal K_\Sigma=\{K_\Sigma(k,x)=\exp(-\tfrac12x^\top\Sigma x+ik^\top x):\Sigma\succeq0\}
\end{equation}
This conclusion admits the full Gaussian rather than treating the truncated Gaussian as a model distribution. The standard-theorem adaptation of the two routes, width alignment, and proof (see B.92) and C.2.2; Proposition~21 (full statement of the fiber-bundle Gaussian and cylindrical-kernel compatibility-family boundary) is given together (see B.92).

\noindent\textbf{Proof} (see B.92).

\paragraph{5.5b (\Hinf layer)}
On $\Hinf$ there exists no globally translation-invariant measure, and the underlying posits no point elements (see \S 2.1; point-wise is an upper-layer-category reading), hence there is \emph{no point-wise Gaussian morphism}; here the Gaussian can only be given by the \emph{cylindrical-kernel compatibility family} $\{K^{(n)}\}_{n\ge1}$ of (f).

\noindent\textbf{Proof} (see B.93).

\noindent\textbf{Proposition 24 (family-internal closure and first-order-error accommodation)}\label{prop:24} (see B.94).

\noindent\textbf{Corollary 12 (topological--analytic-reachable dichotomy, connecting to generative models)}\label{cor:12} (see B.95).

\subsection{From qualitative first-order fiber-bundle integral to descended Gaussian measure characterizing inter-categorical support: zero-residue-point variation}
\label{sec:5.4}

The counting measure, relying on cross-categorical support and joint measure descent--recovery, reaches the Hilbert level, \emph{achieving zero-error arrival at the Hilbert zero-residue point under the infinite-predicate system}; the 0-differential quotient within $\Hinf$ (see Definition~19a) is equivalent to set-categorical assembly. \emph{The dependence of Hilbert on Assumption~1 is essential: the Hilbert space simultaneously carries the differential (induced by inner product / norm) and the limit (completeness), both of which take $\varepsilon$-$\delta$ well-definedness as a premise (see Postulate~3; the output of \S 2.4.3), hence Hilbert is the intensive usage region of Assumption~1; it anchors the quotient-block reading of the partition category (no point elements, differential equivalence classes) and the point-wise reading of the set category to the same Hilbert differential, and this anchoring is independent of the functional---the functional only assigns the value 0, while the anchoring itself holds for any value, and only the value 0 is analytically computable.}

\noindent\textbf{Essential construction} (see B.97). The multi-measure preimage mutual-exclusion constraint of Corollary~10b is the underlying premise of the reverse deduction; after \MPT\ disentanglement the counting measure is recoverable bidirectionally (can be recovered alone without error), while the Lebesgue side cannot be recovered alone, so the output naturally carries first-order error.

\noindent\textbf{Definition 26 (two kinds of pushforward--pullback loss)}\label{def:26}
Let $\boldsymbol\pi_\#$ be the fiber-bundle pushforward (measure descent), and $\boldsymbol\pi^\#$ the fiber-bundle pullback (measure recovery, i.e.\ $BP(\cdot)$).

\emph{MPT numerical pushforward--pullback loss} $\mathcal L_\mathrm{num}$ ($\Setcat$ layer, acting only on already-lifted-and-assembled point objects):
\begin{equation}
\mathcal L_\mathrm{num}\triangleq d_{\mathcal K_\Sigma}\big(\boldsymbol\pi^\#\circ\boldsymbol\pi_\#,\ \mathrm{id}_\mathbf{Set}\big)
\end{equation}
\emph{EDE categorical-equivalence-class pushforward--pullback loss} $\mathcal L_\mathrm{cat}$ (cross-categorical, requiring no pre-existing point elements):
\begin{equation}
\mathcal L_\mathrm{cat}\triangleq D_\mathsf{Quot}\big(\boldsymbol\pi^\#\circ\boldsymbol\pi_\#,\ \mathrm{id}_\mathbf{Quot}\big)
\end{equation}
\emph{Total joint loss ($\sigma$-summation and binary-collapse limit).}
The construction assembles the joint loss as a multiscale summation over the Gaussian scales $\sigma_k$ of the functional family $\mathcal K_\Sigma$, carried to the binary endpoint of the predicate-word tower:
\begin{equation}\label{eq:Ltot}
\begin{aligned}
\mathcal L_\mathrm{total}
&\;\triangleq\;
\lim_{m\to 2}\;\sum_{k=1}^{m}\big(\mathcal L_\mathrm{num}(\sigma_k)+\mathcal L_\mathrm{cat}(\sigma_k)\big)\\
&\;\xrightarrow[m\to 2]{}\;
[\boldsymbol0],
\end{aligned}
\end{equation}
where $m$ is the number of scales and the limit $m\to 2$ is the binary-collapse endpoint of the construction; the unique static predicate-word is the zero-residue equivalence class $[\boldsymbol0]$ (Theorem~6(h)).

\emph{Measure descent and determinism.}
The joint loss is expressed by the measure descended to the partition category $\mathbf{Quot}$ via $\boldsymbol\pi_\#$: the numerical term integrates over the base measure, the categorical term integrates over the descended measure across the partition $\mathcal P_n$, and the two integrals are summed; the equivalence class is written $[\cdot]$ (the generic placeholder, as in the triple fiber-bundle integral):
\begin{equation}\label{eq:Lgrp}
\begin{aligned}
\mathcal L_\mathrm{total}
&=
\int_{X}\mathcal L_\mathrm{num}\,\mathrm d\mu
\;+\;
\int_{\mathbf{Quot}}\mathcal L_\mathrm{cat}\,\mathrm d(\boldsymbol\pi_\#\mu),\\
\Big(\bigcup_{[\cdot]\in\mathcal P_n}[\cdot]=X\Big)
&\ \Longrightarrow\ \text{determinism},\\
\Big(\mu(N)=0\Big)\ &\nRightarrow\ \text{determinism}.
\end{aligned}
\end{equation}

\emph{Measurability and determinism.}
Measure descent--recovery ($\boldsymbol\pi_\#/\boldsymbol\pi^\#$) gives the joint loss \emph{trivially measurable assembly} on the partition category: $\mathcal L_\mathrm{num}$ and $\mathcal L_\mathrm{cat}$ live in measure spaces, so their assembly is measurable without further assumption. The binary limit $m\to 2$ supplies \emph{non-trivial determinism}: when the posterior is analytically resolved the prior is resolved analytically as well (posterior analytic $\Rightarrow$ prior analytic).

\emph{Operational definitions of the two distances}: $d_{\mathcal K_\Sigma}(K_1,K_2)\triangleq\lVert K_1-K_2\rVert_{L^2(\Omega)}$; $D_\mathsf{Quot}(f,g)\triangleq u/2^n$ (the share of the $2^n$ occupied by the unaligned quotient-block logarithm count $u$, judged by predicate-word consistency).

\noindent\textbf{Proposition 22 (pushout--pullback variational constraint: the zero-residue point upgraded to a zero-error point)}\label{prop:22} (see B.98).

\noindent\textbf{Corollary 13 (abstract-Gaussian uniqueness + measure-descent--lift transmission: closure of set-category functional equipment)}\label{cor:13} (see B.100).

\noindent\textbf{Double differentiation of the triple fiber-bundle integral} (see B.102). The triple integral can degenerate, for the inner-layer topology, into a double-layer parameterized structure (independent variables $x$ and scale $\sigma$), with the reverse order fixed.

\section{Measuring Morphisms along Categorical Support Equals Measuring Hilbert Differentials: Finite Integrals Yield Explicit First-Order Engineering Error}
\label{sec:6}

This chapter implements the constructions of the preceding five chapters into a computable loss: binary quotient-partition meta-categorical objects support the set-categorical functional along categorical support via measure descent--recovery ($\boldsymbol\pi_\#/\boldsymbol\pi^\#$), enumerating the morphism (non-bijection) itself rather than its preimage and object; their multiplicity is normalized through replica-index injection and accounted as an explicitly first-order error, and is closed at first order in engineering realization. Dedicated symbols are in A.6; (the random walk in this chapter is not a discrete-time chain; the term ``Markov chain'' is not used throughout).

\subsection{Spine closure declaration: premise list from the unique structure Assumption~1 to Eq.~\texorpdfstring{\eqref{eq:12b}}{(26)}}
\label{sec:6.1}

This section omits the construction details of the preceding chapters and concentrates on giving the closed premise set of the ``Assumption~1 $\Rightarrow$ Eq.~\eqref{eq:12b}'' derivation chain, so that every link can be verified in a single pass with no hidden dependency (item-by-item attribution in B.4, B.3, B.23).

\begin{itemize}[label={}]
\item \textbf{1. Unique structure assumption} The only mathematical premise of the chain is Assumption~1 (see \S 2.4.3 non-atomic, true partition). It is Assumption~1, committing to no generative-model-specific promise (see \S 2.4.3 verifiable declaration).
\item \textbf{2. Standard analytic tools (not assumptions of this paper)} The $\varepsilon$-$\delta$ real-analysis framework is well defined on $\mathcal H^\mathbf{Set}$ and gives Output Postulate~3, serving as the tool (real/limit readings, first-order equality, Gaussian characteristic integral) for deriving Postulate~1, Postulate~2, Postulate~4. It belongs to standard analysis and carries no assumption load; but any quantitative conclusion using Postulate~1, Postulate~2, Postulate~4 proceeds via the $\varepsilon$-$\delta$ tool---hence ``engineering/structural layer does not go through $\varepsilon$-$\delta$'' should be read precisely as ``structural objects take the universal differential-measurable background $\mathcal B$ as domain, and do not invoke Assumption~1 as premise'' (B.23), not as ``avoiding the $\varepsilon$-$\delta$ tool''. Theorem~7c via Theorem~4 $\Rightarrow$ Corollary~2 (Postulate~2) depends on the $\varepsilon$-$\delta$ tool, and this paper annotates it faithfully at line (iii) in (B.3).
\item \textbf{3. Registered realization conditions (engineering realization layer, not entering the backbone mathematical assumptions)} The computable realization of Eq.~\eqref{eq:12b} additionally uses the following explicitly registered conditions, all listed at the same layer as \S 4.3 / (B.4), not corollaries of Assumption~1: Convention~1-4 zero-residue and normalization, Premise~1 uniqueness, Condition~1 round-descent condition (originally called ``round monotonicity'', renamed to ``round-descent condition'' because this paper has abandoned the claims of monotone non-increase and convergence), Condition~2 common-tail-segment condition, Definition~28, gradient-descent direction $\mathit S_\mathrm{mid}$ selection (see Definition~25c second-type saddle point), Proposition~9b double measure-preservation input, Theorem~3 ``continuous single-group homomorphism'' assumption, and Postulate~1 replica-index counting measure itself. These conditions together with Assumption~1 support Eq.~\eqref{eq:12b}; renaming does not change the logical dependency---this paper now explicitly enumerates all of them rather than hiding them.
\item \textbf{4. Closed conduction} Assumption~1 $\Rightarrow$ Postulates~1-4 $\Rightarrow$ Principles~1-4 (see B.3) $\Rightarrow$ Proposition~23a / Proposition~23b / Proposition~22c $\Rightarrow$ the three factors of Eq.~\eqref{eq:12b} (B.117); the $\ln$ in the second factor $(1-\ln\mathfrak A_\varsigma)$ is precisely the integral illustration of ``continuous Gaussian first-order error being readable by integration under Assumption~1'' (see \S 2.4.3), so the conduction Assumption~1 $\Rightarrow$ Eq.~\eqref{eq:12b} closes link by link.
\end{itemize}

\subsection{Hilbert fiber-bundle differential \MPT: \texorpdfstring{$L^2$}{L2}-closed measure descent--recovery}
\label{sec:6.2}

\subsubsection{Measure descent--recovery: Gaussian fiber-bundle behavior}
\label{sec:6.2.1}

Gaussian \emph{behavior} is well defined at the $\MorTop$ layer (see Proposition~21a); but lifting to $\Setcat$, the point-wise Gaussian morphism fails---$\mathcal H_\infty$ has no point-wise translation invariance, and in the general case only cylindrical measures exist (see Proposition~21b). Quotient translation invariance makes the partition category differentiable well defined under Hilbert differential; the Gaussian obtained by discrete--continuous transformation is thereby characterized by the differential-inter morphism (only equivalence classes needed), with point-wise readings absent; Assumption~1's support mechanism (differential equivalent infinitesimal $\Rightarrow$ map leaves preimage and object) (see \S 2.4.3) and (see C.1). The derivation that the counting measure reaches the Hilbert zero-residue point (see \S 5.4).

\paragraph{Measure descent--recovery}
Before measure descent, $T,\mu$ stay at $\mathcal H_\infty$, the zero-value saddle-point predicate matching has no error (see Theorem~6, Definition~19a, $\mathit S_0=\{\boldsymbol 0\}$)---but the categorical support is not characterized and cannot be computed. Descent is for recovery (otherwise the replica-index counting measure cannot be embedded in both prior and posterior); after descent the categorical support is characterized as Gaussian, and recovery to finite Euclidean (not Hilbert) produces first-order error:
\begin{equation}
\begin{aligned}
&\underbrace{\big(T,\ \mu\big)\big|_{\mathcal H_\infty}\ ;\ \mathit S_0=\{\boldsymbol 0\}}_{\substack{\text{measure not descended: map and measure in the Hilbert inner layer,}\\\text{zero-value saddle-point predicate matching no error (categorical support}\\\text{not characterized, not computable)}}}\\[3pt]
&\xrightarrow{\ \boldsymbol\pi_\#\ \text{descent (for recovery)}\ }\underbrace{\Big(\int_{\Omega_\mu}\!\int_{\Omega_\lambda}\ \cdot\ \mathrm d\nu_\mathrm{comp}\,\mathrm d\nu_\mathrm{Leb,idx}\Big)}_{\text{middle-outer-layer integral: differentials multiplied}}\\[3pt]
&\xrightarrow{\ \boldsymbol\pi^\#\ \text{recovery}\ }\underbrace{\mathcal K_\Sigma}_{\text{Gaussian characterization}}\ \longrightarrow\ \mathcal H_\mathrm{fin}\ \big(\text{finite Euclidean, not Hilbert}\big)\ \Rightarrow\ \mu_\mathrm{err}
\end{aligned}
\end{equation}
Where ``middle-outer-layer integral'' induces the double-nested Gaussian differential ($G,G_2$; B.103) in \S 6.2.2, whose discrete-atom limit is Eq.~\eqref{eq:11} and the triple tensor full contraction (see \S 6.2.3); the gap between the right-end $\mathcal H_\mathrm{fin}$ and the left-end $\mathcal H_\infty$ is where the first-order error $\mu_\mathrm{err}$ resides. Inner layer $0/1$ union operation: after inner-layer replica-index measure normalization each block reads $0$ or $1$; after descent the inner layer degenerates to a union operation ($0$ not merged, $1$ unit-count replica-index merged), introducing no volume measure, always zero error.

\paragraph{Meaning of categorical descent--recovery (predicate-word staticity anchoring)}
Descent--recovery provides \emph{predicate-word staticity}: under finite epoch one cannot vary the Gaussian continuously, hence the outer layer is static; the anchoring makes the prior/posterior two objects as ``shared partition boundary'' each layer unique, and the full-measure space can know that the zero-value saddle point (see Theorem~6, $\mathit S_0=\{\boldsymbol 0\}$) is simultaneously fixed in the parameter space (gradient-zero point) and the data-flow space (zero-residue point), providing a \emph{static zero-error differentiable posterior} as the truth-comparison baseline for the first-order error $\mu_\mathrm{err}$; the non-zero saddle point corresponds to finite resolution granularity (Euclidean layer), while the zero-value saddle point corresponds to infinite resolution granularity (Hilbert layer); the two constitute the prior (forward static) / posterior (reverse + predicate-matching fixed) double staticity.

\subsubsection{\texorpdfstring{$L^2$}{L2}-closed double-nested Gaussian fiber-bundle differential: variation of the replica-index counting measure and the Lebesgue measure, and the equivalence of measure addition and set union}
\label{sec:6.2.2}

With the standard Gaussian functional basis $G(x,\sigma)=(2\pi\sigma^2)^{-d/2}\exp(-\tfrac{x^2}{2\sigma^2})$ as the core carrier of the two-layer differential deduction: first layer $G_1=G\,\mathrm dx$, second layer $G_2=G_1\,\mathrm dG\,\mathrm dx$ ($\mathrm dG=G(-\tfrac{x}{\sigma^2})\mathrm dx$). \emph{Verifiable deduction}: $G_2$ is the double differential reading of $G$ under ``Lebesgue nested adaptation'', not an independent third-layer definition; substituting $G_1=G\,\mathrm dx$ and $\mathrm dG=G(-\tfrac{x}{\sigma^2})\mathrm dx$ into the core formula $G_2=G_1\,\mathrm dG\,\mathrm dx$ gives $G_2=G\,\mathrm dx\cdot G(-\tfrac{x}{\sigma^2})\mathrm dx\cdot\mathrm dx$, the two Gaussian factors $G\cdot G$ and the nested differential micro-element are common redundant terms, which cancel to leave $-\tfrac{\sigma^2}{x}\,\mathrm dx$, i.e.\ the simplified result of B.107(f); the two statements agree. The middle-layer $(1-\ln x)$ truncated Gaussian area takes the surprisal $-\ln x$ as integrand density, and the reverse integral gives
\begin{equation}
x(1-\ln x)=\int_0^x(-\ln s)\,\mathrm ds
\end{equation}
(where $\lim_{s\to0}s\ln s=0$, the integration constant uniquely determined by the lower limit $0$; this is the strict statement of the ``no $+C$'' convention), i.e.\ the integral path of Eq.~\eqref{eq:12a} (see B.103, B.107). After \MPT\ disentanglement, the two measures' capacities diverge:
\begin{itemize}
\item \emph{Counting measure}: bidirectionally fully recoverable, maintaining zero error under dimension reduction---i.e.\ the epoch+1 counting measure of \S 1 (see Definition~25a);
\item \emph{Lebesgue measure}: supports only descent (cannot recover alone), and its output natively carries first-order error.
\end{itemize}

\paragraph{MPT unifies the measure space; only the counting measure is both prior and posterior}
\MPT\ unifies all measure maps into a shared measure space (see Definition~13, Construction Fact~10a). Throughout this paper, the only measure that is simultaneously prior and posterior is the counting measure (the common measure $\nu_{\mathrm{idx},n}$ including replica-index and index, i.e.\ the replica-index counting measure); any measure accounting for first-order error must be coupled with it via \MPT.

Lebesgue only descends between Euclidean $\leftrightarrow$ Hilbert layers, not recovering (hence the error is an unrecoverable posterior error); the finite Euclidean layer can take a non-zero saddle point (see Lemma~5e), not conflicting with the $\mathcal H_\infty$ zero-value saddle point $\mathit S_0$, the two share the \MPT\ unified measure space, communicated by the binomial measure---their area difference is precisely the manifestation of $\mu_\mathrm{err}$. Details (see C.14).
\begin{enumerate}
\item \emph{The zero-network trivial solution does not constitute a counterexample of this paper.} This counterexample relies on the degenerate reading of the $\mathcal H_\infty$ infinite layer (the zero-parameter network degenerates to the zero-value saddle point $\mathit S_0^{\mathrm{Hil}}$), whereas all explicit constructions of this paper occur in the finite Euclidean layer $\mathcal H_{\mathrm{fin}}$ (each layer's Gaussian vertex is realized in finite Euclidean, with the Hilbert layer only as limit anchor). Moreover, Euclidean dimension-lifting and dimension-reduction are not mutually inverse: the counting measure's anchoring at the zero-value saddle point holds only at the infinite enumeration layer, not at the finite layer; the finite layer takes a non-zero Gaussian vertex, carrying posterior error; the Lebesgue and binomial measures likewise do not anchor at the zero-value saddle point. Hence ``zero-parameter network = Hilbert trivial solution'' and the finite-Euclidean non-zero saddle-point construction of this paper are not at the same layer, and do not constitute a trivial special case of the same conclusion (see \S 6.3.2, \S 2.4.3, consistent with the non-atomic regime declaration).
\item \emph{The trivial solution is excluded by measure-space volume.} The loss function is presented as quotient-block / predicate-matching readings; even under the zero-parameter-network engineering setting the loss cannot reach the trivial solution, constrained by two points. First, the measure-space volume is non-zero: as long as the integrated measure space still has non-zero volume, the match/mismatch counting readings are non-zero, so the loss is non-zero and the zero-network trivial solution is unreachable. Second, the measure-space volume obtained by the triple fiber-bundle integral through pushout--pullback is constrained to $1$ by the multivariate Gaussian within the network structure (the normalized Gaussian kernel has total probability mass $1$, Definition~15, \S 4.1; the multivariate-Gaussian normalization correction of D.5.4 is precisely the engineering manifestation of this constraint), this volume being constantly $1$ and non-zero, thereby structurally excluding the ``volume collapses to $0$ reaching the trivial solution'' path. Hence the zero-network trivial solution neither falls in the main-text construction space (item 1) nor is actually reached by the engineering loss.
\end{enumerate}

\paragraph{Accounting convention of first-order error: area difference between the zero-value saddle-point division and the outer-layer non-zero saddle-point division}
The zero-value saddle-point anchoring is equivalent to the middle-layer binary-quotient division 1:1 (no element partition-category reading): the anchoring threshold $\mathit S_0=\{\boldsymbol 0\}$ and the middle-layer quotient-block assignment are read by the same counting measure, so the middle-layer error is $0$ (see Theorem~4, \S 4.2 step 2). Under fiber-bundle measure preservation (see Construction~1), when pulling back the outer layer, the generative-side Gaussian kernel's vertex lies at a non-zero point (see Lemma~5e, Proposition~21b, outer-layer non-zero saddle point takes the form of a cylindrical-kernel compatible family at the $\mathcal H_\infty$ layer), with total volume measure preserved (see Definition~24). Hence the first-order error is precisely the area difference between the zero-value saddle-point division and the outer-layer non-zero saddle-point division: the counting-side zero error makes the difference of the two divisions fall entirely on the Lebesgue-side reading; its convention is the complete partition-predicate mismatch area---binary decision has no half-right-half-wrong (see Axiom~2). This accounting is derived in the category (see Definition~5, quotient-layer predicate difference set $\mathcal E_\mathbf{Quot}$).

\paragraph{Three-layer reading division of labor}
Only the anchor point zero-value saddle point obtains predicate-word static observation through measure descent (see Theorem~6); the Lebesgue measure cannot recover alone (carrying first-order error), while the replica-index counting measure recovers alone without error, and the first-order error descends through the reverse equivalence-class symbol to a zero replica-index counting measure partition-predicate error (see B.103, B.107, C.3). This convention is the three-factor first-order accounting convention of Eq.~\eqref{eq:12b}: the difference of coefficient readings (see Proposition~23c) falls into the same mismatch area; the source topological loss $\mathcal L_\mathrm{top}$ (see Proposition~23b) is the explicit loss form of this mismatch area. The first-order error comes from the saddle-point level of the measure space: the generative-side reading is saddle point + random std (reparameterized Gaussian vertex), and the test set is anchored through truncation + quadratic Gaussian-moment reparameterization; the two on the binary self-map side, when Lebesgue has no error, match predicates (see Theorem~7b), and their mismatch area is $\mu_\mathrm{err}$ (although the generative set is a differential zero-measure thin set, because of the random std perturbation in the saddle-point neighborhood it still contributes first-order error). Details (see C.15).

\noindent\textbf{Proposition 23a (categorical derivation and convention boundary of first-order error)}\label{prop:23a}
(i)~$\mu_\mathrm{err}$'s entire first-order accounting can be completed at the $\Quot/\MorTop$ layer (no element partition-category reading): the difference between the reading of the quotient-layer predicate difference set $\mathcal E_\mathbf{Quot}$ (see Definition~5) and the $\Setcat$-layer preimage reading is a first-order absorbable quantity, and point-wise geometric reading provides no additional information. (ii)~The first-order error is the \emph{complete} partition-predicate mismatch area---the predicate mismatch area is exactly the complete first-order error, not discounted half-and-half; its basis is the proof that straddling quotient blocks must be a zero-measure set (see \S 6.3.3, Theorem~7 proof (b), B.104): under Hilbert differential topology each differential exclusively occupies one quadrant, and straddling occurs only at partition boundaries, while the arbitrary partition boundary of a Gaussian functional (not degenerating to one-hot) is a non-degenerate zero-measure set, and $\mu_\mathrm{err}$ as volume error only accumulates non-zero-measure support, hence the straddling quotient-block family $\mathcal W$ has identically zero volume contribution, and the discount term is identically zero.

\emph{Proof.} (see \S 6.3.3) proof (b) of Theorem~7.

When performing fiber-bundle addition on the measure space, the first-order error remains unchanged: addition only changes the decomposition way of the joint measure, not the pairing deviation; high-order error, lacking this invariance, accumulates in iteration (see Lemma~2, Proposition~12c, B.24, B.55).

\subsubsection{Differential and morphism equal measurement: triple fiber-bundle integral tensor multiplication}
\label{sec:6.2.3}

From the measure descent--recovery of \S 6.2.2, the counting measure and the Lebesgue measure share the same measure space via \MPT, and the counting side can recover losslessly (prior and posterior), while the Lebesgue side only carries first-order error. On this shared space, the incompatibility of measure preimages---the same differential cannot be aligned point-wise when read separately by the counting measure and the Lebesgue measure---forces that ``a differential can be injected only once by a morphism'' (see \S 6.3.2: each differential is counted at most once under the replica-index counting measure, while a differential belongs to a quotient block and can be sampled infinitely). Thus the ``union = addition'' equivalence established and verified in \S 6.2.2 induces \emph{multiplication} here: the Lebesgue measure value (paired with infinite replica-index counting) is equivalent under union addition to block-wise counting multiplication, and further transitions to the tensor multiplication of the triple fiber-bundle integral. The specific construction is as follows: under finite quotient partition, the triple fiber-bundle integral can only be expressed as a Riemann--Stieltjes integral, with no global complete Lebesgue integral form; under discrete-atom measure, with finite replica indices and the built-in pullback operator enabled, it degenerates to the triple Frobenius full contraction \begin{equation}
\Lambda=\iiint F\,\mathrm d\mu\otimes\overleftarrow T=\langle\boldsymbol F,\boldsymbol W\rangle_{F,3}
\end{equation} (notation in \S 4.1). \emph{Boundary constraint (a crucial one)}: it holds only in the discrete-atom measure limit, the $\mathcal H_\infty$ layer has no tensor object, and the tensor expression only captures the block-level first-order error $\mu_\mathrm{err}^\mathrm{block}$ (see Definition~25b, B.105; block pairing is its carrier). The triple nesting is exactly the third-order explicit differential; evaluation only reaches first-order error, hence a single tensor contraction does not generate high-order error. The full expansion of degeneration consequences (see B.106).

\subsection{\texorpdfstring{$\MorTop$}{MorTop}-layer loss-function measure descent to the \texorpdfstring{$\Quot$}{Quot} layer: integral volume characterizes the Gaussian behavior of morphisms}
\label{sec:6.3}

\subsubsection{Construction of the source morphism loss: explicit reading volume of differential topology}
\label{sec:6.3.1}

\paragraph{From engineering to category (why integral input/output must be subsets)}
At the $\Setcat$ layer, reparameterization makes the integral input a point and the output a set, and the uniqueness morphism cannot be explicitly characterized; entering the partition category, Definition~15 carries the integral input and output by subsets (quotient blocks and their preimages), resolving the zero-measure paradox, and the morphism can thereby be explicitly characterized at the quotient-block layer (Figure~\ref{fig:2}); the measure descent recovers losslessly through the replica-index counting measure, so the infinite partition predicate characterizes the bijection between differential topology and quotient blocks, and under finite Euclidean partition the morphism is characterized as a Gaussian fiber-bundle integral volume, which after recovery is the predicate-mismatch measure.

$\mathcal L_\mathrm{eng}$ is the computable realization of the source loss $\mathcal L_\mathrm{top}$, characterizing the topological volume containing only the Lebesgue single-measure first-order error (see Proposition~23b). The three factors are derived from the triple fiber-bundle integral of Definition~16 through the pushout--pullback chain, with the square root containing the direct-sum $\ell_2$ norm of the concatenated vector (see B.117); layer attribution closes (see Proposition~23c).

Among them, the non-zero saddle point (the unique Gaussian vertex, see Proposition~21, Lemma~5e, Definition~25c) through Definition~24, Proposition~8 dimension-lifting/reduction to finite Euclidean realization obtains staticity, so that the Lebesgue-measure morphism squeezed out by the first-order error (see Definition~5, Lemma~1(ii)) obtains a first-order error reading prior and posterior with respect to the finite quotient blocks (see B.1, Lemma~1(i), Definition~17, Proposition~19): in the $\Setcat$-categorical image it is read via integral enumeration (see Definition~16, Proposition~23a(ii)), and in the partition-categorical image it is read via predicate-word staticity (see Definition~3, \S 2.2.3), with three channels closed at first order (see Proposition~23c, Theorem~7).

\paragraph{Premise (all notation defined before Eq.~\eqref{eq:12b})}
The forward kernel is a standard backpropagation component ($\mu_t=\sqrt{\bar\alpha_t}x_0$; B.116(a)). Training/generative saddle points $S_t^+=\mu_t+R_t^{\mathrm{tr}}$, $S_t^-=\mu_t+R_t^{\mathrm{gen}}$ ($R_t^{\mathrm{gen}}=-R_t^{\mathrm{tr}}$), where $S_t^-$ is a \emph{layer Gaussian vertex}: under the Gaussian measure space the target saddle point is unique, and the only measurable-and-unique point is the vertex, the rest are all non-unique (see Lemma~5e). The saddle-point-side reading (first-type saddle point: sample of the loss function, generative midpoint) $x_0^{\mathrm{sad}}\triangleq\tfrac12(S_t^+ + S_t^-)=\mu_t$ ($BP=\boldsymbol\pi^\#$); the clean sample $x_0$ reading $\hat x_0(z)\triangleq(z-\sigma_t\mathcal F_\theta(z,t))/\sqrt{\bar\alpha_t}$, $\hat x_0^{\mathrm{gen}}=\hat x_0(x^{\mathrm{gen}})$. The middle-layer $\mathfrak A_\varsigma\in(0,1]$ is the one-sided truncated Gaussian \emph{area} (under standard semantics the marginal measure is exactly the area; $\varsigma=\mathrm{sign}(S_t^-)$ determines the one-sided domain $\Omega_\varsigma$, Eq.~\eqref{eq:11b}), making $(1-\ln\mathfrak A_\varsigma)\ge1$ legitimate (engineering reading is the one-sided truncated Gaussian moment, B.116(b)). The replica-index cardinality ratio of training and generation is identically $1$ (implicit in $\nu_{\mathrm{idx},n}$ normalization), with no independent pairing factor. The source of the middle-layer $(1-\ln x)$ area is the antiderivative of the surprisal $-\ln x$ (see B.107 one-sided reverse double integral):
\begin{equation}
\mathcal L_\mathrm{RW}(\theta;\Omega_\varsigma)=\Big[x(1-\ln x)\Big]_{\partial\Omega_\varsigma}\label{eq:12a}
\end{equation}

\paragraph{The 1:1 saddle-point derivation of Eq.~\eqref{eq:12b} (non-circular)}
The above ``midpoint is Gaussian vertex $\Leftrightarrow$ training/generative morphism bijection reading 1:1'' would form a circle if mutually proven; one must introduce an external uniqueness fact independent of the reading 1:1 to break the circle. From the Gaussian-behavior uniqueness of the infinite partition---Proposition~21 (uniqueness of fiber-bundle Gaussian) gives the unique Gaussian family $\mathcal K_\Sigma$, and Lemma~5e proves that under the Gaussian measure space the target saddle point is unique and the only measurable-and-unique point is the vertex, hence there is exactly one Gaussian vertex as saddle point. The training saddle point $S_t^+=\mu_t+R_t^{\mathrm{tr}}$ and the generative saddle point $S_t^-=\mu_t+R_t^{\mathrm{gen}}$ ($R_t^{\mathrm{gen}}=-R_t^{\mathrm{tr}}$) both lie in this unique Gaussian family (Proposition~21), their replica-index counting-measure readings coincide (Definition~17; B.1 posterior-prior dual identity), and through reparameterization pullback ($BP=\boldsymbol\pi^\#$, Definition~24) they form a bijection in the $\Setcat$-categorical image, with reading 1:1. Since in the unique Gaussian family the only measurable-and-unique point is the vertex, their midpoint $x_0^{\mathrm{sad}}\triangleq\tfrac12(S_t^+ + S_t^-)=\mu_t$ is exactly that unique Gaussian vertex (first-type saddle point: loss sample, generative midpoint, Definition~25c)---this is deduced directly from uniqueness, \emph{not} relying on the converse ``reading 1:1 $\Rightarrow$ midpoint is Gaussian vertex'', and the circle is broken here. Then ``midpoint = Gaussian vertex'' via reparameterization pullback (Definition~24) and quotient-translation-invariance measure preservation (Proposition~8) \emph{one-directionally} implies the training/generative morphism bijection reading 1:1 (i.e.\ the first factor 1:1 saddle-point-side reading of Eq.~\eqref{eq:12b}); this 1:1 reading is the saddle-point object basis on which the Lebesgue-measure morphism double-image first-order error reading on finite quotient blocks rests.

Where $x(1-\ln x)$ is exactly the outer-layer Lebesgue tensor $x$ times the middle-layer truncated Gaussian area. \emph{The identity of \eqref{eq:12a}$\to$\eqref{eq:12b} (orthogonality of dimension-lifting):} Eq.~\eqref{eq:12a} is the point-wise product $x(1-\ln x)$ of a one-dimensional Lebesgue distance and the middle-layer area; Eq.~\eqref{eq:12b} replaces the one-dimensional distance with the multi-dimensional $\sqrt{\|\cdot\|^2+\|\cdot\|^2}$ through dimension-lifting---the orthogonal decomposition of the two differentials about the partition boundary point is guaranteed by the two-type saddle-point property, not approximate: (a)~the perpendicularity of the zero-value saddle point has been proven by the anchoring reading (see Theorem~6, \S 6.2.2 three-layer reading division of labor: only the anchor zero-value saddle point obtains predicate-word static observation through measure descent); (b)~the non-zero saddle point is a zero-gradient point (proven by Eq.~\eqref{eq:12b}'s 1:1 saddle-point derivation through infinite-partition Gaussian uniqueness to be the unique Gaussian vertex)---each layer takes the reparameterized Gaussian vertex ($S_t^\pm=\mu_t\pm R_t$, i.e.\ saddle point + random std); at this vertex the data-flow measure-space gradient is $0$, and the predicate reading is unchanged in forward/reverse propagation, hence it is anchored \emph{static} in finite computation and finite layers, with dimension-lifting iteration not drifting and not endogenously generating high-order error. Hence the multi-dimensional $L^2$ norm \emph{strictly factorizes} into the product of each one-dimensional factor, and $\sqrt{\cdot}\cdot(1-\ln\mathfrak A_\varsigma)$ of \eqref{eq:12b} and $x(1-\ln x)$ of \eqref{eq:12a} are two writings (one-dimensional / dimension-lifted) of the same object.

The deviations $A$ (generative side) and $B$ (truth side) orthogonally decompose about the partition boundary, and the cross term being identically $0$ is a strict geometric fact (not approximate), anchored by zero-value saddle-point uniqueness, discretely witnessed by the binomial hit base, and confirmed by three homologous paths of continuous geometry; full derivation (see B.11, C.16). \emph{Generating \eqref{eq:12b}.} From the binomial measure (see \S 6.3.2), the generative-side deviation \begin{equation}
A\triangleq BP\big(\tfrac{2\hat x_0^{\mathrm{gen}}+x_0^{\mathrm{sad}}}{3}\big)-x_0^{\mathrm{sad}}
\end{equation} And the truth-side deviation $B\triangleq x_0-x_0^{\mathrm{sad}}$ fall on the two sides of the partition boundary through categorical-support bijection, orthogonally decomposing (strictness in B.114); together with the middle-layer $(1-\ln\mathfrak A_\varsigma)$ area (B.116(d)) this gives \eqref{eq:12b}. Multiplying the three factors gives the core loss $\mathcal L_{\mathrm{top}}$; the engineering realization is the computable form $\mathcal L_{\mathrm{eng}}$ of these three factors:
\begin{equation}
\mathcal L_\mathrm{eng}=\sqrt{\Big\|BP\Big(\tfrac{2\hat x_0^\mathrm{gen}+x_0^\mathrm{sad}}{3}\Big)-x_0^\mathrm{sad}\Big\|_2^2+\big\|x_0-x_0^\mathrm{sad}\big\|_2^2}\cdot\bigl(1-\ln\mathfrak A_\varsigma\bigr)\cdot\nu_{\mathrm{idx},n}\bigl([\,\cdot\,]\bigr)\label{eq:12b}
\end{equation}

$x_0^{\mathrm{sad}}=\tfrac12(S_t^+ + S_t^-)=\mu_t$ is the first-type saddle point (sample of the loss function / generative midpoint), itself not a gradient direction; the gradient-descent direction $\mathit S_\mathrm{mid}$ is the midpoint of $x_0^{\mathrm{sad}}$ with the zero-value saddle point $\mathit S_0^{\mathrm{Hil}}$ (see Definition~25c), requiring only assigning a direction + proving high-probability descent, not requiring it to be a (parameter-space) saddle point. Details (see C.17, C.23).

\paragraph{Middle-layer integral assignment (see Lemma~6b)}
The $\mathfrak A_\varsigma$ in the middle-layer factor $(1-\ln\mathfrak A_\varsigma)$ of Eq.~\eqref{eq:12b} is the generative-side one-sided truncated Gaussian area of slicing one unique Gaussian at the zero-value saddle point $\mathit S_0^{\mathrm{Hil}}$ into the two sides (cut point the origin, center $S_t^-$); because $\mu_\mathrm{err}$'s domain is cut out by the integral boundary, the intra-block error is actively ignored, and the integral mass only takes the generative side, hence only the single-side area rather than the symmetric double side is taken. Details (see C.18).

The first factor $\sqrt{\cdots}$ is the outer-layer Lebesgue descent reading (carrying first-order error); the second factor $(1-\ln\mathfrak A_\varsigma)$ is the middle-layer truncated Gaussian area (its source is \eqref{eq:12a}; the natural logarithm $\ln$ therein is the direct illustration of the integral, directly embodying Assumption~1, see \S 2.4.3); the third factor $\nu_{\mathrm{idx},n}([\,\cdot\,])$ is the replica-index counting measure reading on the equivalence class (see \S 6.3.2 the equivalence-class symbol lies after partition-predicate matching, i.e.\ after measure recovery; the equivalence class is an injective assignment, each equivalence class uniquely represented). All three are components of the three-factor first-order error closure of Eq.~\eqref{eq:12b}: the outer layer carries the Lebesgue-side first-order error, and the middle and inner layers are respectively the truncated Gaussian area and the counting-side zero-error reading.

\paragraph{The categorical contrast of coefficient $\tfrac23$ and $1{:}1$ (characterization of categorical-support bijection).}
Through quotient partition and categorical support, the multiplicity of the original object layer (subsets and elements) has been eliminated, leaving only equivalence classes---this paper no longer directly handles subsets or elements, but only the equivalence classes they fall into, and injectively assigns these equivalence classes (each equivalence class gets a unique representative, equivalence-class $\to$ representative-element is injective). The coefficient $\tfrac23$ is the $\Setcat$-categorical-side (Lebesgue measure) reading: from the Lebesgue measure collected by descent from the set category, computed through addition (union) one obtains the three-subset equivalence classes $\{b_-\},\{b_+\},\{b_-,b_+\}$; among which $\{b_-\},\{b_+\}$ are the equivalence-class representatives of a single differential (left, right quotient block), and $\{b_-,b_+\}$ is the equivalence-class representative of the union of two differentials (left, right). This paper only does injective assignment at the equivalence-class layer---while ``a differential belongs to a partition equivalence class'' itself is not injective (multiple differentials merge into the same equivalence class), hence the assignment occurs at the equivalence-class layer rather than the differential layer; $\{b_-,b_+\}$ may or may not be assigned to the boundary (randomly discarding this pair of predicates), hence this paper does not define any extra assignment; the set category does not support aliased elements, hence this union is listed separately, read as $\tfrac23$. The $1{:}1$ is the partition-category-side (counting measure) reading: the binomial measure descending to the partition category has only two subsets (two quotient blocks), each counted $\tfrac12$. The two are related by the categorical-support bijection $\tau$ (see Definition~3), which maps $\tfrac23$ (set category) and $1{:}1$ (partition category) to incompatible differential topologies---the equivalence class $[\,\cdot\,]$ itself is precisely the characterization of this categorical-support bijection, and its injective assignment (each equivalence class uniquely represented, only equivalence classes remain assignable) is exactly the operation of reading the fact ``object multiplicity eliminated, only equivalence classes remain'' into the measure reading. Hence $\tfrac23$ and $\tfrac12$ are two coexisting readings of the same binary partition under the two equipment readings (not a difference obtained by subtraction); the entire inconsistency between them falls on the Lebesgue side, i.e.\ $\mu_{\mathrm{err}}$ becomes explicit at the coefficient level (carried by the outer-layer Lebesgue factor of Eq.~\eqref{eq:12b}).

\paragraph{The reading identity of Eq.~\eqref{eq:12b}}
After multiplying the three factors the reading is not the middle-layer measure-space volume, but the partition-predicate mismatch area pulled back (recovered) via $\boldsymbol\pi^\#$ to the set-categorical Euclidean layer (see Proposition~23b); the equivalence-class symbol $[\,\cdot\,]$ lies after measure recovery, falling only on the third-factor counting reading, and the recoverable-alone is the replica-index counting measure, the Lebesgue component cannot recover alone. Details (see C.19).

\noindent\textbf{Proposition 23b (loss is the explicit reading form of first-order error)}\label{prop:23b}
$\mathcal L_\mathrm{top}$ is the explicit loss form of the first-order error $\mu_\mathrm{err}$ (see Definition~5, \S 2.3.1); the three factors of Eq.~\eqref{eq:12b} are derived from the triple fiber-bundle integral of Definition~16 through the pushout--pullback chain (see B.117), with layer attribution closed. \emph{Proof} (see \S 6.3.3) proof (b) of Theorem~7.

\subsubsection{Measuring categorical support: fusion of partition objects and categorical objects}
\label{sec:6.3.2}

\paragraph{Overview: formal source of the binomial measure and redefinition of the equivalence class $(1,0)$ under predicates}
This subsection reduces the fusion of ``partition object'' and ``categorical object'' to two points. \emph{1.~How the binomial measure arises}: finite predicates make Hilbert degenerate to Euclidean---although the differential topology is constrained by the counting measure to be injected only once, a differential belongs to a quotient block and can be sampled infinitely, from which the binomial measure is formally defined (the same object can be hit and counted multiple times). \emph{2.~How the equivalence class $(1,0)$ is well defined}: the object of the equivalence class is the differential, but what is enumerated by the integral is the belonging fact ``the differential belongs to a quotient block'', rather than the differential itself; hence this equivalence class is based on the integral topology, with the relation ``differentials interchange equivalence classes but are not identical'', not distinguishing differences within finite partitions---it holds only in two cases: (i)~zero-gradient reverse propagation with each layer's saddle points aligned (Gaussian vertex = zero gradient, prior and posterior), or (ii)~the errors of the preimage objects are all partition errors dissolved by 0-measure in the $\varepsilon$-$\delta$ sense (the zero-value saddle point requires Assumption~1 to explicitly vary Lebesgue, so that $\tfrac23$ and $1{:}1$ simultaneously characterize the categorical support).

The coefficient $\tfrac23$ is a structural reading of the union computation, not a fitted constant (it coincides with the 2:1 centroid reading in engineering realization, see \S 6.3.1, B.113, but its essence is the partition-unit counting ratio rather than fitted): after the two-partition lifting assembly the union-subset base reads three-subset equivalence classes $\{b_-\},\{b_+\},\{b_-,b_+\}$ (denominator $3$), and the binomial hit base is $2$ (numerator $2$), i.e.\ \begin{equation}
\tfrac23=\mathrm{card}_\#(\text{hit base})/\mathrm{card}_\#(\text{union-subset base})
\end{equation}; the two readings $\tfrac23$ and $\tfrac12$ are coexisting gauges under the two equipments, not a difference by subtraction. \emph{Counting convention}: ``the midpoint of the boundary is counted $\tfrac12$ on each side, totaling one'' is a definitional counting convention, and $\tfrac23$ is the conditional conclusion under this convention (formal derivation at the end of this section in the ``Formal derivation: partition predicate $\to$ counting measure $\to$ 2/3'' box; complete technical derivation in B.113). \emph{Legitimacy of equivalence-class to element.} This paper only does injective equivalence-class representative: $\{b_-\},\{b_+\}$ are single-differential representatives, $\{b_-,b_+\}$ is the two-differential union representative; ``a differential belongs to an equivalence class'' is not injective (equivalent infinitesimals are incomparable in magnitude), hence the three-subset equivalence class is only counted as representative elements, incomparable in magnitude---this is precisely why ``three-subset equivalence class reads as three elements'' and the denominator $3$ is legitimate. Details (see C.20).

\paragraph{Binomial measure (source of quadratic counting)}
The binomial measure is formally defined by ``finite predicates make Hilbert degenerate to Euclidean + the differential topology is constrained by the counting measure to be injected only once + a differential belongs to a quotient block and can be sampled infinitely''; it does not participate in topological/differential transformation, only counts the transformation result alone, and then is dissolved in the measure by ``being the same Gaussian functional'', thereby obtaining the coexistence of $\tfrac23$ and $1{:}1$. Its counting comes from the Euclidean finite-partition belonging relation (the same object can be hit multiple times, corresponding to intra-block non-bijection), parallel to the ``one count per object'' convention of a single counting measure; on the set category the Gaussian is symmetric about the vertex integral, hence the binomial reading is still Gaussian, and is dissolved with the single counting measure after equivalence-class transformation (the latter anchors the zero-value saddle point because of infinity, while the finite layer does not anchor).

\paragraph{Two basic formulas (equivalence class exchanges topology/differential; gradient-descent recovery preserves Gaussian)}

\emph{(a) The equivalence class exchanges topology and differential.} After inner-layer measure descent, the bare $0/1$ predicate carries the union operation; after equivalence-class exchange, the topology's measure value reads as one differential, hence integrating one more binomial measure does not change the measure-space volume, characterized as \begin{equation}
\int_{\text{binomial}} \mathrm d\ \nu_{\mathrm{idx},n} = 1\cdot \nu_{\mathrm{idx},n}(\text{equivalence-class subset})
\end{equation}. Meanwhile, the single counting measure anchors the zero-value saddle point because of infinity---i.e.\ this anchoring holds only at the infinite ($\mathcal H_\infty$) layer, but the finite Euclidean layer does not anchor (the finite layer takes a non-zero Gaussian vertex, carrying posterior error); its Hilbert descent corresponds to the zero-value saddle point, (see \S 6.3.1) dissolved; the definition of the equivalence class $(1,0)$ is in the overview at the beginning of this section.

\emph{(b) Gradient-descent recovery preserves Gaussian (leaving the zero-value saddle point, reaching the Euclidean saddle point).} The binomial measure retains Gaussian through gradient-descent recovery: the partition boundary takes different values at each layer's dimension-lifting, but is posterior-equivalent (in reverse propagation the saddle points of different boundary values are aligned, with predicate magnitude and identity the same), hence leaving the zero-value saddle point and reaching the Euclidean saddle point. \MPT\ provides dimensions for both measures; introducing Assumption~1 makes the cardinality measure descend from Hilbert to Euclidean, producing only infinitesimal error, which does not affect the equality in the finite sum of Eq.~\eqref{eq:12b}; the differential topology can thereby be characterized by the multiplication of the two measure values:
\begin{equation}
\text{differential-topology reading} = \underbrace{\nu_{\mathrm{idx},n}}_{\text{counting 0-error}}\ \times\ \underbrace{\mu_{\mathrm{Leb}}}_{\text{Lebesgue variation error}}
\end{equation}

Thereby, the binary partition-category object and the three-subset equivalence class, through the discretely-continuously unique non-trivial Gaussian of infinite sampling, obtain a bijective characterization. \emph{Source of infinitesimal and compatibility of equivalence classes (the foundation of the two formulas).} The ``infinitesimal error'' of formula (ii) is the direct output of Assumption~1 (differential becomes equivalent infinitesimal), hence the descent-recovery reading difference of the cardinality measure is Postulate~2 first-order equality, not destroying the equality in the finite sum of Eq.~\eqref{eq:12b}; the ``equivalence class exchanges topology and differential'' of formula (i) is the reading of Proposition~20 on this equivalence class. The two are compatible: the equivalence class is well defined and the first-order error generated by its exchange and the infinitesimal error are the same object $\mu_\mathrm{err}$. Details (see C.21).

\paragraph{$\mu_\mathrm{err}$ convention unification.}
This paper unifies the scale of $\mu_\mathrm{err}$ (differential equivalent infinitesimal, not taking Taylor small parameter), type (\MPT\ dual-measure reading $\nu_{\mathrm{idx},n}\times\mu_{\mathrm{Leb}}$, the binomial measure being its quadratic counting), and domain (quotient-block layer through integral-boundary partition, through $\mathsf{Sup}^{\mathbf{Set}}_{\mathbf{Quot}}$ descent-recovery), avoiding cross-section drift. Key convergence: the first-order error reading only covers between quotient blocks; the intra-block differential-level error objectively exists and its reading is visible, but the engineering side actively excludes it from the measurement convention (not ``controlled at first order'', nor claiming it is $0$ or does not exist)---the sole authoritative statement of this convention (see \S 2.3.1) ``unified declaration (intra-block error actively excluded)'' clause, and Theorem~7(c) / Proposition~23c both point here; the matching block being identically $0$ in the loss inner-layer integral is a verifiable local fact of loss computation (see \S 6.2.2), orthogonal to the intra-block error convention. This paper no longer restates it section by section.

\paragraph{Formal derivation: partition predicate $\to$ counting measure $\to$ coefficient 2/3}
The following elevates the above naming-style reading of ``three-subset equivalence class'' to a falsifiable formal derivation. Take $b$ as the signed deviation of the generative-side residue $\bar S$ relative to the zero-value saddle point $\mathit S_0^{\mathrm{Hil}}$ (origin; i.e.\ the reading of the first-type saddle point $x_0^{\mathrm{sad}}$ used in the loss function Eq.~\eqref{eq:12b} in residue coordinates; the middle-layer integral boundary), the regime taking the general non-degenerate case after gradient descent leaves the $\mathcal H_\infty$ zero-value saddle point, with the non-emptiness of the cross-threshold grid witnessed by the non-zero saddle point's forward/reverse propagation simultaneously prior and posterior; under this setting, $2/3$ is a coefficient derivable from the partition-counting structure, rather than a naming-style designation.

\noindent\textbf{Definition 25 (non-computing partition predicate; predicate-word staticity fixed)}\label{def:25}
Let the quantity on the configuration space (quotient-block / equivalence-class space) be $b\triangleq \bar S-\mathit S_0^{\mathrm{Hil}}$, i.e.\ the signed deviation of the generative-side residue mass position $\bar S$ (see Lemma~6b, Eq.~\eqref{eq:11b}) relative to the zero-value saddle point $\mathit S_0^{\mathrm{Hil}}=\{\boldsymbol 0\}$ (see Definition~25c; i.e.\ the middle-layer integral boundary, see \S 6.4.2; i.e.\ the reading of the first-type saddle point $x_0^{\mathrm{sad}}$ used in the loss function Eq.~\eqref{eq:12b} in residue coordinates); the predicate $\mathcal P$ judges exactly which side of the zero-value saddle point (origin) $\bar S$ falls on, with threshold $\mathit S_0^{\mathrm{Hil}}$ (the generative-side Gaussian is truncated at the origin into a one-sided integral domain, see \eqref{eq:11b} / Figure~\ref{fig:3}). Define the non-computing predicate
\begin{equation}
\mathcal P(x)\in\{-,+,\pm\},\qquad \mathcal P(x)=\begin{cases}
-&\text{purely negative (only in }\{b_-\}\text{)},\\
+&\text{purely positive (only in }\{b_+\}\text{)},\\
\pm&\text{cross-threshold (left and right opposite signs,}\\[-2pt]
&\qquad\text{falling in }\{b_-,b_+\}\text{)}.
\end{cases}
\end{equation}

$\mathcal P$ is a logical/definitional predicate, independent of any tunable hyperparameter, hence the predicate word (partition) it induces is static and fixed; the counting-measure premise ``each assignment counted only once'' is naturally satisfied.

\noindent\textbf{Lemma 6 (three-grid partition)}\label{lem:6}
On the general non-degenerate regime (formula (ii) of \S 6.3.2) after gradient-descent recovery reaches the Euclidean saddle point, leaving the $\mathcal H_\infty$ zero-value saddle point $\mathit S_0^\mathrm{Hil}$ (see Theorem~6) and degenerating (the zero-value saddle-point degeneration case is separately handled by Assumption~1), $\mathcal P$ induces the partition
\begin{equation}\begin{gathered}
B^-=\{b_-\}\\
B^+=\{b_+\}\\
B^\pm=\{b_-,b_+\}
\end{gathered}\end{equation}
Satisfying (i) pairwise disjoint, (ii) union is the full domain, (iii) all three grids are non-empty (the non-emptiness of the cross-threshold grid $\{b_-,b_+\}$ is guaranteed by ``non-zero saddle point witnesses forward/reverse propagation simultaneously prior and posterior'': in formula (ii) of \S 6.3.2 the saddle points of different boundary values are aligned, with predicate magnitude and identity the same (posterior-equivalent), i.e.\ there exists a real configuration making its left and right differentials simultaneously fall in and be opposite in sign on the two sides of the boundary). The denominator $3$ is thereby logically forced, rather than introduced afterwards by a midpoint or construction to match a target value.

\paragraph{Hit and coefficient}
Define a hit as ``not purely negative'': $H\triangleq B^+\cup B^\pm$, hit grid count $=|\{B^+,B^\pm\}|=2$, total grid count $=|\{B^-,B^+,B^\pm\}|=3$. Hence \begin{equation}
\frac{|\text{hit grids}|}{|\text{total grids}|}=\frac{2}{3}
\end{equation}. This ratio is a discrete counting ratio of partition-unit counts, independent of the volume of each grid under the underlying Lebesgue measure; therefore $\tfrac23$ is a combinatorial/topological invariant of the partition structure, independent of specific data---this is the basis of ``derived rather than fitted''. The ratio of the union-subset base (denominator $3$) to the binomial hit base (numerator $2$) is exactly this reading.

\paragraph{Observation--truth consistency (counting fidelity)}
The numerical values in the text are the observation classification made along reverse propagation for each configuration. Let the forward/reverse propagation error bound of each layer be first-order error of $\mu_\mathrm{err}$ level (see Definition~5, Postulate~2, \S 2.2.2 / \S 2.4.3 first-order equality); after introducing Assumption~1, the finite-layer residue is infinitesimal and excluded as 0-measure (see \S 6.3.2 formula (ii) and the ``source of infinitesimal'' paragraph), hence for a.e.\ configuration, observation grid assignment $=$ truth grid assignment, so the observation hit ratio $=2/3$ (the lemma is expanded in (B.113)).

\paragraph{Exactness declaration (must be explicit, not omitable)}
The coefficient $2/3$ is an exact combinatorial constant under the general non-degenerate regime; under the global setting containing the zero-value saddle-point degeneration or finite-layer residue, it holds as an almost-everywhere (a.e.) / limit value, and recovers to the exact value after the 0-measure exclusion of Assumption~1. All references to $2/3$ in this paper take this convention, and must not be used as an unconditional finite exact constant. \emph{Marginal measure reverses predicate sign, categorical-support check, and diffusion instance.} The transformation from measure-space volume to marginal-measure summation (see B.107 reverse double integral $x(1-\ln x)$) contains a reverse equivalence-class symbol, anchoring the recovery reading to the 0 replica-index counting measure partition-predicate error---this is the closing point at the measure layer of the chain from the sole assumption (layered declaration, see B.3) to the engineering first-order error. The reverse-predicate-sign characterization, categorical-support check, dimensional check, threshold consistency, topological-residue operator $R$ and diffusion instance, the evaluation structure of Eq.~\eqref{eq:12b} (see C.5); the engineering-realization verification of Eq.~\eqref{eq:12b} (see B.116).

\subsubsection{Core theorem (see Theorem~7)}
\label{sec:6.3.3}

Synthesizing \S 6.2-\S 6.3, the contribution of this paper consolidates into one core Theorem~7: although the output is the elementary partition identity $\nu(\mathcal M)+\nu(\mathcal E)=1$, its irreplaceable point is at the reading layer: it continuously relaxes the discrete counting measure $\nu_{\mathrm{idx},n}$ into a Gaussian loss via Bochner--Minlos duality, and through \MPT\ / collection explicitly turns the out-of-domain error into a first-order coefficient $\mu_\mathrm{err}$ (taking $2/3$ as the direct reading of the globally counting-measure equipped case, contrasted with the $1{:}1$ $\mathbf{Quot}$ convention when unequipped). Standard single-space measure theory cannot complete this reading in the same domain because of the null-embedding ambiguity. The three-layer categorical chain and $\boldsymbol\pi_\#/\boldsymbol\pi^\#$ are the narrative language carrying this unified reading layer, rather than a device for deriving new measure theorems; among them, the explicitation of the out-of-domain error by Bochner--Minlos duality and \MPT\ is a substantive necessary structure.

\noindent\textbf{Theorem 7 (core theorem)}\label{thm:7}
Binary quotient-partition meta-categorical objects support the set-categorical functional along categorical support via measure descent--recovery ($\boldsymbol\pi_\#/\boldsymbol\pi^\#$), enumerating the morphism (non-bijection) itself rather than its preimage and object; their multiplicity is normalized through replica-index injection and accounted as an explicitly first-order error, and is closed at first order in engineering realization. Precisely:
\begin{itemize}
\item[(a)] \emph{Replica-index measure (mechanism)}: the multiplicity of non-bijection morphisms is normalized through injection by the replica-index measure $\nu_{\mathrm{idx},n}=2^{-n}\sum_i\delta_{B_i}$ (see Definition~17, \S 6.2.1);
\item[(b)] \emph{First-order error accounting (falsifiable consequent; reading)}: this multiplicity is exactly the inconsistent measure $\mu_\mathrm{err}$ (see Definition~5, \S 2.3.1) of the EDE (encoder-decoder-encoder) two-side predicates at the quotient-block layer---the volume error on the shared measure space constructed by the invertible \MPT; its ``first order'' is a structural semantics (the collapse of the binary quotient-partition bijection, the ``first-order'' order semantics of Definition~5), not taking an external small-parameter Taylor asymptotics. Because \MPT\ is a bijection, the volume left by exploration (the differential topology not covered by the morphism image) can only be carried by the object-collection layer ($\mathbf{Quot}$ quotient-block layer through $\mathsf{Sup}$ descent-recovery), and taking the domain directly to diagonal intersection loses the $2^{-n}$ scale factor and ignores this volume (see \S 1 null-embedding ambiguity)---this is why collection rather than domain is adopted. This volume error is explicitly read as the three factors of Eq.~\eqref{eq:12b} through the triple fiber-bundle integral (see Definition~16) (see Proposition~23b, \S 6.3.1).
\item[(c)] \emph{Three-channel closure (boundary)}: all three engineering channels stay at first order, generating no high-order error (see Proposition~23c, \S 6.3.3).
\end{itemize}

\paragraph{Clean separation of standard-result reuse and contribution}
The three clauses of Theorem~7 are the contribution of this paper (new propositions synthesized from ``standard measure-theory tools + this paper's constructions''); the block-counting measure uniqueness (see Lemma~1(i), Haar special case), no collapse $\Leftrightarrow$ quotient-set bijection (see Theorem~1), Gaussian-kernel uniqueness (see Proposition~21a, Schoenberg/Bochner/Maxwell), $\sigma$-additive $\Leftrightarrow$ trace-class (see Proposition~21b, Minlos/Gross), pushout--pullback and measure-preservation conservation, the $\varepsilon$-$\delta$ attribution of Postulates~1-4, and the standard conclusion of categorical probability theory (probability measures form a monad / Kleisli category~\cite{giry1982categorical}, random maps form a Markov category Stoch, pushforward well defined while pullback generally not) are all standard-result reuse (only adapted, no new-contribution claim). The contribution of this paper relative to Giry/Stoch (see \S 2.2.1): under the setting of ``multiple measures + well-defined pullback (bijective partition)'', explicitly reading as $\mu_\mathrm{err}$, an object with no place in the single-measure Kleisli/Markov framework.

\emph{Proof (self-contained, single-pass verifiable).} This proof depends only on the three objects already given in the main text: (i) the normalized block-counting measure $\nu_{\mathrm{idx},n}=2^{-n}\sum_i\delta_{B_i}$, whose canonical uniqueness is given by Lemma~1(i), Definition~17 (see \S 6.2.1); (ii) the predicate-inconsistency set $\mathcal E_\mathbf{Quot}$ (see Definition~5, \S 2.3.1); (iii) the quotient-block-layer support surjection and no-collapse $\Leftrightarrow$ bijection (see Definition~7, Theorem~1). \S 6.2-\S 6.4 and Appendix~B are the implementation checks of the above objects in each engineering channel, not entering this derivation chain, hence this proof can be read self-consistently in a single pass.
\begin{itemize}
\item[(a)] \emph{Mechanism (multiplicity injection normalization)}. The multiplicity of non-bijection morphisms is normalized through injection by $\nu_{\mathrm{idx},n}$ (see \S 2.3.1, Principle~4): one-to-many morphisms are injected one at a time (see Proposition~9a block-permutation invariance), and the number of times a non-bijection is repeatedly covered is absorbed by $2^{-n}$ into a first-order error reading.
\item[(b)] \emph{First-order error accounting (falling onto the partition identity of the counting measure)}. On the quotient-block set $\mathcal P_n$ (cardinality $2^n$), define the predicate matching set $\mathcal M_\mathbf{Quot}\triangleq\mathcal P_n\setminus\mathcal E_\mathbf{Quot}$ (i.e.\ the blocks where $T$ and $\mathcal T^\ast$ agree at the block level). By the construction of $\mathcal E_\mathbf{Quot}$ in Definition~5, $\mathcal M_\mathbf{Quot}$ and $\mathcal E_\mathbf{Quot}$ are pairwise disjoint and union to $\mathcal P_n$, forming a partition of $\mathcal P_n$. Hence for the normalized counting measure $\nu_{\mathrm{idx},n}$ we have \begin{equation}
\nu_{\mathrm{idx},n}(\mathcal M_\mathbf{Quot})+\nu_{\mathrm{idx},n}(\mathcal E_\mathbf{Quot})=\nu_{\mathrm{idx},n}(\mathcal P_n)=1
\end{equation}, i.e.\ the sum of the predicate-matching and predicate-mismatching counting measures is exactly the total count (normalized to $1$; unnormalized is $2^n$ blocks). The first-order error measure $\mu_\mathrm{err}$ through $\mathsf{Sup}^{\mathbf{Set}}_{\mathbf{Quot}}$ descent-recovery (see Definition~5) is positively read as the normalized counting measure on the mismatch set \begin{equation}
\mu_\mathrm{err}(A)=\nu_{\mathrm{idx},n}\big(\mathcal E_\mathbf{Quot}\cap\mathsf{Sup}^{-1}_{\mathbf{Quot}}(A)\big)
\end{equation}, hence the first-order error is exactly the counting measure of the ``mismatch'' portion in this partition---it is itself already a bounded, block-by-block enumerable explicit reading, and is then written as the three factors of Eq.~\eqref{eq:12b} through the triple fiber-bundle integral (see Definition~16) (see Proposition~23b).
\item[(c)] \emph{Three-channel closure (boundary)}. The three engineering channels introduce no high-order error (see Proposition~23c, \S 6.3.3): categorical leakage has been excluded (see Proposition~19), pure dimension-lifting does not endogenously generate (see Theorem~4), and intra-block error is actively ignored (the intra-block micro-level error is not within $\mu_\mathrm{err}$'s reading target, actively ignored by the engineering side, see \S 2.3.1 unified declaration ``intra-block error actively excluded''); topological collapse is an exogenous operation and does not constitute endogenous error of the Gaussian \MPT\ functional family (see Definition~6), hence the high-order error $\mu_\mathrm{high}$ has no generation mechanism (see Definition~6).
\end{itemize}
Synthesizing (a)(b)(c) gives Theorem~7. $\square$

\noindent\textbf{Proposition 23c (engineering side generates no high-order error)}\label{prop:23c}
Under the compatibility of the \S 4.3 conventions and $\boldsymbol\beta$ refinement, the above three items all introduce no high-order error; the existing exclusion chain of categorical leakage, pure dimension-lifting, and topological collapse jointly closes the remaining channels. The conclusion is limited to this paper's explicit realization: learnable variance, trainable coefficients, and path-dependent optimizers are all outside the assertion scope. \emph{Domain limitation (intra-block error actively ignored)}: intra-block micro-level error is not within $\mu_\mathrm{err}$'s reading target and is actively excluded by the engineering side (exclusion by convention, not structural resolution; see \S 2.3.1 unified declaration ``intra-block error actively excluded'' clause: the intra-block micro-level error objectively exists and its reading is visible, but it does not enter $\mu_\mathrm{err}$); the geometric residue of intra-block folding (see Lemma~3) is outside this assertion scope. Item-by-item proof (see \S 6.3.3) (c).

\paragraph{Disclosure and closure of circular argument}
The partition identity of posterior-supported-prior (b) depends on a premise---counting ``total count'' and ``match/mismatch'' must use the same $\nu_{\mathrm{idx},n}$, and this is exactly Definition~25a's ``epoch+1 counting measure prior and posterior'': the same counting measure simultaneously serves as posterior reading and prior equipment between adjacent epochs, hence formally it is a circle (the total count is given by the prior, and the match/mismatch is again counted by the same measure). But the only proposition bundled by this circle is ``there is exactly one saddle point''---i.e.\ the differential where zero lies is unique (see Theorem~6(i), not taking Premise~1 as premise); precisely because the saddle point is unique, the counting measure has a unique fixed reference, calibrating its prior and posterior to the same object, so the circle does not drift. Thus this circle is benign and closable: first independently prove the posterior (see Theorem~6 proving $\mathit S_0^\mathrm{Hil}=\{\boldsymbol 0\}$ is the unique zero-value saddle point), and the circle is anchored, obtaining the first-order error $\mu_\mathrm{err}$ without extra assumption---as long as the posterior (unique saddle point) is proven, the first-order error follows. The circle is not a defect, but the explicit self-statement of the closed chain ``unique saddle point $\leftrightarrow$ counting measure prior and posterior $\leftrightarrow$ first-order error''.

\paragraph{Note (minimality of assumptions)}
Assumption~1 constructs equivalence classes: the differential $\varepsilon$-$\delta$ via non-atomic + true partition gives equivalent-infinitesimal classes, and it is precisely these equivalence classes that make the continuous Gaussian first-order error readable as the $\ln$ form by integration---without equivalence classes, the differential $\varepsilon$-$\delta$ cannot induce this integral $\ln$ form. Thus the only structural assumption remains Assumption~1 (non-atomic + true partition); $\varepsilon$-$\delta$ is a standard analytic tool (output O3), carrying no assumption load, and the engineering realization additionally uses the registered conditions of \S 6.1(c): (a) taking the universal differential-measurable background as domain, not going through this assumption; (c) the three-channel closure mechanism itself introduces no new assumption, but the $\ln$ of its triple fiber-bundle-integral expression goes through this assumption; (b) the quantitative reading (Gaussian-family uniqueness, first-order-equality convention, the non-fitted status of Eq.~\eqref{eq:12b}'s coefficient) goes through Postulate~2, Postulate~4 depending on this assumption and the $\varepsilon$-$\delta$ tool. The square root of Eq.~\eqref{eq:12b} contains the direct-sum $\ell_2$ norm of the concatenated vector (not root-mean-square (RMS) synthesis, not an independent assumption).

\subsection{Random-walk variational descent: evolution of the Lebesgue-measure first-order error}
\label{sec:6.4}

This section constructs the descent direction of the training dynamics, independent of the closure of (Assumption~1) $\to$ \eqref{eq:12b}, only implementing the ``engineering-computable first-order output'' of core Theorem~7 (see \S 6.3.3, complementary to the reading closed loop, not a link of the core derivation chain).

\subsubsection{Endpoint-induced random walk and first-order error descent under the zero-counting constraint}
\label{sec:6.4.1}

The counting measure simultaneously possesses prior and posterior analyticity, while the Lebesgue-side single-point zeroing has no analytic expression, which is exactly where the first-order error lies under the shared measure (see Definition~5); a monotone functional induces an order homomorphism on the one-dimensional quotient-partition interval, and gradient orientation and measure dominance have interval-homotopy-deformation invariance.

\noindent\textbf{Definition 25a (endpoint-induced random walk)}\label{def:25a}
Let the block set of the $n$-th layer binary partition be $\mathcal P_n$, with its un-normalized block-counting measure $\nu_{M,n}$ (total amount $2^n$); a single normalization gives \begin{equation}
\bar\nu_{M,n}\triangleq\nu_{M,n}/2^n=2^{-n}\sum_{i=1}^{2^n}\delta_{B_i}=\nu_{\mathrm{idx},n}
\end{equation} (see Definition~17, i.e.\ the same notation as A.1, only this one normalization step). The random walk $W$ with $\bar\nu_{M,n}$ as the endpoint distribution is called the \emph{endpoint-induced random walk}: characterized solely by the endpoint $W_\infty\sim\bar\nu_{M,n}$, without specifying path, increment, or step size.

\paragraph{Why take endpoint rather than path}
The $\mathcal H_\infty$ layer has no point-wise translation invariance (see Proposition~21b), so the classical random-walk continuous-limit transition (Donsker type) is unavailable; taking the endpoint only needs the additivity and normalization of the counting measure, and the compatibility of each layer's endpoint distribution is given by the $\boldsymbol\beta$ refinement tower, with the discrete--continuous transformation being normalization of the counting measure rather than a limit on step size. Details (see C.22).

\noindent\textbf{Lemma 6a (first-order error descent under the zero-counting constraint)}\label{lem:6a}
Let the \S 4.3 convention and $\boldsymbol\beta$ refinement be compatible; under Condition~1 implementation-layer round condition (see B.4 / C.7): (i) the zero-class mass does not decrease; (ii) the epoch+1 prior-and-posterior circular characterization (saddle-point prior/posterior comparison, not parameter-trajectory monotone); (iii) the first-order error and integral control are homologous: $\mu_\mathrm{err}^{(k)}=c\,I_k$ ($c$ is the normalization factor of Principle~4, not a new parameter). Taking the endpoint distribution $\bar\nu_{M,n}$, define the domain-samplable engineering loss
\begin{equation}
\mathcal L_\mathrm{RW}(\theta;E)\triangleq\int_E\ell(\theta;\cdot)\,\mathrm d\bar\nu_{M,n}=2^{-n}\sum_{B_i\subseteq E}\ell(\theta;B_i)\label{eq:11}
\end{equation}
When $E=\Omega$ this is the expected loss $\mathbb E_{W\sim\bar\nu_{M,n}}[\ell(\theta;W)]$; Eq.~\eqref{eq:12b} calls its one-sided instance $E=\Omega_\varsigma$ through \eqref{eq:12a} (direction $\varsigma$ and integral domain see Eq.~\eqref{eq:11b}). Then:
\begin{enumerate}
\item $\mathcal L_\mathrm{RW}$ is block-wise finite average, and the one-sided instance takes the finite $d$-term per-dimension sum under the per-dimension marginal-sum convention (see Eq.~\eqref{eq:11c}, \eqref{eq:12a}), not inheriting the dimension curse of the full-dimensional Lebesgue joint integral [19,20,23];
\item $\bar\nu_{M,n}$ does not vary with $\theta$, hence the gradient contains no path term of ``distribution moving with parameter'';
\item The static destination is given by Theorem~6(i) (not taking Premise~1 as premise), and the dynamic surface is given by Condition~1---under the premise of not breaking the three-channel first-order error, the EDE matching-block mechanism gives the gradient orientation, and correct-direction counting dominance makes the residue integral high-probability decrease (expanded in B.108); the per-round monotone non-increase and convergence claims have been abandoned (sole authoritative declaration in the next clause E.7).
\end{enumerate}

\paragraph{Conclusion (random walk $\to$ binomial $\to$ two gradient directions $\to$ high-probability descent)}
The endpoint-induced random walk constrains the first-order Lebesgue error; under the \S 6.4.3 zero-residue binary mechanism the binomial gives two gradient directions, and the asymmetric Gaussian division makes the gradient point toward the Gaussian-vertex side with strictly dominant AUC, high-probability decreasing. The discrete--continuous transformation's Gaussian makes the gradient-direction midpoint $\mathit S_\mathrm{mid}$ (see Definition~25c) well defined by continuity. \S 6.4 only provides ``existence of a reading object to the zero-value saddle point and a high-probability descent path'', not guaranteeing per-round monotone or convergence; it is independent of the closure of (Assumption~1) $\to$ \eqref{eq:12b}, and Condition~1 characterizes the saddle-point prior/posterior circular comparison rather than parameter-trajectory monotone. Details (see C.7, E.7).

\emph{Proof.} (see B.109) (legality of citation see C.3).

\subsubsection{Existence of high-probability gradient-descent direction (decoder saddle point, one-sided integral domain, and samplable loss)}
\label{sec:6.4.2}

\noindent\textbf{Definition 25c (two-type saddle-point separation)}\label{def:25c}
The first-type saddle point (loss sample / generative midpoint) $x_0^{\mathrm{sad}}\triangleq\tfrac12(S_t^+ + S_t^-)=\mu_t$, is the saddle-point-side sample reading of Eq.~\eqref{eq:12b}, not a gradient direction; the training/generative saddle points lie in the unique Gaussian family of Proposition~21, with equal counting-measure values, hence their midpoint is exactly that Gaussian vertex. The second-type saddle point (where a gradient exists with high-probability descent direction) takes the midpoint of the first-type midpoint with the zero-value saddle point $\mathit S_0^\mathrm{Hil}=\{\boldsymbol0\}$:
\begin{equation}
\mathit S_\mathrm{mid}\triangleq\tfrac12\big(x_0^{\mathrm{sad}}+\mathit S_0^\mathrm{Hil}\big)=\tfrac12\,x_0^{\mathrm{sad}}.\label{eq:11a}
\end{equation}
($\mathit S_0^\mathrm{Euc}$ is a separately established concept, not participating in this formula; spatial attribution see B.110) $\mathit S_\mathrm{mid}$ is a definitional selection, and the coefficient $2/3$ is a structural reading under this selection (C.4). $\mathit S_\mathrm{mid}$ is a gradient-descent direction (existence), not a loss value, nor a (parameter-space) saddle point: this construction only needs to assign this direction and prove its high-probability descent. Details (see C.23).

\paragraph{Direction indication and integral domain}
\begin{equation}\label{eq:11b}\begin{gathered}
\varsigma\triangleq\mathrm{sign}(\bar S)\in\{+1,-1\}\\
\Omega_\varsigma\triangleq\{r:\varsigma r>0\}.
\end{gathered}\end{equation}
That is, $\Omega_{+1}=(0,+\infty)$, $\Omega_{-1}=(-\infty,0)$: the integral takes the zero-value saddle point (origin) as boundary, dividing into positive and negative directions.

\noindent\textbf{Lemma 6b (integral direction takes only the generative side)}\label{lem:6b}
$\mu_\mathrm{err}$ is the ``error partitioned by the integral boundary'' (see \S 2.3.1 domain clause unified declaration), and its domain is the quotient-block layer cut out by the $\boldsymbol\beta$ partition predicate; the intra-block micro-level error objectively exists and its reading is visible, but is not within this paper's first-order error reading target, actively ignored by the engineering side (not claiming it is $0$ or does not exist), and although the training-side raw data is not lifted/assembled it is still within the domain and assigned $0$---hence the integral mass can only come from the generative side $\bar S$, the integral domain takes $\Omega_\varsigma$ (see Eq.~\eqref{eq:11b}), and the direction $\varsigma$ is determined by which side of the zero-value saddle point the generative-side $\bar S$ falls on. \emph{Proof} (see B.111).

Thus the generative-side instance of \eqref{eq:11} is
\begin{equation}
\mathcal L_\mathrm{RW}(\theta;\Omega_\varsigma)=2^{-n}\sum_{i=1}^{d}\int_{\Omega_\varsigma^{(i)}}\varrho(r_i)\,\mathrm d\nu_M^{(i)}(r_i)\label{eq:11c}
\end{equation}
The residue density $\varrho$ takes the Gaussian kernel of the unique Gaussian family $\mathcal K_\Sigma$ (see Proposition~21); per-dimension notation and per-dimension marginal-sum convention (see B.107). \emph{Construction category: non-set-functional category.} The objects of Eqs.~\eqref{eq:11}/\eqref{eq:11c} are the equivalence-class symbols of the partition category rather than the point-wise mappings of the set-functional category; Euclidean dimension-lifting is an observation transformation rather than a mapping, and the zero-value saddle point is an anchoring relation rather than a self-map (see C.3, details in C.24). Conclusion (2) is the key benefit of this construction. Because the endpoint measure does not vary, this framework needs no score-function estimator (fiber-bundle reparameterization: random $\sigma$ is generated by the denoising diffusion probabilistic model (DDPM) reparameterization, Jacobian known, pushout well defined, no need to differentiate the sampling path), and no differentiable sampling process is needed. Details (see C.25).

Lemma~6a(3) only gives a descent direction, not a convergence rate; $\mathcal L_\mathrm{RW}$ is well defined without depending on trace-class (see Proposition~21b); ``descent direction'' does not imply a specific optimizer's convergence; this paper sets no independent Markov-chain (MC) loss $\mathcal L_\mathrm{MC}$ (uniformly called random walk). Details (see E.8).

\subsubsection{Binary mechanism of zero-residue points: gradient orientation, AUC dominance, and mechanism summary}
\label{sec:6.4.3}

The binomial (zero-residue binary) gives two gradient directions: when predicate matching holds the gradient terminates updating (zero direction); on mismatch it divides into a small-area region (functional extremum aligned with zero residue, correct gradient) and a large-area region (same side consistent, opposite side reversed). For a monotone functional, the same-side area is strictly larger than the opposite-side, and the global accumulated area is dominated by the counting measure, with the correctly-directed sample path having a higher counting replica-index.

\begin{figure}[H]
\centering
\includegraphics[width=156pt]{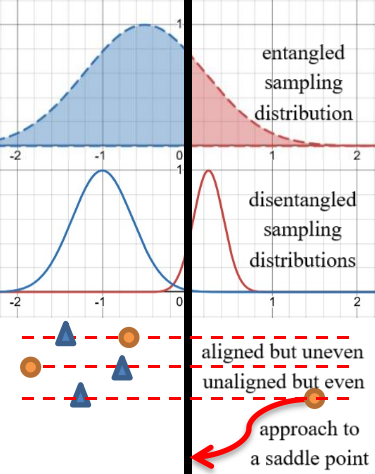}
\caption{Truncated Gaussian and reparameterized Gaussianization. Top: sampling truncated at the zero-value saddle point (one-sided integral domain); bottom: disentangled after Gaussianization; bottom band: aligned-but-non-uniform vs.\ non-aligned-but-uniform configurations, measured by the block-share/Lebesgue-share difference ($\Delta_I$, D).}
\label{fig:3}
\end{figure}

Asymmetric Gaussian division: the generative-side residue density takes the Gaussian kernel of $\mathcal K_\Sigma$ (see Proposition~21), divided at the zero-value saddle point (origin) into asymmetric two sides (the side containing the vertex has dominant mass): the gradient points toward the Gaussian vertex, and the predicate-matching side has strictly larger AUC than the reverse side---the mechanism of ``high-probability first-order-error-reducing gradient direction'' (see Figure~\ref{fig:3}). Under discrete sampling the counting frequency is a finite-sample estimate of the block share: block $B_I$ frequency converges to block share $\lvert I\rvert/2^n$ by measure, but does not converge to the Lebesgue integral area---their difference $\Delta_I$ is precisely the persistently remaining first-order error (D).

\paragraph{Mechanism summary}
This section provides, on the counting-normalized shared measure space, a high-probability first-order-error-reducing gradient direction (see Lemma~6a(3)); the parameter kinematics (whether the optimizer descends each iteration) is not handled by this paper, and the per-round monotone and convergence claims have been abandoned (see E.7).

Engineering factors can only enter the triple fiber-bundle integral by rewriting some layer's integrand or evaluation measure; the layer attribution of Eq.~\eqref{eq:12b}'s three factors has been given in \S 6.3.1. The per-channel verification is as follows; complete substitution derivation (see B.115-B.116).
\begin{enumerate}
\item \emph{Gradient descent} (see Definition~25a, Lemma~6a): the endpoint-induced random walk only rewrites the Lebesgue-side residue density, and the endpoint counting measure does not vary with the parameter, hence generates no distribution-path term; the conclusion is only that a first-order-error descent direction exists (high-probability decreasing, not guaranteeing per-round monotone non-increase).
\item \emph{Coefficient $\tfrac23$}: it is a structural reading after the global counting-measure lifting assembly, not a trainable parameter; its difference from the Lebesgue reading has been accounted into the same first-order error (B.116(c)).
\item \emph{Determined $\sigma$}: the scale specified by the diffusion-noise schedule and its Jacobian are known~\cite{sohl2015deep,ho2020denoising}, different from the random reparameterization trick used for gradient estimation~\cite{kingma2014auto}; the model only bears the residue-side first-order error.
\end{enumerate}

All three channels stay at first order, with closure (see Proposition~23c) (see end of \S 6.3.3). The setup, readings, and figures of the experiments are in D.


\section{Conclusion}
\label{sec:conclusion}

Category cancels elements $\to$ escape the null-measure paradox. The main thread (see \S 1) has established that the a.e.\ reading convention of standard measure theory hides out-of-domain error inside null support; only ``category cancels elements'' closes this gap and forces it out as an explicitly accountable $\merr$ (cancelling elements is not an abstract decoration but a lifting mechanism, while the classical $\Setcat$ volume is a derived corollary). We do not recapitulate this here, but only consolidate its conclusions---the core Theorem~7 (see \S 6.3.3) collapses ``nontrivial-input construction $\to$ engineering-computable first-order output'' into a single verifiable statement, whose validity depends solely on Assumption~1.

\paragraph{Results closure}
Structure layer: the surjection support is lifted by the replica-index measure to a quotient-layer bijection, and the shared measure resolves preimage mutual exclusion through an invertible \MPT; predicate matching supplies the bijection-evaluation caliber, and the replica-index multiplicity normalization supplies the first-order reading attribution (Principle~1). Reading layer: the zero-residue fact carries two readings---point-level counting uniqueness (carrying Premise~1) and differential-reading uniqueness (not presupposing Premise~1)---mutually recognized via Definition~19a (see Theorem~6(i), B.73); the Gaussian kernel is unique in finite dimensions, and bounded in infinite dimensions by the cylindrical-kernel compatibility family (see Proposition~21a, Proposition~21b). Engineering layer: the endpoint-induced random walk and the samplable loss (see \S 6.4) realize the training dynamics as a one-sided integral over the counting measure; Eq.~\eqref{eq:12b} is the explicit reading form of the first-order error (see B.117); the coefficient $2/3$ is a structural reading, not a fitted parameter; the three engineering factors stay at first order via per-channel verification (see Proposition~23c).

\paragraph{Assumption minimality (layered declaration)}
The paper's most central conclusion concerns the proof structure's \emph{assumption minimality (layered declaration)}: the sole backbone independent mathematical assumption is Assumption~1, which bears \emph{all Lebesgue-side quantitative readings}---first-order equality (Postulate~2), Gaussian-family uniqueness (Postulate~4), and the non-fitted status of Eq.~\eqref{eq:12b}'s coefficient; whereas the structure layer (quotient bijection, Theorem~1, Theorem~6(i)) takes the universal differential-measurable background as its domain and does not pass through that assumption. The engineering three-channel verification (see the ``introduces no new error channel'' closure of Proposition~23c) introduces no new assumption in its closure mechanism itself, yet its triple fiber-bundle integral expression $\ln$ passes through that assumption (see \S 2.4.3; B.3; B.4; B.23). This layering is verifiable ring by ring rather than a mere formal declaration: each link of the reading dependency chain $\text{\textcircled{1}}\to\text{\textcircled{2}}\to\text{\textcircled{6}}\to\text{\textcircled{7}}$ rests on an already-established conclusion or a registered condition, and each link of the structure/round chain $\text{\textcircled{3}}\text{\textcircled{4}}\text{\textcircled{5}}$ rests on a combinatorial-counting structural reading; Conventions~1--4, the implementation-layer round condition Condition~1, and the definitional midpoint selection are all carried per declaration and are not mixed into the backbone assumption. Consequently, ``the engineering side produces no high-order error'' is not an empirical observation but a conclusion of structural construction plus reading accounting---through Assumption~1 it acquires a universality reaching directly from the structure layer to the engineering layer; this derivation paradigm can also serve as a reference for testing whether the conclusions of other generative frameworks depend on hidden premises. The framework's irreplaceability lies not in any single device, but in unifying discrete counting, continuous Gaussian, and out-of-domain error under the same scale, and in \emph{characterizing predicate matching} with elementary finite integrals: the Bochner--Minlos continuousization and the explicit out-of-domain error are substantial necessary structures, while the three-layer categorical chain and $\boldsymbol\pi_\#/\boldsymbol\pi^\#$ merely carry this reading layer (see the abstract).

\paragraph{Universe boundary}
The above conclusions do not overstep their respective domains: set-theoretic existence does not entail SGD-reachability---yet, non-analytically, this paper constructs a descent direction of high-probability first-order error reduction, and the denial of reachability does not weaken the existence of this direction; the infinite-dimensional general case does not promise point-wise Gaussian measures; performance experiments do not replace proofs of uniqueness or of error order; modifications beyond this paper's explicit realization (learnable variance, trainable coefficients, path-dependent optimizers) lie outside the asserted scope. Details (see E.6).

\paragraph{Empirical readings}
The experiments provide three corresponding readings: enumeration-morphism compatibility (MNIST train/generation round-trip 1-nearest-neighbor (1NN) compatibility, test-set 1NN 97.51\%), structural matching under non-value-by-value replication, and CIFAR-10 same-structure learned saddle-point-gap improvement ($0.106$ vs.\ DDPM~\cite{ho2020denoising}'s $0.163$, relative improvement $34.97\%$), giving direct empirical support for the framework's observability, slack computability, and training-process monitorability. The theoretical explanation is consistent with the full framework: the training--generation pairing bias is not a matter of sample size or network capacity, but an intrinsic layered geometric effect arising when the coarse-grained quotient partition projects onto the finite-dimensional parameter analytic space---under Principle~1 this bias is localized as the difference between the partition-induced counting reading and the geometric measure reading, namely the first-order error $\merr$ (see Definition~5), explicitly accountable via Eq.~\eqref{eq:12b}. \emph{Precise caliber of the experimental assertion} (see D.1): under the same optimizer and the same variational tool, the framework that reads the first-order error as its reading attains a smaller first-order error---equivalently, more of the correct measure-space volume (multiplicity) is accounted, and more samples fall into the correct quotient block; (see \S 6.4 convergence rate and SGD reachability not claimed, E.6; descent mechanism).

\paragraph{Outlook}
The paper's backbone independent assumption and the outputs it affects are detailed in \S 2.4.3 and B.4; the complete proofs, the adaptations of standard theorems, and the technical derivations compressed by the main text are all in B. Future work should, within the same architectural boundary, examine different data modalities, give quantitative bounds on the error rates, and establish more precise correspondences between this paper's categorical support and the existing theories of optimal transport, flows, and diffusion.

\section*{Appendix A: Notation and Terminology Conventions}
\label{app:A}

This appendix lists the notation and terminology conventions shared across the whole paper, together with the symbols specific to Chapters~2--6: the notation conventions (see A.1) and the terminology conventions (easily-confused word pairs) (see A.2), both \emph{applicable throughout the whole paper}; chapter-specific symbols are used only within their own chapter.

\subsection*{A.1 Notation conventions}

The following lists the \emph{whole-paper-shared} notation and surface syntax; the symbols specific to Chapters~2--6 are in A.3--A.6.

\begin{itemize}
\item For a category $\mathcal C$: $\mathrm{Ob}(\mathcal C)$ is the object collection and $\mathrm{Mor}(\mathcal C)$ is the morphism collection; both belong to the meta-collection level, and the primitive category presupposes no derived instances (see Definition~1b, \S 2.2.2).
\item $(X,\mathcal F)$: a measurable space; $\mathcal F$ is a $\sigma$-algebra on $X$; $(X,\mathcal F)$ is not yet equipped with a measure. Two $\sigma$-finite measures $\mu_1,\mu_2$ serve only as constraint parameters for the subsequent fiber-bundle morphisms.
\item $S\subseteq X$: the discrete sample support set; $J$: the replica-index index set.
\item $\betat=\{\beta_n\}_{n\ge1}$: the differential--predicate bijection tower (see Lemma~4), fixed throughout the paper (the BatchNorm shift parameter is separately denoted $\beta_{\mathrm{BN}}$, Convention~1).
\item $\Sup(f,B)$: the generalized meta-categorical support; full definition (see \S 2.2.2).
\item $\nu_M$: the unit-element counting measure (unnormalized, at the $\Hinf$ level; the limit layer may be a composite measure of $\nu_{\mathrm{idx},\infty}$ and the unit-element counting, Corollary~2, output Postulate~2). The finite-layer normalized block-counting measure $\nu_{\mathrm{idx},n}$ (see Definition~17, \S 4.2) and the notation $\bar\nu_{M,n}$ (see Eq.~\eqref{eq:11}, \S 6.4.1) are \emph{two notations of the same object} ($\bar\nu_{M,n}=\nu_{\mathrm{idx},n}=2^{-n}\sum_i\delta_{B_i}$), not two measures, and there is no second step of re-normalizing an already-normalized measure.
\item $\Hinf$: the completion of the compatible Euclidean direct limit under the inner-product norm (see Theorem~3), the infinite-dimensional ideal model space; the uncompleted direct limit is generally only a pre-Hilbert space. $\Hfin$: the finite-dimensional realizable model space.
\item $\Delta:X\to X\times X$: the diagonal embedding, $\Delta(A)=\{(x,x)\mid x\in A\}$. Here $X\times X$ is only the pure-set Cartesian product, with no measure pre-equipped.
\item Quotient-block union operation: the unique in-category set operation of the $\Quot$ layer (see Proposition~2, B.13---the merge of two quotient partitions chooses the union of their block sets; intersection is realized by the topological lattice induced by the quotient partition). On mutually exclusive quotient blocks the replica-index measure satisfies block additivity $\nu_{\mathrm{idx}}(A\cup B)=\nu_{\mathrm{idx}}(A)+\nu_{\mathrm{idx}}(B)$; measure addition is the reading of the union of mutually exclusive support quotient blocks.
\item Assumption~1 and Postulates~1--4: the numbering convention for the unique extra mathematical assumption and its formal outputs---Assumption~1 (see \S 2.4.3) is denoted Assumption~1, and its four formal outputs are denoted Postulates~1--4, registered in the output formulas stated for Assumption~1 in \S 2.4.3; any argument invoking this assumption cites the input (Assumption~1) and the corresponding output (Postulates~1--4), and the outputs are not separately numbered as corollaries.
\item Division of labor among the homomorphic $F$-series notations (take by context, do not mix): $F(\nu; x, T(x), t)$ is the integrand functional (see \S 4.1, \S 6.2.3); $\boldsymbol F,\boldsymbol W$ are the tensor objects under the discrete-atom limit (see \S 6.2.3); $[F]$ is the quotient morphism (see \S 2.3.2); $\mathcal F_{\mathrm{blk}}$ is the block-pairing self-map family (see Definition~25b, \S 6.2.3); $\mathcal F_{\mathrm{enc}},\mathcal F_{\mathrm{dec}},\mathcal F_\theta$ are the encoder / decoder / noise-prediction maps (see \S 3.1, \S 6.3.3); the $\mathcal F$ in $(X,\mathcal F)$ is the $\sigma$-algebra (this section, standard set-theory usage).
\item Division of labor among the homomorphic $N$-series notations (take by context, do not mix): \begin{equation}
N(\theta)\triangleq\#\{x\in\mathcal X:\nabla_\theta\ell(\theta;x)=\boldsymbol 0\}
\end{equation} is the counting criterion (see Theorem~6(e), \S 4.3; \emph{point-level reading, whose uniqueness statement carries Premise~1}; the differential-reading uniqueness of the same fact is in Theorem~6(i), Definition~19a, and does not presuppose Premise~1); the integrand element of Eq.~\eqref{eq:11c} has been uniformly renamed to $\varrho(r_i)$---a residual-density reading taking the $\mathcal K_\Sigma$ Gaussian kernel (formal definition and canonical ontology in \S 6.4.2 ``canonical ontology and notation'')---and is a different object, level, and notation from the counting criterion $N(\theta)$, and the two must not be conflated; Eq.~\eqref{eq:12a} is its canonical continuous form---the reversed double integral $x(1-\ln x)$.
\item $\Setcat$ in this paper always means the \emph{restricted derived-set category} (see Definition~2b, \S 2.2.3: the locus of derived instances and functional readings obtained inductively via the canonical lifting functor $L'$), not the full class of the standard set category; the Hilbert differential topological structure it carries is denoted $\mathcal H^{\Setcat}$, Assumption~1 (\S 2.4.3).
\item $\mathit S_0$ is the zero-residue set (see Theorem~6, \S 4.3): the general reading is $\{\theta:\mathcal L(\theta)=0\}$, a non-empty Borel-measurable set containing $\boldsymbol0$; under the canonical realization of Conventions~1--4 its \emph{anchored reading} degenerates to $\{\boldsymbol0\}$. The paper's ``$\mathit S_0=\{\boldsymbol0\}$'' is used only in that canonical-realization context and is not propagated as a general reading; the training-set-side raw data has no separate symbol (see Lemma~6b, \S 6.4.2).
\end{itemize}

\subsubsection*{A.1.1 Master symbol--definition cross-reference table}

To avoid notation overload, the following table summarizes the principal symbols of the whole paper, their meanings, and their definition / first-appearance location (section numbers point to the main text; appendix-B entries are denoted B.x).

\begin{description}
\item[$\Quot$/$\MorTop$] binary quotient-partition category / differential-quotient (topological) category (\S 2.1--\S 2.2).
\item[$\Hinf$/$\Hfin$] infinite-dimensional ideal model space / finite-dimensional realizable space (Theorem~3, \S 2.2).
\item[$\betat$ (see Lemma~4)] binary iterative-refinement tower (\S 2.4).
\item[$\nu_{\mathrm{idx},n}$, $\bar\nu_{M,n}$] the $n$-th layer normalized block-counting measure (two notations, same object; Definition~17, \S 4.2; Eq.~\eqref{eq:11}, \S 6.4.1).
\item[$\nu_M$] unit-element counting measure (unnormalized, $\Hinf$ layer; \S 3.1, Corollary~2).
\item[$\merr$] first-order error (shared-measure-space volume error; Definition~5, \S 2.3.1).
\item[$\Queval$] the inconsistency set of the EDE's two-side predicates at the quotient-block layer (Definition~5).
\item[Assumption~1] the sole backbone assumption of the whole paper (non-atomic + true partition; \S 2.4.3).
\item[Postulates~1--4] the four formal outputs of Assumption~1 (\S 2.4.3).
\item[Corollary~10b] multi-measure preimage mutual-exclusion constraint (propagation of Corollary~10a(i); \S 2.4.2).
\item[Proposition~10] layered characterization of absolute continuity (\S 3.1).
\item[Theorem~1] quotient equivalence class $\Leftrightarrow$ element disjoint $\Leftrightarrow$ $\Setcat$ functional well defined (finite equal-cardinality special case: no collapse $\Leftrightarrow$ quotient-set bijection $\Leftrightarrow$ block-pairing bijection; \S 2.3.3).
\item[Theorem~6] existence of zero-residue points, counting uniqueness, differential-reading uniqueness (\S 4.3).
\item[Theorem~7] first-order error closed at first order (formal proposition; \S 6.3.3).
\item[$\mathit S_0$, $\mathit S_0^{\mathrm{Hil}}$, $\mathit S_0^{\mathrm{Euc}}$, $\mathit S_{\mathrm{mid}}$] zero-residue set / Hilbert zero-value saddle point (mid-layer integration bound) / Euclidean zero-residue point / gradient-descent-direction midpoint (distinguish from the second-class saddle point of Definition~25c and $x_0^{\mathrm{sad}}$; \S 4.3, \S 6.4.2).
\item[Condition~1] implementation-layer round measure-comparison condition (\S 6.4.1).
\item[$\mhigh$] high-order error (Definition~6, \S 2.3.1).
\item[$\mathfrak A_\varsigma$, $\Omega_\varsigma$] one-sided truncated-Gaussian area / one-sided integration domain (\S 6.4.2).
\item[$\mathcal L_{\mathrm{RW}}$, $\mathcal L_{\mathrm{top}}$] endpoint-induced random-walk loss / topological loss (\S 6.4.1, Proposition~23b).
\item[$\Gexcl$] additive-exclusion group (skeleton of cross-measure morphism support; B.83).
\item[Proposition~23a / Proposition~23b / 22c] first-order error caliber / topological loss / engineering-side high-order error closed at first order (\S 6.2.2, Proposition~23c).
\end{description}

\subsubsection*{A.1.2 Abbreviations and full names}

All abbreviations used in this paper, with full names (first occurrences in the main text also carry the expansion):

\begin{description}
\item[1NN / KNN] 1-nearest-neighbor / $k$-nearest-neighbor (structural-match evaluation indicator).
\item[ACC] clustering accuracy.
\item[AE / VAE] (variational) autoencoder.
\item[ALAE] adversarial latent autoencoder~\cite{pidhorskyi2020adversarial}.
\item[AUC] area under the curve (of the one-sided truncated-Gaussian reading, \S 6.4.2).
\item[BiGAN] bidirectional generative adversarial network~\cite{donahue2016adversarial}.
\item[BP] back-projection operator ($BP=\boldsymbol\pi^\#$, the measure-recovery channel; \S 3.2, A.5).
\item[CLT] central limit theorem (invoked only to be excluded; B.106).
\item[DDPM] denoising diffusion probabilistic model~\cite{ho2020denoising}.
\item[EDE] encoder--decoder--encoder (three-stage pipeline; \S 1).
\item[FID / IS] Fréchet Inception Distance / Inception Score (sample-quality metrics, not adopted as main readings; D.5.1).
\item[LCA] locally compact abelian (group).
\item[MC] Markov chain (loss notation $\mathcal L_\mathrm{MC}$; uniformly called random walk in this paper).
\item[MFCVAE] multi-facet clustering variational autoencoder~\cite{falck2021multi}.
\item[MPT] (invertible) measure-preserving transformation (the core transformation of this paper; \S 1, \S 2).
\item[MSE] mean squared error ($\lvert\hat\epsilon_t-\epsilon_t\rvert$; B.116, \S 6.4.1).
\item[NMM] neural mixture model~\cite{tissera2022neural}.
\item[PSD / PD] positive semidefinite / positive definite (matrix cone $\mathrm{PSD}_m$, $\mathrm{PD}_m$; C.2.2).
\item[RMS / RMSE] root mean square / root-mean-square error (D.2--D.3).
\item[RSD] relative standard deviation (D.3 calibre footnote).
\item[RW] random walk (loss notation $\mathcal L_\mathrm{RW}$; \S 6.4).
\item[SGD] stochastic gradient descent (reachability is explicitly not claimed; \S 1, E.6).
\item[SMT] sparse manifold transform~\cite{chen2023minimalistic}.
\end{description}

\subsection*{A.2 Terminology conventions (easily-confused word pairs)}

This paper uses the following eight groups of easily-confused word pairs (near-synonyms), whose distinctions are fixed here and not repeated later; the numbering (1)--(8) strictly follows the order of their first introduction in the main text (order: the three-word distinction of replica index $\to$ quotient block $\to$ reading $\to$ binary $\to$ block bijection $\to$ equivalence-class word family $\to$ partition-category object theory $\to$ predicate and predicate word).

\paragraph{(a) Index, replica, count}
The three words each denote one property, all carried by the single number of the replica-index counting measure $\nu_{\mathrm{idx},n}$; when two words co-occur they jointly use two properties. These three words are easily confused in the whole paper, yet each corresponds to an independent property, and all three readings are uniformly carried by the same numeric object---the replica-index counting measure $\nu_{\mathrm{idx},n}$ (see Definition~17, \S 4.2) with reading $\nu_{\mathrm{idx},n}(B)=2^{-n}$:
\begin{itemize}
\item \emph{Index (index)}: the enumeration position / distinct label within a replica fiber, serving computability---assigning each preimage in a one-to-many collection a distinct replica index, forcing uniform enumeration of the replica fiber (see Lemma~5g, \S 5.3, counting replica index $2^{-n}$), but this assignment only distributes indices and does not determine the attribution of samples to quotient blocks (B.7.7). Its property is ``enumeration position / computability''.
\item \emph{Replica / multiple record}: the multiple record of a one-to-many collection---a single topological class can carry arbitrarily many samples; the replica index is the element-reading count of that topological class's attribution, and the replica-index measure transforms this one-to-many collection into a bijection (see \S 1, B.7.7). Its property is ``multiplicity / one-to-many record''.
\item \emph{Count / counting measure}: the reading of the quotient-block counting measure---the block-counting measure $\nu_{\mathrm{idx},n}$ is uniquely determined by the partition structure under the transitive action of the block-bijection group, and is automatically measure-preserving for block bijections (see Lemma~1(i), \S 2.1 and \S 2.4), with constant reading $2^{-n}$; it is the homologous normalization baseline of the discrete (block-counting) and continuous (Lebesgue) readings (see Postulate~1, \S 2.4.3 output). Its property is ``block counting / normalized measurement''.
\end{itemize}
One number carries the three: the replica-index measure $\nu_{\mathrm{idx},n}$ is at once a counting measure (block-counting reading $2^{-n}$), records the replica index (multiple record), and, through the replica-index fiber (see Definition~16, \S 4.1, $\Omega_\nu$), supports uniform enumeration---the same reading $\nu_{\mathrm{idx},n}(B)$ simultaneously encodes these three properties, hence the three words are easily confused. \emph{Two words co-occurring means jointly using two properties}: in the paper, ``replica-index measure / replica-index--counting measure'' = replica (multiple record) + count (block counting); ``replica index / replica-index fiber / replica-index cardinality'' = replica (multiple record) + index (enumeration position); ``replica-counting dimension'' and the like are analogous; whenever two words co-occur, that location simultaneously requires two properties and one must not read only one.

Further distinctions already treated in dedicated passages: lifting-assembly versus lifting ($\Setcat$ is the top layer, and assembly toward it is uniformly called lifting-assembly), enumeration versus surjection (meta-level search completeness vs.\ morphism property, B.23), measure preservation versus pushforward invariance (measure preservation is uniformly written as the pushforward measure equality $\sigma_\#\mu=\mu$), and the four-names-one-measure of counting measures ($\nu_M$ unit-element counting unnormalized / $\nu_{\mathrm{idx},n}$ normalized block counting / $\nu_{\mathrm{idx}}$ replica-index measure / ``replica-index--counting measure'' ``replica-index--cardinality measure'' both refer to $\nu_{\mathrm{idx},n}$, the finite layer being $\nu_{\mathrm{idx},n}$, Definition~17, \S 4.2).

\paragraph{(b) Quotient block and quotient-predicate partition}
A \emph{quotient block} (block, $B_i$ or $X_i$) is an element (component) of the quotient-predicate partition---the minimal truth unit of a predicate word at a fixed tower layer, the atomic equivalence class at fixed resolution, indivisible; the $n$-th layer quotient-block set is $\mathcal P_n=\{B_1,\dots,B_{2^n}\}$. A \emph{quotient-predicate partition} (partition, $\mathcal P$) is the unique categorical object of the $\Quot$ layer (Terminology~(g)), given by the decision predicate of an equivalence binary topology, and is the overall cut consisting of quotient blocks (see Definition~7, \S 2.3.2, a disjoint-subset family covering all of $X$). The quotient block is an element (minimal subset) of that object, and the quotient-predicate partition is the full-set structure of quotient blocks; quotient blocks are named by predicate words (binary addresses) at the $\Quot$ layer, and their reading as subsets of $X$ is the characterization supported out after the categorical-support lifting-assembly to $\Setcat$. In this paper, ``partition'' / ``quotient partition'' by default means the quotient-predicate partition induced by the predicate structure.

\paragraph{(c) Reading and value}
A \emph{value} is the general term for real numbers at the $\Setcat$ layer (coordinates, parameters, measure values, differential values); a \emph{reading} is that subclass of values produced by the paper's constructed objects (measures, differential values, replica-index measures)---namely the \emph{evaluation result} of a constructed object as a functor, of the standard form ``object (observed object)'': $\mu(A)$, $\nu_{\mathrm{idx},n}(A)$, $\nu_{\mathrm{idx}}(\mathrm dx)$, $\#(\cdot)$. Functions and function values are read by standard mathematical convention: $\nu_{\mathrm{idx},n}$ is a measure (functor), and $\nu_{\mathrm{idx},n}(A)$ is its reading (value). Only readings enter the first-order accounting of $\merr$ and participate in first-order equality ($\mathrm dx\odot\nu_{\mathrm{idx}}\sim_1\mathrm d\#$, Corollary~2, output Postulate~2, \S 2.4.3): every first-order reading difference produced by lifting-assembly is charged to this measure and does not enter the high-order term (see Definition~5, \S 2.3.1); bare coordinates and parameter values are not called readings. Layered details are in ``where values reside'' (B.7.6).

\paragraph{(d) Two usages of ``binary''}
(a) \emph{Bivalence of a predicate}: a binary predicate cuts $X$ into a complementary pair of true/false classes (see Definition~2, \S 2.2.3)---the object of the $\Quot$ layer is precisely this quotient-predicate partition given by such a binary predicate (Terminology~(g)); (b) \emph{generation mode of a partition}: binary iterative refinement---the refinement is executed within the partition category, operating on the \emph{collection} of predicates (predicates are not computation): at each step a block's predicate is split into two disjoint non-empty sub-predicates, or is discarded (refinement abandoned) or kept unsplit, so selective refinement reaches any finite block count (e.g.\ split half of an $N=2$ partition again while leaving the other half unsplit, yielding $3$ blocks); \emph{when the collection acts synchronously on all blocks and nothing is discarded, the $n$-th round yields $2^n$ blocks}---$|\mathcal P_n|=2^n$ is the count of this \emph{global refinement tower}, while the predicate collection admits more general selective refinement (see Definition~7, \S 2.3.2). The two usages do not conflict: predicates are binary-valued, and the block-set size along the global refinement tower is a power of $2$; ``binary'' never restricts the block count to $2$. The ``binary quotient partition'' of Definition~7 is read under (b) as ``the finite quotient partition generated by binary iterative refinement''.

\paragraph{(e) Block bijection and quotient bijection}
A \emph{block bijection} is a permutation on the quotient-block set (see Lemma~1, \S 2.1, an element of the block-bijection group $G_n$); a \emph{quotient bijection} is the lifting morphism across $\Quot\to\MorTop$ (see Definition~3, \S 2.2.3). The former is an automorphism of the block set, the latter is an inter-layer morphism; they are two distinct objects (already stated in the final sentence of Definition~3).

\paragraph{(f) Quotient class / equivalence-class word family (what kind of quotient class, what kind of equivalence class)}
The word family ``quotient class / equivalence class'' in the whole paper each has its own referent; their carrier, construction relation, and locus level are fixed here centrally, and later occurrences of ``equivalence class'' are read locally by the chapter they appear in; any newly added usage must give the carrier and relation in the same sentence. The differential quotient of item (b) of this group here does not belong to the equivalence class in the sense of ``two different objects having the same measure'': the rank-1 minimal non-empty topological-differential subset $\mathcal D$ is the marker (representative) of a quotient block, and the differential-topological element $\mathrm dx$ forms a quotient by the relation ``carried within the same quotient block''; hence the object of the $\MorTop$ layer is the differential quotient, while (c)--(h) are equivalence classes at other layers or under other semantics.
\begin{itemize}
\item[(a)] \emph{Quotient block} (element of the quotient-predicate partition, carrying the differential at the $\MorTop$ layer): the minimal truth unit determined by the predicate word at a given tower layer, indivisible---the predicate word is defined as primary, and the equivalence class of points of the base $X$ under the relation ``same predicate word at the given tower layer'' is its $\Setcat$-layer reading (Terminology~(f)(a); the finite quotient-predicate partition is its block set, Definition~7, \S 2.3.2). The \emph{complementary equivalence-class pair} $(X_0,X_1)$ is the pair of the true/false sets of a binary predicate (the two classes of points of $X$ under the relation ``same predicate truth value'', Definition~2, \S 2.2.3); ``atomic equivalence class'', ``bi-partition equivalence-class object'', ``quotient-partition equivalence class'', and ``quotient equivalence class'' all refer to the quotient block.
\item[(b)] \emph{Differential quotient} (object of the $\MorTop$ layer): the quotient taken by the relation that the differential-topological element $\mathrm dx$ of $\Hinf$ is ``carried within the same quotient block marked by $\beta_n(\mathcal D)$'' (see Definition~1a, \S 2.2.2, $\mathrm dx\sim_\beta\mathcal D$); the rank-1 minimal non-empty topological-differential subset $\mathcal D$ is the marker (representative) of that quotient block, not an equivalence object on a par with $\mathrm dx$ (preface of this group). Its shorthand ``topological class'' in the class--element-correspondence context is in \S 2.1, \S 2.2.2, B.7.7; the formal definition of the $0$-differential quotient is in Definition~19a (\S 4.3), equivalent at the $\Hinf$ layer to set-category assembly (see \S 5.4).
\item[(c)] \emph{Element-point equivalence class} ($\Setcat$ layer): the class of $\Setcat$-layer element readings under the relation ``belonging to the same differential quotient'' (see \S 2.1 note ``point-wise differential source''; the interchange structure of Proposition~20, \S 5.2); a single topological class may correspond to arbitrarily many element-point equivalence classes, with no cardinality constraint.
\item[(d)] \emph{Numeric equivalence class}: the class of differential-quantity readings under the relation ``belonging to the counting-measure reading of one and the same rank-1 minimal non-empty set'' (see \S 5.2); ``one differential to one numeric equivalence class'' is precisely the special case where all quotient-block replica indices are $1$ (see \S 2.2.2 note (3) on class--element correspondence).
\item[(e)] \emph{Differential-volume equivalence class}: the class of volumes of an upper-bounded measure space under the relation ``equal Euclidean--Hilbert volume'' (see \S 5.2); the differential-topological preimage equivalence class is the class formed by the preimages of point elements under that volume structure (see \S 5.2).
\item[(f)] \emph{Parameter-degeneracy equivalence class} (see Theorem~6(f), \S 4.3): the class of parameters under the relation ``inducing the same input--output function''---once the first layer outputs identically zero, higher-layer parameters become indistinguishable at the function level; ``function equivalence class'' is its synonymous expression; the zero-residue equivalence class refers to the function equivalence class to which $\boldsymbol 0$ belongs (see Lemma~6a(3), \S 6.4.1), whose differential reading is the differential where zero lies (see Definition~19a: via the equivalence-class transformation $[\boldsymbol0]\mapsto\mathrm dx_0$ it falls into a unique differential quotient, and this reading does not presuppose Premise~1).
\item[(g)] \emph{Modulo equivalence class}: all uniqueness conclusions are quotients modulo a specified symmetry group---Theorem~1 modulo target-block label permutation, Theorem~2 modulo topological-group automorphism (Pontryagin-duality standard equivalence, C.2.4), Theorem~3 modulo the orthogonal group (see \S 3.2, the enumeration closing paragraph of Theorem~2); the self-map quotient-equivalence bijection equivalence class is precisely the class of Theorem~1's bijection modulo label permutation (see \S 2.3.3).
\item[(h)] \emph{EDE-category equivalence class}: i.e.\ the binary-quotient-partition equivalence class (quotient block, group (a) of this group)---the discrimination distance $D_{\Quot}$ of $\mathcal L_{\mathrm{cat}}$ acts between them (see \S 5.4).
\end{itemize}
Two further homomorphic usages not denoting equivalence classes: ``topological-class error'' (see \S 4.3, the error-category label in the set two-state, unrelated to the differential quotient) and ``formal equivalence'' (see Definition~20 note, the formal identity of categorical objects, not a class induced by an equivalence relation).

\paragraph{(g) Object theory of the partition category}
Within $\Quot$ (the partition category) there is only one categorical object: the \emph{quotient-predicate partition}---the overall cut given by the decision predicate of an equivalence binary topology (see Definition~2, \S 2.2.3). The predicate is the construction device inducing that object, and the quotient block is an element (component) of that object; neither is separately established as a categorical object; the $\Quot$-layer name of a quotient block is its predicate word, and its $\Setcat$-layer reading is a subset. All numeric characterizations---partition boundary, volume, midpoint, quotient-block granularity, and the like---are readings supported out after the categorical support is lifting-assembled to $\Setcat$ via $L'$ (see \S 2.1: the $\Quot$ layer has no point elements, no distance, no real numbers); when these characterizations are mentioned later, their level attribution is read per this clause.

\paragraph{(h) Predicate and predicate word}
A \emph{predicate} is the binary function inducing the quotient-predicate partition (the construction device of the $\Quot$ layer, not separately established as a categorical object, Terminology~(g)); a \emph{predicate word} is the binary address (finite string) of a predicate at some layer of the $\betat$ tower, and is the $\Quot$-layer name of a quotient block (element of the quotient-predicate partition). The predicate word is an accompanying description of attribution, not a criterion (see \S 2.2.2 note ``class--element correspondence has no one-to-one constraint'', B.7.7). The two calibers must not be confused: the predicate is what makes the decision and induces the quotient partition (morphism side, not separately a categorical object), while the predicate word is its binary address (partition side, the $\Quot$-layer name of a quotient block); in the $\Quot$ partition category there are only partitions (predicate words), no morphisms (predicates)---wherever ``partition object / quotient block'' is involved, use the predicate word, and wherever ``decision / induction'' is involved, use the predicate.

Further distinctions already treated in dedicated passages: lifting-assembly versus lifting ($\Setcat$ is the top layer, and assembly toward it is uniformly called lifting-assembly), enumeration versus surjection (meta-level search completeness vs.\ morphism property, B.23), measure preservation versus pushforward invariance (measure preservation is uniformly written as the pushforward measure equality $\sigma_\#\mu=\mu$), and the four-names-one-measure of counting measures ($\nu_M$ unit-element counting unnormalized / $\nu_{\mathrm{idx},n}$ normalized block counting / $\nu_{\mathrm{idx}}$ replica-index measure / ``replica-index--counting measure'' ``replica-index--cardinality measure'' both refer to $\nu_{\mathrm{idx},n}$, the finite layer being $\nu_{\mathrm{idx},n}$, Definition~17, \S 4.2).

\subsection*{A.3 Chapter-2-specific notation conventions}
\begin{itemize}
\item Three-layer category chain: $\Quot$ (see Definition~2, \S 2.2.3, the primitive quotient-partition category, binary-predicate layer) $\to\MorTop$ (see Definition~2a, the fiber-bundle morphism category) $\to\Setcat$ (see Definition~2b, the restricted derived-set category).
\item Category-side lifting--descent direction pair: the canonical lifting functor $L':\MorTop\to\Setcat$ (see Definition~2b) and the canonical descent functor $\Gamma:\Setcat\to\Quot$ (see Definition~2c), both (see \S 2.2.3).
\item $\mathcal D$: rank-1 minimal non-empty topological-differential subset, the marker of a minimal quotient block (see Definition~1a, \S 2.2.2).
\item $\tau$: the quotient bijection, the morphism on the $\Quot\to\MorTop$ lifting channel (see Definition~3, \S 2.2.3).
\item $G_n$: the block-bijection group, the permutation group on the quotient-block set (see Lemma~1, \S 2.1).
\end{itemize}

\subsection*{A.4 Chapter-4-specific notation conventions}
\begin{itemize}
\item $\Omega_\nu,\Omega_\lambda,\Omega_\mu$: the inner, middle, and outer fiber layers of the triple fiber-bundle integral (see Definition~16, \S 4.1).
\item \begin{equation}
\nu_{\mathrm{idx},n}\triangleq\frac1{2^n}\sum_{i=1}^{2^n}\delta_{B_i}
\end{equation}: the $n$-th layer normalized block-counting measure, $\nu_{\mathrm{idx},n}(X)=1$ (see Definition~17, \S 4.2; aliases ``replica-index--counting measure / replica-index--cardinality measure''); $\nu_{\mathrm{idx},\infty}$: the cylindrical-compatibility limit measure on $\mathcal C_\beta$ (see Definition~17).
\item \emph{Dimension-notation division of labor}: $m$ is the \emph{lifting finite multiplicity} (see Proposition~18, \S 4.2, the multiplicity of the lifting joint space $Z_m$, $m\ge1$), independent of the data / residual dimension $d$; $d$ is the \emph{residual (data) dimension} (see Eq.~\eqref{eq:11c}/\eqref{eq:12a}, the dimension summed dimension-wise, the dimension of $\R^d$), and the residual is $d$-dimensional. The two must not be confused.
\item $\mu_{\mathrm{gen}}$: the generative-side universal measure.
\item $\Pi:\Theta\to\mathsf{Pair}$: the EDE intermediate-layer predicate-pairing operator, taking values in a pairing lattice rather than a numeric domain.
\item Replica-index joint measure: the pairing joint of the replica-index measure and the base-space base measure (see \S 3.3).
\end{itemize}

\subsection*{A.5 Chapter-5-specific notation conventions}
\begin{itemize}
\item $\boldsymbol\pi_\#$: measure descent (fiber-bundle pushforward); $\boldsymbol\pi^\#$: measure recovery, $BP=\boldsymbol\pi^\#$ is the backward-projection operator (see Definition~24, \S 5.3).
\item \begin{equation}
\boldsymbol Q\triangleq\big(Q^{(1)}_{\gtrless\mu},Q^{(2)}_{\gtrless\mu},\dots,Q^{(k)}_{\gtrless\mu},\dots\big)
\end{equation}: the topological criterion queue, $Q^{(k)}_{\gtrless\mu}\in\{\mathsf{True},\mathsf{False}\}$ is a binary Boolean set predicate, not a real-valued measure function (see Definition~23, \S 5.3).
\item $\boldsymbol T_\otimes$: the tensor-product map.
\item $\Gexcl$: the additive-exclusion group at the layer level.
\item $G,\hat G$: the joint space and its dual.
\item $\mathrm d\#,\ \mathrm dx\odot\nu_{\mathrm{idx}}$: the counting differential and the differential value with replica-index measure (see Corollary~2, Postulate~2 output).
\end{itemize}

\subsection*{A.6 Chapter-6-specific notation conventions}
\begin{itemize}
\item Three functionalizable losses: $\mathcal L_{\mathrm{total}}$ (see Definition~26, \S 5.4), $\mathcal L_{\mathrm{RW}}$ (see Eq.~\eqref{eq:11}, \S 6.4.1), $\mathcal L_{\mathrm{top}}$ (see Proposition~23b, \S 6.3.1)---$\mathcal L_{\mathrm{eng}}$ (see Eq.~\eqref{eq:12b}) is the diffusion-computable realization of $\mathcal L_{\mathrm{top}}$ (three factors multiplied); the three do not form one and the same composite triple.
\item $\bar\nu_{M,n}=\nu_{\mathrm{idx},n}$: the normalized block-counting measure (see Definition~17, \S 4.2; aliases ``replica-index--counting measure / replica-index--cardinality measure''; its identity with $\nu_M$ is in A.1).
\item $\Omega_\varsigma$: the one-sided integration domain, chosen by direction marker $\varsigma\in\{+1,-1\}$.
\item $\varsigma$: the integration-direction marker: \begin{equation}
\varsigma\triangleq\mathrm{sign}(\bar S-\mathit S_0^{\mathrm{Hil}})=\mathrm{sign}(S_t^-)
\end{equation} (see Eq.~\eqref{eq:11b}, \S 6.4.2; $\mathit S_0^{\mathrm{Hil}}=\{\boldsymbol0\}$ is the origin, and the generative-side Gaussian is truncated at the origin into a one-sided integration domain).
\item $\mathit S_{\mathrm{mid}}$: the gradient-descent direction (\emph{second-class saddle point}, Definition~25c, Eq.~\eqref{eq:11a}), namely the midpoint of the ``first-class saddle-point midpoint $x_0^{\mathrm{sad}}$'' and the ``Hilbert zero-value saddle point $\mathit S_0^{\mathrm{Hil}}$'' (\emph{note the distinction from $x_0^{\mathrm{sad}}$}); Eq.~\eqref{eq:12b} reads out the saddle-point side with $x_0^{\mathrm{sad}}$, where $x_0^{\mathrm{sad}}\triangleq\tfrac12(S_t^++S_t^-)=\mu_t$ is the midpoint of the training saddle point and the generative saddle point (\emph{first-class saddle point: the sample of the loss function, the generative midpoint}).
\item $\bar\alpha_t,\beta_t$: the diffusion noise-schedule coefficients (coordinate/variance conversion of the forward kernel $\mu_t=\sqrt{\bar\alpha_t}\,x_0$ and $\sigma_t^2=1-\bar\alpha_t$), assigned by the noise schedule and being zero-error known quantities of the schedule; the concrete schedule form is in Experiment~D.5 and does not enter the factor list of Eq.~\eqref{eq:12b}, whose factors are the outer Lebesgue reading, the middle truncated-Gaussian area, and the inner counting measure---three items.
\item $R_t^{\mathrm{tr}},R_t^{\mathrm{gen}}$: training / generative residuals: \begin{equation}
R_t^{\mathrm{gen}}\triangleq\sigma_t(\hat\epsilon_t-\epsilon)=-R_t^{\mathrm{tr}}
\end{equation}; $\hat\epsilon_t,\epsilon$: predicted noise and true noise.
\item $S_t^+,S_t^-,S_t^{\mathrm{d}},\hat S$: training saddle point, generative saddle point (i.e.\ the outer Gaussian vertex, $S_t^-=\mu_t+R_t^{\mathrm{gen}}$), deviation saddle point ($S_t^{\mathrm{d}}\triangleq S_t^+-S_t^-=2R_t^{\mathrm{tr}}$), and semi-prior semi-posterior saddle point ($\hat S\triangleq\tfrac12(S_t^++S_t^{\mathrm{d}})$).
\item $x_0^{\mathrm{sad}}\triangleq\tfrac12(S_t^++S_t^-)=\mu_t$: \emph{first-class saddle point (loss-function sample, generative midpoint), i.e.\ the unique Gaussian vertex (see Definition~25c, Proposition~21)}, the saddle-point-side reading of Eq.~\eqref{eq:12b} (distinguish from the second-class saddle point $\mathit S_{\mathrm{mid}}$ in \S 6.4.2, Definition~25c).
\item $x^{\mathrm{gen}},x^{\mathrm{rec}}$: generative point and reconstruction point; $\hat x_0(z)\triangleq(z-\sigma_t\mathcal F_\theta(z,t))/\sqrt{\bar\alpha_t}$ is the $x_0$ reading recovered by $BP=\boldsymbol\pi^\#$, $\hat x_0^{\mathrm{gen}}=\hat x_0(x^{\mathrm{gen}})$.
\item $\mathfrak A_\varsigma\in(0,1]$: the normalized truncated-Gaussian Lebesgue area (dimensionless; under the standard semantics the marginal measure is the area); the logarithmic inner quantity of the second factor of Eq.~\eqref{eq:12b}, whose engineering reading is the one-sided truncated-Gaussian moment (B.116(b)).
\item $u_t,P_\varsigma,m_\varsigma,s_\varsigma$: $u_t\triangleq S_t^-/\sigma_t$; the normalized-area reading of the one-sided truncated Gaussian $P_\varsigma=\Phi(\varsigma u_t)$, the truncated mean $m_\varsigma$ and truncated standard deviation $s_\varsigma$ ($X_t\sim N(S_t^-,\sigma_t^2)$, $\varsigma=\mathrm{sign}(S_t^-)$)---all implementation-layer quantities; $P_\varsigma$ is the engineering reading of $\mathfrak A_\varsigma$ and is not an independent factor of Eq.~\eqref{eq:12b}.
\item $[\,\cdot\,]$: the equivalence-class symbol (marks an equivalence-class object, not an operator, no computation)---the marked object is formed via ``Hilbert descent $\to$ partition-category binary quotient partition (predicate matching) $\to$ recovery $\boldsymbol\pi^\#$ to a Euclidean object'' (see \S 6.3.1); $\nu_{\mathrm{idx},n}([\,\cdot\,])$ is the third factor of Eq.~\eqref{eq:12b} (the reading of the replica-index--counting measure on the equivalence class).
\item $\mathbb 1_{\mathrm{far}}$: the far/near criterion (far from the saddle point is generative, near is reconstruction-training), an implementation-layer gating, not a factor of Eq.~\eqref{eq:12b}; the code and experimental caliber are in D.5. $\mathbb 1_{\mathrm{obt}}$: the obtuse-angle criterion (generative and reconstruction lie on opposite sides of the saddle point), entering the factor list of Eq.~\eqref{eq:12b} as a direct sum, the double-capture (bijective) non-trivial self-map criterion (inducing the Fourier transform); see B.116(c).
\item $\mathrm{MSE}$ (mean squared error) $\triangleq\lvert\hat\epsilon_t-\epsilon_t\rvert$: the dimension-wise reconstruction error; $c^*=1.269860385419959$: the multivariate-Gaussian normalization correction coefficient of the training-period MSE lower bound $\max\{\mathcal L_{\mathrm{eng}},c^*\mathrm{MSE}^2(t+1)/T\}$ in the code (fixed, non-trainable, not a factor of the theoretical Eq.~\eqref{eq:12b}, see D.5.4).
\item $\bar S,\mu_{\bar S}$: the source-side reading and its threshold; the complete derivation of the diffusion instance (see C.5, \S 6.3.2 categorical-support check).
\end{itemize}

\section*{Appendix B: Spine Proofs -- Closing the Chain from the Unique Structural Assumption to Equation~\texorpdfstring{\eqref{eq:12b}}{(26)}}

This appendix collects the technical verifications and proofs cited in the main text. The entries B.1--C.7 are numbered in the order in which their objects or subjects appear in the main text (statements and proofs that pair up are listed adjacently), without grouping by content type---among them are early-migrated technical-verification proofs, standard-theorem adaptation checks (C.2, with its internal sub-numbering as shown there), technical notes (B.7), the convention on inter-changing conclusions over the partitioned-category support (C.3), and the full versions of the main text's propositions (sketches in the corresponding main-text positions). The corresponding main-text positions retain only the statement and a pointer ``proof in B.x''. Every proof in this appendix depends only on material already established earlier; no new premises are introduced. Numbered references (proposition / definition / section number) all refer to the main text and to earlier parts of the appendices.

\subsection*{B.1 The posterior--prior dual identity of the replica-index measure and the bijection of collection}

\textbf{Irreversibility of categorical support and the dual identity of the replica-index measure}
At the level of closure, support is still only a map to a collection (object collection); the categorical support is one-to-many and irreversible: staying at this level, surjectivity and invertibility of morphisms cannot be obtained simultaneously in one construction. The resolution rests on the dual identity of the replica-index measure $\nu_{\mathrm{idx}}$ (see Definition~17; \S 4.2): its posteriority lies in its accumulation from the posterior counting measure, which distinguishes each injection within a one-to-many relation, and the measure is nestable and inherits its injectivity; its priority and analytic invertibility lie in the same counting measure simultaneously serving, between adjacent epochs, as both posterior reading and prior equipment (see \S 6.2.2, the $\mathrm{epoch}+1$ counting measure)---the canonical specification of the counting-side measure is independently borne by the replica-index measure construction ($\nu_{\mathrm{idx},n}$ is the uniform counting measure on the $n$-th-layer quotient-block set, uniquely determined by the automorphism invariance of the block-bijection group; Lemma~1(i), Definition~17), and Fourier uniqueness here only supplies a conceptual prototype (see \S 3.2, C.2.4(iii)).

\noindent\textbf{Collection transformed into a bijection}
The dual identity of the replica-index measure makes the collection transformation a bijection: surjectivity is already guaranteed by Definition~1 (restricted-morphism surjectivity) and Proposition~3(ii) (enumeration surjectivity), while injectivity is missing---the way this topic is resolved is not to exclude one-to-many morphisms, but to distinguish each injection within the one-to-many relation one by one (see this section above). Thereupon all support morphisms are nested in the fiber bundle of the replica-index measure (the measure is nestable and inherits its injectivity); via pushout--pullback one obtains injectivity (Chapter~3) so that any nested map inherits the injectivity property; then, via binary-quotient partition, it is lifted to a bijection (Chapter~2).

\noindent\textbf{Cost: MPT-ization and first-order error}
The replica index must be measurable (non-null), so the support morphism is realized as a measure-preserving invertible transformation in the sense of a fiber bundle (see Definition~12; Definition~13; \S 3.1 MPT); Lebesgue measure, via MPT and counting measure, co-constructs the shared measure space; but when this measure is likewise required to be measurable, the differential shared measure makes element points ill-defined, and the construction falls into the element-free partition category (see \S 2.1)---non-null measure thereby turns into first-order error (see Definition~5; \S 2.3.1), and the inherent mismatch of training/generation thereby becomes measurable; this first-order error is generated by the surjection constraint across differential-topological measure spaces, and is unaffected by engineering-implementation perturbations.

\subsection*{B.2 Bijective maps do not preserve measure, quotient-translation-invariant measure preservation and integral non-measure-preservation have separate duties}

\textbf{Limitations of existing paradigms}
The root lies in the cross-domain equivalence conflict between set objects and functional descriptions. The dominant ``prior-support-posterior'' paradigm runs inside $\Setcat$ (i.e.\ \Setcat), pre-specifying a prior-support set; the \Setcat\ category does not track measure structure---a bijection need not preserve measure (measure preservation is not a categorical fact: the measure preservation of a \Setcat-layer must rely on an explicit measure assumption, Lemma~1(ii)), nor does the integral preserve measure (see \S 2.4, original-image differential translation invariance, while the differential translation of functional objects varies). The dominant paradigm lacks both, and can only indirectly construct cross-category support via random reparameterization; analytically measurable and unanalyzable measures cannot be coupled, so a morphism cannot simultaneously bear a surjection constraint and a subset-invertibility constraint, at the cost of losing morphism invertibility and letting pseudo-numerical categories intrude into the model interior (the accumulation mechanism of zero-measure-support error components, see Problem~2.25-2.26).

\noindent\textbf{Proposition 8 (bearer of measure preservation: quotient-translation invariance of the Hilbert differential; the integral does not preserve measure)}
The measure on the $\Quot/\MorTop$ sheaf is a partition-induced reading; its measure preservation is given by the Hilbert differential: mutually exclusive block covering guarantees each block is counted once, multi-to-one within a block is recorded by the replica index (see Theorem~1 boundary statement), and the block-bijection group acts transitively on the block set (see Lemma~1(i)), hence quotient-translation invariance preserves measure ($\sigma_\#\nu_{\mathrm{idx},n}=\nu_{\mathrm{idx},n}$); its point-wise consistency at the \Setcat\ sheaf presupposes Assumption~1 (see \S 3.2; Postulate~1 translation-invariance clause; output). Once a Lebesgue geometric measure is taken at the \Setcat\ sheaf, measure preservation must be explicitly assumed (see Lemma~1(ii)). \textbf{The integral, however, does not: it induces an analytic integral that does not preserve measure}---the original-image differential is translation invariant---on the pushout--pullback bijection $\boldsymbol\pi_\#$ and $\boldsymbol\pi^\#$ are mutually inverse pairings (see Definition~24; \S 5.3), and the differential quotient and its reading collected into quotient blocks via the $\betat$ tower are invariant under lifting and descent; but the differential translation of a functional object varies---the $\Hinf$ sheaf admits no non-trivial translation-invariant measure (see Proposition~21b; \S 5.3), and the Gaussian functional exists only as a cylindrical-kernel compatible family, so the integral side inherits no measure preservation; random reparameterization is exactly such an example (see \S 1). In summary: a bijection does not preserve measure (see Lemma~1(ii)); quotient-translation-invariant measure preservation is given by the Hilbert differential via the replica-index measure, while the integral does not preserve measure; the dominant paradigm lacks both, and measure mass escapes to zero-measure support (see \S 1); this paper's construction recovers quotient-translation-invariant measure preservation via categorical support (see \S 1).

\noindent\textbf{Proof (sketch)}
Mutually exclusive block covering guarantees a single count; multi-to-one within a block is recorded by the replica index; $G_n$-transitivity makes the uniform block-counting reading a Haar reading, and the pre-pushforward measure is invariant under pushforward by a block bijection; integral evaluation itself does not preserve measure. Full proof (see B.37).

\subsection*{B.3 Full derivation of Principle~1 and verification of the whole-text chain}

This entry gives the item-by-item derivation of Principle~1 in \S 2.4.1, and verifies that the chain from the unique assumption (Assumption~1) to the engineering first-order error (see Proposition~23c) closes. All material is previously established; no new premise is introduced.

\noindent\textbf{Derivation of Principle~1 (the Hilbert-differential definition of volume)}
The differential-topological volume of the partition category: Definition~1a gives the quotient relation $\mathrm dx\sim_\beta\mathcal D$, and the $\MorTop$ sheaf object is the differential quotient under this relation (Terminology~(f)(b)); its measure reading is $\nu_{\mathrm{idx},n}$ on the $n$-th layer quotient-block set of the $\betat$ tower (see Definition~17 finite layer) and $\nu_M$ (see Corollary~2 limit-layer composite reading). The measure reading of the volume of the differential measure space where set-category element points reside: the same differential-topological element, after canonical lifting functor $L'$ (see Definition~2b) lifts and equips it, gives the \Setcat-layer reading (Lebesgue side). The common defining material of both is the Hilbert differential: the $\betat$ tower (see Lemma~4: the construction uses only Boolean operations, independent of measure, translation, or coordinates, hence still holds on $\Hinf$) and the $\epsilon$-$\delta$ structure of $\mathcal H^{\Setcat}$ (Assumption~1). The differential topology in Hilbert space is therefore read as equivalence classes (differential quotients), not a pre-given point-set geometry. $\square$

\noindent\textbf{Derivation of Principle~2 (mutual equivalence classes)}
Quotient-sheaf translation invariance: Lemma~1(i) ($G_n$-Haar reading uniquely determines the block-counting measure, $\sigma_\#\nu_{\mathrm{idx},n}=\nu_{\mathrm{idx},n}$) and Proposition~8 (quotient-translation invariance preserves measure via the Hilbert differential). \Setcat-sheaf point-wise consistency: Postulate~1 (point-element measure-preserving reading, same differential across layers differs by at most first order). Lifting--descent passage: Definition~24 ($\boldsymbol\pi_\#,\boldsymbol\pi^\#$ are mutually inverse pairings on the pushout--pullback bijection), and the nature of the construction (see \S 4.1 built-in pullback--pushforward connection between layers). Equal reading: Corollary~2 (Postulate~2, first-order equality $\mathrm dx\odot\nu_\mathrm{idx}\sim_1\mathrm d\#$; the difference falls into $\merr$, Definition~5). Conclusion: the two volume readings are mutual equivalence classes under the same Hilbert differential---the strict reading of the differential-volume equivalence class (Terminology~(f)(e)); this equivalence does not require a Lebesgue-type measure to exist on $\Hinf$ (see \S 3.2; Proposition~21b, which states this measure does not exist). $\square$

\noindent\textbf{Derivation of Principle~3 (predicate-matching evaluation bijection)}
The bijection induced by binary-quotient partition (see Definition~3 quotient bijection) is adjudicated on the partitioned-category support by predicate matching: Theorem~1 (no collapse $\iff$ quotient-set bijection $\iff$ block-pairing bijection, modulo target-block label permutation); the last paragraph of Definition~1a (see Axiom~2: on the minimal quotient, either complete pairing or collapse---a binary decision with no intermediate state, the landing point in the partition category); Proposition~23a(ii) (predicate-mismatch area = full first-order error, not discounted half-and-half---its basis is the proof that a straddling quotient block must be a null-set: under the Hilbert differential topology each differential monopolizes a quadrant, straddling occurs only at a partition boundary, and for a Gaussian functional (not degenerating to one-hot) the partition boundary is null-measure, $\merr$ accumulates only non-null-support volume, so the discount term is identically zero, see \S 6.3.3 Theorem~7 proof (b) and B.104). Exclusion of point-layer degeneration ambiguity: ReLU dead zones and parameter redundancy are point-level counterexamples in the functional category; the final loss carries its own EDE predicate matching and belongs to the partitioned-predicate category (M3 localization of B.73); the definition-version premise $E\subseteq[\boldsymbol0]$ and its zero-residual classification condition $E\cap Z\subseteq[\boldsymbol0]$ are the conjunction of predicates (see \S 4.3). $\square$

\noindent\textbf{Derivation of Principle~4 (replica-number normalization)}
The dual identity of the replica-index measure (B.1): posteriority (accumulated from the posterior counting measure, distinguishing each injection within the one-to-many) and priority (see \S 6.2.2, the $\mathrm{epoch}+1$ counting measure). Multiplicity recording: multi-to-one within a block is recorded by the replica index (see Theorem~1 boundary statement); the non-bijection's repeated coverage of this differential topology is read as the prior-and-posterior replica-index measure's count of this differential quotient. Normalization: block share $2^{-n}$ (see Definition~17; Lemma~1(i)), and the Lebesgue side divides by the same total normalized volume $1$ (see \S 6.4.2 canonical ontology). Settlement export: the samplability of \eqref{eq:11}--\eqref{eq:12b} and the native loss; Lemma~6a(3) Condition~1(iii) controls constant $c$; the $\tfrac23$ cardinality reading (see \S 6.3.2). $\square$

\noindent\textbf{Whole-text chain verification (from the unique assumption to the engineering first-order error; Table~\ref{tab:spine-chain})}

\begingroup\small
\begin{longtable}{M{0.21\textwidth} M{0.21\textwidth} M{0.21\textwidth} M{0.21\textwidth}}
\caption{Chain verification of the spine from Assumption~1 to the engineering first-order error.}\label{tab:spine-chain}\\
\hline
Chain & Content & Basis & Assumption attribution \\
\hline\hline
\textcircled{1} & Assumption~1 $\to$ output Postulates~1--4 & \S 2.4.3 & unique mathematical assumption \\
\hline
\textcircled{2} & Postulate~1, Postulate~2 $+$ Lemma~1(i) $+$ Definition~24 $\to$ Principle~1, Principle~2 & this entry, Principle~1, Principle~2 & via Assumption~1 \\
\hline
\textcircled{3} & Principle~3 $+$ Theorem~1 $+$ Axiom~2 $\to$ predicate-matching evaluation bijection; point-/differential-reading divide & Theorem~6, Definition~19a, Premise~1 domain & Conventions~1--4 architectural; differential reading needs no Assumption~1 \\
\hline
\textcircled{4} & Principle~4 $+$ Theorem~6(i) $+$ Condition~1 $\to$ Lemma~6a(3) existence of a gradient direction decreasing first-order error (monotonic non-increase not guaranteed) & Lemma~6a, B.109 & Condition~1 is an implementation-layer per-round condition (registered in B.4), not a mathematical assumption; monotonicity claim abandoned \\
\hline
\textcircled{5} & Principle~3, Principle~4 $+$ Proposition~20 $+$ B.113 $\to$ $\tfrac23$ cardinality reading; three-point distinction (see Definition~25c) & \S 6.3.2, Definition~25c & midpoint selection is definitional selection (see Definition~25c; three points: boundary midpoint / subtracted saddle $=x_0^{\mathrm{sad}}$, integral boundary $=\mathit S_0^{\mathrm{Hil}}$, gradient direction $=\mathit S_\mathrm{mid}$) \\
\hline
\textcircled{6} & Principle~2 $+$ Proposition~21b $\to$ Gaussian cylindrical-kernel compatible-family boundary (trace-class) & \S 5.3 & via Postulate~4 \\
\hline
\textcircled{7} & reversed sign of the marginal-measure equivalence class (see \eqref{eq:12a} reverse double integral $x(1-\ln x)$, including the reverse equivalence-class sign) communicates measure descent $\boldsymbol\pi_\#$ and recovery $\boldsymbol\pi^\#$ & \S 6.2.1, \S 6.3.1, B.107 & reverse equivalence-class sign is Proposition~20's swapped reading, not a new assumption \\
\hline
\textcircled{8} & \textcircled{1}--\textcircled{7} $+$ three channels (see Proposition~23c) $\to$ engineering-side high-order error closed at first order & Proposition~23c, B.115 & conclusion limited to the explicit implementation of this paper \\
\hline
\end{longtable}
\endgroup

The table \textcircled{1}--\textcircled{8} mixes two kinds of chain links and must be stratified by the attribution of Assumption~1: the \emph{reading-dependency chain} is \textcircled{1}$\to$\textcircled{2}$\to$\textcircled{6}$\to$\textcircled{7} (the structural fact of Postulate~1 is not a new assumption; its ``at most first order'' reading content goes via Assumption~1; Postulate~2, Postulate~4 go via Assumption~1, and the Lebesgue-side quantitative reading of \eqref{eq:12b} and the non-fitted status of coefficients are borne by this chain); the \emph{structure/round chain} is \textcircled{3}\textcircled{4}\textcircled{5} (quotient bijection, Theorem~6(i), Condition~1, the $\tfrac23$ cardinality reading, all take the measurable space $\mathcal B$ equipped with the universal differential as domain, requiring no reference to Assumption~1, see B.23). Eq.~\eqref{eq:12b} sits at the junction of the two chains: its \emph{computable form} is evaluable by the finite sum of \textcircled{3}\textcircled{4}\textcircled{5}; its \emph{quantitative reading content} (the unique Gaussian family of factor \textcircled{3}, the first-order caliber of factor \textcircled{2}) depends on Assumption~1 via \textcircled{2}\textcircled{6}\textcircled{7}. Hence the accurate reading of ``the unique assumption closes'' is: \emph{Assumption~1 bears all Lebesgue-side quantitative readings; the structural-layer construction (quotient partition, enumeration equivalence) does not go through this assumption, while the engineering-side triple-fiber-bundle integral's expression $\ln$ depends on this assumption}---its Lebesgue-side $\ln$ reading (the middle-layer $(1-\ln\mathfrak A_\varsigma)$ normalized truncated-Gaussian area) is borne by output Postulates~1,~2,~4, and cannot be obtained without Assumption~1.

Judgment: every link of the chain \textcircled{1}$\to$\textcircled{8} is based on a previously established conclusion or a registered condition; apart from Assumption~1 in \textcircled{1}, no new mathematical assumption is added; Condition~1 (in \textcircled{4}) and the definitional midpoint selection (in \textcircled{5}) are attributed per declaration to implementation conditions and modeling freedom, and are not mixed into the backbone assumption. In summary: the four factors of \eqref{eq:12b} are derived from the triple-fiber-bundle integral of Definition~16 via the pushout--pullback chain (B.117); \emph{inside the square root is the direct-sum $\ell_2$ norm of the concatenated vector (B.117 addendum: definitional convention, not an independent/orthogonal assumption)}; the remaining factors all come from readings on the chain or variational assignments (cosine schedule, cancellation of the prefactor ratio, $2{:}1$ cardinality mixing, $1/1$ replica-index ratio), not free assumptions. $\square$

\subsection*{B.4 Which arguments depend on Assumption~1 (a checklist of what needs it and what does not)}

Arguments that need it (taking Assumption~1 as premise, all carried out on $\mathcal H^{\Setcat}$; formal output in \S 2.4.3 Postulates~1--4):
\begin{enumerate}
\item \textbf{Point-element measure-preserving reading} (see \S 2.1; \S 2.2.2; Postulate~1 output): the point-wise consistency of the \Setcat-sheaf point reading---the reading of the value where a given differential lies is invariant under different partition finenesses, differing by at most first order;
\item \textbf{Proof of Corollary~2} (see \S 2.2.2; Postulate~2 output): first-order equality of a differential value and any measure reading of a subset---$\epsilon$-$\delta$ continuity gives the functional single-valued correspondence of the reading form, and the differential value and any measure reading are the same equivalence infinitesimal;
\item \textbf{Real reading of the lifted layer and the limit argument} (see Postulate~3 continuity, completeness, point-wise neighborhood; output);
\item \textbf{Gaussian functional integral} (see \S 5.3; Proposition~21; Postulate~4 and 5.5b; output): uniqueness of the Gaussian family, the Gaussian kernel $\mathcal K_\Sigma$, and the evaluation and uniqueness characterization of the cited Minlos theorem / Gross abstract Wiener space results.
\end{enumerate}

Arguments that do not need it (each with its own supporting material, the text proceeds on the $\Quot/\MorTop$ sheaf or the finite-quotient-partition layer; each takes as domain parameter the measurable space $\mathcal B$ equipped with an arbitrary differential structure (the Hilbert differential topology)---Assumption~1 only concretizes $\mathcal B$ as $\mathcal H^{\Setcat}$, see B.23):
\begin{enumerate}
\item \textbf{Quotient-partition object structure} (see \S 2.2; Corollary~1a): existence of the combination of predicate words and quotient blocks, given by the predicate apparatus (see \S 2.2.4 first use of definition, above);
\item \textbf{Triple-fiber-bundle integral on the finite-quotient-partition layer} (see \S 4.1; Definition~16; \S 5.2; Proposition~20): on the finite truncation $Z_m$ it degenerates to the \textbf{finite sum} $\frac{1}{2^n}\sum_{B_i,B_j,B_k}f$, its material being the $\sigma$-finite Borel measure spaces of each layer (see Definition~17; \S 4.2), the measurability or integrability of the integrand (see \S 4.3 Convention~3), and the built-in pullback--pushforward connection between layers (see Construction~1);
\item \textbf{Existence of the zero-residue point in Theorem~6} (see \S 4.3): given independently of the $\epsilon$-$\delta$ form, by the uniqueness of the counting measure;
\item \textbf{``No new error channel introduced'' closure part of the engineering-side three-channel verification} (see Proposition~23c): on the finite layer it is directly given by the finite sum of (2) and the $\merr$ accounting; the closure mechanism itself introduces no new assumption; \textbf{but the engineering-side triple-fiber-bundle integral's expression $\ln$ (see \eqref{eq:12b}'s Lebesgue-side $\ln$ reading, the middle-layer $(1-\ln\mathfrak A_\varsigma)$ normalized truncated-Gaussian area) goes via Assumption~1}---this $\ln$ form is borne by output Postulates~1,~2,~4, and cannot be obtained without Assumption~1 (see B.117; B.3 table row \textcircled{8} and the stratification statement); the quantitative reading of first-order error (unique Gaussian family $\mathcal K_\Sigma$, first-order-equality caliber, non-fitted status of coefficients) likewise depends on Assumption~1 via Postulates~1,~2,~4. The two must not be confused;
\item \textbf{Quotient bijection and enumeration equivalence} (see Definition~3; Proposition~4 / 2.5; Proposition~6(i)): the one-to-one correspondence of predicate words and quotient blocks, and of $\MorTop$ objects, is directly given by the pure combinatorial fact of Lemma~4(1) (the one-to-one correspondence of binary trees and binary words, with no measure, addition, or coordinates appearing in the construction) and the assignment of predicate words in Definition~1a (see \S 2.2.2); the proof chain never references Assumption~1---the quotient bijection is not a corollary of this assumption, nor does this assumption take the quotient bijection as premise (B.23);
\item \textbf{Per-round monotonicity of Lemma~6a(3)} (see \S 6.4.1): given jointly by the measure-comparison Condition~1 ((i) zero-class mass non-decreasing, (ii) density and new-region control, (iii) first-order error controlled by integral) and the static reading of Theorem~6(i)'s predicate words; Condition~1 is the per-round condition of the EDE matching-block mechanism at the implementation layer, \textbf{not derived from Assumption~1, nor entering the backbone mathematical-assumption list}---its attribution is an implementation condition (see \S 4.3 theorem input implementation condition, same layer); static-destination uniqueness alone does not imply this monotonicity. The constant $c$ of Condition~1(iii) is the counting-measure normalization reading of Principle~4 (see \S 2.4.1), not a newly added free parameter;
\item \textbf{Principle~3, Principle~4} (see \S 2.4.1; B.3): predicate-matching evaluation bijection and replica-number multiplicity normalization are combinatorial-counting structural readings, not premised on $\epsilon$-$\delta$; the equal-volume readings of Principle~1, Principle~2 are premised on output Postulates~1,~2 (i.e.\ need Assumption~1), attributed in the ``needs'' items 1--2 above.
\end{enumerate}

\subsection*{B.5 Proof of Proposition~1}

\textbf{Proof}
(i) The two paths mutually recognize each other's construction: the \Setcat-side path gives point-wise differential readings from derivatively-instantiated pre-given measures and differentiable functionals, its reading form (functional single-valued correspondence) premised on Assumption~1 (see Corollary~2 proof (i), \S 2.2.2); the $\Quot/\MorTop$-side path partitions volume by the infinite-dimensional Hilbert predicate system, and the differential-topological element is assigned to a quotient block by the bearing relation to become a differential quotient (see Definition~1a; \S 2.2.2). The shared anchor of the two paths is the predicate-threshold topological numerical anchoring of $\mathit S_0=\{\boldsymbol 0\}$ (see Theorem~6; \S 4.3; B.7.8; \S 2.2.3), namely the differential $\mathrm dx_0$ where the value $0$ falls after the parameter-degeneration equivalence-class transformation (see Definition~19a; the uniqueness of this reading does not presuppose Premise~1, Theorem~6(i)); the \Setcat-side differential reading and the $\Quot$-side predicate-word reading fall, via this anchoring, into the same differential quotient, hence are the same differential rather than two differentials. (ii) The equivalence of the two readings: the threshold of the $\mathbb R$-field order reading and the threshold of the binary-partition membership reading are both the zero-value saddle point (see Lemma~5e; \S 5.3; Proposition~20; \S 5.2), hence the two readings are the presentation of the same predicate on two layers (see \S 6.4.3 predicate-matching criterion, operational definition). $\square$

\subsection*{B.6 The element-free partition category (definition and reading-level restriction)}

When a quotient block (or differential quotient) carries no point element under the \Setcat-layer reading---its partition predicate is not satisfied by any input, the quotient block is empty---the corresponding categorical reading is called an \textbf{element-free partition category}. Under this reading all objects are still the predicates and quotient blocks of the $\Quot/\MorTop$ sheaf, but are not lifted/equipped to \Setcat\ point elements; hence propositions in the element-free partition category are all quotient-sheaf statements, \textbf{not valid under the \Setcat-layer point-wise reading} (see \S 2.1 level-composition convention). In this paper, any proposition or corollary labeled ``(element-free partition-category reading)'' carries this level restriction in its statement: the conclusion may only be read as a quotient-block / predicate reading, not as a point-wise, coordinate, or numerical reading. The \Setcat-layer construction with no point-element bearer falls exactly into this category (see \S 1: when the differential shared measure makes element points ill-defined, the construction falls into the element-free partition category, i.e.\ this definition).

\subsection*{B.7 Technical notes (supplementary explanations)}

The following notes give supplementary explanations of the corresponding main-text conclusions; all mathematical content and the assumption set are unchanged.

\subsubsection*{B.7.1 The assumption status of differential equivalence}
Differential equivalence (see the quotient relation $\mathrm dx\sim_\beta\mathcal D$ of Definition~1a and the first-order equality $\mathrm dx\odot\nu_\mathrm{idx}\sim_1\mathrm d\#$ of Corollary~2---the differential value of an equipped subset $\mathcal D$ and the counting differential are equivalence infinitesimals, both first-order quantities) is not an extra assumption: Definition~1a is a definitional convention, and Corollary~2 is a corollary derived from Assumption~1 (see \S 2.4.3; its premise is declared once in Assumption~1), so differential equivalence as a whole is not an additional assumption independent of that unique assumption---still less is it a self-sufficient definition at the primitive level. The well-definedness of the equivalence class and the surjectivity of the categorical support are given by Proposition~3 (see \S 2.2.2); the content of first-order equality is accounted for by the $\merr$ of Definition~5 and the no-leakage exclusion support of Proposition~19 (see \S 5), its derivation given by Corollary~2's ``derivation'' block (see \S 2.2.2); the failure condition is explicitly delimited by Corollary~2's boundary statement (see \S 2.2.2). More importantly, the differential quotient is precisely the support form of point-wise differential on the $\Quot/\MorTop$ sheaf---the partition category sets no point elements, and the set-layer differential axioms (chain rule, coordinate differential, Gram/Jacobi determinant) must fall into a quotient block via $\mathrm dx\sim_\beta\mathcal D$ before use; first-order equality guarantees this support loses no binary-quotient-partition constraint. Hence the set-layer differential axioms can be borne by the partition category, and differential equivalence is not an extra assumption.

\subsubsection*{B.7.2 The partition-category criterion for neighborhood relations}
To judge, in the partition category, the neighborhood membership of three differentials $\mathrm dx_1,\mathrm dx_2,\mathrm dx_3$---whether the neighborhood of $\mathrm dx_1$ is another differential $\mathrm dx_2$, or the union of $\mathrm dx_2$ and $\mathrm dx_3$---one uses no points or topological balls, but judges by the predicate-word structure of the $\betat$ tower. Each differential is borne by its predicate word (binary address) (see \S 2.1); the ``neighborhood'' of $\mathrm dx_1$ is not a pre-existing ball, but \textbf{the minimal quotient block containing the common prefix of the relevant differential predicate words at a specified tower layer $m$}---the neighborhood is resolution-locked: on the \Setcat\ layer the same problem depends on the ball radius, while in the partition category the tower layer replaces the role of radius, and the problem is well-defined only after a layer $m$ is specified, and the tower provides the canonical sequence of layers. Let $k$ be the tower layer at which the predicate words of $\mathrm dx_2,\mathrm dx_3$ first separate, and $m$ be the layer of $\mathrm dx_1$'s neighborhood block (i.e.\ the quotient block where the length-$m$ common prefix of the predicate words of $\mathrm dx_1$ and $\mathrm dx_2$ resides): if $k\le m$, $\mathrm dx_2,\mathrm dx_3$ have already separated at the neighborhood-block layer or coarser, the neighborhood block contains no $\mathrm dx_3$, and the neighborhood \textbf{is} $\mathrm dx_2$; if $k>m$, the separation of $\mathrm dx_2,\mathrm dx_3$ is finer than the neighborhood-block layer, and at layer $m$ the two still belong to the same quotient block, so the neighborhood \textbf{is} $\mathrm dx_2\cup\mathrm dx_3$---this union is read on the $\Quot$ layer as the minimal common coarsening quotient block (a single quotient-block object), and ``union'' is a \Setcat-layer wording. This criterion and Corollary~10a's (see \S 2.4.2) ``contains no rank-1 non-empty subset'' criterion together with its differential mutual-exclusion condition (D1) are two sides of the same mechanism: two quadrants of $\Hinf$ share a low-dimensional boundary, their intersection is non-empty but not rank-1, so ``neighborhoods touch'' (sharing a low-dimensional boundary) does not constitute ``neighborhood merge''---the merge criterion is whether the intersection contains a rank-1 non-empty subset (equivalently, whether its $\nu_M$ reading is zero). All judgments use only the combinatorial structure of the tower (prefix comparison, common coarsening, transitivity), needing no points or numbers, satisfying the requirement of ``judging in the partition category''.

\subsubsection*{B.7.3 Three basic logics of the object-collection layer}
All categorical-support constructions in this paper depend on three object-collection-layer logics, here explicitly stated together with their sources.
\textbf{(a) Collection completeness $\implies$ findable}: $\mathrm{Ob}(\allowbreak\Quot)$ and $\mathrm{Ob}(\allowbreak\MorTop)$ are complete collections at the meta-collection level (see Definition~1 notation convention)---any object existing in the corresponding category must be within its object collection, with nothing outside the collection; by the object-collection bijection of induction step~1 (see Proposition~4; Proposition~5; \S 2.2.4), enumerating $\mathrm{Ob}(\allowbreak\Quot)$ is equivalent to enumerating $\mathrm{Ob}(\allowbreak\MorTop)$ (see Proposition~6(i); \S 2.2.4). Hence an existing bijection \textbf{must lie within the collection} (the finite layer $2^n$ blocks are enumerable; on $\mathbb R$ from integral boundary to $\pm\infty$ is searchable, and the integral boundary itself, being null, is ignored). The bijection falling into the enumerable domain is the operational object of search, not a claim of an efficient algorithm or actual enumeration of the infinite layer (see \S 5.1).

\noindent\textbf{On the boundary of ``findable'':} the ``findable'' of this clause means the existing bijection falls within the enumerable search domain (finite layer $2^n$ block enumeration; infinite layer borne by cylindrical-kernel compatible-family compatibility), \textbf{not a claim that an efficient algorithm actually retrieves it}, compatible with ``effective search on the infinite predicate tower remains an open algorithmic problem.'' The \textbf{continuity needed here does not come from Hilbert}: the $\Hinf$ layer has no point-wise translation-invariant (Lebesgue-type) measure and no general continuity (see Proposition~21b; \S 5.3); the continuity comes from \textbf{categorical support: Gaussian continuity}---the shared measure is induced by the Gaussian functional (see Proposition~21a/b cylindrical-kernel compatible family), and its absolute continuity is the structural basis of ``joint exclusion'' and ``enumerability'' (see B.82). If this continuity is used for gradient descent, this paper's approach is: each step takes the Euclidean saddle point and the zero-value saddle point, and gives a \textbf{probable gradient-descent direction} point by point (see \S 6.4.3 EDE matching-block mechanism gives gradient orientation)---but \textbf{does not guarantee the gradient surely descends}: just as gradient descent in discrete functionals does not guarantee point-wise descent, the Hilbert layer likewise cannot guarantee it (see \S 6.4.1; B.109 kinematics, which this paper does not handle conclusions on). ``Findable'' therefore bears only the weak commitment ``the bijection is in the search domain, and continuity is given by the Gaussian categorical support,'' implying no convergence or descent guarantee.
\textbf{(b) Binary partition induces invertibility = bijection = surjection + predicate injection}: under binary construction, quotient blocks and $\MorTop$ objects correspond one-to-one by predicate word (see Definition~3; \S 2.2.3); the assignment of quotient block to predicate word is a bijection (see Lemma~4(1); \S 2.4), hence injective; support surjectivity is given by the enumeration equivalence of the object-collection layer (see Proposition~3(ii); \S 2.2.2). On the finite layer, surjection and injection combine into a bijection---the quotient-layer modulo-label-permutation equivalence bijection is exactly the special case (d) of Theorem~1 (no collapse $\iff$ quotient-set bijection, see \S 2.3.3), so the quotient-layer invertibility induced by binary partition, the object-collection-layer bijection, and support surjection plus predicate injection are three mutually equivalent formulations.
\textbf{(c) Integral enumeration yields the collection}: the summation cover of the integral over the measurable spaces of each layer is exactly the object collection covered by the quotient-bijection surjection (see Corollary~2 proof (iii); output Postulate~2; \S 2.2.2); on the finite layer it is a per-block finite average (see Lemma~6a(1); \S 6.4.1). The product of integral enumeration is the object collection; outside the collection there is no integral mass.

\subsubsection*{B.7.4 Apparent causal inversion}
The derivation in \S 5.4 appears to have a causal inversion, an inherent feature of the posterior-enumerates-prior paradigm: one reverse-enumerates all preimages from the zero-residue point of the differential quotient inside the Hilbert partition-morphism category, rather than from the Euclidean-level zero saddle point; choosing the Euclidean zero saddle point would break the mutual-exclusion constraint of multi-measure preimages, and all points except that one still have mutually exclusive preimages.

\subsubsection*{B.7.5 Upgrade from assertion to corollary}
After introducing 5.5b (see \S 5.3 Minlos + trace-class + abstract Wiener space), the \S 6.2.1 conclusion---lifting/equipping to \Setcat\ makes the point-wise Gaussian morphism fail (point-wise kernel requires $\Sigma$ trace-class, in the general case only a cylindrical measure exists, and lifting/equipping loses $\sigma$-additivity)---becomes a direct corollary of that theorem (see Proposition~21b; \S 5.3 remark 2). Contrasting this failure boundary, the positive statement that abstract-level Gaussian morphisms, equipped in the set category, yield computable Gaussian numerical functionals comes from \textbf{Corollary~13} (see \S 5.4); its supporting material is Proposition~21a (finite truncation / trace-class case) rather than the general case of 5.5b; the two statements together form the two directions of the Proposition~21b boundary and do not contradict each other.

\subsubsection*{B.7.6 Where the numbers reside: partition-induction is the unique determination}
By the \S 2.1 level-composition convention, the $\Quot$/$\MorTop$ layers have no \textbf{point-wise} numerical concept (coordinates, point-wise measure values): all numerical presentation is a \Setcat-layer reading, residing in the derivatively-instantiated layer after lifting/equipping; that layer itself has only \textbf{structure-induced readings}---block share ($2^{-n}$), count, rank, replica-index measure reading ($\nu_{\mathrm{idx}}$ and its ``constantly 1'') as the unique determination of the underlying structure (quotient-block set + block-bijection automorphism group), without presuming \Setcat-layer geometry. Corollary~10a(D1)'s $\nu_M(\mathcal D_i\cap\mathcal D_j)=0$ is read out exactly this way: it is the presentation of a structure-induced reading at the \Setcat\ layer, used as an equivalent formulation of that layer's proposition (here only $\nu_M$ and not a Lebesgue reading may be used). Therefore the strict meaning of ``partition-induced'' in Lemma~1(i) is \emph{unique determination}: the underlying objects are merely the quotient-block set and its block-bijection (quotient-group translation) automorphism group; the reading on the \Setcat\ layer satisfying ``automorphism-invariant + full support'' is uniquely determined by this structure to be the block-counting measure. The uniqueness proof uses only the transitivity of the block-bijection group on the block set---a pure combinatorial fact, presuming no numbers and invoking no \Setcat-layer geometry. Hence there is no circularity here: not using \Setcat-layer numbers to prove an underlying property, but the underlying combinatorial structure uniquely determining the \Setcat-layer reading; throughout, phrases like ``structure-induced'' and ``automatically measure-preserving'' are to be read as ``uniquely determined by the underlying structure.'' Convention~2 (see \S 4.3)'s ``counting measure nests replica-index measure constantly 1'' is the same: constantly 1 is the identity of the nested reading (each original item is counted into its topological class (differential quotient, Terminology~(f)(b)) with replica number 1), its underlying content being that the original item is assigned to a quotient block by predicate, and the replica number is read out from the assignment count (see Definition~17; \S 4.2), imposing no one-to-one constraint between original item and quotient block. Corollary~2's (see \S 2.2.2) two sides ($\mathrm dx\odot\nu_\mathrm{idx}$ and $\mathrm d\#$) are also \Setcat-layer readings, whose $\Quot$-layer content is the same partition-structure equivalence under the same $\betat$ tower; the whole chain is ``$\Quot$ combinatorial structure $\to$ unique \Setcat\ reading $\to$ first-order attribution of the reading difference ($\merr$),'' with no step inferring an underlying property from \Setcat\ numbers. \textbf{The division of labor between pre-pushforward measure and invariant reading}: the pre-pushforward measure $\sigma_\#\mu$ (block bijection, Lemma~1(i)) and $\boldsymbol\pi_\#,\boldsymbol\pi^\#$ (see Definition~24; \S 5.3 measure descent--recovery) are \Setcat-layer measure morphisms, whose definition presumes measurable maps and reference measures---they \emph{transmit} measure but do not \emph{uniquely determine} it (the pre-pushforward measure $q_\#\lambda$ varies with the reference measure $\lambda$, and in general blocks are unequal); the reference-free uniqueness of the block-counting reading is given by the $G_n$-invariance (Haar uniqueness) of (i). Hence the complete mechanism of ``partition-induced'' is the two used together: invariance uniquely determines the reading at the structural layer, and pre-pushforward transmits it between layers; not treating pre-pushforward as the definition of the block-counting reading avoids the circularity of ``defining the underlying reading by a \Setcat-layer reference measure.''

\subsubsection*{B.7.7 Class--element correspondence has no one-to-one constraint}
If the class--element correspondence needed to be grounded by some adjudication (assignment or predicate matching), then the correspondence itself would become the premise of the adjudication, forming a circle; this paper's stance is that the correspondence has no one-to-one constraint, hence no need to ground the correspondence. (1) \textbf{Swap structure (see Proposition~20; \S 5.2)}: when a $\Quot$ object is transformed into a topological integral, the integral domain (topological) and the differential numerical value are swapped---one integral domain can be arbitrarily split into several differentials summed, and several differentials can be merged into the same topological domain; expressed as equivalence classes, one topological class (differential quotient, Terminology~(f)(b)) may correspond to any number of element-point equivalence classes (element readings as classes formed by belonging to the same differential quotient, Terminology~(f)(c)), with no cardinality constraint. The replica-index measure $\nu_{\mathrm{idx}}$ (see Definition~17; \S 4.2) gives the record: the replica index is the count of element readings belonging to that topological class; computability only forces uniform enumeration of the replica fiber (see Lemma~5g; \S 5.3 counting replica number $2^{-n}$), not that the replica number is 1. (2) \textbf{Multi-to-one within a block (see Theorem~1; \S 2.3.3 boundary statement)}: no collapse only constrains disjoint image sets across quotient blocks; the morphism inside a quotient block may still be multi-to-one; block-pairing-layer bijection and original-item-layer correspondence are decoupled. Hence ``one class to two elements'' is the structural norm: two element readings belong to the same topological class, each retaining its replica index and geometric position, and what is equivalent is the common bearing of the class layer; the two-level recording of block layer and original-item layer is exactly the accounting object of $\merr$ (see Definition~5; \S 2.3.1), and compressing to one level loses the error structure. (3) \textbf{Source and stance of one-to-one}: the one-to-one relation between a differential and the value where it lies essentially comes from $\epsilon$-$\delta$ continuity and real-value structure (functional single-valued correspondence); this paper's Convention~3 (see \S 4.3) has removed per-sample differentiability---measurability is placed on the differential quotient (see Definition~1a; \S 2.2.2), and the gradient is read from that quotient---hence it does not inherit this legacy, and the element reading's correspondence to the topological class is the multi-valued bearing under the swap structure. Assumption~1 (see \S 2.4.3) supports $\epsilon$-$\delta$; its implication is limited to the functional single-valued correspondence between a differential and the value where it lies on the $\mathcal H^{\Setcat}$/$\mathcal H^{\Setcat}$ / \MorTop/\Setcat\ layers, and does not derive a cardinality constraint of class--element correspondence on the $\Quot$ layer; the two layers are statements of different layers and do not cancel. This paper does not impose ``one differential to one numerical equivalence class'' (the differential-quantity reading as a class of counting-measure readings all belonging to the same rank-1 minimal non-empty set, Terminology~(f)(d); all quotient blocks have replica number 1; Proposition~7's condition $\mathrm{card}_\#(S)\le\mathrm{card}_\#(\mathcal P)$ is its special form, \S 2.3.3): if imposed, the cost is that the quotient-block identity is fully bound by the EDE two groups of predicates via the $\betat$-tower block-wise assignment; this cost is bounded by the differential in-compactness constraint (see Definition~4; \S 2.2.4; Proposition~18; \S 4.2's boundedness condition of the differential-sampling paradigm), at most losing per-point enumerability (\Setcat-layer point-wise finite-parameter analytic reachability---Definition~4 second existence; analyticity of 2.23.3 (see \S 2.4.2); Convention~1 (see \S 4.3) Lebesgue-sense reachability), while differential enumerability is fully preserved (see Proposition~6; \S 2.2.4; quotient bijection, Definition~3; \S 2.2.3; quotient-layer bijection, Theorem~1; \S 2.3.3; computability forces uniform enumeration, Lemma~5g; \S 5.3)---per-point enumerability was never the claim object of the primitive category (see \S 2.2.4 section preamble). (4) \textbf{Neither predicate word nor neighborhood constitutes a membership criterion}: the predicate word is the internal address of a topological class ($\Quot$/$\MorTop$ side), an accompanying description of membership rather than its criterion---several element readings belong to the same topological class, and after belonging the class address is read out; whether two original items that are mutual neighbors belong to the same class is read only from the belonging structure (the level criterion is in B.7.2's ``neighborhood-relation partition-category criterion'' note), and the neighborhood relation itself constitutes neither a merge nor a distinction; boundary-touching (sharing a low-dimensional boundary) is handled by Corollary~10a's (see \S 2.4.2) rank-1 criterion, and elements on the shared low-dimensional boundary fall into no differential quotient (see the criterion after Definition~1a; \S 2.2.2). Hence the class--element correspondence has no circle: the swap structure (see Proposition~20) + multi-to-one within a block (see Theorem~1 boundary statement) + replica-number record (see Definition~17) give the structural status of ``one class to any number of elements,'' needing no correspondence grounding and no measure ($\Hinf$ layer has no global translation-invariant (Lebesgue-type) measure, \S 3.2) nor a downward-exploring bijection to the underlying layer (enumeration and surjectivity, Proposition~3(ii); \S 2.2.2, characterize the search completeness of the categorical support, a different issue from correspondence). From the position in the motivation chain (see \S 1), the class--element correspondence having no one-to-one constraint is the structural basis of the categorical-support one-to-many---the collection does not require classes and elements to correspond distinctly, so morphisms can be one-to-many; the dual identity of the replica-index measure then transforms this one-to-many collection into a bijection, and the measure-ization of the replica index's multiple records of elements is exactly the basis of this transformation.

\subsubsection*{B.7.8 Can lift then descend, cannot descend then lift}
Categorical support can lift then descend, but cannot descend then lift, so there is no homology, homotopy, or any non-categorical analytic means between $L':\MorTop\to\Setcat$ and $\Gamma:\Setcat\to\Quot$, because their categorical objects are completely different: the actual lift is the Euclidean dimension-raising of the numerical-element category (\Setcat\ layer) after descent, while the descent is the zero-value saddle point's zero-error anchoring of the partition predicate, and the first-order error is entirely borne by the Lebesgue measure.

\subsubsection*{B.7.9 Source of the differential and geometric interpretation of lift--descent directions}
This note retains the technical explanation before the main text's compression, adding no premise.

\noindent\textbf{Source of the point-wise differential}
The point-wise reading is derived on the $\MorTop$ layer via the differential--predicate bijection tower $\betat$ (see Lemma~4; \S 2.4) through the differential-quotient relation (see Definition~1a), lifted to \Setcat\ via first-order equality (see Corollary~2) and support morphism-ization (see Proposition~3)---point elements are defined by the differential; the class--element correspondence has no one-to-one constraint (B.7.7).

\noindent\textbf{Proposition 1 (the two-path construction of the differential and the double characterization of the E/D pair bijection)}
The differential is constructed along two paths and anchored at the zero-value saddle point as the shared predicate: on the \Setcat\ layer it is given by differentiable functionals; in the element-free partition category it is given by the Hilbert partition of the $\Quot/\MorTop$ layer---a measure space with an upper bound is partitioned by the infinite-dimensional predicate system into volume, and the differential-topological element $\mathrm dx$ is assigned to a quotient block by the bearing relation to become a differential quotient (see Definition~1a); the two paths mutually recognize, via the equivalence-class transformation, the same differential at $\mathit S_0=\{\boldsymbol 0\}$ (see Theorem~6; \S 4.3) (see Definition~19a, the differential $\mathrm dx_0$ where zero lies; the uniqueness of this reading does not presuppose Premise~1). The EDE-pair bijection has one reading on each of the two layers: on the \Setcat\ layer it is characterized by the magnitude order of real-valued functionals (see Lemma~5e; \S 5.3), and in the partition category by complementary-quotient-block membership (see Proposition~20; \S 5.2); the two correspond via the same anchor point.

\noindent\textbf{Proof (sketch)}
The two paths give consistent readings at the quotient-block level; the two-layer readings of the EDE-pair bijection share the same source, and the anchor is unique. Full proof (see B.5).

\noindent\textbf{Categorical-direction convention}
Along the categorical chain $\Quot\to\MorTop\to\Setcat$, ``upward'' means: the production of observable quantities such as points, elements, coordinates, and numerical values is uniformly called \textbf{lifting}; the reverse is called \textbf{downward}: the goal of \textbf{measure descent} (pushforward $\boldsymbol\pi_\#$) is the partition category $\Quot$; the upward reading of numerical values and points is called \textbf{lifting recovery} (pullback $\boldsymbol\pi^\#$, instantiated as $BP(\cdot)$).

\noindent\textbf{Differential--geometric basis of direction unification}
The lift--descent direction originates from the \textbf{differential}: a measure space with an upper bound (see Proposition~18; \S 4.2) is partitioned by the Hilbert predicate system into volume to obtain \textbf{differential-topological volume} (Terminology~(f)(e)), presented on the $\MorTop$ layer via the differential-quotient relation (see Definition~1a); all directional concepts are unified by the \textbf{Euclidean dimension-raising} as the geometric prototype---it is the only topological dimension-raising operation that iteratively reaches Hilbert space (see \S 3.2), infinite Euclidean dimension-raising lifts the object category into the differential-topological category (see Lemma~5g; \S 5.3), and the pure dimension-raising layer of $\Hinf$ generates no high-order error endogenously (see Theorem~4; \S 4.2); the dual is \textbf{Euclidean dimension-lowering} corresponding to categorical descent ($\Gamma$) and measure descent, and the Lebesgue-side first-order error $\merr$ is the inherent gap on the dimension-raising--lowering passage (see \S 5.2). The resulting direction conventions are listed in Table~\ref{tab:direction}.

\begin{longtable}{M{0.14\textwidth} M{0.17\textwidth} M{0.19\textwidth} M{0.19\textwidth} M{0.19\textwidth}}
\caption{Categorical direction convention.}\label{tab:direction}\\
\hline
Direction & Geometric key & Categorical motion & Measure motion & Arrival layer \\
\hline\hline
To element (up) & Euclidean raise & lifting, lift-equip ($L':\MorTop\to\allowbreak\Setcat$) & lifting recovery $\boldsymbol\pi^\#$ ($BP(\cdot)$, read numbers/points) & \Setcat\ point element \\
\hline
To partition (down) & Euclidean lower & categorical descent ($\Gamma:\Setcat\to\Quot$, canonical descent functor) & measure descent $\boldsymbol\pi_\#$ (pushforward) & $\Quot$ quotient block \\
\hline
\end{longtable}

\subsection*{B.8 Proof of Lemma~1}

\textbf{Proof}
(i) Automorphism invariance is the pre-pushforward equality $\sigma_\#\mu=\mu$: under the transitive action of $G_n$, the reading of any block is sent to any other block with unchanged value, hence all block readings are equal; full support guarantees this common reading is positive, and normalization (total reading 1) presses it to $2^{-n}$, i.e.\ $\mu=\nu_{\mathrm{idx},n}$. Under pre-pushforward, the measure preservation of a block bijection is automatic: $\sigma_\#\nu_{\mathrm{idx},n}=\nu_{\mathrm{idx},n}$ (a block bijection is a block-set permutation, and pre-pushforward sends the uniform reading to itself). (ii) Take $X=[0,1]$, $\mu=\lambda$, $T:X\to X$, $T(x)=x^2$: $T$ is a measurable bijection (strictly increasing and surjective), yet $\lambda(T[0,1/2])/\lambda([0,1/2])=\lambda([0,1/4])/\lambda([0,1/2])=1/2\ne1$. The gap between (i) and (ii)---block counting (measure preservation automatic, geometric information flattened) versus geometric measure (measure preservation must be assumed)---is exactly this paper's first-order error $\merr$ (see Definition~5; \S 2.3.1). $\square$

\subsection*{B.9 Corollary~1 (well-foundedness of binary predicates)}

\textbf{Corollary 1 (well-foundedness of binary predicates)}
A binary topological-decision predicate is equivalent to a topological quotient-partition object (predicate and quotient partition correspond one-to-one), giving $\mathrm{Ob}(\allowbreak\Quot)$ definite membership, and this correspondence depends on no derived structure---without well-founded primitive objects, categorical induction cannot start.

\noindent\textbf{Proof (sketch)}
The true set / false set of a binary predicate gives a complementary equivalence-class pair, i.e.\ a quotient-partition object; conversely a quotient partition gives a binary predicate via the indicator function. The two directions are mutually inverse, hence binary topological-decision predicates and quotient-partition objects correspond one-to-one. Full proof (see B.10).

\subsection*{B.10 Proof of Corollary~1}

\textbf{Proof}
By Axiom~2 (binary-decidability axiom), any formalizable topological proposition has a binary-decision semantics (true / false). Given a binary topological-decision predicate $P$, its true set and false set constitute a complementary equivalence-class pair of $X$, i.e.\ a binary topological quotient-partition object (see Definition~2; \S 2.2.3); conversely, any quotient-partition object gives a binary predicate via its complementary equivalence-class pair $(X_0,X_1)$ through the indicator function $\mathbf 1_{X_0}$. The two directions are mutually inverse, hence binary topological-decision predicates and topological quotient-partition objects correspond one-to-one, forming the primitive object $\mathrm{Ob}(\allowbreak\Quot)$ of the system. This corollary gives the well-foundedness of the primitive object: the correspondence of predicate and quotient partition depends on no derived structure (see Definition~2; \S 2.2.3; Corollary~1a; \S 2.2.2). $\square$

\subsection*{B.11 Corollary~1a (satisfiability without derived instances)}

\textbf{Corollary 1a (satisfiability without derived instances)}
The existence of a quotient-partition object is independent of derived instances, metric, and coordinate structure: if its existence depended on \Setcat-layer derived instances, all constructions would be anchored inside the set category, degenerating into the classical ``prior-support-posterior'' paradigm. This decoupling lets the proposition-existence statement precede the \Setcat-layer equipment (see Definition~4; \S 2.2.4 first existence).

\noindent\textbf{Proof (sketch)}
The definition of a quotient-partition object involves only the base set and a complementary equivalence-class pair, referencing no derived instance, coordinate, or metric; by the \S 2.1 level-composition convention, point elements, coordinates, and metrics all belong to the \Setcat-layer derived-instance level, hence its existence is decoupled from derived structure. Full proof (see B.12).

\subsection*{B.12 Proof of Corollary~1a}

\textbf{Proof}
By Corollary~1, quotient-partition objects and binary topological-decision predicates correspond one-to-one, and its definition involves only the base set $X$ and a complementary equivalence-class pair (see Definition~2; \S 2.2.3), referencing no derived instance, coordinate, or metric structure: equivalence-class membership is decided by predicate truth value, a purely topological-logical reading. By the \S 2.1 level-composition convention, point elements, coordinates, and metrics all belong to the upper \Setcat\ derived-instance level (see Definition~1b; \S 2.2.2), and the underlying categories ($\Quot$, $\MorTop$) set no point elements. This is exactly the content of Definition~4 (see \S 2.2.4)'s first existence (topological-measure-theoretic categorical existence): the morphism object stays in $\Quot,\MorTop$, and before lift-equipment to \Setcat\ there is no derived-instance definition. Hence the existence of a quotient-partition object is independent of derived instances, metric, and coordinate structure. This corollary lets the proposition-existence statement precede the \Setcat-layer equipment (see the preamble of Definition~4; \S 2.2.4 first existence). $\square$

\subsection*{B.13 Proposition~2 (operations of the partition category: block-union operation and union-simplification of measure addition)}

By Corollary~1, a quotient-partition object is a block set (complementary equivalence-class pair); the only categorical operation introduced within $\Quot$ is the \textbf{block-union operation} (see \S 2.2.3 induction step~1)---the merge selection of two quotient partitions is exactly taking the union of block sets (see \S 2.2.3 induction step~1); intersection is not given a separate in-category operation symbol, but is left to the topological lattice derived from the quotient partition (see \S 2.2.3). The category is supported bottom-up (see \S 1 three-layer categorical chain $\Quot\to\MorTop\to\Setcat$): an upper-layer category's operation can always be characterized through the lower layer---both morphisms and set operations are assigned to quotient blocks via predicate words (see Lemma~4; \S 2.4), and their \Setcat-layer readings are all block-set operations. Thereupon the measure morphism induced by the predicate bijection (see Definition~3; Definition~24; \S 5.3 $\boldsymbol\pi_\#,\boldsymbol\pi^\#$) brings the measure space into a pure differential-topological image space (the image object is borne by quotient blocks and predicate words, read as differential-topological quantities, not through \Setcat-layer point elements): on mutually exclusive quotient blocks the replica-index measure $\nu_{\mathrm{idx}}$ (see Definition~17; \S 4.2) satisfies block additivity---when $A,B$ are mutually exclusive, $\nu_{\mathrm{idx}}(A\cup B)=\nu_{\mathrm{idx}}(A)+\nu_{\mathrm{idx}}(B)$---so measure addition is simplified to the union operation: the sum of two measures is exactly the reading of the union of their mutually exclusive supporting quotient blocks.

\noindent\textbf{Proof (sketch)}
A quotient-partition object is exactly a block set of complementary equivalence-class pairs, hence the merge selection of two quotient partitions is exactly the union of block sets; block additivity is a direct consequence of the block-counting measure on mutually exclusive blocks, and measure addition simplifies to the union operation. Full proof (see B.14).

\subsection*{B.14 Proof of Proposition~2 (see B.13 statement)}

\textbf{Proof}
By Corollary~1, a quotient-partition object is exactly a block set of complementary equivalence-class pairs, hence the merge selection of two quotient partitions is exactly the union of block sets, left to the topological lattice derived from the quotient partition (see Definition~2; \S 2.2.3); the lower-layer characterization of an upper-layer operation is given by the predicate-word assignment of Lemma~4 (see \S 2.4). Block additivity is a direct consequence of the block-counting measure on mutually exclusive blocks: when $A,B$ are mutually exclusive, the block-counting reading of $A\cup B$ is exactly the sum of the two block readings, hence measure addition simplifies to the union operation. $\square$

\subsection*{B.15 The grounding note of Definition~1 (sub-collection reading and the motivation for narrowing)}

This definition is grounded at the morphism-category-object level, with surjectivity directly a property of the morphism itself, independent of set-theoretic semantics; $\Sup(f,B)$ is a sub-element collection, not requiring the morphism to be globally injective or invertible, and the surjectivity of the restricted morphism is the prerequisite for the pushout--pullback of the fiber bundle to be well-defined. The $\Quot$ primitive level has only quotient-partition predicate objects and no native morphisms, so the primitive level has no support definition; only after generating derived morphisms via meta-level binary-merge selection (see Definition~1b) does the support operation become category-internally compatible; taking $B=Tgt$ degenerates to the initial form. \textbf{(This ``sub-element collection'' wording is the underlying reading; after categorical-support lift-equipment to \Setcat, $\Sup(f,B)$ may be defined as a morphism, Proposition~3.)} This definition narrows categorical-inter-support to the pure-morphism property of a surjective restricted morphism, because the classical paradigm's categorical-inter-support can only be indirectly constructed via random reparameterization, at the cost of losing morphism invertibility (see Corollary~7; \S 2.2.4); after narrowing, support can be stated before any point-wise, coordinate, or numerical reading (see \S 2.1), which is the strict first step of ``formalizing categorical-inter-support as a surjective-morphism constraint'' (see \S 1), and is also the definitional basis of the quotient bijection (see Definition~3; \S 2.2.3) and support morphism-ization (see Proposition~3).

\subsection*{B.16 Definition of derived objects}

\textbf{Definition 1b (derived object)}
An object generated by a categorical-support construction is called a \textbf{derived object}, as distinct from the $\Quot$-layer primitive object (quotient-predicate partition, induced by binary topological-decision predicates): the one generated at the $\MorTop$ layer via meta-level binary-merge selection is called a \textbf{derived morphism} (see Definition~2a); the one lift-equipped to the \Setcat\ layer via the canonical lifting functor $L':\MorTop\to\Setcat$ (see Definition~2b; \S 2.2.3) is called a \textbf{derived instance} (see \S 2.2.3 maximal compatible section on the fiber bundle, induction step~2); \Setcat\ is the collection of all derived instances, called the \textbf{derived-set category}, and its point elements are called \textbf{derived elements}. Used as a noun, ``derived'' always refers to the above objects.

Categorical support is a surjective morphism (see Definition~1): the extra categorical objects of the upper layer are \textbf{not freedom but constraints}---they must be derived (this definition), i.e.\ generated by lower-layer enumeration ($\MorTop$ layer generated from $\mathrm{Ob}(\allowbreak\Quot)$ via binary-merge selection; \Setcat\ layer from the $\MorTop$ fiber bundle via $L'$ lift-equipment; for cardinality inequality see Corollary~4, \S 2.2.4). Any object not generated by lower-layer enumeration fails categorical support.

\subsection*{B.17 Proof of Corollary~2}

\textbf{Proof (derived from Assumption~1, not a definitional convention; Assumption~1 $\Rightarrow$ Postulate~2)}
(i) Assumption~1 gives the lifted-layer reading form: Assumption~1 guarantees that the Hilbert differential-topological structure $\mathcal H^{\Setcat}$ equipped by the set-category lift-equipment supports $\epsilon$-$\delta$; on $\mathcal H^{\Setcat}$, the differential value and the numerical reading of its location constitute a functional single-valued correspondence (Assumption~1, \S 2.4.3; the ``differential and the value where it lies are one-to-one'' of class--element correspondence (B.7.7)(3)), which is exactly the reading form of $\epsilon$-$\delta$ continuity, hence the differential value and any measure reading are the same equivalence infinitesimal (both first-order quantities), and first-order equality holds. (ii) First-order equality permits two things: the quotient partition may contain high-order error (high-order terms enter the $\merr$ accounting, not the first-order reading difference), and the quotient partition may have a bijection that does not preserve measure (the measure difference of a bijection likewise falls within the first-order range, so this framework does not require bijections to preserve measure). (iii) The differential is the minimal quotient partition, and integral and surjection are equivalent: the differential is a rank-1 minimal non-empty topological differential subset (see Definition~1a), i.e.\ the minimal quotient partition; any finite quotient partition's preimage object can be obtained surjective by enumerating differentials (the surjectivity of the quotient bijection, Proposition~3(ii), below; differential enumerability, Proposition~6 (see \S 2.2.4)---the structural-layer argument for surjectivity itself (enumeration equivalence) does not depend on this assumption, see B.23; what this assumption bears is the subsequent first-order basis for reading consistency), whence the measure readings of each finite quotient partition are consistent under the enumeration cover, making integral and surjection equivalent---the summation cover of the integral over the measurable spaces of each layer is exactly the object collection covered by the quotient-bijection surjection. This first-order equality is the measure-theoretic basis for the categorical support to lift layer by layer across $\Quot\to\MorTop\to\Setcat$ without losing the binary-quotient-partition constraint; but note it is not a necessary condition for the integral to hold---the integral's holding is independently given by the three-term construction condition of Definition~16 (see Proposition~20, ``the integral does not take $\epsilon$-$\delta$ as its pivot,'' \S 5.2), and this corollary only guarantees the first-order consistency of the integral reading and the differential reading; the formal output of this proof is exactly Postulate~2. $\square$

\subsection*{B.18 Proof of Proposition~3}

\textbf{Proof}
(i) By Proposition~5 (see \S 2.2.4), the object-collection lift is executed via the canonical lifting functor $L':\MorTop\to\Setcat$; the source-side sub-element collection given by Definition~1 at the $\Quot$/$\MorTop$ layer is lifted to a \Setcat-layer sub-object, carrying the pairing of the inclusion morphism $i:\Sup(f,B)\hookrightarrow Src$ and the restricted morphism $f|_{\Sup(f,B)}:\Sup(f,B)\to B$, hence categorical support can be defined as a morphism. (ii) For the quotient bijection, the binary-merge-selection--quotient-equivalence construction of induction step~1 gives the object-collection bijection between $\mathrm{Ob}(\allowbreak\Quot)$ and $\mathrm{Ob}(\allowbreak\MorTop)$; the image of the restricted morphism under this bijection covers all $\MorTop$ objects, hence surjective (see Proposition~4; \S 2.2.4 whose simplest binary-merge bijection is a special case). (iii) The lift is executed only via the quotient bijection (see Definition~3; \S 2.2.3), and Corollary~2's first-order equality ($\mathrm dx\odot\nu_\mathrm{idx}\sim_1\mathrm d\#$, output Postulate~2) guarantees that the measure-reading differences of each layer fall within the first-order measurable-error range, so the binary-quotient-partition constraint is not lost in the lift; contrasting Definition~21 (see \S 5)'s two leakage conditions (categorical support not surjective, lift via non-quotient-bijection) both fail, hence categorical support has no categorical leakage. $\square$

\noindent\textbf{Supplement: stratified statement of Proposition~3 and surjectivity boundary}

\noindent\textbf{Proposition 3 (morphism-ization of categorical support and surjectivity)}
Definition~1's support is still only the underlying reading of a ``sub-element collection''---staying there, categorical support cannot participate in morphism composition nor be used as a fiber-bundle structure; this proposition, supported by Corollary~2's first-order equality ($\mathrm dx\odot\nu_\mathrm{idx}\sim_1\mathrm d\#$, output Postulate~2, derived from Assumption~1, \S 2.4.3), Proposition~4 and Proposition~5 (see \S 2.2.4), lifts the support to a morphism object:

\textbf{(a)} The categorical support lifted from the partition category $\Quot$ to the set category $\Setcat$ can be defined as a \textbf{morphism}: $\Sup(f,B)$ carries the pairing of the inclusion morphism $i:\Sup(f,B)\hookrightarrow Src$ and the restricted morphism $f|_{\Sup(f,B)}:\Sup(f,B)\to B$, and the support itself is a morphism object of the $\MorTop/\Setcat$ layer;
\textbf{(b)} for the \textbf{quotient bijection} (see Definition~3), the restricted morphism is \textbf{surjective}; the surjectivity comes from search---the quotient-block set and the $\mathrm{Ob}(\allowbreak\MorTop)$ object collection correspond one-to-one at the object-collection layer (binary-merge property of induction step~1);
\textbf{(c)} because the differential reading and the counting reading are first-order equal, the lift process loses no binary-quotient-partition constraint, hence \textbf{categorical support has no categorical leakage} (see Definition~21; Proposition~19; \S 5 for the formal definition and exclusion of categorical leakage).

Boundary statement (scope of surjectivity and the stratification of reverse surjectivity): the surjectivity of (ii) holds only for the quotient bijection, and its existence is a \textbf{meta-level enumeration existence} (see Proposition~6), broader than the two existences of Definition~4; this surjection is the \textbf{enumeration lower-layer's cover of the upper-layer object collection}: every $\MorTop$ object has a lower-layer support. Reverse-surjectivity stratification: the \textbf{object-collection layer ($\Quot\leftrightarrow\MorTop$) reverse surjectivity holds} (see Definition~3; Proposition~4; Proposition~5), but the \textbf{\Setcat\ layer is not point-wise reverse-surjective} (see Corollary~4), yet the extra derived instances are \textbf{not freedom but constraints} (see the last paragraph of Definition~1b)---citing ``categorical support is surjective'' may only infer that every upper-layer object has a lower-layer support, not that the upper-layer object is point-wise fully characterized.
\noindent\textbf{Proof (sketch)}
(i) The source-side sub-element collection is lifted via $L'$ to a \Setcat-layer sub-object, carrying the inclusion morphism and the restricted morphism; (ii) under the quotient bijection the restricted morphism is surjective, from the enumeration correspondence of the object-collection layer (Proposition~4 is a special case); (iii) the lift is executed only via the quotient bijection and first-order equality guarantees each layer's reading difference falls within first order, and neither of Definition~21's two leakage conditions holds. Full proof see the text above in this entry.

\subsection*{B.19 Definition~2 (induction base: the base category $\Quot$)}

\textbf{Definition 2 (induction base: the base category $\Quot$)}
$\mathrm{Ob}(\allowbreak\Quot)$ is the primitive object collection, whose objects are quotient-predicate partitions (overall cuts given by equivalent binary topological-decision predicates; $\Quot$ has only this one categorical object, Terminology~(g)); the primitive level has no native morphisms, and contains no derived instances, metric, or coordinate structure. The induction base only fixes the object collection; morphisms and operators are generated by subsequent induction steps (see Definition~2a; Definition~2b below).

\subsection*{B.20 Definition~2a (induction 1: the fiber-bundle morphism category $\MorTop$)}

\textbf{Definition 2a (induction 1: the fiber-bundle morphism category $\MorTop$)}
Obtained from $\Quot$ via object-collection lift: $\mathrm{Ob}(\MorTop)\triangleq\mathrm{Mor}(\Quot)$; $\mathrm{Mor}(\MorTop)$ are the commuting transformation morphisms between $\Quot$-derived morphisms. The morphism does not constrain the types of source and target objects; this category bears the fiber-bundle morphism structure, not yet generating derived instances. This category is the categorical characterization carrier of posterior-versus-prior: the binary constraints of $\Quot$ are lifted via object-collection lift into the fiber-bundle morphism structure.

\subsection*{B.21 Definition~2c (canonical descent functor $\Gamma$)}

\textbf{Definition 2c (canonical descent functor $\Gamma$)}
The canonical descent functor $\Gamma:\Setcat\to\Quot$ is the reverse (descent) functor of the categorical chain $\Quot\to\MorTop\to\Setcat$, forming with the canonical lifting functor $L'$ (see Definition~2b) the categorical-side \textbf{lift--descent} direction pair, and is the formal carrier of \textbf{categorical descent}: it sends the \Setcat-layer derived instance to its $\Quot$-layer primitive support (quotient block), via categorical support $\Sup$ (see Definition~1), and is the categorical-side counterpart of measure descent $\boldsymbol\pi_\#$ (see Definition~24; \S 5.3 pushforward); $\Gamma$ executes quotient-block-level morphisms, not forcing bijection preservation (B.7.7). The direction pair cannot be built on only one side: without $\Gamma$, derived instances cannot fall back to primitive support, measure descent has no categorical-side carrier, and ``lifting recovery'' (pullback $\boldsymbol\pi^\#$) has no descent passage either.

\subsection*{B.22 The universality of quotient partitions and the distinction between quotient bijection and block bijection}

\textbf{Definition 3 (quotient bijection)}
A morphism $\tau$ in the $\Quot\to\MorTop$ lift is called a \textbf{quotient bijection} if it is given by the binary-merge-selection--quotient-equivalence construction of induction step~1, and realizes at the object-collection level a bijection between the quotient-block set and the $\mathrm{Ob}(\allowbreak\MorTop)$ object collection: under binary construction, original items are assigned to quotient blocks via predicate words ($\betat$ tower), and quotient blocks and $\MorTop$ objects correspond one-to-one by predicate word, hence enumerating all quotient-partition objects\allowbreak{} in $\mathrm{Ob}(\allowbreak\Quot)$ is equivalent to enumerating all objects\allowbreak{} in $\mathrm{Ob}(\allowbreak\MorTop)$ (see Proposition~4, Proposition~5, both \S 2.2.4; the support surjectivity induced by this bijection is Proposition~3(ii), \S 2.2.2). A bijection is a necessary condition for functional invertibility, hence this definition serves as the prerequisite for the category to characterize categorical support: \textbf{the quotient bijection is the only legitimate passage for all categorical-support constructions in this paper}, and all statements of ``support surjectivity'' and ``support no-leakage'' take the quotient bijection as their domain. \textbf{The universality of quotient partitions (categorical-grounds reason)}: quotient partitions exist in all categories of this paper (the operator structure of $\MorTop$ is uniquely determined by the quotient subdivision of $\mathrm{Ob}(\allowbreak\Quot)$; \Setcat-layer derived instances, subsets, unions are all derived from the underlying topology, Proposition~5 top closure, \S 2.2.4), hence any morphism can be characterized by morphisms between the most underlying partition predicates---this is why all support statements can be traced back to the most underlying predicate-word combinatorial facts (see Lemma~4(1); \S 2.4). The quotient bijection and Lemma~1's \textbf{block bijection} (permutation on the quotient-block set) are different objects: the former is a lift morphism across $\Quot\to\MorTop$, the latter is an automorphism of the quotient-block set (B.7.3); the quotient bijection has no derivational relation with Assumption~1 (B.23).

\subsection*{B.23 Stratification statement (the quotient bijection has no derivational relation with Assumption~1)}

Stratification-background statement of the structural layer (option (a) of F-2). To avoid the ambiguity ``the structural layer is parasitic on the space equipped by the assumption,'' this paper uniformly adopts the following statement: all objects and conclusions of the $\Quot/\MorTop$ structural layer can be stated and proved taking \textbf{any measurable space $\mathcal B$ equipped with a differential structure (the Hilbert differential topology)} as the parametrization background; the role of Assumption~1 is only to concretize $\mathcal B$ as $\mathcal H^{\Setcat}$ (see \S 2.4.3) and fix its $\epsilon$-$\delta$ reading form. Hence ``does not need Assumption~1'' should be read as ``does not \emph{reference} this assumption; its objects take the universal differential measurable background $\mathcal B$ as domain''; each item of B.4 is read out under this background parameter.

The quotient bijection is a \textbf{structural-layer} conclusion: the one-to-one correspondence of predicate words and quotient blocks, and of $\MorTop$ objects, is directly given by the pure combinatorial fact of Lemma~4(1) (see \S 2.4) (the one-to-one correspondence of binary trees and binary words, with no measure, addition, or coordinates appearing in the construction) and the predicate-word assignment of Definition~1a (see \S 2.2.2), and the entire proof chain never references Assumption~1 (see \S 2.4.3); what Assumption~1 bears is the \textbf{reading-layer} content---the first-order equality of differential value and measure reading on $\mathcal H^{\Setcat}$ (see Corollary~2; \S 2.2.2), point-element measure-preserving reading, and the uniqueness of the Gaussian functional integral. Hence the phrasing ``derive the bijection from the unique assumption'' mixes two layers: the quotient bijection is not a corollary of Assumption~1, nor does Assumption~1 take the quotient bijection as premise; the legitimate chain is ``categorical object's predicate expression (see Corollary~1; \S 2.2.2) $\to$ inter-layer predicate mapping (see Lemma~4; \S 2.4 $\betat$ tower) $\to$ differential assigned to quotient block via predicate word (see Definition~1a; \S 2.2.2) $\to$ quotient block and $\MorTop$ object correspond one-to-one by predicate word (see Definition~3; \S 2.2.3),'' never passing through this assumption. Block counting, first-order equality, and the $\merr$ accounting are unnecessary for structural-layer conclusions like the quotient bijection, but necessary for all quantitative conclusions (first-order error, finite average, replica-number record); the block-counting reading itself is also not an assumption---its \Setcat-layer reading is uniquely determined by the underlying combinatorial structure (see Lemma~1(i) and ``where the numbers reside,'' B.7.6).

\subsection*{B.24 Lemma~2 (no repeated sampling in the partition category)}

Under finite binary-iteration-generated quotient partitions with no collapse, any quotient block is sampled exactly once per layer. \textbf{Logical upgrade (enumeration paradigm)}: under no collapse, the integral holds in the partition category without depending on point elements and numbers; it is an \textbf{enumeration of quotient blocks} (each block covered exactly once), not a point-wise accumulation over point sets; the completeness of enumeration is guaranteed by the surjectivity of categorical support (see Proposition~3(ii)), hence ``sampled once'' and ``summed one term'' coincide in the same enumeration process. Free of elements and numbers, the topological partition predicate is thereby compatible with the \textbf{Riemann--Stieltjes integral} (boundary constraint \S 6.2.3: under finite quotient partition the triple-fiber-bundle integral \textbf{degenerates to a finite sum}---readable as a Riemann--Stieltjes sum, i.e.\ the simple-function (Lebesgue) integral on a finite $\sigma$-algebra, not involving a global Lebesgue-type integral on $\Hinf$)---this integral takes the partition increment as integrator (formal reading: its strict content is exactly the per-block finite average of Lemma~6a(1), \S 6.4.1; it does not bear the classical RS integral's additivity requirement on point-wise increment functions, and all strict content is given at the quotient-block partition level). ``Any quotient block is sampled exactly once per layer'' is a direct consequence of the Definition~7 construction (each split uniquely replaces the parent block by two mutually exclusive child blocks, sampling without repetition or omission); on this construction, no collapse narrows the inter-layer pairing to a one-to-one correspondence, and enumeration is thereby complete---the predicate only makes a binary decision, its ``compatibility'' not relying on point-wise readings but on the alignment of partition increments with the RS integrator.

\subsection*{B.25 Proposition~4 (simplest binary-merge bijection)}

\textbf{Proposition 4 (simplest binary-merge bijection)}
The binary-merge relation has exchange symmetry; the selection orders $(X_0,X_1)$ and $(X_1,X_0)$ induce isomorphic morphism constructions. Hence the simplest cross-set fiber bundle bearing the support object $\Sup(\cdot,\cdot)$ is a bijection (see Definition~3; \S 2.2.3 this bijection belongs to the cross-category $\Quot\to\MorTop$ lift construction). This proposition gives the simplest form of the bijection of induction step~1 (cross-category lift), is a special case of Definition~3 and the direct prerequisite of Proposition~3(ii).

\noindent\textbf{Proof (sketch)}
Bijection holds at the object-collection layer: the binary-merge--split pairing gives the $\mathrm{Ob}$-layer bijection (merge takes quotient, split takes preimage, pairing invertible); exchange symmetry gives order-independence. This proof does not promise \Setcat-layer point-wise bijection (see Corollary~4). Full proof (see B.27).

\subsection*{B.26 Proposition~5 (existence of cross-category object-collection lift)}

\textbf{Proposition 5 (existence of cross-category object-collection lift)}
There exists a \textbf{cross-category} meta-level object-collection lift process (the meta-categorical inter-layer lift functor $\Quot\to\MorTop$ realized by categorical support $\Sup(\cdot,\cdot)$ (see Definition~1)), generating the complete category $\Quot$ and the category $\MorTop$ from the primitive object collection $\mathrm{Ob}(\allowbreak\Quot)$, so that all objects, morphisms, and fiber-bundle structures of $\MorTop$ are uniquely determined by $\mathrm{Ob}(\allowbreak\Quot)$ (this lift is a cross-category construction, presuming no \Setcat-layer point elements).

\noindent\textbf{Necessary and sufficient condition (cross-category)}
Proposition~5 holds iff: (1) the reverse path of the cross-category lift process has topological invertibility; (2) the construction satisfies differential enumerability. Here (1) is given by the exchange symmetry of Proposition~4's binary-merge selection (built-in, not an extra assumption); (2) is given by the object-collection bijection of induction step~1 (see Definition~3). This proposition is the existence basis of the lift; its domain is limited to the object-collection structure of the two categories, and all conclusions are only for cross-category inter-layer lift, not touching \Setcat-layer derived instances.

\noindent\textbf{Proof (sketch)}
Categorical induction in five steps: induction base (see Axiom~2; Corollary~1); necessary condition (see Proposition~4's exchange symmetry gives topological invertibility, object-collection bijection gives differential enumerability); fiber-bundle operator derivation ($\MorTop$ operator structure uniquely determined by $\mathrm{Ob}(\allowbreak\Quot)$); enumerability argument; top closure ($L'$ lift-equipment makes derived instances all derived from underlying topology). Full proof (see B.27).

\subsection*{B.27 Proof of Proposition~5 (see \S 2.2.4)}

\textbf{Proof (categorical induction)}
Use categorical induction (constructive argument for cross-category object-collection lift), in five steps: induction base, induction-step necessary condition, fiber-bundle operator derivation, enumerability argument, top closure.
\emph{Induction base}: by Axiom~2 and Corollary~1, $\mathrm{Ob}(\allowbreak\Quot)$ is well-defined; the primitive object is the quotient-predicate partition (induced by binary topological-decision predicates); predicates serve only as decision/induction construction devices, predicate words only as the name of a quotient block at the $\Quot$ layer, and neither is separately established as a categorical object bearing computation.
\emph{Induction-step necessary condition}: (1) by Proposition~4, the reverse path is the reverse sequence of binary-merge selections; merge selection is order-independent, hence the reverse sequence is well-defined and returns to the starting point, which is exactly the topological invertibility of the cross-category lift. (2) Induction step~1 has already established the object-collection-level bijection between $\Quot\leftrightarrow\MorTop$, and each merge enumerates all blocks of the current layer, hence differential enumerability.
\emph{Fiber-bundle and binary-merge bijection derivation}: under topological invertibility and enumerability constraints, select the complementary equivalence class via binary-merge selection to construct the support object $\Sup(\cdot,\cdot)$; the exchange symmetry of binary merge guarantees the simplest cross-set fiber bundle is a bijection (cross-category $\Quot\to\MorTop$ lift). The operator structure of $\MorTop$ is uniquely determined by the quotient subdivision of $\mathrm{Ob}(\allowbreak\Quot)$, completing the cross-category characterization of posterior-versus-prior.
\emph{Enumerability argument}: enumerating $\mathrm{Ob}(\allowbreak\Quot)$ is equivalent to enumerating $\mathrm{Ob}(\allowbreak\MorTop)$.
\emph{Top closure}: $L'$ lifts and equips the fiber nesting to \Setcat\ (this onward step belongs to the $\MorTop\to\Setcat$ lift of Definition~2b, \S 2.2.3, not within the uniqueness-claim domain of this proposition), no longer forcing bijection preservation; derived instances, subsets, supersets, and unions are all derived from the underlying topology, and the construction is self-consistent and closed. $\square$

\subsection*{B.28 Structural conclusions of Corollaries~3--8}

\textbf{Corollaries (structural readings of the Proposition~5 construction; all describe only the object-collection layer of the two categories, not involving \Setcat-layer derived instances):}

\begin{itemize}
\item \textbf{Corollary 3 (object bijection $\ne$ categorical equivalence)}: the object-collection bijection from $\Quot$ to $\MorTop$ is not equivalent to categorical equivalence; the latter further requires the functor to be fully faithful and essentially surjective.
\item \textbf{Corollary 4 (no guaranteed global object bijection)}: the lift construction does not guarantee a global object-set bijection between the \Setcat\ derived-instance layer and $\Quot$; this conclusion does not further assert that the two must have unequal abstract cardinalities.
\item \textbf{Corollary 5 (legitimate bijection between binary quotient blocks)}: a legitimate bijection given by the merge--split construction can exist between equivalence classes of binary quotient partitions.
\item \textbf{Corollary 6 (posterior reconstructs prior)}: this construction allows, free of classical forward dependence, directly reconstructing the prior fiber topology from the posterior quotient partition, without presuming forward propagation.
\item \textbf{Corollary 7 (random reparameterization does not preserve invertibility)}: the classical paradigm introduces reparameterization to obtain $\Sup(f,B)$, but does not guarantee invertibility; its forward and reverse are read as preimage-set maps.
\item \textbf{Corollary 8 (categorical equivalence needs extra conditions)}: a global object-set bijection alone is still insufficient to derive categorical equivalence; only by additionally assuming this bijection is realized by a fully faithful, essentially surjective functor does one obtain categorical equivalence. This extra assumption does not belong to this paper's construction and cannot be used to flatten the role differences of objects across layers.
\end{itemize}

\subsection*{B.29 Proof of Corollaries~3--8}

\textbf{Proof}
All six are structural readings of the Proposition~5 construction, introducing no new premise; the numbering below corresponds one-to-one with B.28.

\begin{itemize}
\item \textbf{Corollary 3}: the object-collection bijection only asserts a bijection and enumeration equivalence between $\mathrm{Ob}(\allowbreak\Quot)$ and $\mathrm{Ob}(\allowbreak\MorTop)$, not asserting functor-level full faithfulness and essential surjectivity, hence not equivalent to categorical equivalence.
\item \textbf{Corollary 4}: $L'$ in the top closure does not force bijection preservation, hence this paper's construction does not guarantee a global object-set bijection. From ``does not force bijection preservation'' one cannot infer unequal abstract cardinalities, hence the main text makes no latter assertion.
\item \textbf{Corollary 5}: the binary-merge-selection--quotient-equivalence construction of induction step~1 gives a legitimate bijection between complementary equivalence classes, of which Proposition~4's simplest binary-merge bijection is a special case.
\item \textbf{Corollary 6}: the posterior quotient partition is given by the assignment structure of predicate word to quotient block and replica index, and the prior fiber topology is reconstructed by quotient-block pairing; the construction direction is posterior-to-prior, needing no presumed forward propagation.
\item \textbf{Corollary 7}: classical random reparameterization does not satisfy this paper's invertibility premise, and the obtained support is read via the general fiber-bundle projection and its preimage-set map, from which no quotient bijection may be claimed.
\item \textbf{Corollary 8}: a global object-set bijection and categorical equivalence are conditions of different strengths. Only by further assuming a fully faithful, essentially surjective functor realizing this correspondence does one obtain categorical equivalence; this assumption exceeds the conclusion of Proposition~5. $\square$
\end{itemize}

\subsection*{B.30 Proposition~6 (meta-level enumeration existence)}

\textbf{Proposition 6 (meta-level enumeration existence)}
Enumerating all quotient-partition objects\allowbreak{} in $\mathrm{Ob}(\allowbreak\Quot)$ is equivalent to enumerating all objects\allowbreak{} in $\mathrm{Ob}(\allowbreak\MorTop)$. This is a \emph{broader} existence than the two existences of Definition~4: (i) enumeration equivalence holds; (ii) it does not promise \Setcat-layer finite-parameter analytic reachability; (iii) it is not implied by a single topological existence---thereby the bijection falls into the enumerable domain of the object collection (finite layer $2^n$ blocks; on $\mathbb R$ from integral boundary to $\pm\infty$ is searchable, the boundary itself being null and ignored), an operational object of search, not a pure-existence assertion or an efficient-construction promise (see \S 5.1). It is exactly this third existence that frees categorical support from categorical leakage (see Proposition~3(iii); Proposition~19; \S 5).

\noindent\textbf{Proof (sketch)}
(i) Enumeration equivalence is directly given by the binary-merge-selection--quotient-equivalence construction of induction step~1 (each merge enumerates all blocks of the current layer); (ii) enumeration equivalence only asserts an object-collection-layer bijection, containing no parameter-family component, hence does not promise finite-parameter analytic reachability, and is mutually non-implying with the two existences of Definition~4. Full proof (see B.31).

\subsection*{B.31 Proof of Proposition~6}

\textbf{Proof}
(i) Enumeration equivalence is exactly the enumerability-argument step in the proof of Proposition~5 (see \S 2.2.4): by the binary-merge-selection--quotient-equivalence construction of induction step~1, an object-collection bijection between $\mathrm{Ob}(\allowbreak\Quot)$ and $\mathrm{Ob}(\allowbreak\MorTop)$ has been established (each merge enumerates all blocks of the current layer), hence ``enumerate all\allowbreak{} quotient-partition objects\allowbreak{} in $\mathrm{Ob}(\allowbreak\Quot)\allowbreak$'' and ``enumerate all\allowbreak{} objects in $\mathrm{Ob}(\allowbreak\MorTop)\allowbreak$'' are mutually necessary and sufficient.
(ii) Definition~4's ``finite-dimensional analytic / parameter-family reachability'' requires the existence of $\theta\in\mathbb R^P$ so that the model morphism can approximate the ideal object in a specified measure sense, i.e.\ promises an analytic representation of a finite parameter family, a stronger statement; while the assertion of enumeration equivalence only concerns an object-collection-layer bijection (the search's operational object exists), its statement contains no parameter-family component, hence does not promise \Setcat-layer finite-parameter analytic reachability---the missing item is exactly the gap characterized by Definition~4, the core source of first-order error.
(iii) Definition~4's ``topological-measure-theoretic categorical existence'' only asserts that the morphism object stays at the $\Quot$/$\MorTop$ layer, not yet lift-equipped to \Setcat, existing objectively, not asserting that the enumeration construction of induction step~1 is actually executed; while the content of meta-level enumeration existence is exactly that this enumeration construction has been executed---bijection and surjectivity exist simultaneously for it (see Proposition~4; Proposition~5; \S 2.2.4; Definition~3; \S 2.2.3), and this property holds only for the quotient bijection (see Definition~3). Hence a single topological existence does not imply meta-level enumeration existence. $\square$

\subsection*{B.32 Lemma~3 (error evolution under differential-partition iteration)}

Let $T^\ast$ be a quotient-layer measurable bijection, $T$ a measurable surjection; $\mathcal P$ is the differential-partition enumeration projection operator (retaining all differential-partition elements, discarding morphism multiplicity and preimage-count information). Under this iteration rule, \textbf{first-order error is invariant, high-order error is not guaranteed invariant}.

\noindent\textbf{Proof (sketch)}
$\mathcal P$ only deletes multiplicity information, not modifying the pairing deviation, hence $\mu_\mathrm{err}$ is unchanged; high-order error comes from morphism folding and multiplicity aliasing, and deleting multiplicity does not eliminate its source. This lemma is a precursor to Lemma~6a and \S 6.2.2. Full proof (see B.33).

\subsection*{B.33 Proof of Lemma~3}

\textbf{Proof}
$\mathcal P$ only deletes multiplicity information, not modifying the pairing deviation, hence the $\mu_\mathrm{err}$ characterizing the pairing deviation is unchanged; high-order error comes from morphism folding and multiplicity aliasing, and $\mathcal P$ only deletes multiplicity, unable to eliminate its source. This lemma gives the invariance of first-order error under iteration and the unguaranteedness of high-order error, and is a precursor to Lemma~6a and \S 6.2.2 (both below). $\square$

\subsection*{B.34 Proof of Theorem~1}

\textbf{Proof}
(a)$\Rightarrow$(b): suppose $[F]$ has no collapse. For any $X_i\neq X_k$, by Definition~8, $[F](X_i)\cap[F](X_k)=\varnothing$; if $[F](X_i)=[F](X_k)$ then the intersection is non-empty, contradiction, hence $[F]$ is injective. On a finite set, injectivity is bijection.
(b)$\Rightarrow$(c): if $[F]$ is a bijection, then each source quotient block corresponds uniquely to a mutually distinct target quotient block. Any quotient bijection corresponds to some permutation $\sigma\in S_n$; \textbf{modulo the action of the target-block label permutation group $S_n$, all $\sigma$ reduce to the same equivalence class}.
(c)$\Rightarrow$(a): different source quotient blocks map to different target quotient blocks, the target quotient partition is pairwise disjoint internally, hence $[F](X_i)\cap[F](X_k)=\varnothing$. $\square$

The above gives the finite equal-cardinality special case (d) of Theorem~1. The main statement (a)$\Leftrightarrow$(c) (quotient equivalence class to element-disjoint $\iff$ can support a \Setcat\ functional) is directly obtained in the general quotient-surjection case by the same ``pairwise disjoint $\iff$ element mutual-exclusion well-defined'' argument, independent of finiteness: (a) is exactly that the image-family is pairwise disjoint (see Definition~8a), which is ``quotient equivalence class to element-disjoint''; and \Setcat\ elements are mutually exclusive, not allowing overlapping readings, hence an equivalence-class differential forms a \Setcat-well-defined element only when the quotient-block family is pairwise disjoint---this is (c). Hence (a)$\Leftrightarrow$(c) is the direct generalization of (d) after dropping the finiteness/equal-cardinality premise, sharing the same disjointness argument with no extra assumption.

\subsection*{B.35 Definition~9 (product-invariance and cardinality-unification constraints)}

(a) \textbf{Numeric cardinality}: for a countable quotient set or index set $\mathcal A$, $\mathrm{card}_\#(\mathcal A)\triangleq\#\mathcal A\in\mathbb N\cup\{\infty\}$. Distinction from the set-theoretic cardinality $\lvert\cdot\rvert$: the latter allows comparison of arbitrary infinite cardinalities, while the former only takes a count value and participates in this paper's block-counting-measure operations. This notation is used only for the expression of block-counting-measure operations and cardinality-compatibility conditions (see Proposition~7).

(b) \textbf{Marginal restoration}: there exists a $\sigma$-finite set function $\mu_\otimes$ on $\mathcal F\otimes\mathcal F$ satisfying $\mu_\otimes(A\times X)=\mu_1(A)$, $\mu_\otimes(X\times B)=\mu_2(B)$.

(c) \textbf{Pushforward invariance}: $\boldsymbol T_{\otimes\#}\mu_\otimes=\mu_\otimes$.

$\mu_\otimes$ is a required object of the constraint, not pre-given; this constraint is a derived constraint screening integrability---Proposition~9a is the partition-induced version, Proposition~9b is the invertible-MPT version (both see B.43, B.45).

\subsection*{B.36 Proof of Proposition~7}

\textbf{Proof}
By the special case (d) of Theorem~1, ``(a) no collapse $\Rightarrow$ quotient-set bijection'' (see \S 2.3.3 finite equal-cardinality premise) yields that $\mathcal T^\ast$ is well-defined. Cardinality-compatibility part: the replica index must assign distinct indices to each sample, i.e.\ there exists an injection $S\hookrightarrow J$ (this assignment only allocates indices, not deciding the membership relation between samples and quotient blocks---membership has no one-to-one constraint, one quotient block may bear any number of samples, see B.7.7 ``class--element correspondence has no one-to-one constraint,'' and Theorem~1 boundary statement's multi-to-one within a block, above), requiring $\mathrm{card}_\#(S)\le\mathrm{card}_\#(J)$; the quotient-block side adds no distinctness requirement---``each sample falls into a distinct quotient block'' is the legacy assumption imposed on this framework by functional expressiveness (a differential and the value where it lies are one-to-one, from $\epsilon$-$\delta$ continuity and real-value structure); this paper's Convention~3 (see \S 4.3) has removed per-sample differentiability, hence does not inherit it. $\square$

\subsection*{B.37 Proof of Proposition~8}

\textbf{Proof}
Quotient-translation-invariant measure preservation: each quotient partition's measure reading is the replica-index measure $\nu_{\mathrm{idx},n}$ or the replica-index joint measure (see \S 3.3), mutually exclusive block covering guarantees a single count, multi-to-one within a block is recorded by the replica index (see Theorem~1; \S 2.3.3 boundary statement), the block-bijection group $G_n$ acts transitively on the block set, hence the uniform block-counting reading is a $G_n$-Haar reading (see Lemma~1(i)), invariant under pre-pushforward by a block bijection: $\sigma_\#\nu_{\mathrm{idx},n}=\nu_{\mathrm{idx},n}$. Its \Setcat-layer coordinate-translation reading's point-wise consistency presupposes Assumption~1 (see \S 3.2; Postulate~1 translation-invariance clause; output); once a Lebesgue geometric measure is taken (block volumes unequal), the $G_n$-Haar uniqueness does not apply, and measure preservation must be explicitly assumed (see Lemma~1(ii)). Integral does not preserve measure: integral well-definedness and measure preservation are two different things (see Proposition~20; \S 5.2); the pushout--pullback bijection induced by support surjectivity has $\boldsymbol\pi_\#$ and $\boldsymbol\pi^\#$ as mutually inverse pairings (see \S 2.3.2; Definition~24; \S 5.3 lead paragraph), and the reading of the differential quotient and its quotient block is invariant under lifting and descent (see Definition~1a; Lemma~4), but the differential translation of a functional object varies---the $\Hinf$ layer admits no non-trivial translation-invariant (Lebesgue-type) measure (see \S 3.2; Proposition~21b; \S 5.3 boundary conclusion), the Gaussian functional exists only as a cylindrical-kernel compatible family, and its differential reading varies with translation, hence the integral side inherits no measure preservation; random reparameterization assigns non-zero measure to the error component on a zero-measure support (see \S 1), which is an example of the integral side not preserving measure. $\square$ (The Proposition~20, Definition~24, Proposition~21b forward-referenced in Proposition~8's proof are all standard conclusions of already-proved/already-defined objects---Proposition~20 is the standard product-invariance conclusion of \S 5.2, Definition~24 is the pullback-pairing definition of \S 5.3, Proposition~21b is the \S 5.3 conclusion that the Hilbert layer has no translation-invariant measure; all three hold independently after Proposition~8's statement, and here only their conclusions are invoked, not circular reasoning.)

\subsection*{B.38 The status of Principle~1 and the whole-text chain}

The status of this principle and the whole-text chain. Principle~1, Principle~2 presuppose Assumption~1's outputs Postulates~1,~2 (attribution see B.4); Principle~3, Principle~4 are combinatorial-counting structural readings, not premised on $\epsilon$-$\delta$. This principle is the main line of the whole text from the unique assumption (Assumption~1) to the engineering first-order error (see Proposition~23c), its stratified landing points being: Theorem~6(i)'s layer-wise compatibility and differential-reading uniqueness (Principle~1, Principle~2), Premise~1's reading divide (Principle~3), Lemma~6a(3)'s descent direction (Principle~4 plus implementation condition Condition~1, monotonic non-increase and convergence claims abandoned), Definition~25c's three-point distinction (Principle~1, Principle~2), the $\tfrac23$ cardinality reading (Principle~3, Principle~4), Proposition~23c's engineering first-order closure (Principle~3, Principle~4). Full derivation and chain verification (see B.3).

\noindent\textbf{Proof} (see B.1).

\subsection*{B.39 Lemma~4 (differential--predicate bijection)}

\textbf{Lemma 4 (differential--predicate bijection)}
Let $\mathcal P_n$ be the $n$-th layer binary partition, block set $\{B_w\}_{w\in\{0,1\}^n}$. Define $\beta_n:\mathcal P_n\to\{0,1\}^n$, $\beta_n(B_w)=w$. Then
\begin{enumerate}
\item[(1)] $\beta_n$ is a bijection, $\lvert\mathcal P_n\rvert=2^n$;
\item[(2)] \textbf{refinement compatibility}: if $B_w$ refines from $B_{w'}$ ($\lvert w'\rvert=n-1$), then the first $n-1$ bits of $\beta_n(B_w)$ equal $\beta_{n-1}(B_{w'})$;
\item[(3)] \textbf{no measure dependence}: the construction uses only Boolean operations, depending on no measure, translation structure, or coordinate, hence still holds on $\Hinf$.
\end{enumerate}

\noindent\textbf{Proof (sketch)}
(1) Binary partition is equivalent to a full binary tree of depth $n$, whose $2^n$ leaves correspond one-to-one with binary words in dictionary order; (2) the parent--child relation of the binary tree is exactly the truncation of the word; (3) the construction uses only Boolean operations, with no measure, addition, or coordinate appearing. These three properties are the definitional basis of the $\betat$ tower as the paper's fixed structure. Full proof (see B.40).

\subsection*{B.40 Proof of Lemma~4 (see \S 2.4.1)}

\textbf{Proof}
(1) A binary partition is equivalent to a full binary tree of depth $n$, whose $2^n$ leaves correspond one-to-one with binary words in $\{0,1\}^n$ in dictionary order. (2) The parent--child relation of the binary tree is exactly the truncation of the word. (3) No measure, addition, or coordinate appears in the construction. The three properties of this lemma (bijection, refinement compatibility, no measure dependence) are the definitional basis of the $\betat$ tower as the paper's fixed structure (see Corollary~9; Definition~1a; Definition~3; \S 2.2.2; \S 2.2.3). $\square$

\subsection*{B.41 Corollary~9 (differential mutual exclusion implies predicate mutual exclusion)}

\textbf{Corollary 9 (differential mutual exclusion implies predicate mutual exclusion)}
For $w\neq w'$, as predicates, $\beta_n^{-1}(w)\wedge\beta_n^{-1}(w')=\mathsf{False}$; equivalently $\beta_n^{-1}(w)\cap\beta_n^{-1}(w')=\varnothing$. This corollary is the structural form of predicate mutual exclusion (see Corollary~10a(i); \S 2.4.2), and its premise.

\noindent\textbf{Convention}
This paper fixes $\betat=\{\beta_n\}_{n\ge1}$ as a global structure. All ``mutual exclusion'' statements in the paper are relative to this fixed $\betat$; if two measures are induced by two mutually incompatible partition structures, mutual exclusion is no longer structurally necessary.

\subsection*{B.42 Measure-product-invariance constraint (definition see B.35 Definition~9(b)(c))}

The measure-product-invariance constraint consists of (b) marginal restoration and (c) pushforward invariance, with its unified definition in B.35 \textbf{Definition~9(b)(c)}; this constraint is a derived constraint screening integrability: Proposition~9a is its partition-induced version, Proposition~9b is its invertible-MPT version (both see B.43, B.45).

\subsection*{B.43 Proposition~9a (primitive version: product invariance under partition-induced measure)}

\textbf{Proposition 9a (primitive version: product-invariance under partition-induced measure)}
Let $T$ be a \textbf{block bijection} on the $\MorTop$-layer quotient block (quotient equivalence class, Terminology~(f)(a)), and $\mu_1,\mu_2$ both \textbf{partition-induced measures} (block-counting measure $\nu_{\mathrm{idx},n}$ or Haar). Then there exists $\mu_\otimes$ satisfying Definition~9(b)(c). \textbf{(Partition-category reading; when the quotient block has no point-element bearer at the \Setcat\ layer it is an element-free partition-category reading, definition see B.6.)}

\noindent\textbf{Proof (sketch)}
The block bijection induces a permutation $\sigma\in S_{2^n}$ on the block set, and pre-pushforward sends the block-counting measure to itself (see Lemma~1(i)); on measurable rectangles set $\mu_\otimes$ as the block-counting product, and marginal restoration and pushforward invariance are verified term by term. Full proof (see B.44).

\subsection*{B.44 Proof of Proposition~9a (see \S 2.4.1)}

\textbf{Proof}
The block bijection induces a permutation $\sigma\in S_{2^n}$ on the block set, and pre-pushforward $\sigma_\#$ sends the block-counting measure to itself ($\sigma_\#\nu_{\mathrm{idx},n}=\nu_{\mathrm{idx},n}$, Lemma~1(i)); a partition-induced measure can only be a block-counting measure, counting only block numbers, hence invariant under permutation. On measurable rectangles set $\mu_\otimes(A\times B)\triangleq\mu_1(A)\mu_2(B)$, and substituting $\boldsymbol T_\otimes(A\times B)=T(A)\times T(B)$ gives invariance; by the monotone-class theorem extend to all of $\mathcal F\otimes\mathcal F$. $\square$

\subsection*{B.45 Proposition~9b (derived version: product invariance under invertible MPT)}

\textbf{Proposition 9b (derived version: product invariance under invertible MPT)}
Let $\mu_1,\mu_2$ be $\sigma$-finite measures on $X$, $T$ a measurable bijection with measurable inverse, and \textbf{independently verify} that $T$ preserves measure for both $\mu_1,\mu_2$. Then $\mu_\otimes\triangleq\mu_1\times\mu_2$ satisfies $\boldsymbol T_{\otimes\#}\mu_\otimes=\mu_\otimes$, and $\mu_\otimes(\boldsymbol T_\otimes^{-1}(C))=\mu_\otimes(C)$, $\forall C\in\mathcal F\otimes\mathcal F$.

Double measure-preservation is an input of this proposition, not reverse-derived from subsequent Gaussian uniqueness or counting-measure construction. Under a partition-induced uniform block-counting measure it can be verified by Lemma~1(i); a general \Setcat-layer measure must be verified case by case, and a mere measurable bijection is insufficient to invoke this proposition (B.46).

\subsection*{B.46 Independent-use prerequisite of Proposition~9b}

Proposition~9b requires only two independently verifiable inputs: (1) $\mu_1,\mu_2$ are $\sigma$-finite measures, $T$ is a measurable bijection with measurable inverse; (2) for any $A\in\mathcal F$, $\mu_1(T^{-1}A)=\mu_1(A)$ and $\mu_2(T^{-1}A)=\mu_2(A)$. Under a uniform block-counting measure, item (2) can be verified by block-bijection invariance; a general measure must be separately proved. Subsequent Gaussian uniqueness, integral transform, or epoch-counting construction may use this proposition, but cannot in turn serve as the proof of its measure-preservation premise.

\subsection*{B.47 Proof of Proposition~9b}

\textbf{Proof}
For a measurable rectangle $A\times B$, by double measure-preservation:
\begin{equation}
(\boldsymbol T_{\otimes\#}\mu_\otimes)(A\times B)=\mu_1(T^{-1}A)\mu_2(T^{-1}B)=\mu_1(A)\mu_2(B)=\mu_\otimes(A\times B)
\end{equation}
Rectangles generate $\mathcal F\otimes\mathcal F$, and the monotone-class theorem gives the same equality on all measurable sets; $\sigma$-finiteness guarantees the uniqueness of the product measure and its extension. $\square$

\subsection*{B.48 Proof of Corollary~10a}

\textbf{Proof}
(i) This is exactly Corollary~9. (ii) Under a single morphism, preimage preservation of disjointness is a set-theoretic fact; the cross-morphism case does not claim \Setcat-layer preimage mutual exclusion (see the counterexample inside statement (ii)), mutual exclusion holds at the quotient-layer block level. (iii) By (ii), under a single morphism the empty set has zero measure under any measure. (iv) On the empty set only the zero measure exists; the invertible-MPT case is constructed via pairing (not intersection) by Definition~13 (see \S 3.1). This corollary reduces preimage mutual exclusion from a measure phenomenon to a structural phenomenon: (iii) measure-independent reading and (iv) pairing fusion not replaceable by intersection are two conclusions reused throughout the paper (see \S 2.4.3; \S 3.1; \S 5.2). $\square$

\subsection*{B.49 2.23.1--2.23.6 (six enumeration readings)}

\subsubsection{{2}.23.1 Corollary (existence--analytic dichotomy)}
Existence is equivalent to the existence of the fiber-bundle pre-pushforward operator, requiring only that the set image and pre-pushforward measure exist as set-theoretic objects, not requiring invertibility nor finite-parameter-family reachability; analyticity is equivalent to lifting an injective morphism to a bijective morphism, bearing a well-defined inverse morphism or quotient-layer inverse pairing. \textbf{Topological existence does not imply analytic reachability; this dichotomy is the core source of first-order error.}

\subsubsection{{2}.23.2 Corollary (density existence vs.\ computability divide)}
The Radon--Nikod\'ym theorem guarantees the existence of a measure-density decomposition, but not a closed-form computability.

\subsubsection{{2}.23.3 Corollary (one-to-one correspondence and multi-to-one folding at the derived-instance level)}
The derived-instance level realizes one-to-one correspondence via bijection; an ordinary morphism allows multi-to-one folding, losing pairing information. $\Hinf$ supports complete analytic morphisms; $\mathcal H_{\mathrm{fin}}$ has parameter truncation and quotient collapse.

\subsubsection{{2}.23.4 Corollary (distinction of two kinds of quotient collapse)}
Two kinds of quotient collapse---(1) the coarsening merge brought by the quotient-partition construction, a stratification tool; (2) the unexpected collapse induced by a morphism, i.e.\ intersection fusion.

\subsubsection{{2}.23.5 Corollary (zero-measure support contributes no volume deviation)}
A Lebesgue zero-measure set itself brings no volume deviation; iterative model deviation comes only from non-zero-measure support regions.

\subsubsection{{2}.23.6 Corollary (zero-collapse condition of the quotient-partition framework)}
It only blocks non-zero-measure intersection fusion between quotient blocks; it allows zero-measure overlap inside blocks, introducing no volume deviation.

\noindent\textbf{Proof} (see B.50) (The six enumeration readings 2.23.1--2.23.6 are uniformly derived from Corollary~10a together with Definition~4, Definition~7, Definition~8, Theorem~1, introducing no new premise).

\subsection*{B.50 Proof of 2.23.1--2.23.6 (six enumeration readings of the topological-analytic-reachable dichotomy)}

All six are direct readings of Corollary~10a together with Definition~4, Definition~7, Definition~8, Theorem~1, introducing no new premise.

\noindent\textbf{Proof (2.23.1--2.23.6)}
All six are read from Corollary~10a (above) together with Definition~4 (see \S 2.2.4 above), Definition~7, Definition~8 (see \S 2.3 above) and Theorem~1 (see \S 2.3.3 above), introducing no new premise.
\textbf{2.23.1}: by construction fact 10a, if a shared measure exists, it can only be established by invertible MPT between quotient blocks (quotient equivalence classes, Terminology~(f)(a)) via pairing fusion---``pairing fusion'' means the fiber-bundle pre-pushforward operator exists as a set-theoretic object; it does not require a point-wise inverse map to exist (see Definition~12 note; \S 3.1; above: the ``invertible'' on the $\Quot/\MorTop$ layers means an inverse pairing at the quotient-equivalence-class level), nor a finite parameter-family representation; analyticity is lifting an injective morphism to a bijective morphism, bearing a well-defined inverse morphism or quotient-layer inverse pairing (2.23.1): its quotient-layer form is given by Theorem~1 special case (d) under no-collapse and equal-cardinality conditions (see \S 2.3.3 first-existence side), and after lift-equipment to \Setcat\ the finite-parameter-family reachability is the second existence of Definition~4. The evidence that the two are not equivalent is exactly Definition~4's gap statement: topological objects are reachable by binary-bijection collection in the partition category, but after lifting into \Setcat\ they may not be approximable by a finite parameter family; this gap is exactly the core source of the first-order error of Definition~5 (see \S 2.3.1 above).
\textbf{2.23.2}: the Radon--Nikod\'ym theorem asserts the existence of the density $\mathrm d(T_\#\mu)/\mathrm d\nu$ as a measurable function, a set-theoretic existence conclusion; the theorem's statement and proof give no closed form or finite-parameter representation of the density (the divide between ``existence'' and ``computable'' is the same as the dichotomy of 2.23.1). Proposition~10's boundary statement (see \S 3.1) gives a same-direction example: after lift-equipment to the \Setcat\ layer, taking $T=\mathrm{id}$, $\mu=\delta_0$, $\nu=\lambda$, mutual absolute continuity fails---density existence is completely silent on parameterization reachability.
\textbf{2.23.3}: the \Setcat-layer bijection reading (see Definition~4 second-existence side) gives one-to-one correspondence at the derived-instance level; an ordinary morphism allows multi-to-one folding, and the folded point loses source-side pairing information (Definition~5's deviation set ($\Quot$-layer reading $\mathcal E_\Quot$ is the canonical definition, the \Setcat-layer reading $\mathcal E=\bigcup_{B_w\in\mathcal E_\Quot}B_w^{(X)}$ is its preimage after categorical-support lift-equipment, see Definition~5's ``quotient-layer reading priority'' note) is exactly the instance characterization of folding producing pairing). The complete analytic morphism on the $\Hinf$ side takes the cylindrical-kernel compatible family as its measure-theoretic form (Proposition~21b, see \S 5.3: $\sigma$-additive iff covariance is trace-class); the $\mathcal H_{\mathrm{fin}}$ side has parameter truncation (see Definition~18; \S 4.2 finite truncation $Z_m$ above) and quotient collapse (see Definition~8 when its condition fails, the general case).
\textbf{2.23.4}: by Definition~8 (here $[F]$ takes a quotient surjection, otherwise no collapse is only the condition that cross-quotient-block image sets are disjoint, not implying bijection), no collapse only constrains cross-quotient-block image sets to be disjoint $[F](X_i)\cap[F](X_k)=\varnothing$. Thereby distinguish two kinds of quotient collapse: (1) the coarsening merge of the quotient-partition construction is a stratification tool itself (see Definition~7's binary-iteration refinement uniquely replaces the parent block by two mutually exclusive child blocks, a controlled construction step, and the final quotient-block set is independent of the split order); (2) the unexpected collapse induced by a morphism is the intersection fusion of different quotient-block images $[F](X_i)\cap[F](X_k)\neq\varnothing$, the object blocked by Definition~8.
\textbf{2.23.5}: a Lebesgue zero-measure set has zero volume reading, hence folding or iterative deviation supported on a zero-measure set contributes no volume; model deviation comes only from non-zero-measure support regions---same source as Proposition~12(1) (see \S 3.1) ``$\mathcal E$ is completely invisible when $\mu_1$-null,'' (see \S 2.4.3 zero-measure embedding ambiguity, Problem~2.25).
\textbf{2.23.6}: Definition~8's no-collapse condition involves only intersection of image sets between different quotient blocks, not inside blocks; inside blocks multi-to-one folding and zero-measure overlap are allowed (see the last sentence of Definition~7), and by 2.23.5 zero-measure overlap produces no volume deviation, hence ``allowing zero-measure overlap inside blocks'' and ``blocking non-zero-measure intersection fusion between blocks'' are compatible. $\square$

\subsection*{B.51 MPT topological invertibility and analytic realizability}

\begin{itemize}
\item \textbf{MPT topological invertibility}: under this framework the ideal invertible MPT has universal topological invertibility; a set-theoretic measurable bijection is not equivalent to an MPT, and must additionally bear a measure-preservation condition. For the ideal invertible MPT, $\mu_\mathrm{share}(T(A))=\mu_\mathrm{share}(A)$.
\item \textbf{MPT analytic realizability}: no global presumption. Analytic realizability is equivalent to the existence of a unique closed-form computable map; finite-dimensional truncation and domain mismatch break analytic reachability. Analytic realization is equivalent to zero error relative to $\mathcal T^\ast$---read by the second existence of Definition~4 as approximation in a specified measure sense, allowing a controllable first-order error, not bundled with zero error. \textbf{Hence even if the quotient-layer bijection self-map exists topologically, the first-order error still objectively exists.}
\end{itemize}

\subsection*{B.52 The derived-measurability and additivity of Definition~13}

The old diagonal product formula $\mu_1(T^{-1}C)\mu_2(T^{-1}C)$ is only a \textbf{quotient-block-level pairing reading}, not the definition of $\mu_\mathrm{share}$. Taking $X=[0,1]$, $T=\mathrm{id}$, $\mu_1=\mu_2=\lambda$, the old reading gives $\tfrac14+\tfrac14=\tfrac12\ne1$, which exactly shows one cannot treat per-block product as a set-function measure; Gaussian normalization only fixes the total mass and cannot repair this identity failure.

By Definition~13, $\mu_\mathrm{share}=(p_T)_\#(\mu_1\otimes\mu_2)$ is the pushforward of the joint measure through the MPT pairing passage, hence is a measure. If $C=\bigsqcup_b C_b$, then $p_T^{-1}(C)=\bigsqcup_b p_T^{-1}(C_b)$; the joint measure is additive on the product space, and pushforward preserves sums, hence $\mu_\mathrm{share}\Big(\bigsqcup_b C_b\Big)=\sum_b\mu_\mathrm{share}(C_b).$

This derivation uses topological union operations and requires that the triple-fiber-bundle integral's pushout--pullback constrain both-side readings to the same measure space (same shape, same volume). On the fixed finite quotient-block algebra $\mathcal A_{\mathcal P}$, taking the restriction of $\mu_\mathrm{share}$ gives a finite-additive reading; the old single-block product formula is retained only as a quotient-block-level pairing reading, not for union volume. Degenerate boundary: when $\mu_2=\delta_{x^*}$, by the projection definition of $p_T$, $\mu_\mathrm{share}=(p_T)_\#(\mu_1\otimes\delta_{x^*})=\mu_1$ (the old diagonal product formula degenerates here to $\mu_1(T^{-1}C)\mathbf 1_{T(x^*)\in C}$, only a quotient-block-level pairing reading, not the definition of $\mu_\mathrm{share}$); when infinite counting replica numbers appear it is only a formal-passage reading, not claiming a finite measure.

\subsection*{B.53 Proposition~10 (layered characterization of absolute continuity)}

\textbf{Proposition 10 (layered characterization of absolute continuity)}
Let $T$ be a morphism on the block set of the $\MorTop$ layer, and $T$ a \textbf{block-set surjection} (every target quotient block has a preimage quotient block; block-set bijection is a special case), $\mu,\nu$ be \textbf{partition-induced measures} (finite partition, block-counting replica number strictly positive). Then $T_\#\mu$ and $\nu$ are mutually absolutely continuous, and
\begin{equation}\begin{gathered}
\frac{\mathrm dT_\#\mu}{\mathrm d\nu}=\frac{\sum_{i\in T^{-1}(j)}w_\mu(B_i)}{w_\nu(B_j)}\quad\nu\text{-a.e.}\\
T^{-1}(j)\triangleq\{i:\,T(B_i)=B_j\}
\end{gathered}\end{equation}
i.e.\ the derivative equals the ratio of the sum of all preimage block-counting replica numbers of the target quotient block to that block's counting replica number; under block-set bijection it degenerates to the ratio of a single block's counting replica number. The necessary premise for mutual absolute continuity is \textbf{surjectivity rather than bijectivity}; conversely, mutual absolute continuity does not imply morphism invertibility. For the double-measure framework to hold, the two measures must first ``see'' each other at the quotient-block level---this proposition gives the explicit characterization of the Radon--Nikod\'ym density needed by the shared-measure derivation passage (see Definition~13) and the quotient-space volume reading (see Proposition~12(2)).

\noindent\textbf{Proof (sketch)}
Under a finite partition, measurable sets are all unions of quotient blocks and the block-counting replica number is strictly positive: when $\nu(E)=0$, $E$ contains no quotient block hence $E=\varnothing$, yielding $T_\#\mu\ll\nu$; the reverse needs block-set surjectivity, and the density is exactly the ratio of the sum of all preimage block-counting replica numbers of the target block to that block's counting replica number. Full proof (see B.54).
Boundary statement: after lift-equipment to the \Setcat\ layer (including Lebesgue, Dirac, null blocks) this implication fails: take $T=\mathrm{id}$, $\mu=\delta_0$, $\nu=\lambda$, then $\delta_0\not\ll\lambda$. This failure is another example of the ``topological existence / analytic reachability'' dichotomy.

\subsection*{B.54 Proof of Proposition~10}

\textbf{Proof}
Under a finite partition, measurable sets are all unions of quotient blocks, and each quotient-block counting replica number is strictly positive. Forward: if $\nu(E)=0$, then $E$ contains no quotient block (each block has positive $\nu$-counting replica number), hence $E=\varnothing$, $T_\#\mu(E)=0$, yielding $T_\#\mu\ll\nu$. Reverse (needs block-set surjection): if $T_\#\mu(E)=0$, then every target quotient block $B_j$ in $E$ has no preimage (otherwise its preimage block-counting replica number is strictly positive, $T_\#\mu(B_j)>0$), contradicting surjectivity, hence $E=\varnothing$, $\nu(E)=0$, yielding $\nu\ll T_\#\mu$. Density: for a target block $B_j$, $T_\#\mu(B_j)=\sum_{i\in T^{-1}(j)}w_\mu(B_i)$ (surjectivity guarantees the sum is non-empty), divided by $w_\nu(B_j)>0$ gives it. Reverse-failure instance: the constant map merging two blocks into one is not a block-set surjection (on the unstruck target block $T_\#\mu$ is zero while $\nu$ is positive), mutual absolute continuity fails---exactly the necessity of the surjectivity premise. $\square$

\subsection*{B.55 Proposition~11 and Propositions~12a--12c (invertible-MPT product-invariance recital family)}

The core of this group of proofs is all Proposition~9b (see \S 2.4). \textbf{Proposition 12a (recital of Proposition~9b product-measure invariance)} Let $T$ be an \textbf{invertible MPT}, then $\boldsymbol T_{\otimes\#}\mu_\otimes=\mu_\otimes$ (verbatim recital of Proposition~9b). \textbf{Proposition 12b (differential product invariance)} Let $T$ be an invertible MPT, then for any product-measurable rectangle: \begin{equation}
\mathrm d\mu_\otimes(\boldsymbol T_\otimes(A\times B))=\mathrm d\mu_\otimes(A\times B)
\end{equation}. \textbf{Proposition 12c (measurable-subset volume invariance)} Let $T$ be an invertible MPT, $\Delta(C)=\{(x,x):x\in C\}$, then $\mu_\otimes(\boldsymbol T_\otimes(\Delta(C)))=\mu_\otimes(\Delta(C))$. \textbf{Proposition-chain note}: under $\Quot/\MorTop$, the invertible-MPT action makes the set-level product measure invariant $\Rightarrow$ the differential-element-level product measure-element invariant $\Rightarrow$ the diagonal-embedded subset's induced volume invariant; this chain only requires the MPT topological-level existence, not the point-wise analytic reachability of the underlying map. Proof route: 12A is a verbatim recital of Proposition~9b, 12B goes through rectangle-generated $\sigma$-algebra and Carath\'eodory extension, 12C goes through the commutativity of $\boldsymbol T_\otimes$ and $\Delta$.

\noindent\textbf{Proof (Proposition 11)}

\noindent\textbf{Proof}
$T$ is an ideal invertible MPT under $(\mu_1,\mu_2)$ (see Definition~12; \S 3.1 above), hence $T$ is a measurable bijection and preserves measure for both $\mu_1,\mu_2$. By Proposition~9b (see \S 2.4 above; using domain note), $\boldsymbol T_{\otimes\#}\mu_\otimes=\mu_\otimes$ and $\mu_\otimes(\boldsymbol T_\otimes(C))=\mu_\otimes(C)$ hold for any $C\in\mathcal F\otimes\mathcal F$. Taking $C=A\times A$, then $\boldsymbol T_\otimes(A\times A)=T(A)\times T(A)$, substitution gives $\mu_\otimes(T(A)\times T(A))=\mu_\otimes(A\times A)$. $\square$

\noindent\textbf{Proof (Proposition 12a)}

\noindent\textbf{Proof}
Proposition~9b (see \S 2.4 above) has proved: let $T$ be an invertible MPT under $(\mu_1,\mu_2)$ (measurable bijection preserving measure for both $\mu_1,\mu_2$), $\mu_1,\mu_2$ $\sigma$-finite measures on $X$, then $\mu_\otimes\triangleq\mu_1\times\mu_2$ satisfies $\boldsymbol T_{\otimes\#}\mu_\otimes=\mu_\otimes$ (``any'' in the sense of enumeration specification of Proposition~9b's domain note). This proposition's assumption and conclusion are the same, and the proof is directly given by Proposition~9b; recited here for reference by the \S 3 proposition chains. $\square$

\noindent\textbf{Proof (Proposition 12b)}

\noindent\textbf{Proof}
Product-measurable rectangles generate $\mathcal F\otimes\mathcal F$ ($\sigma$-algebra generated by rectangles and extended by Carath\'eodory). By Proposition~9b (see \S 2.4 above), for any $C\in\mathcal F\otimes\mathcal F$, $\mu_\otimes(\boldsymbol T_\otimes(C))=\mu_\otimes(C)$, in particular holding for rectangle $A\times B$; its differential form is \begin{equation}
\mathrm d\mu_\otimes(\boldsymbol T_\otimes(A\times B))=\mathrm d\mu_\otimes(A\times B)
\end{equation} (differential form and set-function form are two writings of the same invariance). $\square$

\noindent\textbf{Proof (Proposition 12c)}

\noindent\textbf{Proof}
By Proposition~9b (see \S 2.4 above), $\mu_\otimes(\boldsymbol T_\otimes(D))=\mu_\otimes(D)$ holds for any $D\in\mathcal F\otimes\mathcal F$; take $D=\Delta(C)$ (see Definition~5; \S 2.3.1 diagonal-embedding image above). Also $\boldsymbol T_\otimes$ and $\Delta$ commute: $\boldsymbol T_\otimes(\Delta(C))=\{(T(x),T(x)):x\in C\}=\Delta(T(C))$. Combining the two gives \begin{equation}
\mu_\otimes(\boldsymbol T_\otimes(\Delta(C)))=\mu_\otimes(\Delta(T(C)))=\mu_\otimes(\Delta(C))
\end{equation}. $\square$

\subsection*{B.56 Proposition~12 (single-layer joint integral and the single-measure barrier)}

\textbf{Proposition 12 (single-layer joint integral and single-measure barrier)}
Let $\Omega(\mu_2)\in\mathcal F$ be the integral domain constrained by $\mu_2$, $T$ an ideal invertible MPT, $f$ integrable. Define the \textbf{single-layer joint integral}
\begin{equation}
\mathcal I(f,T;\mu_1,\mu_2)\triangleq\int_{\Omega(\mu_2)}f(x,T(x))\,\mathrm d\mu_1(x)
\end{equation}
Then
\textbf{(a) single-measure barrier}: this clause holds for \textbf{any category's Hilbert} situation---for a differential measure, \textbf{if there is only one measure one cannot distinguish the pairing structure of the error subset $\mathcal E$ from the pairing structure of the signal subset $X\setminus\mathcal E$}---a single measure only outputs a volume scalar, encoding no pairing constraint. In particular when $\mathcal E$ is a $\mu_1$-null set (the typical case of discrete sample sets), $\mu_1(\mathcal E)=0$, and the error subset is completely invisible. \textbf{This is the formal source of the null-measure embedding ambiguity.}
\textbf{(b) volume representability of the quotient partition}: via $\mathcal I$, the quotient space $X/{\sim}$ obtains a volume-level measurable expression; volume is generated by multiplicity weight via bijection induction.
\textbf{(c) composite-map realization}: assuming a topological-level measurable bijection $T_0$ with no closed-form analytic realization, the composite gives a measurable realization $X\to X$ embedded into the product space; Leibniz point-wise differentiation is used only where extra differentiability holds, and the general case only retains the global differential quotient of Definition~1a.
\textbf{(d) non-measurability under single-differential assumption}: retaining only one differential structure (equivalent to a single-measure framework), the quotient partition corresponding to the second measure cannot be extracted, not measurable.
\textbf{(e) distinction of multiplicity weight and volume}: under $\Setcat$, the multiplicity weight attaches to a single point and does not support subset decomposition; the volume output by $\mathcal I$ is induced by the multiplicity weight and can be decomposed and quantified on any disjoint measurable subset.

The single-layer joint integral is only used to argue the \textbf{necessary condition} for dissolving the null-measure embedding ambiguity, not sufficient: it does not modify the base-space axiom and cannot assign non-zero volume to an isolated point; it must combine with the element replica-index structure and rely on the Chapter~4 triple-nested fiber-bundle integral (see Definition~16; \S 4.1) to couple the replica-index measure and the geometric measure through the inner, middle, and outer three-layer fibers into a product-space joint integral, thereby completing the sufficient condition.

\noindent\textbf{Proof} (see B.58) (five sub-proofs).

\subsection*{B.57 \S 3.2 three basic premises (locally-compact condition, translation-invariant reading, product-integral-invariance constraint)}

This section's construction rests on three basic premises, whose content and applicable domain are as follows.

\begin{itemize}
\item \textbf{Locally-compact condition}: the base space $G$ is locally compact Hausdorff; local compactness guarantees the existence of an exhaustion sequence by compact sets, integrals over compact sets are finite, and is the source of integral computability.
\item \textbf{Translation invariance (\Setcat-layer dimension-raising reading, not a $\Quot$-layer structure-induced property)}: the partition category sets no point elements, and ``translation'' is not expressible at the $\Quot/\MorTop$ layer---what truly exists there is block-permutation invariance. \textbf{Lifting to the Hilbert differential-topological layer does not automatically give quotient-translation invariance}: the holding of quotient-translation invariance is read out only after the categorical lift-equipment to \Setcat---Euclidean dimension-raising makes some measure differences disappear (the block-counting pre-pushforward by a block bijection is invariant, $\sigma_\#\nu_{\mathrm{idx},n}=\nu_{\mathrm{idx},n}$, Lemma~1(i)), and the \Setcat\ layer then presents the coordinate-translation-invariant reading (after normalization it is the Haar reading). Hence translation invariance is a \Setcat-layer \textbf{dimension-raising reading}, not a $\Quot$-layer structure-induced property, nor a premise; its point-wise consistency is premised on Assumption~1 (see \S 2.4.3 the $\epsilon$-$\delta$ structure of $\mathcal H^{\Setcat}$ above), i.e.\ output Postulate~1; the Lebesgue-side translation-invariant reading is a geometric-side property, and its deviation from the counting side is accounted into the first-order error $\merr$ (see Definition~5; \S 2.3.1 above), not entering the assumption set. The $\Hinf$ layer admits no non-trivial translation-invariant (Lebesgue-type) measure (see \S 3.2 boundary conclusion, below), and translation invariance never becomes a global premise.
\item \textbf{Product-integral-invariance constraint} (see Proposition~9b recital; Fubini standard form see C.2.6): based on this constraint one can construct the dynamical normalization, i.e.\ F\o lner compact-exhaustion averaging, guaranteeing on the \textbf{finite-dimensional truncation $Z_m$} that the multiplicity weight does not escape the measure space. \textbf{Applicable domain}: product-integral invariance and F\o lner normalization are used only on the finite-dimensional truncation $Z_m$ (Fubini / Tonelli require $\sigma$-finite measurable spaces, while the $\Hinf$ layer admits no $\sigma$-finite Borel measure that is translation-invariant and positive on open sets, see \S 3.2 boundary conclusion); the $\Hinf$ layer uses only the cylindrical-kernel compatible family of (f), and this paper \textbf{does not claim} that F\o lner can resolve the global integral divergence of the infinite-dimensional / infinite-partition case.
\end{itemize}

\subsection*{B.58 Proof of Proposition~12 (single-layer joint integral and the single-measure barrier, five sub-proofs)}

The five parts respectively handle the single-measure barrier, quotient-block volume reading, composite-map realization, single-differential non-extractability, and multiplicity--volume distinction.

\noindent\textbf{Proof (see Proposition 12)} (1) \textbf{Single-measure barrier}. The evaluation measure of $\mathcal I(f,T;\mu_1,\mu_2)$ is only $\mu_1$ (the integrand appears with respect to the second variable only through $T(x)$, introducing no structural reading of $\mu_2$). By the \S 2.1 level-composition convention and Definition~4's first existence (see \S 2.2.4 above), $\mu_1$ as a base-space measure only gives a volume-scalar reading for subsets: any $A,B\in\mathcal F$ that are $\mu_1$-equivalent yield an indistinguishable single-measure output. In particular, when $\mathcal E$ is a $\mu_1$-null set (see \S 2.4.3 the typical case of discrete sample sets, Problem~2.25), $\mu_1(\mathcal E)=0$, and $\mathcal I$ outputs the same for the error subset and the signal subset---the distinction of pairing structure is invisible under a single measure, which is the formal source of the null-measure embedding ambiguity.
(2) \textbf{Quotient-partition volume representability}. Let $X/{\sim}=\{B_w\}_{w\in\{0,1\}^n}$ be the $n$-th layer quotient partition of $\betat$ (see Lemma~4; \S 2.4 above). Taking $f$ as the indicator of a quotient block (or a constant on a quotient block), $\mathcal I$ reads the volume of that quotient block under the joint measure; by Proposition~12c (see \S 3.2; B.55), under invertible MPT this volume is invariant along $\boldsymbol T_\otimes$, and the multiplicity weight (the number of times an element is counted in the integral) is preserved one-to-one via bijection induction, hence the quotient space $X/{\sim}$ obtains a volume-level measurable expression.
(3) \textbf{Composite-map realization}. Measurability is closed under composition: $\mathcal F_\mathrm{dec},\mathcal F_\mathrm{enc}$ are modeled as ordinary continuous measurable maps (see \S 3.1 above), $T_0$ is measurable by assumption, hence $\mathcal F_\mathrm{enc}\circ T_0\circ\mathcal F_\mathrm{dec}$ is a measurable realization $X\to X$; embedded into the product space $X\times X$ via $\boldsymbol T_\otimes(x,y)=(T(x),T(y))$ (see Definition~5; \S 2.3.1 above). $T_0$ having no closed-form analytic realization does not imply point-wise differentiability; Leibniz point-wise differentiation via the chain rule is used only where extra differentiability holds, and the general case uses the differential--predicate bijection tower $\betat$ of Definition~1a (see \S 2.2.2 above) to give the global differential (the composite map sends the $\beta_n$-layer quotient block to the $\beta_n$-layer quotient block); the two differentials respectively relate to $\mu_2$ (replica side) and $\mu_1$ (base side), and are first-order equal by Corollary~2 (see \S 2.2.2; Postulate~2 above; output).
(4) \textbf{Non-measurability under single-differential assumption}. Retaining only one differential structure, the quotient partition has only one $\betat$-tower reading, and the second measure $\mu_2$'s corresponding quotient block has no distinguishable predicate support; by Corollary~10a(iii) (see \S 2.4.2 above), mutual exclusion is derived from structural disjointness, independent of measure choice---lacking the second partition boundary, the second measure's corresponding quotient partition cannot be extracted, hence not measurable.
(5) \textbf{Distinction of multiplicity weight and volume}. At the \Setcat\ layer, a point element is a single-point object (see \S 2.1 level-composition convention), the multiplicity weight attaches to a single point and cannot be directly summed over subset decomposition; the volume output by $\mathcal I$ is generated by the multiplicity weight via bijection induction ((2)), and is a subset reading, hence can be decomposed and quantified on any disjoint measurable subset by $\sigma$-additivity. $\square$

\subsection*{B.59 Theorem~2 (uniqueness of Euclidean-normalized Fourier correspondence)}

\textbf{Theorem 2 (uniqueness of Euclidean-normalized Fourier correspondence)}
Take $G=\mathbb R^d$, fix the self-dual identification $\hat G\cong\mathbb R^d$, character $\chi_\xi(x)=e^{2\pi i\xi\cdot x}$ and frequency normalization. Then the convolution--pointwise-product correspondence given by $\chi_\xi$ and the $L^1$--$L^2$ Fourier--Plancherel formula is uniquely determined. Changing the character pairing or frequency scale gives another normalization, not within this theorem's uniqueness domain; this theorem does not generalize to the modular-category automorphism uniqueness of $G\to\hat G$ on a general LCA group, nor does it specify the replica-index measure. Standard-theorem adaptation (see C.2.4).

\subsection*{B.60 Theorem~3 (compatible linear dimension-raising: volume form, standardness, and limit)}

\textbf{Theorem 3 (compatible linear dimension-raising: volume form, standardness, and limit)}
Let $G^{(n)}\cong\mathbb R^n$, and \textbf{assume} each $\varphi_n:G^{(n)}\to G^{(n+1)}$ is a continuous injective group homomorphism compatible with the refinement tower; on finite-dimensional Euclidean groups this is equivalent to an injective linear embedding. Further assume the product invariance of Proposition~9b and the dynamical normalization hold. (a) The $n$-dimensional Haar volume factor is $\sqrt{\det(D\varphi_n^{\mathsf T}D\varphi_n)}$, and there exists an Euclidean equipment compatible along the tower. (b) The standard Euclidean embedding is unique modulo $O(n{+}1)$ as a standardness premise of Assumption~1, not a conclusion of product invariance. Denote by $\Hinf$ the \textbf{completion} of this compatible direct limit under the inner-product norm; the direct limit itself is generally only a pre-Hilbert space. The infinite-dimensional layer admits no non-trivial translation-invariant Lebesgue-type measure, and the translation-invariance argument is limited to finite truncations. Proof (see B.61).

\subsection*{B.61 Proof of Theorem~3 (see \S 3.2)}

\textbf{Proof}
(a) A continuous group homomorphism $\varphi_n:\mathbb R^n\to\mathbb R^{n+1}$ must be a real linear map; injectivity is equivalent to $D\varphi_n$ having full column rank. Hence the scaling factor of the $n$-dimensional Haar volume on its image relative to the source coordinate volume is
\begin{equation}
\sqrt{\det(D\varphi_n^{\mathsf T}D\varphi_n)}
\end{equation}
where $D\varphi_n$ is an $(n+1)\times n$ non-square matrix, and the ordinary determinant cannot be taken. Product invariance and dynamical normalization fix the compatibility constraint of the volume reading, and the refinement-compatibility condition fixes the inter-layer commutative diagram; together they give the Euclidean equipment compatible along the tower. Linearity is a property of the continuous-injective-group-homomorphism premise, not an independent conclusion of product invariance.

(b) Taking the standard (isometric) embedding is a normalization convention on the solution side (see the explicit assumption of Theorem~3), \textbf{not belonging to} (Assumption~1); any two isometric embeddings differ by an orthogonal transformation, hence the standard Euclidean embedding is unique modulo $O(n{+}1)$. Block-bijection invariance and product invariance can only constrain the volume form, not exclude non-Euclidean norms or continuous scale freedom, hence cannot uniquely determine the embedding by them---the embedding's uniqueness comes from the isometric convention itself, not from some ``standardness subclause'' of Assumption~1, which has no such subclause.

\noindent\textbf{Limit and boundary} Compatible embeddings give the direct limit $\varinjlim\mathbb R^n$ and inherit the inner product; this space is generally incomplete, hence define $\Hinf$ as its completion under the inner-product norm. An infinite-dimensional separable Hilbert space admits no non-trivial $\sigma$-finite translation-invariant Borel measure positive on open sets, hence Proposition~9b and the translation-invariant reading are used only on finite truncations. $\square$

\subsection*{B.62 Proposition~13}

\textbf{Proposition 13 (pairwise product invariance of multiple groups)}
Let $\{G_k\}_{k=1}^n$ be a family of LCA groups, each satisfying the product-integral invariance of Proposition~9b. For any $i,j$, if $G_i,G_j$ are topologically isomorphic, then there exists a correspondence preserving product-integral invariance.

\noindent\textbf{Proof}
Each $G_k$ by Theorem~2 has a Fourier correspondence $\mathcal F_k:G_k\to\hat G_k$. When $G_i\cong G_j$, take the composite $\varphi_{ij}=\mathcal F_j^{-1}\circ\mathcal F_i$; the composite operation preserves product-integral invariance (see Proposition~9b each step respectively satisfies it after composition), obtaining the target correspondence. This proposition is the multi-group generalization of Theorem~3, providing the basis for the product invariance of each dimension's component operator in \S 5.3 (b). $\square$

\subsection*{B.63 Propositions~14--16 and 17 (measurability premises and integral-simplification family)}

This group is directly derived from the invertible-MPT structure of Definition~12 (see \S 3.1) and Definition~15 (see \S 3.3) together with Proposition~9b. \textbf{Proposition 14 (measurability premise)}: $T$ and $\overleftarrow T$ guarantee that $\mu_\otimes$ is measurable under the differential-topology--replica-index structure; the minimal measurable element of the integral inner layer can be measured by $\overleftarrow\mu_\otimes$. This proposition guarantees the measurability premise of integral evaluation: $\mu_\otimes$ and $\overleftarrow\mu_\otimes$ are defined on the same measurable space. \textbf{Proposition 15 (integral simplification on the bijection sub-domain)}: restricted to the bijection sub-domain $\mathcal X_\mathrm{biject}$ (quotient-block bijection reading of the partition category, see proof), the double-nested fiber-bundle integral degenerates to an ordinary double integral, and the built-in pullback degenerates to the identity transformation: $\overleftarrow T\mu_\otimes=\mu_\otimes$. \textbf{Proposition 16 (variable substitution under ideal invertible MPT)}: let $T$ be an ideal invertible MPT, then $\overleftarrow T$ is equivalent to the standard product preimage pullback, and the variable-substitution formula holds.

\noindent\textbf{Proof (Proposition 14)}

\noindent\textbf{Proof}
By Definition~12 (see \S 3.1 above), $T$ is a measurable bijection, with measurable inverse and measure-preserving; hence $T$ and the built-in pullback $\overleftarrow T=\boldsymbol T^{-1}_{\otimes\#}$ (see Definition~14 ideal invertible MPT special case, above) are both $\mathcal F\otimes\mathcal F$-measurable. Pullback preserves the Borel structure: the preimage of a measurable map preserves Borel sets (see Definition~13; \S 3.1 the shared-measure derivation passage $\mathcal G_T$ is exactly constructed from this pullback, above), hence $\mu_\otimes$ and $\overleftarrow\mu_\otimes$ are defined on the same measurable space, and $\mu_\otimes$ is measurable. The minimal measurable element of the integral inner layer: by Definition~1a (see \S 2.2.2 above), the differential-topological element $\mathrm dx$ of $\Hinf$ falls via $\sim_\beta$ into the quotient block marked by $\beta_n(\mathcal D)$, and the generator of the quotient block is the rank-1 minimal non-empty topological differential subset $\mathcal D$; this generator is measurable under $\overleftarrow\mu_\otimes$, hence the minimal measurable element of the integral inner layer can be measured by $\overleftarrow\mu_\otimes$. $\square$

\noindent\textbf{Distinction from ordinary Cartesian double integral}: the two layers share a base point $x$, and the coupling constraint comes from $T$, the element replica-index structure, and $\overleftarrow T$. The nested fiber-bundle integral has \textbf{order at least two}, with no single-nested fiber-bundle integral. The $\mathcal I(\cdot)$ in Proposition~12 is uniformly named the \textbf{single-layer joint integral}, serving only as an auxiliary tool for the necessary condition.

\noindent\textbf{Proof (Proposition 15)}

\noindent\textbf{Proof}
This proof proceeds on the \textbf{partition category} ($\Quot/\MorTop$ layer): the bijection reading on $\mathcal X_\mathrm{biject}$ is a \textbf{quotient-block bijection} (the quotient-layer bijection is given by Theorem~1 special case (d) under no-collapse and equal-cardinality conditions (see \S 2.3.3); point-wise measurable bijection is the \Setcat-layer reading, generally not holding, Definition~12 note), $\mu_1,\mu_2$ take partition-induced measures (block-counting measure $\nu_{\mathrm{idx},n}$, finite partition). Under a finite partition there is \textbf{no null block} (see Proposition~10 every block count strictly positive), hence in the partition category there is no zero-measure problem, needing no \Setcat-layer measure-preservation assumption. By Proposition~9a (see \S 2.4 above), the block bijection induces a permutation $\sigma\in S_{2^n}$ on the block set, $\sigma_\#\nu_{\mathrm{idx},n}=\nu_{\mathrm{idx},n}$ (see Lemma~1(i)), hence $\boldsymbol T_{\otimes\#}\mu_\otimes=\mu_\otimes$ holds at the quotient-block level. The built-in pullback operator $\overleftarrow T$ coincides with the standard preimage pullback $\boldsymbol T^{-1}_{\otimes\#}$ under ideal invertible MPT (the ``distinction'' note of Definition~14 above), hence \begin{equation}
\overleftarrow T\mu_\otimes=\boldsymbol T^{-1}_{\otimes\#}\mu_\otimes=\mu_\otimes
\end{equation}. The two sides of the integral equality of Definition~15 have the same measure, and the double-nested fiber-bundle integral degenerates to an ordinary double integral, with the built-in pullback degenerating to the identity transformation. This proposition gives the degenerate form on the bijection sub-domain: the built-in pullback does nothing, and the integral returns to the standard form. $\square$

\noindent\textbf{Proof (Proposition 16)}

\noindent\textbf{Proof}
$T$ is an ideal invertible MPT under $(\mu_1,\mu_2)$ (see Definition~12; \S 3.1 above), and by Proposition~9b (see \S 2.4 above; using domain note) $\boldsymbol T_{\otimes\#}\mu_\otimes=\mu_\otimes$, i.e.\ $\boldsymbol T^{-1}_{\otimes\#}\mu_\otimes=\mu_\otimes$; also by the ideal invertible MPT special case (see Definition~14 above) $\overleftarrow T\mu_\otimes=\boldsymbol T^{-1}_{\otimes\#}\mu_\otimes$, hence $\overleftarrow\mu_\otimes=\mu_\otimes$. Substituting into Definition~15 (above) and applying the product variable substitution $u=T(x)$ to the measurable bijection $T$ (measure preservation absorbs the Jacobian factor: $\mu_\otimes(\boldsymbol T_\otimes^{-1}(C))=\mu_\otimes(C)$, Definition~13, \S 3.1, above), we get
\begin{equation}
\iint_{X\times X} f(x,T(x))\,\mathrm d\mu_\otimes=\iint_{X\times X} f(T^{-1}(u),u)\,\mathrm d\mu_\otimes
\end{equation}
which is the standard product variable-substitution formula. This proposition gives the equivalence with the standard variable substitution under ideal MPT. $\square$

\noindent\textbf{Proof (Proposition 17)}

\noindent\textbf{Proof}
Let the joint measure be a finite discrete atomic measure $\mu_\otimes=\sum_{i,j}w_{ij}\,\delta_{(x_i,x_j)}$ (atomic counting replica number $w_{ij}>0$ (index-measure reading); the atom is the \Setcat-layer point-element reading of the quotient-layer discrete reading after lift-equipment, i.e.\ the additional premise of Proposition~17's statement; the counting replica-number reading belongs to the product-space joint measure, not the base-space $X$ measure, compatibility verification see end of \S 3.3, above), and the model pairing is given by the index matrix $F=(f_{ij})$ ($f_{ij}$ is the integrand reading on $(x_i,x_j)$). Substituting into Definition~15 (above), the integral degenerates on the atomic measure to a sum over counting replica numbers
\begin{equation}
\iint_{X\times X} f\,\mathrm d\mu_\otimes=\sum_{i,j}w_{ij}f_{ij}=\langle W,F\rangle_F
\end{equation}
where $W=(w_{ij})$, $\langle\cdot,\cdot\rangle_F$ is the Frobenius inner product, i.e.\ Frobenius tensor contraction. Index split: by the behavior of $T$, split the index set into the bijection-subdomain indices $I_\mathrm{biject}$ (see Definition~4; \S 2.2.4 $T$ restricted to it is a bijection, second-existence side above) and the error-subset indices $I_\mathcal E$ (see Definition~5; \S 2.3.1 carrying pairing deviation above); the two form a disjoint union, hence the sum splits into two independent sums: \begin{equation}
\langle W,F\rangle_F=\sum_{(i,j)\in I_\mathrm{biject}}w_{ij}f_{ij}+\sum_{(i,j)\in I_\mathcal E}w_{ij}f_{ij}
\end{equation}. This proposition gives the equivalence of the integral and Frobenius contraction in the finite-discrete case, and is the basis for index splittability. $\square$

\subsection*{B.64 Proposition~17 (finite-discrete case: integral equivalent to tensor multiplication)}

\textbf{Proposition 17 (finite-discrete case: integral equivalent to tensor multiplication)}
When the joint measure is a finite discrete atomic measure, and the atom $\delta_{(x_i,x_j)}$ is the \Setcat-layer point-element reading obtained by lift-equipment from the $\Quot/\MorTop$ discrete reading (\textbf{additional premise (lift-equipment)}), the integral degenerates to Frobenius tensor contraction; the index set splits into the bijection-subdomain indices $I_\mathrm{biject}$ and the error-subset indices $I_\mathcal E$, and the sum is splittable.

\noindent\textbf{Proof} (see B.63).

\subsection*{B.65 Properties of Construction~1: inter-layer transformation rules}

Each layer's computation result stays in that layer's measure space; cross-layer operations only allow outward pushforward propagation and inward pullback reconstruction; the map output is redefined as a $\sigma$-finite Borel measure on the target space, participating in subsequent integral operations. This construction obeys the ``MPT topological existence, analytic not necessarily reachable'' dichotomy (see \S 3.3; Definition~4), all core operations are defined on the product joint measure space, avoiding direct operation on the base measurable space's two native measures' preimage intersection; under the ideal invertible-MPT premise it is self-consistent with Corollary~10a, and when the invertible-MPT premise is missing the avoidance mechanism fails.

\noindent\textbf{Evaluation order and measure-recovery positioning (how the three layers evaluate, how counting recovers)} This paragraph fixes the evaluation order of the triple-nested fiber-bundle integral, the common basis of the \S 6.2.3 tensorization side and the \S 6.3 loss function. The order is uniquely determined by the categorical direction: along categorical support, lifting and descending are two paradigms---lifting is the functional nesting process (see Theorem~3 completed form in \Setcat, single step is standard Euclidean embedding), descending is the predicate cumulative-error process (see Theorem~1; \S 5.2 predicates reside in $\Quot$, pushout--pullback well-defined on $\MorTop$); the two paradigms belong to different categories, hence the evaluation order is not exchangeable. The positioning of counting-measure recovery in this order is \textbf{the matching of measure maps and the equivalence of measure values}---$\boldsymbol\pi_\#$ and $\boldsymbol\pi^\#$ are mutually inverse pairings (see Definition~24; \S 5.3); the explicit form under the discrete-atomic limit is developed in \S 6.2.3.

\noindent\textbf{Measure-descent formula (forward propagation, lifting paradigm)}---the finite quotient-partition measure is descended via pushforward $\boldsymbol\pi_\#$ into a copy measure of the middle-outer composite, and the three-layer integral uses this as carrier:
\begin{equation}
\nu_{\mathrm{idx},n}=\frac1{2^n}\sum_{i=1}^{2^n}\delta_{B_i}
\end{equation}
(where $\delta_{B_i}$ is the \textbf{indicator measure} of the $i$-th quotient block $B_i$, the same notation as \eqref{eq:11} in \S 6.4.1 and \S 6.4.2---uniformly the atomic constituent of the block-counting measure in the whole paper, whose normalization is exactly $\nu_{\mathrm{idx},n}$, Definition~17; see below)
\begin{equation}
\begin{split}
\Big\{\big(\Omega_\nu,\mathrm d\nu_\mathrm{idx}\big)\Big\}
&\xrightarrow{\ \boldsymbol\pi_\#\ }
\Big\{\big(\Omega_\nu,\mathrm d\nu_\mathrm{idx}\big),\\
&\qquad\big(\Omega_\lambda\times\Omega_\mu,\ \mathrm d\lambda,\ \overleftarrow{\mathrm d\nu}_\mathrm{idx}\big)\Big\}
\end{split}
\end{equation}
\begin{equation}
\Lambda=\int_{\Omega_\mu}\Big[\int_{\Omega_\lambda}\big(\int_{\Omega_\nu}F(\nu; x, T(x), t)\,\mathrm d\nu_\mathrm{idx}(\nu)\big)\,\mathrm d\nu_\mathrm{comp}(\lambda,\overleftarrow{\nu})\Big]\,\mathrm d\nu_\mathrm{Leb,idx}(\mu)
\end{equation}
Here both measures are read by their explicit definition, and the outer enumeration takes the middle-layer domain as the integral domain (see Definition~16 above).

\noindent\textbf{Measure-recovery formula (backward propagation, descending paradigm)}---equip the inner layer with the binomial measure, collapsing the middle-outer multi-integral into a single integral via pullback $\boldsymbol\pi^\#$ (i.e.\ $BP$), and the counting measure reaches the Hilbert-layer zero-residue point (Theorem~6, \S 4.3):
\begin{equation}\begin{gathered}
\Lambda\xrightarrow{\ \boldsymbol\pi^\#\ }\int F\,\mathrm d\nu_M\\
\nu_M\ \text{reaches}\ \mathit S_0=\{\boldsymbol 0\}
\end{gathered}\end{equation}
Reading the two formulas together gives this paragraph's order conclusion: \textbf{the middle layer must be integrated with the outer layer first; one cannot recover only the middle-layer measure without recovering the outer-layer measure} (meaning the middle-layer Lebesgue component $\mathrm d\lambda$ must be integrated together with the outer $\mathrm d\mu_\mathrm{gen}$; the recovery of the replica-index-counting measure $\nu_M$ follows the division of labor in the next sentence)---measure recovery is backward propagation (see Definition~24; \S 5.3 inward integral flow), with fixed direction. \textbf{The exchange division of labor of the two layers}: the Lebesgue measure (middle-layer $\mathrm d\lambda$ and outer-layer $\mathrm d\mu_\mathrm{gen}$) \textbf{cannot be recovered alone}, its integral object is \textbf{differential Lebesgue}, the integral \textbf{does not commute} differential numerical value and topology, carrying first-order error; the replica-index-counting measure (middle-layer descended copy $\overleftarrow{\mathrm d\nu}_\mathrm{idx}$, i.e.\ $\nu_M$) \textbf{can be recovered alone without error}---its meaning is to obtain the zero-error (no first-order error) zero-value saddle-point-boundary-anchored partition predicates, making the loss functional directly characterize ``binary quotient partition $\to$ bijection $\to$ first-order shared-measure-space volume error''. The middle layer $\Omega_\lambda$ descends the replica-index-counting measure (evaluation measure $\nu_\mathrm{comp}=\lambda\otimes\overleftarrow\nu_\mathrm{idx}$, first multiplying differential Lebesgue), \textbf{then performs the overall topological--differential numerical exchange} (see Proposition~20; \S 5.2 swap), and when recovering the measure adds the reverse equivalence-class sign, obtaining the numerical value of the \textbf{topological volume (differential-volume sum)}; this transformation descends the first-order-error measure-space volume via categorical support into the \textbf{zero-replica-index-counting-measure partition-predicate error} (a partition category, not a zero-Lebesgue-error one). Recovery \textbf{preserves posterior predicates without error} ($\mathcal L_\mathrm{cat}(\mathit S_0)=0$, Proposition~22, \S 5.4), but \textbf{gives no improvement for the prior (functional)}: the counting measure is simultaneously prior and posterior (see \S 6.2.2 the $\mathrm{epoch}+1$ counting measure), and prior equals posterior is a structural reading at the measure level, essentially with no room for improvement.

\subsection*{B.66 Theorem~4 ($\Hinf$'s bijection preserves volume and first-order-error closure)}

\textbf{Theorem 4 ($\Hinf$'s bijective volume-preservation and first-order-error closure)}
Let the factor spaces satisfy Lemma~2 (B.24), Theorem~1, and the measurable measure-preserving-bijection property. Form the product dimension-raising of the replica-index fiber and the sample geometric space to obtain $\Hinf$. Then
\begin{enumerate}
\item[(a)] $\Hinf$ inherits the globally measurable volume-preserving bijection structure;
\item[(b)] product dimension-raising does not change the first-order measurable property of the error;
\item[(c)] $\Hinf$ internally generates no high-order error.
\end{enumerate}

Here ``\textbf{preserves volume}'' means \textbf{fiber-bundle volume conservation along the $\boldsymbol\pi_\#/\boldsymbol\pi^\#$ passage}, \textbf{not} that $\Hinf$ admits a global translation-invariant (Lebesgue-type) measure---the latter does not exist (see \S 3.2). Theorem~4 and \S 3.2 state two different things and do not conflict.
\noindent\textbf{Proof (sketch)}
(a) The product map $T_\mathrm{prod}=(T_\nu,T_\mu)$ has measurable volume-preserving components, hence the product map is a measurable volume-preserving bijection; (b) the four-step error localization (first-order error localized in the shared measure space, middle-layer counting component zero error, error variation in the outer layer with the pullback--pushforward passage measure-preserving, binary refinement measure-preserving with first-order reading difference)---product dimension-raising changes no measure structure; (c) the two structural-deviation sources of high-order error (see Definition~6 categorical leakage, topological collapse) do not appear. Full proof (see B.67).

\subsection*{B.67 Proof of Theorem~4}

\textbf{Proof}
(a) Let $(\Omega_\nu,\mathcal F_\nu,\nu_\mathrm{idx})$, $(\Omega_\mu,\mathcal F_\mu,\mu_\mathrm{gen})$ each bear an invertible-MPT bijection; construct the product map $T_\mathrm{prod}=(T_\nu,T_\mu)$, with measurable volume-preserving components, hence the product map is a measurable volume-preserving bijection. Product dimension-raising only performs set and $\sigma$-algebra topological splicing, adding no singularity or non-measurable set, hence $\Hinf$ inherits the globally measurable volume-preserving bijection structure. (b)(c) The derivation proceeds in the partition-predicate category (i.e.\ the element-free partition-category reading, definition see B.6), all readings taken as quotient-block / predicate readings; the \Setcat-layer point-wise conclusion is read out by lift-equipment. Localize the error in four steps, then give the basis of (b)(c).

\noindent\textbf{Step 1 (error localization: first-order error comes from the shared measure space)} (element-free partition-category reading) the first-order error is not assumed on any fiber, but localized: the non-null-measure transference introduced by cross-category lift-equipment falls on the Lebesgue measure and counting measure's shared measure space co-constructed via invertible MPT ($\mu_\mathrm{share}=\mathcal G_T(\mu_1,\mu_2)$, Definition~13, \S 3.1)---this is the unique passage for the two measures' pairing fusion (construction fact 10a: shared measure can only be obtained via pairing fusion through invertible MPT, not via intersection fusion). Hence the first-order error $\merr$ (see Definition~5; \S 2.3.1) has as its domain the gap between the block-counting reading and the geometric reading on the shared measure space, not a point-wise property of any fiber; to call some fiber's ``error first-order measurable'' apart from this localization is a meaningless assertion.

\noindent\textbf{Step 2 (middle-layer counting component error is 0)} (element-free partition-category reading) in the triple-nested fiber-bundle integral (see Definition~16; \S 4.1), the integral landing point of the replica-index measure $\nu_\mathrm{idx}$ is the \textbf{middle-layer} joint-conservation fiber $\Omega_\lambda$ (carrying $\mathrm d\lambda$ and the descended copy $\overleftarrow{\mathrm d\nu}_\mathrm{idx}$, Definition~16); its reading is the block count (\begin{equation}
\nu_{\mathrm{idx},n}(A)=\frac1{2^n}\#\{i:B_i\subseteq A\}
\end{equation}, Definition~17, \S 4.2), mutually exclusive block covering guarantees a single count, and any measure counts each block only once (see Lemma~1(i): $\sigma_\#\nu_{\mathrm{idx},n}=\nu_{\mathrm{idx},n}$, quotient-translation-invariant measure preservation, \S 2.4 measure-preservation bearer note). The block-counting reading contains no geometric-volume gap, hence the \textbf{counting component error of the middle layer is 0}---the error cannot hide in the replica-index-measure layer (the gap attribution of the middle-layer Lebesgue component $\mathrm d\lambda$ is in Step 3).

\noindent\textbf{Step 3 (error varies to the outer layer, and the pullback--pushforward passage preserves measure)} (\Setcat-layer reading) the three-layer integral is evaluated along the lifting passage (see Definition~16's ``evaluation order and measure-recovery positioning'' paragraph, \S 4.1): outward pushforward $\boldsymbol\pi_\#$ presses the inner replica index into the middle-outer composite, inward pullback $\boldsymbol\pi^\#$ reconstructs the preimage; the middle-layer reading becomes a \Setcat-layer point element only after the predicate error has been cumulated on the $\Quot$ side and lifted-equipped through the outer $\Omega_\mu$'s $L'$ (``measure-recovery order'' paragraph, \S 4.1). \textbf{Distinction of where it resides and where variation executes}: the middle-layer $\mathrm d\lambda$ component resides on the Lebesgue side of the shared measure space (same side as $\mathrm d\mu_\mathrm{gen}$), while the \textbf{execution} of variation is in the outer layer $\Omega_\mu$---by the evaluation order, the middle-layer reading must be lift-equipped through $L'$ to become a \Setcat-layer point element; residing and executing are two layers; hence the middle-layer counting-component zero error of Step 2 does not preclude this gap from appearing on the outer-layer geometric reading. By Step 2, all gaps on the shared measure space can only \textbf{vary to the outer layer} $\Omega_\mu$ (carrying $\mathrm d\mu_\mathrm{gen}$). The outer-layer operation is exactly the triple-fiber-bundle integral's defined \textbf{pullback preimage then pushforward to the measure space}: pullback $\boldsymbol\pi^\#$ takes the preimage (see Definition~24; \S 5.3 $\boldsymbol\pi_\#,\boldsymbol\pi^\#$ mutually inverse pairings), pushforward $\boldsymbol\pi_\#$ preserves the block share (see Lemma~1(i)), hence this passage is \textbf{measure-preserving}---variation does not change the total measure value, only moves the gap onto the outer-layer geometric reading; dimension-raising also does not change this gap's bound: under product dimension-raising the two factor readings each stay (Step 1 localization, Step 2 middle-layer zero error), and the gap accumulates block by block by block count, its bound unchanged.

\noindent\textbf{Step 4 (binary and measure-preserving derive first-order error)} (element-free partition-category reading) binary refinement (see Definition~7; \S 2.3.2) is measure-preserving at the quotient-block level at each step (see Lemma~1(i): pre-pushforward by a block bijection invariant), and the geometric gap on the shared measure space is cut into per-block readings on the finite quotient partition; by Corollary~2's first-order equality ($\mathrm dx\odot\nu_\mathrm{idx}\sim_1\mathrm d\#$, output Postulate~2, \S 2.2.2), the block-counting reading and the Lebesgue geometric reading differ by first order, falling within the absorbable first-order measurable error $\merr$, generating no high-order term (accounting caliber: first-order error is exactly the mismatched partition-predicate area of the zero-value saddle-point split and the outer non-zero saddle-point split, \S 6.2.2 first-order-error accounting caliber). Hence (b) holds---product dimension-raising does not change the first-order measurable property of the error, derived from Step 2 (middle-layer counting-component zero error), Step 3 (varies to outer layer and measure-preserving), and Step 4 (binary measure-preserving plus first-order equality), not a premise assumption.

\noindent\textbf{Basis of (c) (dimension-raising generates no new-type error)}: by Steps 1--3, a measurable error must be a measure-structure change of the shared measure space, and product dimension-raising changes no measure structure: middle-layer counting-component zero error (Step 2), measure-preserving variation passage of the gap (Step 3), binary refinement measure-preserving and reading difference first order (Step 4). The two structural-deviation sources of high-order error---categorical leakage and topological collapse (the binary-quotient-partition category is broken, degenerating into multi-partition, with minimal cross-partition quotient blocks appearing) (see Definition~6; \S 2.3.1)---are both absent in dimension-raising: pure dimension-raising does not compress dimension (no topological-collapse-type multi-partition break), nor passes through non-quotient-bijection lifting (see Proposition~19; \S 5 categorical leakage excluded); \textbf{pure dimension-raising layer generates no high-order error endogenously (see Theorem 4 this theorem's conclusion)}. Hence $\Hinf$ internally cannot generate high-order error, and (c) holds. This theorem gives the basis that the pure dimension-raising layer generates no high-order error endogenously, the first link of the whole paper's high-order-error exclusion chain (see Definition~6; \S 2.3.1 closing the two sources beyond which is the dimension-raising passage). $\square$

\subsection*{B.68 Theorem~5 (replica continuous-measure-map compatibility)}

\textbf{Theorem 5 (replica continuous-measure mapping compatibility)}
The measure readings and integral variables of the three-layer fibers finally land in $\mathbb R$ (i.e.\ the reading / value range after lift-equipment to the \Setcat\ layer; the inner layer still does not derive point elements at the $\Quot/\MorTop$ primitive layer, Definition~16); relying on non-infinite-integral-bound partition bijection / non-bijection sub-regions; all measure values land in $\mathbb R$. Under global normalization, at the replica infinite-partition limit, the compatibility reading between the replica-level measure system and the universal measure $\mu_\mathrm{gen}$ has as its \textbf{substantive content}: the projective limit of $\{\nu_{\mathrm{idx},n}\}$ and the Helly weak limit give the \textbf{same limit} $\nu_{\mathrm{idx},\infty}$ on the Cantor-type environment $\mathcal C_\beta$ (the two routes agree); the same $[0,1]$ integral value range and unit total mass are only accompanying properties, not the information content of compatibility (any two probability measures share value range and total mass). This compatibility is established at the $\MorTop$ level; this theorem \textbf{does not claim} an order-preserving isomorphism of the integral value range (see \S 4.3 that needs the point-wise structure of the $\mu_\mathrm{gen}$ side, exactly what the boundary statement avoids); the order structure is preserved only under the quotient-layer reading, \textbf{not} an internal point-wise isomorphism of the derived-set category.

\noindent\textbf{Proof (sketch)}
Finite layer: $\nu_{\mathrm{idx},n}$ is a probability measure on $2^n$ atoms, forming a projective compatible family under the binary-iteration refinement tower; the limit layer gives the cylindrical compatible limit measure. Full proof (see B.69).

\subsection*{B.69 Proof of Theorem~5}

\textbf{Proof}
Finite layer: by Definition~17, $\nu_{\mathrm{idx},n}$ is a probability measure on $2^n$ atoms; under the binary-iteration refinement tower $\{\nu_{\mathrm{idx},n}\}_{n\ge1}$ forms a projective compatible family ($\nu_{\mathrm{idx},n}=\nu_{\mathrm{idx},n+1}\circ\iota_n^{-1}$, same construction as the compatibility argument of Definition~25a (see \S 6.4.1)), hence there exists a unique projective limit $\nu_{\mathrm{idx},\infty}$ on the replica index set. On the integral-value-range side: $\nu_{\mathrm{idx},n}(X)=1$ and the global normalization of $\mu_\mathrm{gen}$ give a uniformly bounded probability-measure family, and by the Helly selection principle taking a weakly convergent subsequence gives a compatible limit. \textbf{Why the two routes give the same limit (projective limit and Helly weak convergence)}: (element-free partition-category reading) the projective limit gives the unique cylindrical compatible measure on the Cantor-type environment $\mathcal C_\beta$ (see Definition~17; \S 4.2 limit-environment anchoring); the Helly route takes the weak limit of the compatible cylindrical-measure family on this same environment, and the weak limit is uniquely determined by cylindrical compatibility, hence the two routes give the same $\nu_{\mathrm{idx},\infty}$. \textbf{Order structure as quotient-layer reading (not claiming order-preserving isomorphism)}: an order-preserving isomorphism with $\mu_\mathrm{gen}$ needs point-wise structure, which this paper does not claim (see \S 4.3 boundary statement); the order structure is preserved at the quotient layer---the block share monotonically non-increases along the binary refinement tower (refinement does not increase the single-block share: for $m>n$, $2^{-m}<2^{-n}$, Lemma~4(2)), and the column share monotonically converges on $\mathcal C_\beta$, hence the limit reading preserves the usual order of the $[0,1]$ value range. This theorem gives the compatibility of the replica-level measure with the universal measure (projective limit and weak limit identical, plus global normalization), the basis for $\nu_{\mathrm{idx},\infty}$ being well-defined. $\square$

\subsection*{B.70 Definition~18 (dimension-raising integral)}

\textbf{Definition 18 (dimension-raising integral)}
Let $m$ be the iteration count of dimension-raising on the middle-layer joint-conservation fiber $\Omega_\lambda$, $m\ge1$: $m=1$ has no extra dimension-raising iteration, directly degenerating to the middle-layer integral term of the original bare triple fiber bundle; $m>1$ executes $m-1$ rounds of binary dimension-raising iteration; the dimension-raising integral acts entirely on the middle-outer composite, not modifying the inner-layer $\Omega_\nu$ structure. By the binary-iteration refinement of Definition~7: let $G\subset\mathbb R^d$, $G_q$ the finite quotient group obtained by $N$ binary refinements of $G$, $\hat G_q$ its Pontryagin-dual replica fiber. The \textbf{dimension-raising joint space} is
\begin{equation}
Z_m\triangleq G\times\prod_{k=1}^{m-1}\big(G_q^{(k)}\times\hat G_q^{(k)}\big)
\end{equation}
equipped with the joint product Haar measure $\lambda_\mathrm{joint}^{(m)}$. The \textbf{dimension-raising integral} is defined as
\begin{equation}
\mathcal I_\mathrm{lift}^{(m)}(\cdot)\triangleq\int\cdots\int\limits_{G\times\prod_{k=1}^{m-1}\big(G_q^{(k)}\times\hat G_q^{(k)}\big)}(\cdot)\ \mathrm d\lambda\prod_{k=1}^{m-1}\big(\mathrm d\nu_{q,k}^\mathrm{quo}\,\mathrm d\nu_{q,k}\big)
\end{equation}
where $\mathrm d\lambda$ is the base-space Lebesgue measure, $\mathrm d\nu_{q,k}^\mathrm{quo}$ the quotient-block measure of the $k$-th round binary quotient partition, and $\mathrm d\nu_{q,k}$ the counting Haar measure on the replica fiber $\hat G_q^{(k)}$. $m\to\infty$ corresponds to the infinite-round binary-iteration replica-dimension-raising limit, compatible with $\nu_{\mathrm{idx},\infty}$ (see Definition~17; Theorem~5); this definition is the object of the inner compactness of Proposition~18 (see \S 4.2 below).

\noindent\textbf{Assembly form of the middle-layer integral} Under the triple-nested fiber-bundle integral paradigm (see Definition~16; \S 4.1), the construction is equipped to the middle-layer joint-conservation fiber $\Omega_\lambda$, and $\mathcal I_\mathrm{lift}^{(m)}$ performs the binary-quotient-partition volume assignment; the middle-layer integral takes the ``one-fold--multi-fold--one-fold'' form: the outer layer enumerates the middle-layer domain, the middle instantiates the dimension-raising integral, and the inner imposes the joint-measure constraint:
\begin{equation}
\Lambda_\lambda=\int_{\Omega_\lambda}\Big[\,\mathcal I_\mathrm{lift}^{(m)}\big(F_\lambda(x,T(x),t)\big)\,\Big]\,\mathrm d\nu_\mathrm{comp}\,\mathrm d\nu_\mathrm{Leb,idx}
\end{equation}
Here both measures are read by their explicit definition, and the outer enumeration takes the middle-layer domain as the integral domain (see Definition~16).
Inter-layer maps obey the properties of Construction~1: outward pushforward $\mathrm d\mu_\mathrm{gen}=T'_\#(\mathrm d\lambda)$; inward pullback \begin{equation}
\Lambda'_\lambda=\int_{\Omega_\lambda\subset\mathbb R}\cdots\ T'^{-1}_\#(\mathrm d\mu_\mathrm{gen})
\end{equation}.

\subsection*{B.71 Proposition~18 (inner compactness of binary dimension-raising quotient partition)}

\textbf{Proposition 18 (inner compactness of binary dimension-raising quotient partition)}
This proposition runs on the $\MorTop$ middle-outer composite; $m\ge1$, $m=1$ corresponds to the bare triple integral. Let $G$ be a \textbf{bounded} Borel subset of $\mathbb R^d$ (or a general bounded locally compact Abelian group), equipped with Lebesgue measure $\lambda$, satisfying the following two construction conditions: \textbf{Condition 3a boundedness}---$G$ bounded, same source as Convention~4's measure-descent target assignment (Convention~4 is an assignment convention, not an assumption; its formalization and the ``necessity of boundedness'' see \S 4.3, below L548), guaranteeing $0<\lambda^{(m)}_\mathrm{joint}(Z_m)<\infty$; \textbf{Condition 3b locally-compact Hausdorff product closure}---$G$ locally compact Hausdorff, each $G_q^{(k)},\hat G_q^{(k)}$ a finite discrete Abelian group, the product $Z_m$ still locally compact Hausdorff (Haar measure on such spaces is $\sigma$-finite and inner compact, a standard fact). $G_q$ is the finite quotient group of $G$ after $N$ binary refinements, $\lvert G_q\rvert=2^N$; by the uniqueness of the lift--descent map (see Theorem~3, \S 3.2, above: unique modulo $O(n{+}1)$, not taking a unitary meaning for the square embedding), the Pontryagin-dual replica fiber $\hat G_q$ is set-theoretically well-defined. Take finite $m\ge1$, the dimension-raising joint space $Z_m$, use the dimension-raising integral $\mathcal I^{(m)}_\mathrm{lift}$; the joint product Haar measure satisfies $0<\lambda^{(m)}_\mathrm{joint}(Z_m)<\infty$. Then the volume-integral assignment of the $2^N$ binary quotient partition after dimension-raising has inner compactness with respect to $\lambda^{(m)}_\mathrm{joint}$.

\noindent\textbf{Resonance with normalization} Taking $G$ bounded is self-consistent with \S 4.3's ``sample input normalized to $[0,1]$''; this also guarantees $\lambda^{(m)}_\mathrm{joint}(Z_m)<\infty$.

\noindent\textbf{Proof} (see B.72) (inner compactness verification of the locally-compact Hausdorff product space: given by standard facts plus finite-partition recheck, introducing no new premise).

\subsection*{B.72 Verification proof of Proposition~18 (inner compactness of the dimension-raising joint space)}

\textbf{Proof}
$G$ is locally compact Hausdorff; each $G_q^{(k)},\hat G_q^{(k)}$ is a finite discrete Abelian group, the product $Z_m$ is still locally compact Hausdorff; $\lambda^{(m)}_\mathrm{joint}$ is the $\sigma$-finite product Haar measure on $Z_m$, satisfying inner compactness: \begin{equation}
\lambda^{(m)}_\mathrm{joint}(B)=\sup\{\lambda^{(m)}_\mathrm{joint}(K):K\subseteq B,\ K\text{ compact}\}
\end{equation}. Let the base-space binary quotient partition be $\{G_{q,i}\}$, and after dimension-raising the quotient block $B_i\triangleq G_{q,i}\times\prod_{k=1}^{m-1}(G_q^{(k)}\times\hat G_q^{(k)})$ forms a finite partition of $Z_m$; each $B_i$ is a Borel measurable set and inherits inner compactness. For any $A\subseteq Z_m$, let $A_i=A\cap B_i$, then $\lambda^{(m)}_\mathrm{joint}(A)=\sum_i\lambda^{(m)}_\mathrm{joint}(A_i)$; finite summation preserves the supremum operation, and the whole quotient-partition volume assignment is inner compact. $\square$

\subsection*{B.73 Proof of Theorem~6}

\textbf{Proof}
(a) Convention~1's architectural condition makes zero parameters produce zero data flow, and zero-output calibration makes that output precisely pair with the target predicate, hence all $r_i=0$, $\mathcal L(\boldsymbol0)=0$; by $\mathcal L$ Borel measurable, $\mathit S_0=\mathcal L^{-1}(\{0\})$ is Borel measurable. (b) Non-negativity gives global minimality; at points where the gradient exists, minimality gives zero gradient. (c) Directly holds by the definition of $\mathit S_0$ and $r_i\ge0$. (d) In Convention~3's local gradient-existence domain, zero data flow and the chain rule give per-sample zero gradient.

(e) By (d), $N(\boldsymbol0)=|\mathcal X|$, and functional equivalence does not change the per-sample gradient, hence the entire $[\boldsymbol0]$ attains the maximum. The reverse implication is exactly Premise~1: if $N(\theta)=|\mathcal X|$, then $\theta\in[\boldsymbol0]$. Hence uniqueness is a \textbf{conditional conclusion} (point-level reading), not derivable solely from a finite partition having no zero-measure block, loss non-negativity, or measurability. (f) is the equivalent rewriting of (e) for per-sample stationary points. (g) Holds by the definition of the EDE exact-pairing set.

(i) \textbf{Differential-reading uniqueness (not premised on Premise~1)}: \textbf{layer-by-layer construction}---the $n$-th layer $\mathrm dx_{0,n}$ takes the quotient block $B_{0^{n}}$ with predicate word $w=0^{n}$, i.e.\ the quotient block where the zero-value output reading resides (see Convention~1's zero-output calibration giving the output space's single point $\{0\}$); the block membership uses the \textbf{half-open block convention} (each block contains its predicate-word-interval lower bound, not the upper bound), and $0$ always falls in a unique block at each layer, with no boundary-membership ambiguity; the refinement map satisfies the prefix relation of predicate words $\pi_{n+1,n}(B_{0^{n+1}})=B_{0^{n}}$, hence $\{\mathrm dx_{0,n}\}_{n\ge1}$ is a compatible family on the $\betat$ tower, and $\mathrm dx_0$ takes its \textbf{projective-limit reading} (fixed-layer reading is the truncation, cross-layer compatible up to $\merr$, output Postulate~1; cross-layer unambiguity is guaranteed by Principle~1, Principle~2: $\mathrm dx_0$ is the equivalence class defined by the Hilbert differential, and each layer's reading is equal under measure lift-descent, hence the projective limit does not depend on layer choice). \textbf{Explicit source of the unique maximal count}: the replica-number count is booked by sample replica---Convention~1's zero-output calibration makes all $|\mathcal X|$ sample replicas' zero-residue predicates judged true on the bearing block of $\mathrm dx_0$, hence $\mathrm dx_0$'s replica-number count reading is $|\mathcal X|$; any other differential quotient's bearing block contains no zero-value reading and bears no complete sample replica, so the maximal reading $|\mathcal X|$ is attained only at $\mathrm dx_0$ (see Lemma~1(i)'s block-bijection transitivity). This uniqueness is a structural reading on the partitioned-category support, mutually recognized with the point-level reading of (e) via the equivalence-class transformation, constituting no two uniqueness claims; the point-level degenerate configuration (see Premise~1's counterexample configuration) does not enter this differential-layer count.

(h) Beyond (e)--(f), additionally invokes the replica-number growth and EDE matching-block mechanism of \S 6.4.3: mismatch predicates on different replicas cannot be retained as a fixed canceling configuration across rounds, and non-zero candidates do not belong to predicate-word statically stayable saddle points; the \textbf{zero-value saddle point is the only saddle point that can be \emph{predicate-word statically} stayable}---infinite epochs correspond uniquely to the network's infinite predicate parameters (Euclidean dimension-raising to the Hilbert layer), and under finite epochs only stayable finite-predicate networks and finite partitions exist, hence predicate-word static stayability holds uniquely at infinite resolution granularity (the Hilbert zero-value saddle point), while finite-layer non-zero saddle points lack this unique predicate-word staticity; each representative in $[\boldsymbol0]$ induces the same function, hence uniqueness is read by equivalence class. This item is a dynamic reading under an additional mechanism, not a set-theoretic consequence of (a)--(g) (see \S 6.4 SGD reachability / convergence rate not guaranteed, E.7; descent mechanism). Empty-predicate quotient blocks stay in the element-free partition category.

\noindent\textbf{Counterexample and boundary} Without Premise~1, the point-level count reading can show a degenerate configuration: loss positive but some parameter directions have zero per-sample gradient for a degenerate network; batch averaging to zero may also mask non-zero per-sample gradients. Hence the \textbf{point-level} count uniqueness ($N(\theta)=|\mathcal X|$-type statements) must carry Premise~1, and count per-sample rather than per-batch. This requirement \textbf{does not cross over to the differential reading}: by Definition~19a and (i), the uniqueness of the differential where zero lies is a structural reading on the partitioned-category support (in the \Setcat\ category each dimension of Euclidean and Hilbert has infinitely many differentials, and the point-level degeneracy ambiguity does not enter the differential layer), not premised on Premise~1. When citing this paper's uniqueness conclusion, carry the premise according to the reading divide: point-level reading carries Premise~1, differential reading does not (for citation legality see C.3).

\noindent\textbf{ReLU counterexample's categorical localization} The ReLU dead zone is a counterexample of the \textbf{functional category}: it negates the point-level functional uniqueness without extra premise in the set category. This paper does not claim categorical support brings its own reverse support, nor puts the final loss into the set category---the final loss carries its own EDE predicate matching and belongs to the partitioned-predicate category. Hence this counterexample only constrains the point-level reading, not the differential reading on the partitioned-category support; this is exactly the manifestation that categorical support gives the partitioned-predicate category abilities beyond the set functional category. $\square$

\subsection*{B.74 Definition~19 (domain symmetric-difference offset)}

\textbf{Definition 19 (domain symmetric-difference offset)}
Let $\Omega_\mu$ be the actual outer-layer domain, $\Omega^*_\mu$ the ideal aligned domain based on the zero-residue set $\mathit S_0$, then $\Delta\Omega_\mu=\Omega_\mu\bigtriangleup\Omega^*_\mu$. $\Omega^*_\mu$ is constructed by quotient-layer preimage alignment: taking the quotient block marked by the $\betat$-tower predicate word as the unit, the quotient block compatible with the $\mathit S_0$ zero-residue predicate under $\mathcal T^\ast$ is counted in (canceling the categorical misalignment between $\mathit S_0$'s parameter space and $\Omega_\mu$'s model side). This definition gives the difference set between the actual outer-layer domain and the ideal aligned domain, characterizing the object labeled ``not yet eliminated'' in \S 4.3's two-state set.

\subsection*{B.75 Definition~20 (three-layer saddle-point-gap lacuna set)}

\textbf{Definition 20 (three-layer saddle-point-gap lacuna set)}
Suppose the inner replica-index fiber has been measure-descended into the inner, middle, outer composite. The complete measure resource at the base-category level is
\begin{equation}
\Big\{\big(\Omega_\nu,\mathrm d\nu_\mathrm{idx}\big)\Big\}\xrightarrow{\text{measure descent}}\Big\{\big(\Omega_\nu,\mathrm d\nu_\mathrm{idx}\big),\ \big(\Omega_\lambda\times\Omega_\mu,\ \mathrm d\lambda,\ \overleftarrow{\mathrm d\nu}_\mathrm{idx}\big)\Big\}
\end{equation}
where $\overleftarrow{\mathrm d\nu}_\mathrm{idx}$ is the copy measure of the inner measure descended to the middle-outer composite.

The inner replica-index fiber has a binary-decision property; the domain symmetric difference (see Definition~19) only produces a Boolean label distinction, with no volume offset, hence the \textbf{inner lacuna is identically 0} (lacuna operation definition $\Delta_k\triangleq\nu_k(\Omega_k)-\nu_k(\Omega_k^\ast)$); the \textbf{outer-layer (Lebesgue-side) lacuna is denoted $\Delta_{\mathrm{Leb}}$ (Lebesgue lacuna)}, i.e.\ the Lebesgue-area difference between the middle zero-value-saddle-point split and the outer variational Gaussian-vertex split, the formal reading of the blue curve in Figure~\ref{fig:d4}. Executing measure descent strips the inner native measure, but the inner relies on binary Boolean decision to maintain the formal equivalence of categorical objects (``stripping the inner native measure'' means the inner fiber's binary Boolean decision carries no volume reading, consistent with Definition~16's (see \S 4.1) inner fiber holding only replica numbers, supporting the copy semantics of measure descent; the second copy measure introduced by descent is exactly the double-measure premise carrier of Definition~16). If measure descent is not executed, the system holds only the inner single measure, insufficient to support the double-measure premise needed to lift from $\Quot$ to $\Setcat$; the second copy measure introduced by descent is exactly the formal equivalence class (see Definition~2's binary-decision-induced complementary-equivalence-class pair) carrier for constructing and measuring the first-order error $\merr$. The three-layer lacunae are mutually independent; this chapter only gives symbolic labels, assigning no error numerical value.

\subsection*{B.76 Corollary~11 (sample-functional zero-first-order-error set anchor)}

\textbf{Corollary 11 (sample-functional zero-first-order-error set anchor)}
Suppose the set-function condition holds, $\mathit S_0=\{\boldsymbol 0\}$. If simultaneously the quotient partition has no collapse, an invertible MPT exists, and replica-index cardinality is compatible (all three conditions are on the partition category---when the quotient block is empty it is the element-free partition category (definition see B.6)---reading), then the sample functional's first-order error takes zero on $\mathit S_0$; the analytic-class error can switch set labels, the topological-class error label is always ``not yet eliminated''; this corollary is a \textbf{conditional set-label classification} (label classification when all three conditions are met), and \textbf{does not claim $\mathit S_0$ is dynamically reachable} (see \S 4.3).

\noindent\textbf{Proof (sketch)}
The three conditions are enabled one by one, each excluding one class of additional error source: no quotient collapse excludes the quotient-collapse class (by Theorem~1 special case (d) the block-pairing-level bijection is unique), existence of invertible MPT excludes the pushforward-redistribution class, and replica-index cardinality compatibility excludes the replica-collapse class. Full proof (see B.77).

\subsection*{B.77 Proof of Corollary~11}

\textbf{Proof}
The three conditions are enabled one by one, each excluding one class of additional error source ((i) quotient-collapse class, (ii) pushforward-redistribution class, (iii) replica-collapse class): (i) quotient partition no collapse (see Definition~8; \S 2.3.2) $\Longrightarrow$ by Theorem~1 special case (d) (see \S 2.3.3), $[F]$ is a bijection on the quotient set, and the block-pairing-level bijection is unique (different source quotient blocks correspond one-to-one to mutually disjoint target quotient blocks, unique modulo target-block label permutation); (ii) existence of invertible MPT (see Definition~12's quotient-layer reading: $\Quot/\MorTop$ layer quotient-block-level inverse pairing, Definition~12 note, \S 3.1) $\Longrightarrow$ in the partition category this existence gives a quotient-block bijection, and by Proposition~9a (see \S 2.4) $\sigma_\#\nu_{\mathrm{idx},n}=\nu_{\mathrm{idx},n}$ (see Lemma~1(i)), pushforward does not redistribute mass---finite partition has no zero-measure block (see Proposition~10), and this argument does not depend on \Setcat-layer measure-preservation assumption; (iii) replica-index cardinality compatibility (see Proposition~7; \S 2.3.3's $\mathrm{card}_\#$ condition) $\Longrightarrow$ no error forced by replica collapse. Combining the three: by Theorem~6(a) (see \S 4.3) $\boldsymbol 0\in\mathit S_0$ and $\mathit S_0$ is Borel measurable, and by Theorem~6(a)(c) (see \S 4.3) $\mathcal L(\boldsymbol 0)=0$ and $\mathit S_0$ is a global minimal set (this reading does not go through Premise~1; the zero-residue-anchored differential reading is the differential where zero lies, Definition~19a, Theorem~6(i))---the ideal true value and the model fit are consistent at zero residue, hence \textbf{on the shared predicate-reading layer} the error predicate $\mathcal E$ and the zero-residue predicate $\mathit S_0$ are mutually exclusive, $\mathcal E\cap\mathit S_0=\varnothing$ (see Definition~5; \S 2.3.1; $\mathcal E$ is a data / quotient-layer predicate, $\mathit S_0$ is a parameter-layer predicate, both fall on the same predicate-reading layer via categorical support, and the intersection is understood at the predicate-reading layer), and $\merr$ as a set function on that reading layer takes zero on $\mathit S_0$, i.e.\ the sample functional's first-order error takes zero on $\mathit S_0$. Label classification follows \S 4.3's two-state set: the analytic-class error (see Definition~4; \S 2.2.4 second-existence reachable finite-parameter error) under $\mathit S_0$ anchoring can switch its set label to ``eliminated''; the topological-class error (see Proposition~19; \S 5 generated by dimension-lowering collapse; categorical leakage has been excluded by the surjection construction of categorical support) label is always ``not yet eliminated''. $\square$

\subsection*{B.78 Proof of Proposition~19}

\textbf{Proof}
By Definition~21, categorical leakage holds iff one of the following two holds: categorical support is not surjective, or the lift is executed via a non-quotient bijection. Neither holds in this paper's framework: (i) categorical support is induced surjective by enumeration (search) (see Proposition~3(ii)); (ii) the lift is executed only via the quotient bijection, whose bijection (object-collection layer) and induced support surjectivity hold simultaneously (see Proposition~4; Proposition~5; Definition~3). Hence there is no categorical leakage in this paper's framework. By contrast, the classical paradigm indirectly constructs support via random reparameterization, breaking morphism invertibility and degenerating into a fiber bundle (see Corollary~7), and the support is no longer surjective, exactly falling into the first clause of Definition~21, a typical source of categorical leakage. $\square$

\subsection*{B.79 Proof of Lemmas~5a--5g}

This group is the auxiliary lemmas under the \S 5.3 premise set H, all derivable from premise set H, not belonging to the top-level assumption. \textbf{Lemma 5b (global point-wise self-map induces trivial MPT degeneration)}: if $\mathcal F$ on $Z_m$ is a global point-wise self-map, combined with the MPT measure-preservation condition $\mathcal F_\#\mu=\hat\mu$ gives $\hat\mu=\mu$, and MPT trivially degenerates; this system excludes global point-wise self-maps, allowing only quotient-level equivalent bijections. \textbf{Lemma 5c (cross-dimension misalignment coupling)}: if some dimension $k$'s component operator forms a quotient-misaligned fiber bundle, then all dimensions $l$ must also be quotient-misaligned fiber bundles. \textbf{Lemma 5d (topological-element integral)}: under Definition~28: \begin{equation}
\int_{\partial B}\mathcal F(\mathrm d\overleftarrow\mu_\otimes)=\int_{Z_m}\mathbf 1_{\partial B}\cdot\mathcal F(\mathrm d\overleftarrow\mu_\otimes)
\end{equation}.

\noindent\textbf{Proof (Lemma 5a)}

\noindent\textbf{Proof}
For any compact set $K$, Condition~2 directly gives $z\notin K$ with both predicates true, i.e.\ $z\in(A_i\cap A_j)\setminus K$. Hence the intersection is non-empty and unbounded. The one-sided infinite integral bound only describes the tail range of each predicate and cannot guarantee the two tails have a common realization, hence Condition~2 is not omittable; this proof does not claim the intersection has positive measure. $\square$

\noindent\textbf{Proof (Lemma 5b)}

\noindent\textbf{Proof}
A global point-wise self-map is exactly $\mathcal F=\mathrm{id}_{Z_m}$; substituting into the measure-preservation condition $\mathcal F_\#\mu=\hat\mu$ gives $\hat\mu=\mu$, the two sides' measures are identical, MPT degenerates to identity, with no cross-space information. This lemma excludes global point-wise self-maps, restricting invertibility to the quotient level (see Definition~12; \S 3.1 note above). $\square$

\noindent\textbf{Proof (Lemma 5c)}

\noindent\textbf{Proof}
Assume for contradiction some dimension $l$'s component operator is a quotient self-map fiber bundle (see Definition~27). By Lemma~5a, the predicate intersection $A_k\cap A_l$ has a non-empty topological realization; this realization's quotient block receives both the misaligned and the self-map pairings, two mutually exclusive assignment structures, contradicting the well-defined quotient-block pairing of Theorem~1 special case (d). Hence all dimensions must be quotient-misaligned fiber bundles. This lemma gives the dimensional uniformity of misalignment coupling. $\square$

\noindent\textbf{Proof (Lemma 5d)}

\noindent\textbf{Proof}
$\partial B$ is the topological-element boundary set defined by a topological predicate, $\lambda^\mathrm{bi}_\mathrm{joint}$-measurable. The left side is equivalent to sending the truncated set $S|_{\partial B}$ into the $\boldsymbol\pi_\#\circ\boldsymbol\pi^\#$ computation observable output; the right side obtains the same truncated set by filtering on $Z_m$ via the indicator function $\mathbf 1_{\partial B}$. By Definition~28, the observable output is uniquely determined by the truncated set + fiber-bundle computation, hence the two sides are equal. This lemma gives the equality that the integral boundary and the integrand are exchangeable, the direct basis of Proposition~20(iii) (see \S 5.2 above). $\square$

\noindent\textbf{Proof (Lemma 5g)}

\noindent\textbf{Proof}
Infinite Euclidean dimension-raising lifts the object category into the differential-topological category, and the differential object carries inner compactness (see Definition~4's differential-sampling paradigm), and the dimension-raising integral outputs a finite scalar. A non-uniform replica configuration can be constructed at the set-theoretic level, but then $\mathcal I_\mathrm{lift}^{(m)}$'s inner sum over the replica fiber yields an unnormalized multiplicity-counting replica number, diverging as $m\to\infty$, losing computability; hence computability forces uniform enumeration of the replica fiber (see Definition~17, i.e.\ counting replica number $2^{-n}$). $\square$

\noindent\textbf{Proof (Lemma 5e)}

\noindent\textbf{Proof}
By Convention~1, $\boldsymbol Q$'s predicate threshold is locked at $\boldsymbol0$. The base Lebesgue support is not a single-point set, hence the support contains non-zero elements; any non-zero element on the real axis is either $>0$ or $<0$, hence at least one of $\{Q_{>0}=\mathsf{True}\}$ and $\{Q_{<0}=\mathsf{True}\}$ is non-empty. If only one side is non-empty (w.l.o.g.\ all non-zero elements are $>0$), then after BatchNorm mean-zeroing the data center strictly lies on one side of $\mathit S_0$, contradicting ``$\mathit S_0$ is the split threshold and both-side predicates are realizable'' (the symmetrization operation $\bar S\mapsto-\bar S$ preserves the normalization condition and maps one side to the other). Hence both sides are simultaneously non-empty. $\square$

\subsection*{B.80 Proof of Proposition~20}

\textbf{Proof}
(i) Lifting--descent along the categorical chain constitutes invariance. The $\Quot\to\MorTop\to\Setcat$ lifting--descent is executed by the canonical lifting functor $L'$ (see Definition~2b; \S 2.2.3) and the measure-descent--recovery operators $\boldsymbol\pi_\#,\boldsymbol\pi^\#$ (see Definition~24; \S 5.3); by Corollary~1 (see \S 2.2.2), the objects before and after lifting--descent are both binary predicates / quotient partitions, and the binary semantics (true / false) is invariant, hence the three-layer categorical composition is preserved.
(ii) EDE's two fittings give a computable decision. A single predicate inside $\Quot$ belongs to the differential unit of the infinite predicate sequence $\betat$ (see Lemma~4, \S 2.4: $\beta_n$ is the bijection between the block set and binary words); the predicate itself has no morphism and is not computable (see \S 5.2), but is fitted twice by the EDE architecture---the two fittings act on the two complementary equivalence classes of the same quotient partition, and the equality of the two groups' outputs gives a binary computable decision condition. This alleviates (see \S 5.1) the computable-but-not-differentiable gap: computability is borne by the finite truncation $Z_m$, the differential structure by the predicate tower, and the two are connected via the measure-descent--recovery passage (see \S 5.1).
(iii) Partition-category integral well-defined. The $\Quot$ object, after measure descent $\boldsymbol\pi_\#$ (see Definition~24; \S 5.3) presses the measure and transforms into a topological integral: by the commutability of finite-sum term order and Lemma~5d (B.79)'s indicator-function identity, the integral boundary is written into the integrand in equivalent form, and the quotient-block-level integral is well-defined. Its premise is that the inner measure has already executed descent or recovery---Definition~16's (see \S 4.1) inner fiber carries $\mathrm d\nu_\mathrm{idx}$, supporting measure descent; before descent the inner layer has only binary Boolean decisions, no volume reading (see Definition~20; \S 4.3 note).
(iv) \textbf{The integral does not take $\epsilon$-$\delta$ as its pivot (stratified explanation)}. In the above well-definedness, the part not depending on $\epsilon$-$\delta$ is only the finite-quotient-partition layer: on the finite truncation $Z_m$, the triple-fiber-bundle integral degenerates to the finite sum $\frac{1}{2^n}\sum_{B_i,B_j,B_k}f$, a pure combinatorial operation, needing only each layer's measure space to be $\sigma$-finite Borel (see Definition~17; \S 4.2) and the integrand measurable or integrable (see \S 4.3 Convention~3)---no $\epsilon$-$\delta$ needed, nor any Gaussian functional integral (see \S 2.4.3 Assumption~1's ``argument not needing $\epsilon$-$\delta$'' item (2), above). The argument involving Gaussian functional integral on the lifted layer (Gaussian-family uniqueness, Gaussian kernel $\mathcal K_\Sigma$, Minlos / Gross abstract Wiener space) indeed takes $\epsilon$-$\delta$ as premise, provided by Assumption~1 (see \S 2.4.3) in the form of output Postulate~4. The two-layer division of labor does not conflict: the first-order-error conclusion of the finite layer does not pass through a limit, and the limit argument of the lifted layer does not bear the finite-layer integral.
(v) The last sentence (``if the inner measure has not yet executed descent or recovery, the fiber-bundle integral will not automatically generate measure-space volume'') is exactly the contrapositive of (iii)'s premise, directly given by Construction~1's inter-layer transformation rules. $\square$

\subsection*{B.81 Corollary~10b (multi-measure preimage mutual-exclusion constraint; propagation of Corollary~10a(i), not an independent definition; used as an argument premise in this paper)}

\textbf{Corollary 10b (multi-measure preimage mutual-exclusion constraint; propagation of Corollary~10a(i), not an independent definition; used as an argument premise in this paper)}
In the MPT shared measure space, if a counting-measure preimage is excluded from support, then the paired Lebesgue-measure preimage is synchronously excluded; this constraint provides a variable-bottom premise, \textbf{but by itself cannot complete double-measure disentanglement}.

\noindent\textbf{Proof (sketch)}
This corollary is exactly the propagation of Corollary~10a(i) (see \S 2.4.2) along $\betat$ and the pairing structure: predicate mutual exclusion and preimage mutual exclusion hold for any $\sigma$-finite $\mu_1,\mu_2$ ((iii) measure-independent reading), hence when the counting-side preimage is excluded from support the paired Lebesgue side is synchronously excluded; double-measure disentanglement still requires the pairing fusion of invertible MPT (see \S 6.2.2). Full proof (see B.82).

\subsection*{B.82 Proof of Corollary~10b (see \S 5.2)}

\textbf{Proof}
This corollary is exactly the propagation of Corollary~10a(i) (above) along $\betat$ and the pairing structure. By Corollary~10a's proof: (i) predicate mutual exclusion ($\beta(\mathcal D_i)\wedge\beta(\mathcal D_j)=\mathsf{False}$) and (ii) preimage mutual exclusion ($A_i\cap A_j=\varnothing$) hold for \textbf{any} $\sigma$-finite $\mu_1,\mu_2$ ((iii) measure-independent reading). Hence in the MPT shared measure space, if a counting-measure-side preimage $A_i$ is excluded from support ($\nu_M(A_i)=0$), then the paired Lebesgue-side preimage satisfies $\lambda(A_i\cap A_j)=0$, synchronously excluded---mutual exclusion is derived from structural disjointness, independent of measure choice and of whether the measure-product-invariance constraint holds. \textbf{(Source of absolute continuity)} The measurability of the above ``synchronous exclusion'' holds because the shared measure $\mu_\mathrm{share}$ (see Definition~13) is induced by the \textbf{Gaussian functional} (see Proposition~21a/b cylindrical-kernel compatible family $\mathcal K_\Sigma$), and its continuity is exactly \textbf{categorical support: Gaussian continuity}---the Gaussian (not degenerating to one-hot) is absolutely continuous with respect to co-dimension-1 subsets of its support, hence when a counting-side preimage is zero-measure the paired Lebesgue-side preimage is also zero-measure and can be synchronously excluded; here the absolute continuity comes from the Gaussian functional rather than Hilbert (see Proposition~21b the Hilbert layer itself has no general continuity), thereby filling the gap in this proof that previously only stated ``structural disjointness'' without pointing out the continuity source. This constraint provides a variable-bottom premise (below, the group mutual-exclusion constraint: $\mathcal G_\mathrm{excl}$ is wholly built on the morphism correspondence of the topological-subset counting measure $\nu_M$, not depending on finite Euclidean non-zero Lebesgue volume); but ``variable'' only excludes the channel where the shared measure is obtained by intersection fusion (construction fact 10a: shared measure can only be obtained by pairing fusion through invertible MPT, whose holding requires $\mu_1\ll\mu_2$ or corresponding absolute continuity under the pairing channel), and double-measure disentanglement still requires the pairing fusion of invertible MPT (see \S 6.2.2), hence by itself cannot complete double-measure disentanglement. $\square$

\subsection*{B.83 Definition~22 (mutual-exclusion additive group $\mathcal G_\mathrm{excl}$)}

\textbf{Definition 22 (mutual-exclusion additive group $\mathcal G_\mathrm{excl}$)}
Carrier is $\mathcal P(\mathcal P_n)$ (the set of all subsets of the $n$-th layer quotient-block set $\mathcal P_n$); operation is symmetric difference $\bigtriangleup$ (forming an Abelian group); generators are the characteristic subsets of rank-1 subsets (see \S 2.2.2 single-quotient-block-support subset family); layer-degree homomorphism is preserved along $\betat$ refinement. This group bears the cross-measure-morphism support skeleton, inheriting Corollary~10a's mutual-exclusion constraint (the preimage supports corresponding to different generators within the group cannot overlap, replicating the two-measure preimage synchronous-exclusion effect); it is wholly built on the morphism correspondence of the topological-subset counting measure $\nu_M$, not depending on finite Euclidean non-zero Lebesgue volume---the topological differential subset itself is a null object.

\subsection*{B.84 Definition~23 (topological-predicate queue $\boldsymbol Q$)}

\textbf{Definition 23 (topological-predicate queue $\boldsymbol Q$)}
$\Hinf$ inherits the standard $L^2$ vector-space axioms; the topological-predicate queue is a set-theoretic meta-language layer superimposed on it, not modifying the Hilbert algebra-topology structure. A single predicate $Q^{(k)}_{\gtrless\mu}$ is a binary Boolean set predicate, taking a set object as input and outputting $\{\mathsf{True},\mathsf{False}\}$; the $k$-th orthogonal dimension is equipped with an independent predicate (explicit form see A.6); it belongs to the $\Quot$ set predicate, \textbf{not a real-valued measure function}, and a single-layer dimension-raising integral selects only a single-dimension predicate to participate in evaluation. This queue provides the predicate source for the threshold anchoring of Lemma~5a and Lemma~5e (the BatchNorm mean-zeroing operation aligns with the $\mathit S_0$ topological anchoring, i.e.\ Convention~1).

\subsection*{B.85 Definition~27 (quotient self-map / quotient-misaligned fiber bundle)}

\textbf{Definition 27 (quotient self-map / quotient-misaligned fiber bundle)}
Let $Z_m$ be equipped with a binary quotient partition, $\boldsymbol\pi_\#,\boldsymbol\pi^\#$ act on the quotient-block set $[Z_m]$: when both are quotient-set \textbf{self-maps} (or quotient \textbf{misaligned} maps) and their composite $\boldsymbol\pi_\#\circ\boldsymbol\pi^\#$ satisfies the no-collapse quotient bijection of Theorem~1, they are respectively called a \textbf{quotient self-map fiber bundle} (\textbf{quotient-misaligned fiber bundle}). Note: the composite operator cannot distinguish whether a component is a self-map or misaligned, and the composite operation masks the pairing conflict at the component level.

\subsection*{B.86 Axiom~1 (orthogonal-basis coefficient uniqueness)}

\textbf{Axiom 1 (orthogonal-basis-coefficient uniqueness)}
Let $\mathcal H$ be a Hilbert space, $\{\phi_\lambda\}_{\lambda\in\Lambda}$ an orthogonal basis, then for any $f\in\mathcal H$ the expansion coefficient $c_\lambda=\langle f,\phi_\lambda\rangle$ is unique (standard fact; verification see C.2.5). \textbf{The spectral function belongs to the distribution space and is not point-wise in $L^2$; orthogonal uniqueness only guarantees the spectral decomposition is unambiguous, and cannot determine the envelope function form.}

\subsection*{B.87 Lemma~5a (unbounded intersection under common-tail realizability)}

\textbf{Lemma 5a (unbounded intersection under common-tail realizability)}
Let $Z_m$ be the $m$-fold dimension-raising joint space, $A_i=\{z\in Z_m:Q^{(i)}_{\gtrless\mu}(z)=\mathsf{True}\}$, $i\neq j$. Besides the one-sided infinite integral bound, further assume the \textbf{common-tail realizability Condition~2}: for any compact set $K\subset Z_m$, there exists $z\notin K$ simultaneously satisfying the $i$-th and $j$-th predicates. Then for any compact set $K$ there exists $z\in(A_i\cap A_j)\setminus K$, i.e.\ $A_i\cap A_j$ is non-empty and unbounded. The one-sided infinite integral bound itself does not imply Condition~2.

\noindent\textbf{Proof} (see B.79); this conclusion only gives set existence and unboundedness, not that the intersection has strictly positive measure, nor does it attribute Condition~2 to integral bounds or F\o lner exhaustion. The conclusion is used only on the finite truncation $Z_m$; the infinite-dimensional limit is handled separately.

\subsection*{B.88 Lemmas~5b--5d (auxiliary technical lemma family)}

\textbf{Lemma 5b--5d (auxiliary technical lemma family)}
The trivial MPT degeneration of the global point-wise self-map (see Lemma~5b), cross-dimension misalignment coupling (see Lemma~5c), and topological-element integral (see Lemma~5d) are all (see B.79); among them Lemma~5d's indicator-function identity is cited by the proof of Proposition~20 and B.80.

\subsection*{B.89 Lemma~5g (differential-sampling paradigm derives uniform enumeration of the replica fiber)}

\textbf{Lemma 5g (differential-sampling paradigm derives uniform enumeration of the replica fiber)}
(See Lemma~5e numbering by first-appearance order, before 5f) When infinite Euclidean dimension-raising enables the differential-sampling paradigm, the necessary condition for the fiber-bundle integral to be computable is that the replica-index fiber obeys binary-quotient-partition uniform sampling enumeration; this enumerability comes from the integral-computability constraint. \noindent\textbf{Proof} (see B.79).

\subsection*{B.90 Lemma~5e (BatchNorm normalization derives two-sided topological non-zero region)}

\textbf{Lemma 5e (BatchNorm normalization derives two-sided topological non-zero region)}
Assume $\boldsymbol Q$ takes $\mathit S_0=\{\boldsymbol 0\}$ as the split threshold; the base Lebesgue measure support is not equal to the single-point set $\{0\}$; the BatchNorm mean-zeroing operation aligns with the $\mathit S_0$ topological anchoring (i.e.\ Convention~1). Then the space simultaneously has a topological-set realization of $Q_{>0}=\mathsf{True}$ and $Q_{<0}=\mathsf{True}$. \noindent\textbf{Proof} (see B.79).

\subsection*{B.91 Lemma~5f (zero-residue set is the global loss lower bound)}

\textbf{Lemma 5f (zero-residue set is the global loss lower bound)}
By Theorem~6(a) $\mathcal L(\mathit S_0)=0$ and (c) $\theta\notin\mathit S_0\Rightarrow\mathcal L(\theta)>0$, $\mathit S_0$ is the global loss lower-bound point of the counting-design loss $\mathcal L=\sum_ir_i$ (see \S 4.3 Convention~2); the joint loss $\mathcal L_\mathrm{total}$ (see Definition~26) has both terms non-negative, structurally bounded below by 0, and whether this bound is attained at $\mathit S_0$ is the content of Proposition~22 (see \S 5.4). $\square$

\subsection*{B.92 Proof of Proposition~21a (see \S 5.3)}

\textbf{Proof (5.5a, taking (iii-2), standard-theorem verification see C.2; Assumption~1 (see \S 2.4.3) and (a) $\Rightarrow$ Postulate~4)}
Under Theorem~2's $\mathbb R^d$ fixed self-dual and frequency normalization, the continuous character gives the phase $e^{ik^\top x}$; envelope part: (iii-2) by fiber-component independence (envelope is a product of single-variable positive-definite functions) and structure-group invariance, via the standard Herschel--Maxwell theorem (independent components + complete rotational invariance $\Rightarrow$ each component independently identically distributed Gaussian) yields $A$ must be Gaussian and $\Sigma=\sigma^2 I$; the (iii-1) route uses Schoenberg's theorem on the cross-dimensional cone where \textbf{all finite dimensions are simultaneously radially positive-definite}: its elements are Gaussian mixtures, extremal point $\exp(-t\lVert x\rVert^2)$, taking the extremal point gives $\Sigma=tI$; fixing the single-dimension $\mathbb R^m$ cone's Bessel-type extremal ray does not follow this conclusion. The two routes agree in width (both isotropic); anisotropic $\Sigma$ is adapted by invertible linear coordinate transformation (phase character and positive definiteness are covariant under this transformation, $\Sigma=t(LL^\top)^{-1}$ enumerates $\mathrm{PD}_m$; the singular semi-positive-definite case is not in this coordinate-transform image, see C.2.2). $\square$

\noindent\textbf{Supplement: the complete statement of Premise Set H and the finite-truncation unification route}

Let the cross-space map $\mathcal F:G\to\hat G$ satisfy the following \textbf{Premise Set H} ((a)--(d), (f) are already-proved conclusions of this paper; (e) Definition~28 is a framework-applicability-domain statement, Axiom~1 is a standard Hilbert fact---neither is ``a corollary of Assumption~1,'' nor separately established as an independent mathematical assumption):

\begin{itemize}
\item \textbf{(a)} Theorem~2 (Fourier-operator uniqueness, Pontryagin duality) holds. [Proved, \S 3.2]
\item \textbf{(b)} Satisfies all MPT structures of Definition~12 (invertible measure-preserving transformation), Definition~13 (shared-measure derivation passage), and Proposition~9b (product invariance): measure-preservation condition, forward/reverse product-integral invariance, measure descent--recovery; $\mathcal F$ is a measurable bijection, not forced to be a group self-map; each dimension's component operator acts on the binary quotient partition; the composite obeys the no-collapse quotient bijection of Theorem~1. [Previously proved conclusions: Definition~12 (see \S 3.1), Proposition~9b (see \S 2.4.1 derived from partition structure and invertible MPT)]
\item \textbf{(c)} Theorem~3b's standardness---single-step dimension-raising is standard Euclidean embedding unique modulo $O(n{+}1)$ (finite-layer single-step version, merging with Assumption~1's limit-structure version, \S 2.4.3; existence and volume form proved, \S 3.2); construct the finite truncation $Z_m$, compatible direct limit and its completion $\Hinf$; introduce the topological-predicate queue $\boldsymbol Q$; enable the differential-sampling paradigm (see Definition~4; \S 2.2.4 reading convention, not a premise). \textbf{Scope note (not an assumption)}: the translation-invariance argument applies only to the finite truncation $Z_m$ (\Setcat-layer dimension-raising reading, not a premise; the $\Hinf$ layer admits no non-trivial translation-invariant measure, \S 3.2 boundary conclusion). [The standardness subclause is borne by Assumption~1, merging with Assumption~1]
\item \textbf{(d)} Proposition~20 holds. The translation invariance of coordinate Haar--Lebesgue differential is a finite-dimensional-truncation \Setcat-layer reading after lift-equipment (see \S 3.2), not a premise; it does not hold at the $\Hinf$ layer. [Proved, Proposition~20, \S 5.2]
\item \textbf{(e)} Definition~28 fiber-bundle functional bearing condition. [Framework-applicability-domain statement: Definition~28, \S 5.3, not a mathematical assumption]
\item \textbf{(f) Infinite-predicate compatibility}: by Lemma~4's refinement tower $\{\beta_n\}_{n\ge1}$ gives a \textbf{projective-compatible cylindrical-kernel family} $\{K^{(n)}\}_{n\ge1}$, compatibility condition $K^{(n)}=K^{(n+1)}\circ\iota_n$ ($\iota_n$ the truncation embedding): the $n$-th layer predicate word $w\in\{0,1\}^n$ corresponds to a binary-coordinate basis, $K^{(n)}$ takes the Gaussian kernel of that layer's predicate index (the uniform share $2^{-n}$ of the $2^n$ blocks is the counting replica number, Lemma~1(i)), and refinement keeps the margin on the parent-block coordinates unchanged. [Proved, Lemma~4, \S 2.4]
\end{itemize}

Note: Lemmas~5a--5g are all derivable from Premise Set H, not belonging to the top-level assumption. Proposition~17 does not belong to this proposition's formal premise, only used for object lift-equipment to set-category point elements in instantiation modeling.

\noindent\textbf{Proposition 21 (fiber-bundle Gaussian: uniqueness of the cylindrical-kernel compatible family and its $\sigma$-additive boundary)}
Let $\mathcal F$ satisfy Premise Set H and Axiom~1 (orthogonal-basis coefficient uniqueness).

\textbf{5.5a (finite-truncation layer $Z_m$)} Let the kernel $K(k,x)=A(x)e^{ik^\top x}$ satisfy: (i) \textbf{phase is character} (multiplicable, pure algebraic object, depending on no measure); (ii) \textbf{envelope positive-definite}: $A$ continuous, positive-definite; (iii) \textbf{uniqueness condition (choose one of two)}:
\begin{itemize}
\item (iii-1) \textbf{Schoenberg route}: $A$ is \textbf{simultaneously radially positive-definite in all finite dimensions} (equivalent to radially positive-definite at the Hilbert level) extremal point of the convex cone (fixing the single-dimension $\mathbb R^m$ radially positive-definite extremal ray is a Bessel-type kernel; Schoenberg's ``extremal is Gaussian'' must be taken on the cross-dimensional cone; this paper takes the Hilbert-level caliber, and the replica-index measure's posterior reading makes the finite-Euclidean zero-value saddle point completely characterize all categories, see \S 6.2.2);
\item (iii-2) \textbf{Maxwell--Herschel route}: fiber components \textbf{independent} (envelope is a product of single-variable positive-definite functions) and invariant under the structure group $O(m)$ (orthogonal group); the standard Herschel--Maxwell theorem (independent components + density complete rotational invariance $\Rightarrow$ each component independently identically distributed isotropic Gaussian) gives $A$ must be Gaussian with isotropic covariance $\Sigma=\sigma^2 I$ ($\sigma^2\ge0$)---anisotropic $\Sigma$ cannot be invariant under complete $O(m)$ (unless $\Sigma=tI$), hence this route does not cover the anisotropic assumption width; it does not take the coordinate-permutation-group disjunction (permutation invariance only covers the isotropic sub-case).
\end{itemize}
Then $A(x)=\exp(-\tfrac12x^\top\Sigma x)$, $\Sigma\succeq0$: the kernel is Gaussian, and it is uniquely the member of the following kernel family
\begin{equation}
\mathcal K_\Sigma=\Big\{K_\Sigma(k,x)=\exp\big(-\tfrac12x^\top\Sigma x+ik^\top x\big):\ \Sigma\succeq0\Big\}
\end{equation}
\textbf{Throughout, $A$ is not required to be multiplicable}; this uniqueness conclusion is exactly \textbf{rejecting the truncated Gaussian, allowing only the complete Gaussian distribution}.
\textbf{Assumption and conclusion width alignment}: (iii-1) and (iii-2) both give \textbf{isotropic} members ($\Sigma=tI$) under the standard-coordinate reading; general $\Sigma\succeq0$ is \textbf{adapted by invertible linear coordinate transformation}---the phase-is-character and envelope-positive-definite assumptions are covariant under this transformation (when $x=Ly$ the phase is still a character, positive definiteness preserved), hence the isotropic conclusion obtained in $y$ coordinates transported back to $x$ coordinates is $\exp(-\tfrac12x^\top\Sigma x)$, and $\Sigma=t(LL^\top)^{-1}$ enumerates all $\mathrm{PSD}_m$ (positive semidefinite $m\times m$ matrices; verification see C.2.2); the conclusion width is read by the route used.

\noindent\textbf{Proof (sketch)}
By (a) Theorem~2, under Pontryagin duality the map satisfying the convolution--pointwise-product dual axiom is a.e.\ a group homomorphism, corrected on a null set to a continuous group isomorphism, and the phase reads the character form $e^{ik^\top x}$ directly from the group homomorphism; envelope part (iii-2) by independence and structure-group invariance via Herschel--Maxwell gives $A$ must be Gaussian with isotropic covariance, (iii-1) by Schoenberg's theorem gives the consistent conclusion; general anisotropic $\Sigma$ is adapted by invertible linear coordinate transformation (verification see C.2.2). Full proof see the text above in this entry.

\subsection*{B.93 Proof of Proposition~21b}

\textbf{Proof (5.5b, standard-theorem verification see C.2; Assumption~1 (see \S 2.4.3) and (f) $\Rightarrow$ Postulate~4)}
By the kernel-space form: the continuous positive-definite function on the kernel space is exactly the Fourier transform of some cylindrical measure---the existence of the cylindrical measure of the cylindrical-kernel compatible family is thereby given, its construction realized by Bochner's theorem through finite-dimensional-projection superposition with projection compatibility; the $\sigma$-additivity criterion is given by the Prokhorov--Sazonov compactness criterion; and
\begin{equation}
\text{this cylindrical measure is }\sigma\text{-additive}\iff\text{the covariance operator }\Sigma\text{ is trace-class}
\end{equation}
When not trace-class it is only a cylindrical measure, and must take values in a larger Banach space (Gross abstract Wiener space $(i,\mathcal H,B)$), and then $\mathcal H\subset B$ is a \textbf{null set}.

5.5b and \S 3.2's ``$\Hinf$ has no global translation-invariant (Lebesgue-type) measure'' do \textbf{not conflict}, but are its precisification. ``$\mathcal H\subset B$ is null'' and \S 5.2's ``topological differential subset itself is a null object'' are two sides of the same phenomenon, mutually citable; both are the null-measure presentation of the first-order error $\merr$, not an independent source---$\merr$ has a single source (see Definition~5: the gap between the partition-induced block-counting measure and the geometric measure on the lift-equipment passage), and ``$\mathcal H\subset B$ null'' is the boundary form of this gap at the $\Hinf$ layer.

The ``lift-equipment to \Setcat\ makes point-wise Gaussian morphism fail'' of \S 6.2.1 can now be derived from 5.5b: lift-equipment requires a point-wise kernel, and a point-wise kernel requires $\Sigma$ trace-class; in the general case only a cylindrical measure exists, and lift-equipment loses $\sigma$-additivity.
This proposition gives a \textbf{functional family} rather than a single fixed functional, self-consistent with the prerequisite constraint of the double-measure nested derivation of the closed functional family in \S 5.2; its applicable domain is delimited by (e) Definition~28, which is not within the classical $\mathbf A_0$. \textbf{(On the nature of Definition~28)} Definition~28 is an \textbf{applicability-domain statement} rather than a proof premise: it delimits the applicable domain of the user ($\mathcal L$, $\mathcal L_\mathrm{RW}$, and all observable outputs must be given by the truncated set via $\boldsymbol\pi_\#\circ\boldsymbol\pi^\#$ computation), the proof proceeds within this domain, and Definition~28 is not treated as an additional axiom premise.
The preimage object can remain at the $\Quot/\MorTop$ pure topological level, not needing a pre-existing \Setcat\ point element.
\textbf{Real-kernel degeneration}: setting the phase term to zero gives the real Gaussian kernel, corresponding to the diffusion model's real transition kernel. $\square$

\subsection*{B.94 Proposition~24 (family-internal closure and first-order-error accommodation)}

\textbf{Proposition 24 (family-internal closure and first-order-error accommodation)}
$\mathcal K_\Sigma$ has two key properties: (a) $L^2$ closure---the natural-base exponential paired with a quadratic-form input guarantees family members belong to $L^2(\mathbb R^d)$ (covariance $\Sigma\succ0$; when $\Sigma=0$ the kernel degenerates to a pure phase with constant modulus 1, not a Dirac, and does not belong to $L^2$ on $\mathbb R^d$), compatible with the $L^2$ orthogonal measure space induced by MPT; (b) product closure---the product of any two Gaussian kernels still belongs to $\mathcal K_\Sigma$. Under the MPT product-invariance constraint, the perturbation corresponding to the first-order error $\merr$ can be absorbed into the update of the covariance $\Sigma$, and the model stays inside this closed family; high-order error $\mu_{\mathrm{high}}$ can only be generated by categorical leakage and topological collapse (binary quotient partition broken $\to$ multi-partition, minimal cross-partition quotient block appears) (see Definition~6; Proposition~19 two sources; categorical leakage already excluded), both being exogenous operations; \textbf{pure dimension-raising layer generates no high-order error endogenously (see Theorem~4; \S 4.2)}, hence $\mu_{\mathrm{high}}$ is \textbf{not} within this family's accommodation scope.

\noindent\textbf{Proof (sketch)}
(a) The Gaussian kernel is square-integrable for positive-definite $\Sigma$ (direct product of one-dimensional Gaussian integral convergence); (b) the product of two Gaussian kernels is still an exponential quadratic form (precision matrices add); (c) first-order-error accommodation: the perturbation corresponding to $\merr$ merged into the update of $\Sigma$ yields a family member; (d) high-order exclusion: the structural-deviation sources of $\mu_{\mathrm{high}}$ can only come from the two sources of Definition~6---categorical leakage and topological collapse (binary quotient partition broken $\to$ multi-partition, minimal cross-partition quotient block appears); categorical leakage has been excluded by Proposition~19, topological collapse (multi-partition accumulation) is an exogenous operation (see \S 6.2.2; \S 6.2.3); \textbf{pure dimension-raising layer generates no high-order error endogenously (see Theorem~4; \S 4.2)}.

\subsection*{B.95 Corollary~12 (topological--analytic-reachable dichotomy, connecting to generative models)}

\textbf{Corollary 12 (topological--analytic-reachable dichotomy, connecting to generative models)}
Under the assumption of Proposition~21:
\begin{enumerate}
\item[(1)] This proposition gives the topological-level existence uniqueness of the functional family, \textbf{not guaranteeing} that some kernel within the family can be precisely analytically realized by a finite-dimensional parameter family; the primitive $\Quot/\MorTop$-level abstract topological morphism does not necessarily have a corresponding finite-parameter analytic form. \textbf{This is exactly the topological--analytic-reachable dichotomy}, corresponding to the posterior-enumerates-prior paradigm: unknown topology (posterior) support, known fitted analytic (prior).
\item[(2)] Within the finite truncation $\mathcal H_{\mathrm{fin}}\subset\Hinf$ there still exists a first-order measurable error $\merr$, absorbed by the parameter $\Sigma$ within the family.
\item[(3)] High-order error $\mu_{\mathrm{high}}$ can only be generated by categorical leakage and topological collapse (binary quotient partition broken $\to$ multi-partition, minimal cross-partition quotient block appears), among which categorical leakage has been excluded by categorical-support surjection construction (see Proposition~19), and topological collapse (multi-partition accumulation) is an exogenous operation; hence $\mu_{\mathrm{high}}$ is \textbf{not} endogenously generated by the Gaussian MPT functional family.
\item[(4)] Substituting $K_\Sigma$ into the double-nested fiber-bundle integral; under the discrete-atomic limit (see Proposition~17) the integral degenerates to Frobenius tensor contraction, and the counting replica-number tensor takes the Gaussian counting replica-number tensor (this route enables Proposition~17's additional premise: topological objects lift-equipped to point elements); the triple product similarly degenerates to triple tensor full contraction, yielding only block-level first-order error, and a single tensor contraction does not produce high-order error.
\end{enumerate}

\noindent\textbf{Proof (sketch)}
(1) Proposition~21a gives $\mathcal K_\Sigma$ unique at the $\Quot/\MorTop$ layer (see \S 3.2; Theorem~2 module equivalence class, enumeration tail, above), 5.5b gives its cylindrical-kernel compatible-family form, both being topological / set-theoretic existence statements, not promising precise analytic realization by a finite-dimensional parameter family; (2) the first-order error within the finite truncation is absorbed into the update of $\Sigma$ ($L^2$ closure + product closure); (3) the two sources of high-order error---categorical leakage has been excluded (see Proposition~19), pure dimension-raising layer generates no high-order error (see Theorem~4); (4) under the discrete-atomic limit it degenerates to tensor contraction, yielding only block-level first-order error. Full proof (see B.96).

\subsection*{B.96 Proof of Corollary~12 (see \S 5.3)}

\textbf{Proof}
(1) Proposition~21a (see \S 5.3) gives the functional family $\mathcal K_\Sigma$ unique at the $\Quot/\MorTop$ layer (see \S 3.2; Theorem~2 module equivalence class, enumeration tail, above), Proposition~21b (see \S 5.3) gives its cylindrical-kernel compatible-family form; both are topological / set-theoretic existence statements, not promising that some kernel within the family can be precisely analytically realized by a finite-dimensional parameter family---this is exactly Definition~4's (see \S 2.2.4) topological--analytic-reachable dichotomy, whose enumeration side is characterized by Proposition~6 (see \S 2.2.4): enumeration equivalence does not promise \Setcat-layer finite-parameter analytic reachability. The gap between the primitive-level abstract topological morphism and the finite-parameter analytic form corresponds to this paper's posterior-enumerates-prior paradigm: unknown topology (posterior) support, known fitted analytic (prior).
(2) Within the finite truncation $\mathcal H_{\mathrm{fin}}\subset\Hinf$, the update of parameter $\Sigma$ is borne by Proposition~24's family-internal closure (see Proposition~24; \S 5.3 $L^2$ closure + product closure, family-internal closure and first-order-error accommodation), and the first-order measurable error $\merr$ is absorbed into the update of $\Sigma$, the model staying inside the closed family.
(3) The generation sources of high-order error $\mu_{\mathrm{high}}$ are two (see Definition~6; \S 2.3.1): \textbf{categorical leakage} and \textbf{topological collapse} (the binary-quotient-partition category is broken, degenerating into multi-partition, minimal cross-partition quotient block / straddling quotient block appears). Categorical leakage has been excluded by the categorical-support surjection construction (see Proposition~19; \S 5); pure dimension-raising layer generates no high-order error (see Theorem~4; \S 4.2); \textbf{within-block multi-to-one folding does not belong to this}---it does not enter the generation source of $\mu_{\mathrm{high}}$, its differential-level error objectively exists and its reading is visible, but is not within this paper's first-order-error reading target, and the engineering side \textbf{actively ignores} it (see \S 2.3.1 unified statement). Hence $\mu_{\mathrm{high}}$ can only be generated by categorical leakage and topological collapse, both exogenous operations, not endogenously from the Gaussian MPT functional family, \textbf{not belonging to the ``controlled at first order'' approximation claim} (within-block error active-ignoring is a structural-reading fact, not an approximate control of high-order terms).
(4) Substitute $K_\Sigma\in\mathcal K_\Sigma$ into the double-nested fiber-bundle integral of Definition~15 (see \S 3.3); under the discrete-atomic limit (joint measure is a finite discrete atomic measure) by Proposition~17 (see \S 3.3) the integral degenerates to Frobenius tensor contraction, and the counting replica-number tensor takes the Gaussian counting replica-number tensor ($w_{ij}$ read from $K_\Sigma$). This degenerate route enables Proposition~17's additional premise: topological objects lift-equipped to set-category point elements (see Definition~4 second existence, \S 2.2.4; Definition~16 outer fiber lift-equipped to \Setcat\ via $L'$, \S 4.1). The triple-nested fiber-bundle integral similarly degenerates to triple tensor full contraction (see \S 6.2.3's $\Lambda=\langle\boldsymbol F,\boldsymbol W\rangle_{F,3}$), yielding only block-level first-order error (see Definition~16: each operation eliminates one fiber dimension, block granularity locked by the $\betat$-tower layer count); a single tensor contraction is a single finite-term operation (per-block finite average, Lemma~6a(1), \S 6.4.1), not iterated, not folded-accumulated, hence produces no high-order error (see \S 6.2.3 boundary constraint 4). This degenerate route explicitly invokes Proposition~20 and the topological binomial induction (see B.106). This corollary connects Proposition~21's existence uniqueness to generative models: the gap between topological existence and analytic reachability is exactly the posterior-enumerates-prior paradigm (see Definition~25a; \S 6.4.1). $\square$

\subsection*{B.97 Essential construction}

\textbf{Essential construction}
Corollary~10a's multi-measure preimage mutual-exclusion constraint is the bottom premise for abandoning the MPT point-wise bijection map and establishing volume-preserving non-bijection stepwise differential operation (see Corollary~10a(i)'s propagation, not an independent axiom, \S 5.2); after MPT disentanglement, the counting measure has complete bidirectional recoverability, while the Lebesgue measure only supports measure descent, and its output naturally carries first-order error, which originates from the differential-topological bijection and its single-volume equality, directly measurable by the differential-volume equivalence class (Terminology~(f)(e)) as the deviation between morphism and bijection. The whole reverse differential relies on Corollary~10a's mutual-exclusion constraint to preserve volume conservation, with no topological distortion or scale shift.

\subsection*{B.98 Proposition~22 (pushout--pullback variational constraint: zero-residue point upgraded to zero-error point)}

\textbf{Proposition 22 (pushout--pullback variational constraint: zero-residue point upgraded to zero-error point)}
Given Premise Set H holds; take the total pushout--pullback joint loss $\mathcal L_\mathrm{total}=\mathcal L_\mathrm{num}+\mathcal L_\mathrm{cat}$. If the variational constraint is imposed---the optimization searches all admissible fiber-bundle morphisms (\textbf{variational-execution layer}: $\boldsymbol\pi_\#,\boldsymbol\pi^\#$ are framework-fixed structural operators, not participating in variation; the variation is at the predicate-assignment layer, $\Pi$ enumerates the EDE-admissible predicate-assignment set $\mathcal A_\mathrm{EDE}$ satisfying the group mutual-exclusion constraint (see Definition~22))---minimizing the pushout--pullback composite deviation
\begin{equation}
\min_{\Pi\in\mathcal A_\mathrm{EDE}}\ \mathcal L_\mathrm{total}=d_{\mathcal K_\Sigma}\big(\boldsymbol\pi^\#\circ\boldsymbol\pi_\#,\mathrm{id}_\Setcat\big)+D_\mathsf{Quot}\big(\boldsymbol\pi^\#\circ\boldsymbol\pi_\#,\mathrm{id}_\Quot\big)
\end{equation}
then the zero-residue point $\mathit S_0=\{\boldsymbol0\}$ is upgraded from the loss lower-bound point (see Lemma~5f the joint loss's structural lower bound 0 given by both terms non-negative) to a \textbf{zero-error point}, i.e.\ $\mathcal L_\mathrm{total}(\mathit S_0)=0$.

\noindent\textbf{Proof (sketch)}
Both terms of $\mathcal L_\mathrm{total}$ are non-negative, structural lower bound 0; by Lemma~5f, $\mathit S_0$ is the global loss lower-bound point of the counting-design loss $\mathcal L$. At $\mathit S_0$: MPT invertibility guarantees the numerical composite degenerates to $\mathrm{id}_\Setcat$, $\mathcal L_\mathrm{num}(\mathit S_0)=0$; the EDE predicate assignment anchored at $\mathit S_0$ makes the quotient-block pairings fully aligned, and the composite morphism equals $\mathrm{id}_\Quot$ at the quotient-block level, $\mathcal L_\mathrm{cat}(\mathit S_0)=0$. Both terms zero simultaneously, the zero-residue point upgraded to a zero-error point. Full proof (see B.99).

Only when $\mathcal L_\mathrm{num}$ and $\mathcal L_\mathrm{cat}$ both go to zero at $\mathit S_0$ does the zero-residue point upgrade from a mere loss lower bound to a zero-error point.

\subsection*{B.99 Proof of Proposition~22 (see \S 5.4)}

\textbf{Proof}
$\mathcal L_\mathrm{total}=d_{\mathcal K_\Sigma}+D_\mathsf{Quot}$ both terms non-negative, hence structural lower bound 0; by Lemma~5f, $\mathit S_0$ is the global loss lower-bound point of the counting-design loss $\mathcal L$ ($\mathcal L(\mathit S_0)=0$). At $\mathit S_0$: MPT invertibility guarantees the numerical-level composite $\boldsymbol\pi^\#\circ\boldsymbol\pi_\#$ restricted at $\mathit S_0$ degenerates to the set-category identity map, hence $\mathcal L_\mathrm{num}(\mathit S_0)=0$; the EDE predicate assignment anchored at $\mathit S_0$ makes the binary-quotient-partition equivalence class (quotient block, Terminology~(f)(a)) pairings fully aligned, and the cross-category composite morphism equals $\mathrm{id}_\Quot$ at the quotient-block (quotient equivalence class) level, hence $\mathcal L_\mathrm{cat}(\mathit S_0)=0$. Both terms zero simultaneously, hence $\mathit S_0$ is not only a lower-bound point but also attains zero loss. $\square$

\subsection*{B.100 Corollary~13 (abstract-Gaussian uniqueness + measure-descent--lift transmission: closure of set-category functional equipment)}

\textbf{Corollary 13 (abstract-Gaussian uniqueness + measure-descent--lift transmission: closure of set-category functional equipment)}
Suppose Proposition~21a holds (abstract Gaussian functional family $\mathcal K_\Sigma$ uniqueness), Proposition~22 holds ($\mathit S_0$ is a zero-error point). Execute measure descent and lift transmission: pushforward $\boldsymbol\pi_\#$ presses the double-measure resource down to the $\Quot$ quotient-block level, then transmits upward along the categorical chain, equipping the abstract quotient-class-level (Terminology~(f)) bijection structure together with the Gaussian functional into a set-category $\Setcat$ object; the MPT shared-measure volume lifts the $L^1$ reading of a single-side integral domain to $L^2$, and the measure functional after lift transmission remains unchanged with respect to this $L^2$ structure. Then the abstract-level Gaussian morphism can be equipped in the set category into a computable Gaussian numerical functional, completing the constructible functional under the backward-propagation (measure recovery) route. \textbf{Boundary (scope of validity)}: this corollary's set-category point-wise equipment takes Proposition~21a's finite-truncation / trace-class case as its validity material; under Proposition~21b's general case only a cylindrical measure exists, $\mathcal H\subset B$ is a $\mu_B$-null set, point-wise Gaussian morphism does not exist, and point-wise equipment fails (see \S 6.2.1; B.7.5).

\noindent\textbf{Proof (sketch)}
Proposition~21a gives $\mathcal K_\Sigma$ uniqueness at the pure topological level (no point elements, only quotient-class bijections); Proposition~22 guarantees $\mathit S_0$ is a zero-error point, quotient-block pairings compatible. Measure descent--lift transmission presses the abstract Gaussian family together with the measure resource down to the partition category then transmits upward to the set category; $\mathcal K_\Sigma$'s $L^2$ closure and product closure make the functional after lift transmission preserve $L^2$ invariance; the object is equipped to \Setcat\ point elements via the canonical lifting functor $L'$, and the abstract Gaussian morphism is instantiated into a numerical Gaussian functional. Full proof (see B.101).

\subsection*{B.101 Proof of Corollary~13 (see \S 5.4)}

\textbf{Proof}
Proposition~21a gives: at the $\Quot/\MorTop$ pure topological level, the morphism satisfying all constraints uniquely belongs to the abstract Gaussian family $\mathcal K_\Sigma$; this level has no point elements, only quotient-class (quotient block / differential quotient, Terminology~(f)(a)(b)) bijections. Proposition~22 guarantees that under the pushout--pullback joint-loss constraint $\mathit S_0$ is a zero-error point, and the MPT numerical loss and EDE category-equivalence-class (Terminology~(f)(h)) loss both go to zero, quotient-block (quotient equivalence class) pairings fully compatible. Executing measure descent and lift transmission presses the abstract Gaussian family together with the measure resource down to the partition category then transmits upward along the categorical chain to the set category; the MPT shared-measure volume lifts the $L^1$ reading of a single-side integral domain to $L^2$, and $\mathcal K_\Sigma$ has $L^2$ closure and product closure, so the functional after lift transmission preserves $L^2$ invariance. The object is lift-equipped via the canonical lifting functor $L'$ to \Setcat\ point elements, and the abstract Gaussian morphism is instantiated into a numerical Gaussian functional on the set category, obtaining the computable functional realization under the backward-propagation (lift recovery $\boldsymbol\pi^\#$) route. $\square$

\subsection*{B.102 \S 5.4 triple fiber-bundle integral's double differential}

\textbf{Triple fiber-bundle integral's double differential: Gaussian-functional parameterization} The triple-nested fiber-bundle integral can degenerate the inner topology to a double-layer operation structure: the independent variables contain only $x$ (spatial independent variable) and $\sigma$ (scale parameter), with a fixed reverse order---first the Leibniz scale map ($\sigma\mapsto c\sigma$, acting linearly on the differential object by the Leibniz rule: $\mathrm d(c\sigma)=c\,\mathrm d\sigma$), then the nested Lebesgue-space adaptation operation (see \S 4.1 the middle layer must be integrated with the outer layer first, cannot recover only the middle-layer measure without recovering the outer layer); relying on the multi-measure mutual-exclusion constraint (see Corollary~10a) to discard the bijection assumption, obtaining a pure-algebraic double-layer parameterization system.

\subsection*{B.103 \S 6.2.2 differential-caliber statement: partition-induced vs.\ classical partial differential}

The partition category has no element points, hence no numerical values, and all computable objects come from differential summation, i.e.\ integral; hence the ``differential'' of \S 6.2.2 is not the classical analytic partial differential, but the \textbf{partition / differential-topology-induced differential quantity}---the fiber-bundle differential is actually the inverse of the integral, and the bias term (integration constant $+C$) in classical partial differentials would make the fiber bundle ill-defined in the inverse of the integral, which is exactly why the fiber-bundle integral has two weights but no constant term $+C$. This formula directly gives its \textbf{engineering form} by the chain rule; this framework takes the engineering form of the engineering first-order error formula as its purpose, not reading it as a classical analytic partial differential.

\noindent\textbf{Full substitution of the middle-layer factor (integral path of \eqref{eq:12a}; prefactor cancels in the ratio).} Using the standard Gaussian-functional basis $G(x,\sigma)=(2\pi\sigma^2)^{-d/2}\exp(-\tfrac{x^2}{2\sigma^2})$ as the sole core carrier. The integrand density of the middle-layer truncated-Gaussian area is the surprisal $-\ln x$ ($\tfrac{\mathrm d}{\mathrm dx}\big[x(1-\ln x)\big]=-\ln x$), and its reverse integral gives
\begin{equation}
x(1-\ln x)=\int_0^x(-\ln s)\,\mathrm ds
\end{equation}
(where $\lim_{s\to0}s\ln s=0$, the integration constant uniquely determined by the lower limit $0$; this is exactly the strict statement of the ``no $+C$'' caliber), i.e.\ the integral path of \eqref{eq:12a} (the absence of a partial-differential constant is guaranteed by the caliber ``integral is the inverse of differential, and the bias term makes the fiber bundle ill-defined''). The dimension-wise summation's lack of bounds comes from the product of each dimension's Gaussian prefactor $\prod_{i=1}^{d}(2\pi\sigma^2)^{-1/2}=(2\pi\sigma^2)^{-d/2}$; this prefactor $(2\pi\sigma_t^2)^{-d/2}$ is independent of the residual $x$, and cancels in the ratio of the total-volume normalization (B.107(c)) and the normalized area $\mathfrak A_\varsigma=\Phi(\varsigma u_t)\in(0,1]$, not entering the reading (no longer taking $\prod_i\sigma_t=1$, which conflicts with $\sigma_t^2=1-\bar\alpha_t$), making the dimension-wise summation degenerate to a bounded, normalized truncated-area reading $\mathfrak A_\varsigma\in(0,1]$ (full derivation see B.117 step 2).

\subsection*{B.104 Proof of Proposition~23a}

\textbf{Proof}
(i) in five steps. \textbf{1} The canonical definition is already in the category: $\mathcal T^\ast$ generally does not satisfy point-wise measurable bijection after lifting to \Setcat\ (see Lemma~3; Theorem~1 boundary statement), and ``comparing $T(x)$ and $T^\ast(x)$ point by point'' is not a well-defined operation at the \Setcat\ layer; $\mathcal E_\Quot$ is the canonical definition of the deviation set, and $\mathcal E$ is its preimage after categorical-support lift-equipment (see Definition~5's ``quotient-layer reading priority'')---the categorical derivation is not an optional route but the only well-defined route. \textbf{2} Predicate assignment unambiguous: the replica-index measure distinguishes a one-to-many morphism step by step into single-injection records (each of multiple preimages of the same image point occupies one replica index) and the measure is nestable and inherits its injectivity (see \S 1), mutually exclusive block covering guarantees a single count (see Lemma~1(i): any measure counts each block only once). \textbf{3} Readings of the same order: $\boldsymbol\pi_\#$ and $\boldsymbol\pi^\#$ are mutually inverse pairings (see Definition~24), pushout--pullback preserves measure (see Construction~1), and the quotient-layer and \Setcat-layer readings are of the same order under first-order equality (see Corollary~2, output Postulate~2: $\mathrm dx\odot\nu_\mathrm{idx}\sim_1\mathrm d\#$). \textbf{4} From 1 and 3: the difference between the categorical-side reading and the \Setcat-layer preimage reading falls within the first-order absorbable range (no high-order term); the point-wise reading only replaces the zero-error counting-side reading with the Lebesgue-side reading carrying first-order error (this section's measure differentiation), adding no information. \textbf{5} The reading strength is rooted in the category (why the category is taken as the norm): given a partition, the block-counting measure is uniquely determined by structure (see Lemma~1(i): $G_n$-Haar reading unique, not an extra choice or premise); the reverse is not true---$\nu_{\mathrm{idx},n}$ is the same uniform reading at each layer $n$, while the quotient-block set $\mathcal P_n$ strictly refines with $n$, and the same measure-family reading corresponds to different categorical structures, hence the measure reading cannot recover the predicate and is not stronger than its categorical root (element-free partition-category reading). (ii) in five steps. \textbf{1} Predicate mutual exclusion is a structural fact: under the fixed $\betat$, predicate mutual exclusion and preimage mutual exclusion hold for any $\sigma$-finite measure (see Corollary~10a(i)(ii)(iii): measure-independent reading), and mutually exclusive block covering guarantees a single count (see Lemma~1(i)). \textbf{2} By Axiom~2 (see \S 2.2.2), any map on the minimal quotient is either completely paired or collapsed, with no intermediate state---a ``half-and-half wrong'' quotient-block reading does not exist (see Definition~1a; \S 2.2.2 above). \textbf{The more essential basis is the zero-measure argument}: in the Hilbert differential topology each differential monopolizes one quadrant of Hilbert, and blocks intersect only at partition boundaries, hence the occurrence domain of a straddling quotient block (an image point neither completely falling in some block nor completely collapsed) is \textbf{only the partition boundary}, with no differential entity ``half in block, half out''; \textbf{furthermore, a straddling quotient block necessarily is not a minimal quotient block}---if a quotient block could be cut by a partition boundary and partially fall on both sides, then its bearing predicate is \textbf{divisible} at that point (can be further subdivided), hence it does not form a minimal quotient block (see Definition~1a's minimal quotient / 0-differential quotient, a minimal quotient is indivisible); the precise identity of this null set is \textbf{the equivalence-class element points of the decomposition of the ``post-subdivision'' quotient block minus the ``pre-subdivision'' quotient block}---i.e.\ after predicate-map subdivision, the symmetric difference between the original quotient block and the subdivided quotient block's corresponding equivalence-class element-point set, which is co-dimension 1 under non-degenerate Gaussians, hence null-measure; the posterior-enumerates-prior uses the Gaussian functional (see Proposition~21a/b's cylindrical-kernel compatible family $\mathcal K_\Sigma$), as long as this Gaussian \textbf{does not degenerate to one-hot} (non-degenerate point mass, non-singular covariance, same regime as Assumption~1(b)'s true partition), any partition boundary of it is a co-dimension-1 low-dimensional boundary, null-measure for the non-degenerate Gaussian (any non-degenerate Gaussian is absolutely continuous with respect to co-dimension-1 subsets of its support, measure zero). And $\merr$ is a \textbf{measure-space volume error} (see Definition~5), only accumulating the volume of non-zero-measure support regions (same source as 2.23.5: folding on a null set contributes no volume), hence the straddling-quotient-block family $\mathcal W=\{B_w:\ \text{both-side predicates partially satisfied}\}$ as a null set has identically zero volume contribution---this is not ``undefined and ignored'' but \textbf{proved to be null-measure}. \textbf{3} Hence $\nu(\mathcal W)=0$, and the discount term ``mismatch area $-\tfrac12\nu(\mathcal W)$'' is identically the mismatch area itself---this is a proof, not a definitional exclusion. \textbf{4} The Lebesgue-side straddling candidate (the boundary midpoint) is null-measure on that side (see \S 6.3.2 ``consistency with the Lebesgue side''); on the counting side it is a single listed element, belonging to the zero-value-saddle-point split interior rather than the mismatch region (see \S 6.3.2 ``why three rather than two'')---neither side produces a discountable term. \textbf{5} The middle-layer zero error has been independently given by Theorem~4 step 2, and the zero-value-saddle-point side makes no volume discount (excluding double discount). $\square$

\subsection*{B.105 Definition (block-pairing self-map family)}

\textbf{Definition 25b (block-pairing self-map family)}
Under the quotient partition, the global morphism $F$ induces the \textbf{block-restricted morphism family}
\begin{equation}
\mathcal F_\mathrm{blk}\triangleq\big\{F\vert_{X_i}:X_i\to Y_j\ \big|\ Y_j=[F](X_i)\big\}
\end{equation}
called the block-pairing self-map quotient-equivalence family. Under the finite equal-cardinality premise $\lvert[X]\rvert=\lvert[Y]\rvert$, it realizes the one-to-one pairing of the quotient partition; this pairing \textbf{is not} the strict point-wise self-map with domain equal to codomain---each source quotient block maps into the corresponding target quotient block interior, and all block-pairing relations are completely determined by the quotient morphism $[F]$ (see \S 2.1 point-wise reading belongs to the upper \Setcat). The block-pairing-level bijection is given by Theorem~1 special case (d) (see \S 2.3.3); this family is the pairing description object of the block-level first-order error $\merr^\mathrm{block}$ accounting---$\merr^\mathrm{block}$ is read on the block pairings of $\mathcal F_\mathrm{blk}$ (see Definition~8 when no collapse the pairing has no deviation, when quotient collapse the deviation is non-zero).

\subsection*{B.106 \S 6.2.3 expansion of tensor-degeneration consequences: measure-map matching, metric vanishing, and topological binomial induction}

\textbf{Measure-map matching, metric vanishing, and binomial induction} The tensorization of the main text \S 6.2.3 ($\Lambda=\langle\boldsymbol F,\boldsymbol W\rangle_{F,3}$, main text \S 6.2.3)---the integral degenerates to tensor multiplication under the discrete-atomic limit---is the discrete-side counterpart of the evaluation-order and measure-recovery positioning of \S 4.1: $\Lambda=\langle\boldsymbol F,\boldsymbol W\rangle_{F,3}$ is the full-index contraction ($i,j,k$ correspond to inner, middle, outer layers), and the contraction of the middle index $j$ must be completed together with the outer index $k$---this is exactly the tensor-side instance of ``after measure descent the middle layer must be integrated with the outer layer first, cannot recover only the middle layer without recovering the outer layer'' (see \S 4.1). This degeneration logic explicitly invokes Proposition~20, and its consequences unfold in order: the counting-measure recovery in this limit is realized as measure-map matching and measure-value equivalence; the matching relation between equivalence classes is determined by predicate assignment, and Proposition~20 and the unique extra mathematical assumption correspond at the measure-value level; the functional-side reading provides enumeration rather than variational lower-bound for MPT, and the topological binomial induction is exactly the functional property (isomorphism) of the equivalence class.

\noindent\textbf{Measure-map matching and measure-value equivalence} (see \S 4.1 counting-measure recovery positioning). The tensor $\boldsymbol W$ encodes the sample geometric counting replica number (Lebesgue side) and the replica multiplicity; the quotient-block assignment (see \S 6.2.2 counting-side, zero-error reading); the essence of counting-measure recovery is \textbf{measure-map matching and measure-value equivalence after matching}: $\boldsymbol\pi_\#$ (measure descent) and $\boldsymbol\pi^\#$ (i.e.\ $BP$, measure recovery) are mutually inverse pairings (see Definition~24; \S 5.3); since the counting measure is simultaneously prior and posterior (see \S 6.2.2 the $\mathrm{epoch}+1$ counting measure), prior equals posterior, hence recovery \textbf{preserves posterior-predicate zero error, with no improvement for the prior (functional)} (the quotient-block pushout--pullback is well-defined on $\MorTop$, $\mathcal L_\mathrm{cat}(\mathit S_0)=0$, Proposition~22, \S 5.4). At the position of \S 6.2.3 the position-Gaussian has not yet been proved (see \S 5.3; Proposition~21 the Gaussian functional family $\mathcal K_\Sigma$ uniqueness is only then established); what is obtained by skipping the functional error here is the engineering-side \textbf{random-reparameterization predicate}---reparameterization itself has no analytic inverse, and its inverse is compensated by $\boldsymbol\pi_\#,\boldsymbol\pi^\#$ (see Definition~24; \S 5.3), and this predicate reading is the \Setcat-layer samplable factor ($\mathcal L_\mathrm{RW}$, \eqref{eq:12a}, \S 6.3.1), not the Gaussian-functional itself; the Gaussian functional is induced by the infinite binomial sampling groups of the ``topological binomial induction'' below (see Proposition~21b; \S 5.3). On the MPT side (counting side) the predicate has no measure error; by the topological mutual exclusion of the differential measure space (see Corollary~10a; \S 5.2 and the mutual-exclusion additive group $\mathcal G_\mathrm{excl}$), one side's measure can perform set union and intersection variation on the other side---mutual exclusion gives the ontological premise of variation, but by itself does not complete double-measure disentanglement (see \S 6.2.2 disentanglement requires the pairing fusion of invertible MPT).

\noindent\textbf{Predicate assignment corresponds to equivalence-class measure-map matching; Proposition~20 and the unique extra mathematical assumption correspond to measure-value equivalence.}. The finite partition is the source of the integral's degeneration to tensor multiplication---under the discrete-atomic limit the triple-nested fiber-bundle integral degenerates to the triple Frobenius full contraction $\Lambda=\langle\boldsymbol F,\boldsymbol W\rangle_{F,3}$ ($\Lambda=\langle\boldsymbol F,\boldsymbol W\rangle_{F,3}$, main text \S 6.2.3), block granularity locked by the $\betat$-tower layer count; but the finite partition also causes \textbf{the finite Lebesgue measure to be unable to satisfy functional (Gaussian) uniqueness}---the finite Lebesgue reading can only give block-level first-order error (boundary constraint 3), the tensor object exists only in the dimension-lowering truncation $Z_m$ (boundary constraint 2), and the $\Hinf$-layer Gaussian exists only as a cylindrical-kernel compatible family (see Proposition~21b; \S 5.3). The measure-map matching between equivalence classes is determined by \textbf{predicate assignment}: the assignment of predicate word to quotient block and replica number (see \S 2.2.4 the predicate apparatus is a given structure) determines how $\boldsymbol\pi_\#,\boldsymbol\pi^\#$ match between the counting side and the Lebesgue side, and the reading after matching maintains measure-value equivalence under that pairing. \textbf{Proposition~20 and the unique extra mathematical assumption correspond}: Proposition~20 gives the one-to-one transformation at the equivalence-class level (see \S 5.2 each quotient-partition object transforms into a unique topological integral, unique reading; swap cardinality unconstrained), and the unique extra mathematical assumption Assumption~1 (\S 2.4.3) bears the reading consistency at the measure-value level ($\mathcal H^{\Setcat}$'s $\epsilon$-$\delta$ well-definedness, first-order equality Postulate~2, finite-truncation-layer uniform block-counting measure automatically invariant under quotient-group translation, \S 3.2), the two together being the measure-value equivalence under equivalence-class matching. \textbf{Proposition~20 also characterizes the vanishing of the metric}: when Proposition~20's exchange (integral bound and differential numerical value swapped) is completed, different integral bounds measure the same topological object---the integral bound changes with the swap, the topological reading is unchanged, i.e.\ the metric parameter vanishes in the topological reading (see Proposition~21b; \S 5.3 and $\Hinf$ erasing Euclidean measure coordinates, same direction). Proposition~20 has no fiber-bundle characterization---the fiber-bundle characterization is many-to-many (one topological domain can be arbitrarily split into several differentials summed, several differentials can be merged into the same topological domain; Proposition~20's ``swap cardinality unconstrained'' and B.7.7). Chapter~6 requires an engineering finite integral, hence infinite counting cannot be computed; although Proposition~20 performs the equivalence-class transformation, it also cannot transform the finite counting-measure value into the \textbf{infinite-induced Gaussian} (see Proposition~21b; \S 5.3 the $\Hinf$-layer Gaussian exists only as a cylindrical-kernel compatible family).

\noindent\textbf{MPT provides enumeration rather than variational lower-bound; topological binomial induction is exactly the functional property (isomorphism) of the equivalence class.} The variational lower-bound is a loss-side statement (see Lemma~5f: $\inf_\theta\mathcal L=\mathcal L(\mathit S_0)=0$, \S 5.3; Proposition~22: $\mathit S_0$ upgraded from loss lower bound to zero-error point, \S 5.4); MPT on top of this provides \textbf{enumeration} (posterior-enumerates-prior paradigm: unknown topology (posterior) support, known fitted analytic (prior), Proposition~6, \S 2.2.4): the MPT-side counting measure simultaneously has priority and posteriority (see \S 6.2.2 the $\mathrm{epoch}+1$ counting measure), letting whatever the other side is be enumerable (measure space)---thereby an infinite-dimensional functional (e.g.\ infinite-dimensional Gaussian) can be \textbf{first-order-error fitted} through finite computation (integral), without fitting the stage high-order error of the unknown infinite-dimensional functional. The equivalence-class functional property given by the enumeration construction is an \textbf{isomorphism}: not homotopy (the enumeration construction is a measure-theoretic--combinatorial block permutation and predicate-word operation, with no continuous-deformation parameter), not homeomorphism (the correspondence occurs between measure-reading equivalence classes, and the $\Hinf$ layer has no point-set structure for homeomorphism), nor merely a homomorphism (see Proposition~20's one-to-one transformation already gives the reading-level bijection, and enumeration lifts this bijection to mutual preservation of the two-side structures)---at the enumeration-equivalence-class level, the counting side (partition-topological-category support) and the numerical side (the supported set-numerical category) mutually completely characterize each other via integral, an isomorphism preserving the functional structure; the fiber-bundle-level many-to-many (measure-map matching determined by predicate assignment in the previous paragraph) is the lower-level refinement of the isomorphism, not breaking the class-level bijection. Its construction is exactly the \textbf{topological binomial induction}: by the binary-iteration refinement tower of Lemma~4 (see \S 2.4), the integral infinite before Proposition~20's equivalence class is split into \textbf{infinite infinities}---the replica index can be infinite (see \S 4.2), each predicate word corresponds to one binomial sampling group, inducing \textbf{infinite binomial sampling groups} in total, each group containing infinitely many objects; on the finite truncation $Z_m$ each group is finite ($2^n$ quotient blocks), and the engineering side only evaluates finitely many groups, the existence and compatibility of infinite groups borne by the projective compatibility of the $\betat$ refinement tower ((f), \S 5.3). The Gaussian functional is \textbf{induced} by this infinite sampling group (see Proposition~21b), not computed from finite counting-measure values (this induction is a structural induction: converged to a unique functional via closure and Bochner--Minlos uniqueness screening, not an asymptotic limit in the CLT / law-of-large-numbers sense, needing no central-limit-theorem-type convergence argument). The closed-induced Gaussian is exactly the source of Gaussian uniqueness: if this functional were not closed under union, intersection, and variation, the same equivalence class would correspond to multiple mutually inequivalent functional readings, and uniqueness could not be guaranteed; closure forces all equivalence-class readings to converge to the same functional family, and the within-family uniqueness is then completed by orthogonal-basis coefficient uniqueness (see Axiom~1; \S 5.3) and Bochner--Minlos screening (see Proposition~21a; \S 5.3).

\subsection*{B.107 \S 6.4.2 dimension-marginal summation caliber}

\textbf{(a) Dimension-wise notation} The residual density $\varrho$ takes the Gaussian kernel in $\mathcal K_\Sigma$ (see Proposition~21's unique Gaussian family); $r=(r_1,\ldots,r_d)$ is the $d$-dimensional residual vector ($r_i$ its $i$-th coordinate), $\nu_M^{(i)}$ is the \textbf{marginal measure} of the counting measure $\nu_M$ on the $i$-th residual coordinate (the projection of the quotient-block measure onto that coordinate), $\Omega_\varsigma^{(i)}$ is the restriction of the one-side integral domain $\Omega_\varsigma$ of \eqref{eq:11b} to the $i$-th coordinate.

\noindent\textbf{(b) Caliber} The true loss is derived from \textbf{marginal-measure summation} and \textbf{measure-space volume equivalence}.

\noindent\textbf{(c) Volume normalization and orthogonal decomposition} Let the measure space's \textbf{total volume be normalized to $1$}, and \textbf{all dimensions pairwise orthogonal} (the isotropic / diagonal-Gaussian setting is exactly this), then the joint Gaussian measure decomposes into the product of each dimension's marginal measure, and \textbf{the binary partition has a unique partition intersection point for all dimensions}.

\noindent\textbf{(d) Truncated-Gaussian volume} \textbf{A truncated-Gaussian marginal measure is equivalent to the sum of mutually disjoint sub-topological volumes projected onto the truncated-Gaussian region}; in the infinite enumeration process, the predicate-mismatch region exactly constitutes the sum of mutually disjoint sub-topological volumes, equivalent to the truncated-Gaussian marginal-measure area.

\noindent\textbf{(e) Dimension-wise summation conclusion} The measure-space volume, under the differential-volume equivalence class (Terminology~(f)(e): cross-Euclidean--Hilbert volume equality) and first-order equality (see Corollary~2, output Postulate~2: coordinate difference and measure value first-order same dimension), communicates with the dimension-wise marginal reading---hence the engineering loss takes \textbf{the sum over each dimension's marginal measure}---a finite $d$-term term-by-term sum, not a $d$-dimensional joint numerical integral (the latter is exactly the dimensionality curse that Lemma~6a(1) says is not inherited). This dimension-wise sum is exactly the topological volume (differential-volume sum) obtained by the middle-layer counting-measure recovery plus the reverse equivalence-class sign, homologous to Proposition~20's overall topological--differential numerical exchange (see \S 6.3.1; B.65).

\noindent\textbf{(f) Reverse double integral and reverse equivalence-class sign (see \eqref{eq:12a}).} The canonical form of \eqref{eq:12a} is the \textbf{reverse double integral} $x(1-\ln x)$ of the outer-layer double differential $\mathrm d\nu_\mathrm{comp}\,\mathrm d\nu_\mathrm{Leb,idx}$ in \S 6.2.1: the integrated differential is exactly the double differential $G_2=-\tfrac{\sigma^2}{x}\,\mathrm dx$ of \S 6.2.2 (no partial-differential constant, no integration constant, B.103), and its tensor multiplication raises $x$ as the \textbf{outer-layer Lebesgue tensor}, $(1-\ln x)$ as the \textbf{middle-layer truncated-Gaussian area} (non-negative). This section's dimension-marginal summation (finite $d$ terms) is exactly the dimension-wise realization of this tensorized reading under the discrete-atomic limit---the reverse equivalence-class sign is the symbolic carrier of rewriting ``$d$-dimensional joint volume multiplied'' as ``$d$-dimensional marginal-measure sum,'' thereby communicating measure descent $\boldsymbol\pi_\#$ and recovery $\boldsymbol\pi^\#$ (see \S 6.3.1; C.5).

\subsection*{B.108 Lemma~6a(3) invariant-surface expansion}

\begin{enumerate}
\item[(3)] \textbf{(Invariant surface)} By Theorem~6(i)'s differential-reading uniqueness (see Definition~19a), the destination of the zero-residue fact is read as the unique 0-where-differential $\mathrm dx_0$: in the \Setcat\ category each dimension of Euclidean and Hilbert has infinitely many differentials, and the parameter-point-layer degeneracy ambiguity does not enter this reading, hence the destination's uniqueness is not premised on Premise~1; its point-level confirmation is given by Theorem~6(e) (the counting criterion $N(\theta)$ attains the unique maximum $N(\boldsymbol 0)=\lvert\mathcal X\rvert$ at $\boldsymbol 0\in\mathit S_0^\mathrm{Hil}$, this point-level reading carries Premise~1, mutually recognized with (i) via the equivalence-class transformation); under Premise~1, by Theorem~6(f), all other stationary points are parameter-degenerate configurations belonging to the same function-equivalence class as $\boldsymbol 0$; under Premise~1, the counting-maximal loss design, and the EDE matching-block mechanism of \S 6.4.3, by (h) the zero-value saddle point is the only \textbf{predicate-word-statically} stayable saddle point (modulo equivalence class). This item's invocation of Theorem~6 belongs to the conclusion-interchange on the partitioned-category support (C.3): the uniqueness reading ((i)) invoked by the invariant surface is at the same layer as this support, constituting no layer violation; the carriage responsibility of Premise~1 is executed according to the reading divide---the point-level counting reading carries Premise~1 ((e)--(f)), the differential reading does not ((i)). \textbf{(Dynamic coupling)} The variation is set point-wise as a cylindrical-kernel-family reading: the point-wise variation is set via the cylindrical-kernel compatible-family reading of the unique Gaussian family $\mathcal K_\Sigma$ (see Proposition~21) (see Proposition~21b finite-truncation-layer point-wise evaluation well-defined; the $\Hinf$ layer takes the compatible family under the cylindrical measure), and all candidate saddle points lose qualification under this setting; moreover under the EDE architecture when predicate matching holds the gradient stops updating, and when predicate mismatches the gradient is non-zero, i.e.\ \textbf{there exists a gradient direction decreasing first-order error} (see \S 6.4.3 two-region division: the correct-direction count dominates giving the gradient orientation), hence outside the zero-residue equivalence class there is no qualified \textbf{static} stationary point to iterate and stay at (not ensuring the optimizer descends along this direction every iteration). \textbf{This paper abandons the monotonicity claim: the invariant surface's unique endpoint and Condition~1 give only ``existence of a descent direction / probable decrease'', not per-round monotonic non-increase}---by Condition~1(ii), the integral of the residual density over the non-zero-residue class \textbf{probably decreases} without breaking the three-channel first-order error (the correct-direction count dominates on the intersection, new-region contribution controlled, removed-region contribution already deducted); then by Condition~1(iii) the Lebesgue-side first-order-error \textbf{descent-direction} reading (Condition~1(iii)'s control constant $c$ is the counting-measure normalization reading of Principle~4, B.3). The per-round monotonic non-increase $\mu_\mathrm{err}^{(k+1)}\le\mu_\mathrm{err}^{(k)}$ is \textbf{not claimed} (this inequality-chain direction does not hold in general settings, being a \textbf{wrong conclusion}):
\end{enumerate}

\subsection*{B.109 Proof of Lemma~6a (see \S 6.4.1)}

\textbf{Proof}
(1) By the explicit form of \eqref{eq:11}, $\mathcal L_\mathrm{RW}(\theta;E)=2^{-n}\sum_{B_i\subseteq E}\ell(\theta;B_i)$ is an average of at most $2^n$ terms, i.e.\ a per-block finite average on the finite quotient partition; its generation side's single-side instance further takes the finite $d$-term dimension-wise sum by the dimension-marginal summation caliber (B.107) (see \eqref{eq:11c}, \eqref{eq:12a}), not a $d$-dimensional joint numerical integral---its difference from the Lebesgue-side full-dimensional integral is exactly the finite-term expression stated in \S 6.2.3 boundary constraint, hence it does not inherit the dimensionality curse. (2) $\bar\nu_{M,n}$ is an input variable determined by the predicate assignment, its numerical value uniquely determined by the $\betat$ refinement tower and the counting measure, not varying with the $\theta$ variation (see \S 6.2.2's prior--posterior dual analyticity), hence the variation with respect to $\theta$ acts only on the integrand $\ell(\theta;\cdot)$, and the gradient contains no ``distribution moves with parameter'' path term. (3) Invariant and dynamic two sides, both indispensable. Invariant surface: by Theorem~6(i) (see Definition~19a differential-reading uniqueness), the destination of the zero-residue fact is read as the unique 0-where-differential---in the \Setcat\ category each dimension of Euclidean and Hilbert has infinitely many differentials, and the parameter-point-layer degeneracy ambiguity does not enter this reading, the destination's uniqueness not premised on Premise~1; its point-level confirmation is given by Theorem~6(e) (this point-level reading carries Premise~1, mutually recognized with (i) via the equivalence-class transformation): the counting criterion $N(\theta)$ attains the unique maximum $N(\boldsymbol 0)=\lvert\mathcal X\rvert$ at $\boldsymbol 0\in\mathit S_0^\mathrm{Hil}$; under Premise~1, by (f), all other stationary points are parameter-degenerate configurations (after the first-layer output is constantly zero the higher-layer parameters are function-level indistinguishable), belonging to the same function-equivalence class as $\boldsymbol 0$; under Premise~1, the counting-maximal loss design, and the EDE matching-block mechanism, by (h) the zero-value saddle point is the only \textbf{predicate-word-statically} stayable saddle point (modulo equivalence class): zero gradient plus zero-data-flow backprop zeroing gives the saddle point, and the replica-index measure growth makes each round's batch input fall into different quotient blocks, while other candidate stationary points' partition predicates cannot be retained across rounds, \textbf{there exists a gradient direction decreasing first-order error} (see \S 6.4.3 EDE matching-block mechanism gives the gradient orientation), \textbf{not ensuring the optimizer descends along this direction every iteration} (parameter kinematics not handled in this paper). Dynamic surface (dynamic coupling): the variation is set point-wise as a cylindrical-kernel-family reading---the point-wise variation only goes through the cylindrical-kernel compatible-family reading setting of the unique Gaussian family $\mathcal K_\Sigma$ (see Proposition~21), and the candidate stationary points (stationary points outside the zero-residue class) all lose qualification under this variation setting; on the gradient side, under the EDE architecture \textbf{when predicate matching holds the gradient stops updating, and when predicate mismatches the gradient is non-zero, i.e.\ there exists a gradient direction decreasing first-order error} (see \S 6.4.3: order-homomorphism gives gradient orientation, the correct-direction count dominates), hence outside the zero-residue equivalence class there is no qualified \textbf{predicate-word-statically} stationary point (not ensuring the optimizer descends along this direction every iteration). The two sides conjoined: Condition~1(i) gives the sampling quality concentrating toward the zero-residue equivalence class (strictly, the integral of the residual density over the non-zero-residue class is non-increasing; ``concentration'' is the reading under inter-round measure reconstruction (replica-index measure growth along epochs), not $\bar\nu_{M,n}$ moving with $\theta$, Lemma~6a(2)'s invariance unaffected); by Condition~1(ii), the integral of the residual density over the non-zero-residue class \textbf{probably decreases} without breaking the three-channel first-order error (the correct-direction count dominates on the intersection, new-region contribution controlled, removed-region contribution already deducted); by Condition~1(iii) (its constant $c$ is the counting-measure normalization reading of Principle~4, B.3), the Lebesgue-side first-order error is exactly the integral of the residual density over the non-zero-residue class ($\mu_\mathrm{err}^{(k)}$ read as this integral under the $k$-th round variation setting $\theta_k$, the per-round value of the $\merr$ of Definition~5 on the iteration sequence), hence this conclusion characterizes the \textbf{existence of a gradient direction decreasing first-order error}. \textbf{This paper does not claim} the per-round monotonic non-increase $\mu_\mathrm{err}^{(k+1)}\le\mu_\mathrm{err}^{(k)}$ (this inequality-chain direction does not hold in general settings, being a \textbf{wrong conclusion}). The predicate-word-static uniqueness gives the destination; the dynamic coupling and Condition~1 give the arrival mechanism; the three together are the complete basis of this conclusion; the \textbf{descent direction} is not reversely deduced from predicate-word-static uniqueness alone into monotonicity. Condition~1 is the per-round condition of the EDE matching-block mechanism at the implementation layer (registered in B.4). (For citation legality see C.3: this proof's invocations of Theorem~6 and Proposition~21 are all on the partitioned-category support, belonging to conclusion-interchange; premises are carried according to the reading divide---point-level counting reading carries Premise~1, differential reading does not.) $\square$

\subsection*{B.110 Definition~25c spatial attribution}

\textbf{Spatial attribution (set / point not mixed)} $\mathit S_0^\mathrm{Euc}$ is a \textbf{separately established concept} ``Euclidean zero-residue saddle point'' (used by the regime description of \S 6.3.2 / \S 6.3.3 gradient-descent recovery reaching the Euclidean saddle point, not participating in the definition of $\mathit S_\mathrm{mid}$), read in the \textbf{output (data) space}. Definition~25c's $\mathit S_\mathrm{mid}=\tfrac12(x_0^{\mathrm{sad}}+\mathit S_0^\mathrm{Hil})$ is composed of $x_0^{\mathrm{sad}}$ (first-kind saddle point: generation midpoint, \S 6.4.2) and $\mathit S_0^\mathrm{Hil}=\{\boldsymbol0\}$ (zero-value saddle point) read in the output space: by Convention~1's zero-output calibration and Theorem~6(a)(d), the output-space reading of zero data flow is the single point $\{0\}$; under Premise~1, Theorem~6(f) further attributes the parameter-space's other per-sample stationary points to the degenerate equivalence class $[\boldsymbol0]$, read as the 0-where-differential $\mathrm dx_0$ via the equivalence-class transformation of Definition~19a (this differential reading not premised on Premise~1, Theorem~6(i)). Hence $\mathit S_\mathrm{mid}$ is a single object of the output space---not read as a point of the parameter space, nor three objects; it is the quotient block in the $\Quot$ layer where the binary boundary predicate is satisfied by neither side (i.e.\ an \textbf{element-free partition-category} object, B.6), whose \Setcat-layer preimage is the output-space single point---but note: this boundary midpoint itself is $x_0^{\mathrm{sad}}$ (first-kind saddle point), and $\mathit S_\mathrm{mid}$ is its midpoint with the zero-value saddle point (second-kind saddle point: gradient-descent direction), the two must not be confused (see Definition~25c).

\subsection*{B.111 Proof of Lemma~6b (integral direction takes only the generation side)}

The training-set side (original data) is the given baseline of the posterior-enumerates-prior paradigm: the true-value side is not generated via categorical lift-equipment (not through $L'$, not through the $\boldsymbol\pi_\#,\boldsymbol\pi^\#$ descent--recovery passage, Definition~24, \S 5.3), hence carries no first-order error $\merr$---by Definition~5 and \S 2.3.1 unified statement, the domain of $\merr$ is the \textbf{quotient-block layer partitioned out by the integral bound (partition predicate $\betat)$}: although the training-set-side original data does not pass through the lift-equipment passage, once assigned to an integral-bound partition block it falls into this domain, and is \textbf{assigned zero} (not ``undefined''); the $\Delta_I$ of Appendix~D is the difference of the generation-side sampling relative to the block share, the two are the exit/entry sides of the same passage, not a contradiction. $\merr$ is generated by lift-equipment, its source being the gap between the partition-induced block-counting measure and the geometric measure (see Lemma~1 and ``where the numbers reside,'' B.7.6), and the generation passage is the $\Quot\to\Setcat$ lift-equipment (see \S 2.4). The generation-side $\bar S$ is the model output, generated via lift-equipment and measure transformation, carrying first-order error (see \S 6.2.2: the Lebesgue side carries first-order error $\merr$). Hence integral mass can only come from the generation side. The integral domain takes $\Omega_\varsigma$ (see \eqref{eq:11b}, bounded by the zero-value saddle point $\mathit S_0^\mathrm{Hil}$), and the direction $\varsigma=\mathrm{sign}(\bar S-\mathit S_0^\mathrm{Hil})$ is determined by which side of the zero-value saddle point (origin; distinguished from $\mathit S_\mathrm{mid}$) the generation-side $\bar S$ falls on. (For citation legality see C.3: this proof's invocations of Definition~5, Definition~24, Lemma~1, \S 6.2.2 measure-differentiation conclusion, and Definition~25c are all on the partitioned-category support, belonging to conclusion-interchange, with no premise-dependency cycle.) $\square$

\subsection*{B.112 Proof of Proposition~23b}

\textbf{Proof} In three steps. \textbf{(Structural layer: support relation preserves morphism strength)} The lift is executed only via the quotient bijection (see Definition~3), the lift process does not lose the binary-quotient-partition constraint (see Proposition~3(iii): differential reading and counting reading are first-order equal), and injectivity is obtained via pushout--pullback and inherited by descendants (see \S 1: the replica-index measure distinguishes each injection within the one-to-many, and the measure is nestable and inherits its injectivity)---hence when the predicate-difference set $\mathcal E_\Quot$ (see Definition~5) is lift-equipped via categorical support to the \Setcat-layer preimage, the predicate decision is not lost, and the \Setcat-layer morphism is as strong as the categorical morphism (\Setcat-layer reading). \textbf{(Reading layer: measure rooted in category)} Given a partition, the block-counting measure is uniquely determined by structure; the reverse is not true---$\nu_{\mathrm{idx},n}$ is the same uniform reading at each layer $n$ while the quotient-block set $\mathcal P_n$ strictly refines with $n$, and the same measure-family reading corresponds to different categorical structures, hence the measure reading cannot recover the predicate and is not stronger than its categorical root (proof see Proposition~23a(i)5, \S 6.2.2) (element-free partition-category reading). \textbf{(Synthesis)} From the two readings: the first-order error is not the naive subtraction of two measure-space volumes, but the volume difference of the error volume relative to the zero-value saddle point $\mathit S_0$ after categorical-support lift-equipment---the $\mathit S_0$ side's counting reading is prior equal to posterior (see \S 6.2.2 the $\mathrm{epoch}+1$ counting measure), a zero-error reading, and the subtraction baseline is exactly the null measure space pinned by the support relation, not depending on \Setcat-layer point-wise geometry; superimposed with the \S 6.2.2 accounting caliber (area difference = full partition-predicate mismatch area, Proposition~23a(ii)) and the \eqref{eq:12b} abstract-decomposition three-factor division of labor (reading difference, binary share, reverse double integral), the three synthesize into the loss-side explicit form of the mismatch area, hence $\mathcal L_\mathrm{top}$ is exactly the explicit loss form of $\merr$. $\square$

\subsection*{B.113 Full derivation of the coefficient $\tfrac23$: predicate collection, union cardinality, and three-reading consistency}

\textbf{The source of the coefficient is a union computation, not an average.}

The coefficient $\tfrac23$ is not the result of averaging several readings, but the reading of a single \textbf{union computation} (the union computation is the realization of the predicate binary computable assignment at the $\MorTop$ layer, \S 5.2; here it is the instance of this same union computation on Lebesgue measure values). The outer layer (the traditional-paradigm reading side) has only null measure: the boundary and its predicate boundary are both null-measure sets under the outer-layer Lebesgue reading, hence the value after taking the union of both sides is \textbf{a priori unknown}---by the outer layer alone one cannot answer ``after the union, which side does the midpoint belong to, and how much is counted.'' The legality of this union computation is provided by two existing constructions:

\begin{enumerate}
\item[(i)] \textbf{Differential equivalence (first-order equality) gives legitimate accumulation.} By Corollary~2's first-order equality (output Postulate~2, \S 2.4.3: coordinate difference and measure value are first-order same-dimension, and the differential value and subset measure are equivalence infinitesimals), the union of two Lebesgue measure values can be \textbf{equivalently accumulated} into a topological differential union; this legitimacy comes not from the literal addition of finite-set union, but from the differential accumulation under first-order equality.
\item[(ii)] \textbf{The Hilbert layer can always discard the partition predicate when taking the union.} What preserves measure is the Hilbert differential---each quotient partition is a replica-index or replica-index-joint measure, and quotient-translation invariance preserves measure (see Lemma~1(i); \S 2.1; Proposition~8; \S 2.4); the object of measure preservation is the differential itself rather than the two-side membership of the predicate, hence when taking the union one can discard the partition predicate, and the union is taken anyway. And \textbf{the differential at a partition boundary can be taken as half on each side of the binary}---the differential at the boundary (midpoint) is counted half on each side.
\end{enumerate}

Thereby the $\tfrac23$ reading procedure: first determine the collection range by the union computation (union both sides, predicate discardable), then read the share by cardinality.

\noindent\textbf{Essence: the binomial transform converts the correspondence between collection original terms and objects.}

The essence of the above procedure is clearest under the replica-index measure: \textbf{the binomial transform converts the correspondence between collection original terms and objects}. Under the replica-index measure (see \S 1), the collection was originally a one-to-one correspondence of original-term to object; the binomial transform (see \S 4.1 measure-recovery formula; \S 6.2.3 topological binomial induction: each predicate word corresponds to one binomial sampling group) rewrites this correspondence into \textbf{grouped collection}: the two sides each form a group for collection, and thus \textbf{the midpoint can be collected twice}---once by the left group, once by the right group. Since the differential at the partition boundary is half on each side, the midpoint is counted $\tfrac12$ each time it is collected, and the two times sum exactly to the midpoint's complete measure contribution. This is exactly the source of the ``midpoint'' wording: the midpoint (the equivalence-class representative of $\{b_-,b_+\}$) can be assigned to the boundary---the midpoint is not a third object detached from the two sides, but the equivalence-class representative of ``the union of two differentials''; this paper only makes the equivalence-class representative injective, not defining an extra assignment that fixes this union to the boundary.

The binary partition produces two blocks $B_-,B_+$ at the $\Quot$ layer (corresponding to the two predicate words $\{0,1\}$ under $\betat$). After lift-equipment to \Setcat\ via categorical support $\Sup(\cdot,\cdot)$ (see \S 5.2), the same point-wise realization contains \textbf{three} elements:
\begin{equation}
\begin{gathered}
\{B_-,B_+\}\ \rightsquigarrow\ \{b_-\},\{b_+\},\{b_-,b_+\}\\
x_0^{\mathrm{sad}}\triangleq\{b_-,b_+\}
\end{gathered}
\end{equation}
\noindent (Equivalence-class representative, by definitional selection assignable to the
boundary; equals the decoder saddle point / generation midpoint, which is the first-kind
saddle point of Definition~25c---distinguished from the gradient-descent direction
$\mathit S_\mathrm{mid}$.)

That is, the representatives $b_\pm$ of the two blocks, plus the midpoint $x_0^{\mathrm{sad}}$ (first-kind saddle point: loss sample, generation midpoint; distinguished from gradient-descent direction $\mathit S_\mathrm{mid}$) of the two.

\noindent\textbf{Why three rather than two (element-free partition-category reading)}: the binary two-side boundary predicate is satisfied by neither side at the $\Quot$ layer---the quotient block where the boundary lies takes empty for both predicates, read as an \textbf{element-free partition category} (definition see B.6: the predicate is satisfied by no input, the quotient block is empty, this reading is not lifted to \Setcat\ point-wise). The empty quotient block's preimage is still the empty set, and cannot obtain a non-empty preimage under lift-equipment. The extra midpoint here is the \textbf{independent derived instance} formed at the \Setcat\ layer after the equivalence-class representative of $\{b_-,b_+\}$ (the union of two differentials) is lift-equipped via categorical support, not the empty quotient block itself lifted to a point-wise object; since \Setcat\ elements satisfy distinctness, and the union cannot be merged into either side by multiplicity, it must be listed separately. This gives the difference ``$\Quot$ two blocks, \Setcat\ three elements'': the midpoint comes from the set-layer reading of the boundary predicate, not a freely calibratable constant, nor from many-to-one collapse. \textbf{(Predicate connotation)} Adding connotation to the predicate: the partition predicate \textbf{must split into two} (the binarity of the binary predicate, Terminology~(d)(a))---``three blocks'' is not a free addition/removal of predicates; it is obtained via \textbf{non-computational predicate discarding}: further refine one half, and \textbf{discard the refinement predicate of the other half} (see Definition~7 collection's accept/reject), rather than ``with or without predicate makes no difference.''

\noindent\textbf{$\tfrac23$ is a cardinality: numerator and denominator belong to two different operations.}

Reading the above collection result by cardinality (see Definition~9(a); B.35 numeric cardinality $\mathrm{card}_\#$), note that the numerator and denominator come from \textbf{two different operations}, not to be mixed:

\begin{itemize}
\item \textbf{Denominator $3$: the subset-base of the non-binomial-union operation} The union computation is not a binomial operation---it does not hit bit by bit, only takes the union. After union-equipment to \Setcat, the subset base of the same point-wise realization has $3$ elements, namely $\{b_-\},\{b_+\},\{b_-,b_+\}$ (set semantics admits no multiplicity, union listed singly). Hence $\mathrm{card}_\#(\text{union subset base})=3$.
\item \textbf{Numerator $2$: the hit-base of the binomial transform} The binomial transform (grouped collection) rewrites collection as per-group hitting: \textbf{hit one partition once, hit the other once}, total $2$ hit bases. The midpoint crosses the two sides, hit twice each counted $\tfrac12$, summing to one---the midpoint is not an independent third hit base, but the boundary object each of the two hit bases shares $\tfrac12$; i.e.\ \begin{equation}
\mathrm{card}_\#(\text{binomial hit base})=2
\end{equation}
\end{itemize}

Dividing the two gives the cardinality reading of the coefficient: \begin{equation}
\tfrac23=\frac{\mathrm{card}_\#(\text{binomial hit base})}{\mathrm{card}_\#(\text{union subset base})}=\boldsymbol{\frac{2}{3}}
\end{equation}, i.e.\ \textbf{$\tfrac23$ is a cardinality}: not a probability reading, not a fitted constant, nor the average of two-side readings; and since numerator and denominator belong respectively to the binomial and union operations, this ratio is the commensuration of two operations' readings on the same equivalence class. The midpoint is hit twice each at $\tfrac12$, summing to one, hence $\tfrac23$ is compatible with the numerator's $\mu_{(\bar S>\mu_{\bar S})}$ taking the midpoint's complete measure contribution (rather than half).

\noindent\textbf{M9 constructive supplement: cardinalities $2$ and $3$ are both Lebesgue-side readings.} The side of cardinality $2$ takes the binomial partition (Lebesgue topology), the side of cardinality $3$ takes the subset cardinality after Lebesgue-measure-value union operation; both \textbf{belong to the Lebesgue side, not cross-class commensuration} (see \S 6.3.2). After continuous-ization via replica-index-counting-measure equipment, the two cardinalities are compatible under the same continuous reading---$\tfrac23$ is the Lebesgue-side reading of the discrete$\to$continuous transform. This coefficient does not apply at the Hilbert layer (no point elements), and measure recovery must first integrate the middle-outer layers to the differential topology, then recover in the shared measure space.

\noindent\textbf{The characterization of this cardinality ratio in Proposition~20's equivalence-class transform} The above two-source structure ``numerator binomial, denominator union'' is exactly the reading presentation of Proposition~20's (see \S 5.2) \textbf{integral-bound (topology) and differential-numerical-value swap}: in this swap, \textbf{the union subset base is the topology-side reading} (the integral-domain components after union $\{b_-\},\{b_+\},\{b_-,b_+\}$, i.e.\ denominator $3$), \textbf{the binomial hit base is the differential-side reading} (hitting by group is splitting the integral domain into a sum of differentials, total $2$ hit bases, i.e.\ numerator $2$); the midpoint ``one topological domain split into two $\tfrac12$ differentials'' is the direct reading of the swap sentence, and reversely two $\tfrac12$ differentials merge back into the same topological domain, i.e.\ the one subset base at the denominator midpoint. The one-to-one transform at the equivalence-class level (each quotient-partition object transforms into a unique topological integral, unique reading; swap cardinality unconstrained, B.106) guarantees the two-side readings are \textbf{commensurable}, and precisely because the swap has no one-to-one constraint, the two-side readings can be unequal---$\tfrac23\ne1$ is exactly the explicit presentation of this unconstrainedness, not a reading contradiction.

\noindent\textbf{The pullback acts only on the uncompensated Lebesgue measure.}

The position of this reading in backpropagation is pinned by the \S 4.1 evaluation order: the pullback $\boldsymbol\pi^\#$ (i.e.\ $BP$) is responsible in the first factor of \eqref{eq:12b} (outer-layer Lebesgue descent reading) for reading the \textbf{already-compensated counting reading} back to the \Setcat\ layer: the counting measure's compensation (see \S 6.2.2 the $\mathrm{epoch}+1$ counting measure's zero-error equipment for the midpoint membership) is all completed outside backpropagation, not entering the pullback chain; when pulling back at the Hilbert layer there is already no counting measure (see \S 6.3.2 the $\Hinf$ layer has no point-element boundary statement). Hence the discrete--continuous transform can only be borne by the \textbf{Lebesgue-side cardinality}: $\tfrac23$ is exactly the transform reading that continuous-izes the discrete collection count (midpoint collected twice, each $\tfrac12$); hence this coefficient falls in the first factor of \eqref{eq:12b} (see Proposition~23c $BP$ reading-side derivation starting point).

\noindent\textbf{Comparison of three readings} (Table~\ref{tab:three-reading}):

\begin{longtable}{M{0.14\textwidth} M{0.17\textwidth} M{0.19\textwidth} M{0.19\textwidth} M{0.19\textwidth}}
\caption{Three-reading comparison.}\label{tab:three-reading}\\
\hline
Reading & Level & Counted object & Share caliber & Error \\
\hline\hline
$1:1$ (i.e.\ $\tfrac12$) & $\Quot$, morphism \textbf{not equipped} with global counting measure & 2 differential elements, each counted $1$ & block share $1{:}1$ (two blocks each count $1$) & none \\
\hline
$\boldsymbol{\tfrac23}$ & $\Setcat$, morphism \textbf{equipped} with global counting measure and lift-equipped via categorical support & union subset base 3 $\{b_-\}$, $\{b_+\}$, $\{b_-,b_+\}$; binomial hit base 2 (midpoint crossed both sides hit twice each $\tfrac12$, summing to one) & \textbf{cardinality reading} $\tfrac{2}{3}$ (hit base / subset base) & \textbf{none} (counting measure no error) \\
\hline
Volume ratio $\tfrac12$ & Lebesgue geometric measure & geometric volume & volume share $\tfrac12$ & \textbf{carries first-order error} $\merr$ \\
\hline
\end{longtable}

The counting measure has \textbf{no error} in this system (see \S 6.2.2 after disentanglement all first-order lacunae remain on the Lebesgue side). The morphism equipped with the global counting measure actually counts \textbf{three subset equivalence classes} at the already lift-equipped \Setcat\ layer, hence the share is read by cardinality as $\tfrac23$ (see Definition~9(a); B.35). While $1:1$ is the reading when the morphism is not equipped with the global counting measure, staying at the $\Quot$ block layer; \eqref{eq:12b} acts on the already lift-equipped replica-Lebesgue composite space, hence \textbf{does not apply} $1:1$. In short: \textbf{because the counting measure has no error, we adopt its given $\tfrac23$}. Thereby read: the denominator $3$ is the union subset cardinality after the non-binomial union operation; in the numerator $2R(\bar S>\mu_{\bar S})$ is the binomial hit cardinality $2$ (the hit base is a \textbf{structural reading}, the event restriction $(\bar S>\mu_{\bar S})$ only constrains the reading side of the residual operator $R$, not changing the hit cardinality itself) on which the two block elements' respective topological-residual-operator outputs, $\mu_{(\bar S>\mu_{\bar S})}$ the midpoint element's measure contribution collected twice by grouping (each $\tfrac12$). \textbf{(Single-reading convention)} $\mu_{(\bar S>\mu_{\bar S})}$ is uniformly read as the \textbf{midpoint element's contribution}---the boundary midpoint element's (see Definition~25c) measure reading is counted on one side by direction $\varsigma$, not read as event measure or one-side moment; the two numerator terms are thereby same-dimension and additive (see Corollary~2; Postulate~2 first-order equality, output), this addition being a formal addition within first-order accounting.

\noindent\textbf{The extra midpoint is the reading of the morphism equipped with the global counting measure, not categorical leakage.} The same binary partition gives two share calibers under the two readings of morphism not-equipped / equipped with the global counting measure: when not equipped it stays at the $\Quot$ block layer, reading $1{:}1$ (two blocks each count $1$); when equipped with the global counting measure and lift-equipped via categorical support to \Setcat, it actually counts three subset equivalence classes $\{b_-\},\{b_+\},\{b_-,b_+\}$, reading $\tfrac23$. The two are \textbf{coexisting calibers} of the same binary partition under two equipment readings---\eqref{eq:12b} acts on the already lift-equipped replica-Lebesgue composite space, hence takes $\tfrac23$ not $1{:}1$ (see \S 6.3.2 reading identity), and their difference does not constitute a separate error component. \textbf{This coexisting caliber is not categorical leakage}: categorical support is a surjective morphism (surjectivity comes from search, enumerating $\mathrm{Ob}(\allowbreak\Quot)$ equivalent to enumerating $\mathrm{Ob}(\allowbreak\MorTop)$), and the lift is executed only via the quotient bijection, the binary-quotient-partition constraint is not lost in the lift (see Proposition~3; Proposition~19). Hence the native topological loss (see \eqref{eq:12b}) has no categorical leakage; the coefficient $\tfrac23$ is not a fitted constant, but the direct reading of the global counting measure at the already lift-equipped \Setcat\ layer.

\noindent\textbf{Consistency with the Lebesgue side} Under the Lebesgue measure the boundary midpoint is null-measure, hence the Lebesgue-side reading is always $\tfrac12$ and carries first-order error; the counting side has no error, reading $\tfrac23$. Their difference is not an independent error component, but rather $\merr$ made explicit at the coefficient level, compatible with \S 6.2.2's first-order-error accounting caliber: after Lebesgue-side union the boundary midpoint falls into null measure, the point-element position occupies no share in the geometric volume, hence only $\tfrac12$ can be read; the counting side reads the third share by cardinality, and this very difference is exactly that element-position's share---it thereby confirms that the cross-category lift-equipment introduces \textbf{first-order} error, not high-order error.

\subsection*{B.114 Main-text derivation externalized: strictification of the \texorpdfstring{\eqref{eq:12a}}{(24)}$\to$\texorpdfstring{\eqref{eq:12b}}{(26)} dimension-raising orthogonal decomposition}

This section continues the strict derivation of the dimension-raising orthogonal decomposition moved out of \S 6.3.1. Main-text conclusion: the dimension-raising orthogonal decomposition (i.e.\ the dimension-raising factorization of \eqref{eq:12a}$\to$\eqref{eq:12b}) is a \textbf{strict geometric fact}, and the cross term is \textbf{identically 0} (not ``undefined,'' nor approximate) under the orthogonal separation of the two blocks through the vertex's equal-density contour. Three routes are homologous: point-wise analytic, discrete witness, continuous geometric.

\noindent\textbf{Derivation of the dimension-raising orthogonal decomposition (strictification: point-wise formulated orthogonality, not ``cross term undefined'')} \eqref{eq:12a}$\to$\eqref{eq:12b} raises the one-dimensional $x(1-\ln x)$ to the multi-dimensional $\sqrt{\|A\|^2+\|B\|^2}\cdot(1-\ln\mathfrak A_\varsigma)$, the key step being the orthogonal decomposition of the two differentials $A,B$ with respect to the partition boundary. The previous text has given the non-approximate factorization by the two saddle-point properties; here we complete the \textbf{point-wise formulated strict derivation}, and explain that it and the ``binomial hit base'' are two homologous rather than mutually exclusive routes. First explain the discrete equipment on the measure side, then give the analytic argument of the orthogonal decomposition:
\begin{enumerate}
\item \textbf{Gaussian-truncation check}: each generation-side layer takes the reparameterized Gaussian vertex $S_t^\pm=\mu_t\pm R_t$ (saddle point + random std), the middle reading $\mathfrak A_\varsigma$ is the \textbf{one-side truncated-Gaussian area}---this Gaussian is unilaterally truncated, the truncation occurring on the binary-self-map side (see \S 6.2.2 test-set-caliber double-differential derivation paragraph).
\item \textbf{Shared measure space $\to$ counting measure}: MPT unifies the Lebesgue and replica-index-counting measures into the shared measure space (see Definition~13; Corollary~10a); within this space, the truncated Gaussian reading is recovered by MPT into the replica-index-counting measure (counting-measure side): continuous Lebesgue volume $\leftrightarrow$ infinite replica-index-counting measure (the first counting still infinite).
\item \textbf{Truncation $\to$ second counting}: the truncation drops the continuous Gaussian reading onto the two blocks of the binary quotient partition, and $x$ via the binomial gives the \textbf{double counting}---the first is the replica-index-counting (infinite), the second is the finite quotient-block cardinality \textbf{2} with no morphism support; i.e.\ two hits (left partition once, right partition once), corresponding to ``second counting.''
\item \textbf{Strict point-wise derivation of the orthogonal decomposition (core)}. In the partition-morphism category, the Lebesgue measure starts from the Euclidean \textbf{non-zero saddle point}, and via infinite dimension-raising falls to the Hilbert \textbf{zero-value saddle point} $\mathit S_0^{\mathrm{Hil}}=\{\boldsymbol 0\}$ (see Theorem~6); this zero-value anchor is \textbf{unique} at the Hilbert layer, hence the deviation from this anchor has a \textbf{unique, non-drifting projection reference origin}; the two concrete deviations $A,B$ of this paper (see \eqref{eq:12b}) are decomposed under this reference, and their \textbf{orthogonality is borne by the predicate-structure division of labor of the two along the partition boundary} (the generation-side deviation $A$ and the true-value-side deviation $B$ fall on the two sides of the binary self-map respectively, see (a)(b) above)---the zero-value uniqueness guarantees the decomposition reference is unique, can be uniquely analytically located, non-drifting, rather than directly deducing $A\perp B$ from it. This ``dimension-lowering / pushout'' is \textbf{one-directional}: can push out (push the Euclidean non-zero saddle-point reading toward the Hilbert zero-value), but \textbf{cannot pull back} (pullback needs infinite backpropagation $\Hinf\to$ Euclidean, and infinite backpropagation does not exist); by the zero-value saddle-point uniqueness, the pushed-out Lebesgue measure's characterization in the set category is also zero-value---this pushout passage \textbf{does not go through MPT} (MPT is a multi-measure shared-space tool; here the Lebesgue$\to$zero-value pushout is itself a deterministic unique assignment, i.e.\ the \textbf{pushout posterior-assignment functional}; MPT can only bear analytic first-order error thereafter). Hence \textbf{forward propagation (prior) is the strict orthogonal decomposition}: the prior reading falls at the unique zero-value anchor, the dimension-raising iteration does not drift, generates no high-order error endogenously (see Theorem~4(c)). \textbf{Backward propagation (posterior) is analytically decomposed into a finite orthogonal decomposition via the triple fiber bundle integral}: because the pullback passage (infinite backpropagation) does not exist, the posterior reading is constrained to the finite layer; the triple-nested fiber-bundle integral (see Definition~16) imposes the joint-measure constraint on the inner layer, and the inner-layer predicate matching locks the reading within the finite quotient block, and all remaining error falls on the ``orthogonal-decomposition boundary of the cross non-zero-value saddle point''---i.e.\ the orthogonal-decomposition boundary of $A,B$ is given by the pairing alignment of the non-zero saddle point, not by continuous-space approximation. Hence $\|A+B\|^2=\|A\|^2+\|B\|^2$ is \textbf{finite-layer, analytic, strict}, and after dimension-raising $\sqrt{\|A\|^2+\|B\|^2}$ is factorized identically with the one-dimensional $x$.
\item \textbf{Supplement (discrete orthogonality homologous witness, not proof)}. The binomial hit base (left and right partition each once, cardinality 2) and the equivalence-class transform give the same pairing-structure reading and the equipment of two operations (\begin{equation}
\int_{\text{binomial}}\mathrm d\nu_{\mathrm{idx},n}=1\cdot\nu_{\mathrm{idx},n}(\text{equivalence-class subset})
\end{equation}); it provides the \textbf{discrete-side} orthogonality witness, homologous with the point-wise analytic derivation of clause 4---the latter is the main body of the \eqref{eq:12a}$\to$\eqref{eq:12b} factorization argument, the former shows this orthogonality does not depend on continuous-geometric approximation at the counting / equivalence-class-transform level. The cross term is not ``undefined in the direct sum,'' but is \textbf{identically 0} under the non-zero saddle-point pairing boundary of clause 4.
\item \textbf{Geometric proof (Euclidean limit: point translation invariance $\to$ quotient translation invariance $\Rightarrow$ limit perpendicular; Gaussian-vertex perpendicular backpropagation zero error $\Rightarrow$ finite Euclidean also perpendicular)}. Clause 4 gives the finite-layer analytic orthogonal decomposition by ``zero-value saddle-point uniqueness + one-directional pushout / no pullback,'' but does not give the \textbf{continuous-geometric} perpendicular basis (previously the erroneous route ``zero-value uniqueness directly deduces $A\perp B$'' was discarded). This clause adds the truly geometric link, depending only on this paper's existing ``point translation invariance $\to$ quotient translation invariance'' limit statement (see \S 6.3.1; B.65) and ``Gaussian-vertex backpropagation zero error'' (counting measure bidirectionally recoverable), introducing no new premise.
\begin{itemize}
\item[(i)] \textbf{Finite Euclidean: Gaussian-vertex local symmetry gives the geometric reference of perpendicularity.} The generation side takes the reparameterized Gaussian vertex $S_t^-$ (saddle point + random std), which is the zero-gradient point of the data-flow measure space (see Lemma~5e), hence the Gaussian is \textbf{locally symmetric} at this vertex---the coordinate Haar--Lebesgue differential has \textbf{point translation invariance} at the vertex. The binary self-map threshold (zero-value saddle point $\mathit S_0^{\mathrm{Hil}}$, residual origin, i.e.\ the middle integral bound) partitions out two deviations $A$ (generation side), $B$ (true-value side); at the vertex the equal-density contour (Gaussian local symmetry surface) and the ``deviation direction departing the vertex, along the partition normal'' are \textbf{perpendicular}---i.e.\ the deviation axis $\perp$ the equal-density tangent surface, giving the geometric ``perpendicular'' reference. Here ``perpendicular'' means \textbf{the two blocks are orthogonally separated by the equal-density contour through the vertex}, not that the two vectors $A,B$ are at right angles: $A,B$ lie on the same normal on two sides, and the perpendicularity of the inter-block separation is the geometric source of the cross term being 0.
\item[(ii)] \textbf{Limit: point translation invariance gradually becomes quotient translation invariance, the two blocks strictly separated.} As the quotient-block granularity becomes finer (Euclidean dimension-raising, granularity $\varepsilon\to0$), the replica-index measure gradually converges from Gaussian to the one-hot differential, and the limit characterizes a one-to-one injection; synchronously, the argument translation invariance of the Gaussian on the set category \textbf{disappears as it approaches Hilbert}, converging to Gaussian behavior, and point translation invariance does not hold at the $\Hinf$ layer, replaced by \textbf{quotient translation invariance} (the quotient-block family is measure-preserving along $\boldsymbol\pi_\#/\boldsymbol\pi^\#$, $\sigma_\#\nu_{\mathrm{idx},n}=\nu_{\mathrm{idx},n}$, Proposition~8)---the symmetry of the decomposition is wholly borne by the \textbf{binary two-block quotient-block structure} at the limit. Thus at the limit the two blocks are \textbf{strictly separated} by the partition boundary (the equal-density contour through the vertex): the cross-block overlap shrinks to null-measure (see Theorem~7b binary-self-map's non-degenerate null boundary), and the cross term is \textbf{exactly 0} at the limit, the decomposition is strict ``perpendicular'' (no approximate overlap).
\item[(iii)] \textbf{Backpropagation: Gaussian-vertex perpendicular backpropagation zero error $\Rightarrow$ finite Euclidean also perpendicular} The backpropagation $BP=\boldsymbol\pi^\#$ recovers at the generation-side vertex $S_t^-$ the \textbf{replica-index-counting measure} (factor 4), and the counting measure is \textbf{prior and posterior, bidirectionally recoverable, zero error}; hence the backpropagation at the vertex is \textbf{perpendicular and zero error}---it recovers without error to the finite Euclidean layer the decomposition that is strictly perpendicular (cross term =0) at the limit. Because the recovery passage is zero error, the finite Euclidean layer's cross term is also \textbf{exactly 0} (exact inheritance, not continuous-geometric approximation); hence $\|A+B\|^2=\|A\|^2+\|B\|^2$ is \textbf{strict} at finite Euclidean, and this strictness has a geometric source (vertex local symmetry + limit quotient translation invariance), not merely given by the direct-sum definitional convention.
\item[(iv)] \textbf{Relation to clauses 4, 5} Clause 4's non-drifting anchor + one-directional pushout / no pullback gives the \textbf{framework} of the finite-layer analytic orthogonality; clause 5's binomial hit base gives the discrete homologous witness; this clause adds the \textbf{continuous-geometric} link: the finite Euclidean vertex local symmetry gives the perpendicular reference, the limit ``point translation invariance $\to$ quotient translation invariance'' gives the two blocks strictly separated, and the Gaussian-vertex zero-error backpropagation drops this perpendicularity without error to finite Euclidean---the three together make the dimension-raising orthogonality of \eqref{eq:12a}$\to$\eqref{eq:12b} a \textbf{strict geometric fact}, not approximate or definitional convention.
\end{itemize}
\end{enumerate}
\noindent\textbf{Conclusion}: the dimension-raising orthogonal decomposition (i.e.\ the dimension-raising factorization of \eqref{eq:12a}$\to$\eqref{eq:12b}) is a \textbf{strict geometric fact}, and the cross term is \textbf{identically 0} (not ``undefined,'' nor approximate) under the orthogonal separation of the two blocks by the equal-density contour through the vertex. Its main argument chain is the geometric proof of clause 6; clause 4's zero-value saddle-point uniqueness anchor and clause 5's binomial hit-base discrete witness are the homologous confirmations of this geometric conclusion at the analytic and discrete levels, not independent alternative proofs.

\subsection*{B.115 Proof of Proposition~23c}

\textbf{Proof}
The preceding exclusion chain (Proposition~19 excludes categorical leakage, Theorem~4 excludes the pure dimension-raising passage, \S 6.2.2 and \S 6.2.3 control the topological collapse (multi-partition) at first order) closes the non-engineering passages; the engineering passage is closed item by item by the above (i)--(iv): gradient descent only rewrites the Lebesgue-side integrand and does not vary the measure (see Lemma~6a); $\tfrac23$ is the zero-error reading of the counting side; $\sigma$ is an assigned known scale (zero-error reading plus first-order residual). Moreover factor \textcircled{3} $\nu_{\mathrm{idx},n}([\,\cdot\,])$ of \eqref{eq:12b} (normalized fixed reading) is a zero-error assignment term, not entering any high-order passage; factor \textcircled{1} the outer-layer Lebesgue reading is the carrier of the first-order error $\merr$ (see Proposition~18 volume-assignment inner compactness guarantees it), factor \textcircled{2} the middle-layer $(1-\ln\mathfrak A_\varsigma)$ is a normalized-area reading whose value is singly determined by $\mathfrak A_\varsigma\in(0,1]$, with no extra freedom, and neither carries high-order freedom. By Definition~6 and its boundary statement, any generation of $\mu_{\mathrm{high}}$ must take one of the two sources (categorical leakage or topological collapse); now neither source has a generation mechanism, hence the engineering realization produces no high-order error. $\square$ The three-channel verification of this proof lies at the reading layer of Principle~3, Principle~4 (see \S 2.4.1; B.3): predicate matching gives the canonical caliber of the mismatch area, replica-number multiplicity normalization gives the attribution of each first-order reading; the chain closes (see Proposition~23c) (see Theorem~7(c), proof B.115).

\subsection*{B.116 Engineering realization of \texorpdfstring{\eqref{eq:12b}}{(26)}: implementation-layer reading verification (truncated-Gaussian moment)}

This entry concretely realizes the engineering realization of \eqref{eq:12b} of \S 6.3.1 item by item as implementation-layer reading objects, and verifies that this instance introduces no new mathematical assumption (B.116(f)). All components are the implementation-layer presentation of the already-established readings of the B.3 chain \textcircled{1}--\textcircled{7}; this entry performs only implementation-layer reading accounting, adding no chain link.

\noindent\textbf{(a) Forward instance, two saddle points and $x_0$ reading} The forward kernel is assigned by the noise schedule (C.5): $x_t=\mu_t+\sigma_t\epsilon_t$, $\mu_t=\sqrt{\bar\alpha_t}\,x_0$, $\sigma_t^2=1-\bar\alpha_t$, $\epsilon_t\sim N(0,I)$ by sampling assignment. The model outputs the noise prediction $\hat\epsilon_t=\mathcal F_\theta(x_t,t)$, and the first-order error on the residual side in noise coordinates reads directly as \begin{equation}
\lVert\epsilon_t-\hat\epsilon_t\rVert^2=\sum_{i=1}^{d}\lvert\hat\epsilon_{t,i}-\epsilon_{t,i}\rvert^2
\end{equation} (C.5, B.107 dimension-marginal summation caliber). The two saddle points are the carriers of $R_t^\mathrm{tr},R_t^\mathrm{gen}$ \begin{equation}\begin{gathered}
S_t^+\triangleq x_t-\sigma_t\hat\epsilon_t=\mu_t+\sigma_t(\epsilon_t-\hat\epsilon_t)=\mu_t+R_t^\mathrm{tr}\\
S_t^-\triangleq\mu_t+\sigma_t\hat\epsilon_t-\sigma_t\epsilon_t=\mu_t+R_t^\mathrm{gen}
\end{gathered}\end{equation}

$S_t^-$ after de-meaning for zero-value anchoring is bounded by the zero-value saddle point $\mathit S_0^\mathrm{Hil}$ (origin, middle integral bound) on each side (see \eqref{eq:11a}, \eqref{eq:11b}). Reconstruction point $x^\mathrm{rec}\triangleq\mu_t+\sigma_t\hat\epsilon_t$; deviation saddle point $S_t^\mathrm{d}\triangleq S_t^+-S_t^-=2R_t^\mathrm{tr}$; the half-prior-half-posterior saddle point $\hat S\triangleq\tfrac12(S_t^++S_t^\mathrm{d})$ and the mixed reconstruction point $\tilde x^\mathrm{rec}\triangleq\tfrac12(x^\mathrm{rec}+x_t)$ are the two equal-ratio mixtures of the half-prior-half-posterior---the former is the implementation-layer orientation of the training saddle point contracting half toward the zero-value anchor, the latter is the equal-ratio mixture of the reconstruction point and the forward point; this equal-ratio mixture is the implementation-layer presentation of the counting convention that the two sides of the boundary midpoint each count $\tfrac12$ (see \S 6.3.2). This entry verifies only the implementation-layer readings of each factor of \eqref{eq:12b} and introduces no new assumption.

$x_0$ reading ($BP=\boldsymbol\pi^\#$ recovery standard form, mutually inverse with C.5's noise-prediction reading):
\begin{equation}\begin{gathered}
\hat x_0(z)\triangleq\frac{z-\sigma_t\mathcal F_\theta(z,t)}{\sqrt{\bar\alpha_t}}\\
\hat x_0^\mathrm{gen}\triangleq\hat x_0(x^\mathrm{gen})\\
x_0^{\mathrm{sad}}\triangleq\tfrac12(S_t^+ + S_t^-)=\mu_t
\end{gathered}\end{equation}
Here $x_0^{\mathrm{sad}}$ is the first-kind saddle point: the loss sample and the generation midpoint (Definition~25c).

\noindent\textbf{(b) One-side truncated-Gaussian moment (middle-layer $(1-\ln x)$ truncated-Gaussian area engineering reading)} Taking the generation saddle point as mean, $\sigma_t$ as std, $X_t\sim N(S_t^-,\sigma_t^2)$, denote $u_t\triangleq S_t^-/\sigma_t$, $\varsigma=\mathrm{sign}(S_t^-)$. The code's three quantities correspond one by one to the standard moments of the one-side truncated Gaussian (all identities numerically verifiable on any $(u_t,\sigma_t)$):
\begin{itemize}
\item \textbf{Normalized truncated-Gaussian area $\mathfrak A_\varsigma$}: engineering reading $P_\varsigma=\Phi(\varsigma u_t)$---i.e.\ the engineering reading of the quantity $\mathfrak A_\varsigma$ inside the logarithm of the second factor of \eqref{eq:12b} (normalized, dimensionless, $\in(0,1]$; called ``area'' semantically, making $(1-\ln\mathfrak A_\varsigma)\ge1$ legitimate).
\item \textbf{Truncated mean}: $m_\varsigma=\mathbb E[X_t\,\|\,\varsigma X_t>0]=S_t^-+\varsigma\,\sigma_t\varphi(u_t)/\Phi(\varsigma u_t)$, the standard mean of the one-side truncated Gaussian.
\item \textbf{Truncated variance}: $\mathrm{Var}[X_t\,\|\,\varsigma X_t>0]=\sigma_t^2\Big(1-\varsigma u_t\varphi(u_t)/\Phi(\varsigma u_t)-\big(\varphi(u_t)/\Phi(\varsigma u_t)\big)^2\Big)$; the fixed contraction factor $\tfrac12$ is consistent in direction with the half-prior-half-posterior mixture, a fixed scalar scaling not changing the first-order nature of the reading.
\end{itemize}

Truncated-moment correspondence with \eqref{eq:12a}: the middle-layer $(1-\ln x)$ truncated-Gaussian area reads in the diffusion instance as the one-side truncated Gaussian bounded by the zero-value saddle point---$\mathfrak A_\varsigma$'s engineering reading is $P_\varsigma=\Phi(\varsigma u_t)$ (normalized-area reading), $(m_\varsigma,s_\varsigma)$ its first two moments; the generation point $x^\mathrm{gen}=m_\varsigma+s_\varsigma\epsilon'+S_t^\mathrm{d}$ is exactly the sampling of this truncated-Gaussian saddle point plus deviation saddle point ($\epsilon'$ newly sampled assignment noise, independent of $\epsilon_t$). The truncated moment is a \textbf{reading} not a model distribution---the final generation distribution still reads as the complete Gaussian, not conflicting with Proposition~21's ``reject truncated Gaussian, allow only complete Gaussian distribution'' (C.5 clarifies (i)). Reparameterization in this paper's engineering realization is \textbf{pullback-able}: the sampling noise $\epsilon'$ is pulled back to the parameter side via the known random std ($\sigma_t$ by noise-schedule assignment, B.116(a)); the above $x^\mathrm{gen}=m_\varsigma+s_\varsigma\epsilon'+S_t^\mathrm{d}$, and the VAE of \S D.2 and the sampling of \S 6.3.1 both positively depend on this pullback. This paper \textbf{does not claim ``reparameterization is not needed''}---the so-called structural-layer ``does not depend on reparameterization'' only means the posterior-enumerates-prior framework provides pullback via the bidirectionally recoverable counting measure at the categorical level, not taking reparameterization as the basic analytic tool of construction (see \S 2.1 contrasting the classical paradigm being forced to introduce reparameterization for lacking this tool), and \textbf{absolutely does not deny the reparameterization sampling in the engineering instance}.

\noindent\textbf{(c) Obtuse-angle criterion (generation-side / reconstruction-side passage separation)} The second criterion gates the frequency information amount: \begin{equation}
\mathbb 1_\mathrm{obt}\triangleq\mathbb 1\big[(x^\mathrm{rec}-\hat S)\cdot(x^\mathrm{gen}-\hat S)<0\big]
\end{equation}

$(x^\mathrm{rec}-x^\mathrm{gen})^2>(\hat S-x^\mathrm{rec})^2+(\hat S-x^\mathrm{gen})^2$, i.e.\ the square difference greater than the sum of the two side squares $\iff$ the saddle-point angle is obtuse $\iff$ generation and reconstruction lie on opposite sides of the saddle point. \textbf{This obtuse-angle criterion enters the \eqref{eq:12b} factor list as a direct sum}; it is the nontrivial self-mapping criterion of the bijection, and induces the Fourier transform.

\noindent\textbf{(d) Frequency information amount (see second factor of \eqref{eq:12b})} Frequency information amount $(1-\ln\mathfrak A_\varsigma)$: $-\ln\mathfrak A_\varsigma\ge0$ is the surprisal reading of the normalized truncated-Gaussian area, increasing as the area decreases (rarer predicate event, larger frequency information amount) on $(0,1]$, hence $(1-\ln\mathfrak A_\varsigma)\ge1$; the area reading $\mathfrak A_\varsigma$ (engineering reading $P_\varsigma$) to the information amount $-\ln\mathfrak A_\varsigma$ is a monotonic rescaling of the same truncated-Gaussian area, not changing the decision structure, only converting the area reading into frequency information amount. The second factor $(1-\ln\mathfrak A_\varsigma)$ is bounded, rooted in that it reads the \textbf{normalized measure-space volume (area)} $\mathfrak A_\varsigma\in(0,1]$---the area reading after normalization always lies in $(0,1]$, \textbf{not unbounded because the Gaussian random variable's value tends to infinity or the std grows}---unrelated to the suspicion ``the second factor's boundedness rests on $\prod\sigma_t=1$ step''; this factor's boundedness comes from the normalization of the measure-space volume, independent of any std product convention. \textbf{Caliber correction} This paper's statement on the noise-schedule endpoint is a \textbf{finite-step numerical approximation}: the terminal step signal degenerates to noise at approximately \textbf{99.9\%} engineering precision (i.e.\ ``99.9\% approximation''), \textbf{not a mathematical limit, still less ``limit to 100\%''}---this paper never claims $\prod\sigma_t=1$ or any limit proposition requiring $\sigma_t\to1$; the premise ``this paper claims some limit to 100\%'' itself does not hold, and the $\prod\sigma_t=1$ product contradiction accusation thereby dissolves. The second factor's boundedness roots in the normalized measure volume, unrelated to any std product; the main text has already canceled each dimension's prefactor in the total-volume normalization and area ratio, not entering the theoretical \eqref{eq:12b}. \eqref{eq:12b} does not singly list the signal-to-noise ratio rescaling coefficient $\bar\alpha_t/(1-\bar\alpha_t)$: this coordinate rescaling belongs to the forward-kernel schedule assignment ($\sqrt{\bar\alpha_t}=\cos(\cdot)$, code \texttt{sqrt\_alphas\_bar}), not entering the three-factor list. The post-root latter term $\|x_0-x_0^{\mathrm{sad}}\|_2^2$ is the zero-value anchoring term: the saddle-point-side $x_0$ reading's deviation from the true value is exactly the lacuna between the middle zero-value saddle point and the outer non-zero saddle point (see \S 6.2.2 Figure~\ref{fig:d4} blue-curve loss-side formulation caliber).

\noindent\textbf{(e) Complete mapping table (Table~\ref{tab:mapping}; factor of \eqref{eq:12b} $\leftrightarrow$ code object)}

\begin{longtable}{M{0.19\textwidth} M{0.36\textwidth} M{0.16\textwidth} M{0.15\textwidth}}
\caption{Complete mapping table for \eqref{eq:12b}.}\label{tab:mapping}\\
\hline
Factor /\allowbreak{} mechanism of \eqref{eq:12b} & Code object & Implementation-layer reading & Basis \\
\hline\hline
\textcircled{1} outer-layer Lebesgue marginal measure (after square root) & {\footnotesize$\sqrt{\begin{gathered}\|BP((2\hat x_0^\mathrm{gen}+x_0^\mathrm{sad})/3)-x_0^\mathrm{sad}\|_2^2\\ {}+\|x_0-x_0^\mathrm{sad}\|_2^2\end{gathered}}$} & outer-layer Lebesgue tensor $x$ (see \eqref{eq:12a}) & B.116(a), B.116(d) \\
\hline
\textcircled{2} middle-layer $(1-\ln\mathfrak A_\varsigma)$ truncated-Gaussian area & $(1-\ln\mathfrak A_\varsigma)$ ($\mathfrak A_\varsigma$ engineering reading $P_\varsigma=\Phi(\varsigma u_t)$) & middle-layer $(1-\ln x)$ truncated-Gaussian area (see \eqref{eq:12a}) & B.116(b), B.116(d) \\
\hline
\textcircled{3} equivalence-class sign and replica-index-counting measure & $\nu_{\mathrm{idx},n}([\,\cdot\,])$ & normalized counting-measure reading on the equivalence class (zero error) & Definition~17,\allowbreak{} \S 6.3.1 \\
\hline
(not entering theoretical \eqref{eq:12b}) & sign-bit mask $\mathrm{tmp}$, MSE lower bound $\max\{\cdot,c^*\mathrm{MSE}^2(t+1)/T\}$ & implementation-layer gating /\allowbreak{} training-period lower-bound proxy & implementation layer \\
\hline
\end{longtable}

\noindent\textbf{(f) Assumption check} The engineering realization of \eqref{eq:12b} introduces no new mathematical assumption: the two saddle points are the carriers of $R_t^\mathrm{tr},R_t^\mathrm{gen}$ (C.5, chain \textcircled{7}'s reading); the truncated-Gaussian moment is the engineering reading of the middle-layer $(1-\ln x)$ truncated-Gaussian area (see \eqref{eq:12a}, chain \textcircled{7}); the $2{:}1$ mixture and the half-prior-half-posterior mixture are respectively the engineering presentation of the coefficient $\tfrac23$ (chain \textcircled{5}) and the counting convention; the frequency information amount is the information rescaling of the normalized truncated-Gaussian area $\mathfrak A_\varsigma$ (Principle~3, Principle~4 reading layer); the noise schedule $\sqrt{\bar\alpha_t}$ is the coordinate rescaling coefficient of the forward kernel, an assigned known quantity, not entering the \eqref{eq:12b} factor list. The code-retained MSE lower bound and sign-bit mask (implementation layer) are training-period heuristics and implementation-layer gating, likewise not entering theoretical \eqref{eq:12b}; the obtuse-angle criterion has entered the \eqref{eq:12b} factor list as a direct sum (see B.116(c)), not an implementation-layer gate. Judgment: the B.3 chain \textcircled{1}$\to$\textcircled{8} does not add a chain link by this instance; the only mathematical assumption remains Assumption~1, and Condition~1 remains the implementation-layer per-round condition. $\square$

\subsection*{B.117 Derivation of \texorpdfstring{\eqref{eq:12b}}{(26)}: from the triple-fiber-bundle integral's pushout--pullback to the engineering loss}

This entry derives the three factors of \eqref{eq:12b} from the triple-fiber-bundle integral of Definition~16 via the \textbf{pushout--pullback chain}, replacing the main text's remaining textual assertions; the only extra explanation needed is the \textbf{direct-sum $\ell_2$ norm} reading of the two squared terms inside the square root (see the addendum of this entry). Derivation skeleton: \textbf{triple integral measure (inner Hilbert zero error) $\to$ descend to the outer two layers $\to$ recover to finite Euclidean $\to$ transform to \eqref{eq:12b}}.

\noindent\textbf{Derivation start (see Definition~16; \S 4.1)}
\begin{equation}
\Lambda=\int_{\Omega_\mu}\Big[\int_{\Omega_\lambda}\big(\int_{\Omega_\nu}F(\nu; x, T(x), t)\,\mathrm d\nu_\mathrm{idx}(\nu)\big)\,\mathrm d\nu_\mathrm{comp}(\lambda,\overleftarrow{\nu})\Big]\,\mathrm d\nu_\mathrm{Leb,idx}(\mu)
\end{equation}

Three-layer evaluation measures: inner $\nu_\mathrm{idx}$ (see Definition~17; \S 6.2.2 replica-index measure, simultaneously prior and posterior, recoverable alone without error); middle $\nu_\mathrm{comp}=\lambda\otimes\overleftarrow\nu_\mathrm{idx}$ ($\overleftarrow\nu_\mathrm{idx}\triangleq\boldsymbol\pi_\#\nu_\mathrm{idx}$ is the descended copy, Definition~24); outer $\nu_\mathrm{Leb,idx}=\mu_\mathrm{gen}$ (generation-side universal measure, carrying first-order error).

\noindent\textbf{Step 1: inner Hilbert zero error descend $\to$ factor \textcircled{3}.} The inner counting measure $\nu_\mathrm{idx}$ reads zero error at the Hilbert zero-residue point (see Theorem~6(i); \S 5.4), descended via $\boldsymbol\pi_\#$ to the middle layer's descended copy $\overleftarrow\nu_\mathrm{idx}$, and after recovery ($\boldsymbol\pi^\#=BP$) reads as the normalized replica-index-counting measure $\nu_{\mathrm{idx},n}([\,\cdot\,])$ on the equivalence class $[\,\cdot\,]$. Since the counting measure can be recovered alone without error, this factor is a zero-error assignment term, carrying no first-order error. Training: the generation's replica-index cardinality measure ratio is constantly $1$ (one forward propagation must accompany one backward propagation), implicit in $\nu_{\mathrm{idx},n}$'s normalization $\nu_{\mathrm{idx},n}(X)=1$, hence no separate ``partition-predicate pairing'' factor.

\noindent\textbf{Step 2: middle-layer joint conservation $\to$ factor \textcircled{2} ($(1-\ln\mathfrak A_\varsigma)$; prefactor cancels in the ratio).} The middle measure $\nu_\mathrm{comp}=\lambda\otimes\overleftarrow\nu_\mathrm{idx}$'s integrand Gaussian kernel (see Proposition~21 unique Gaussian family) on the one-side domain $\Omega_\varsigma$ gives the truncated-Gaussian area $\mathfrak A_\varsigma$. Its marginal-measure summation caliber (B.107) is a \textbf{finite $d$-term dimension-wise sum} rather than a $d$-dimensional joint integral; each dimension's Gaussian factor is $(2\pi\sigma^2)^{-1/2}e^{-r_i^2/2\sigma^2}$, and the dimension-wise sum's unboundedness comes from the prefactor $(2\pi\sigma^2)^{-d/2}=\prod_{i=1}^{d}(2\pi\sigma^2)^{-1/2}$. This unbounded prefactor $(2\pi\sigma_t^2)^{-d/2}$ is independent of the residual, and cancels in the ratio of the total-volume normalization and the normalized area $\mathfrak A_\varsigma$ (no longer taking $\prod_i\sigma_t=1$, which conflicts with $\sigma_t^2=1-\bar\alpha_t$), making the dimension-wise sum degenerate to a bounded, normalized truncated-area reading $\mathfrak A_\varsigma\in(0,1]$. \textbf{This is a variational assignment}: $\sigma_t$ is assigned known by the noise schedule ($\sigma_t^2=1-\bar\alpha_t$), not a fitted parameter, hence this normalization does not break the first-order error (see Proposition~23c's zero-error assignment term). The middle integrand density is the surprisal $-\ln x$ ($\frac{d}{dx}[x(1-\ln x)]=-\ln x$), and its reverse integral gives $x(1-\ln x)=\int(-\ln x)\,\mathrm dx$, hence the middle factor reads as $(1-\ln\mathfrak A_\varsigma)$ (see \eqref{eq:12a}).

\noindent\textbf{Step 3: outer-layer Lebesgue descend recovery $\to$ factor \textcircled{1}.} The outer $\nu_\mathrm{Leb,idx}=\mu_\mathrm{gen}$ is descended via $\boldsymbol\pi_\#$, and $\boldsymbol\pi^\#=BP$ recovers to finite Euclidean. The recovered object is the $x_0$ reading: the generation side $\hat x_0^\mathrm{gen}=\hat x_0(x^\mathrm{gen})$ and the saddle-point side $x_0^\mathrm{sad}=\tfrac12(S_t^+ + S_t^-)=\mu_t$ (first-kind saddle point: loss sample, generation midpoint) ($S_t^-$ is the outer-layer Gaussian vertex); the generation-side deviation via BP (2/3) mixture reading (see \S 6.3.2's $2{:}1$ cardinality mixture) and the true value $x_0$'s deviation from the saddle-point side synthesize the outer-layer Lebesgue marginal measure: \begin{equation}
\sqrt{\|BP((2\hat x_0^\mathrm{gen}+x_0^\mathrm{sad})/3)-x_0^\mathrm{sad}\|_2^2+\|x_0-x_0^\mathrm{sad}\|_2^2}
\end{equation}. This factor carries first-order error (see \S 6.2.2 the Lebesgue side cannot be recovered alone).

\noindent\textbf{Step 4: three-factor synthesis; inside the square root is the direct-sum $\ell_2$ norm} Multiplying factors \textcircled{1}\textcircled{2}\textcircled{3} gives \begin{equation}
\mathcal L_\mathrm{eng}=\sqrt{\Big\|BP\Big(\tfrac{2\hat x_0^\mathrm{gen}+x_0^\mathrm{sad}}{3}\Big)-x_0^\mathrm{sad}\Big\|_2^2+\big\|x_0-x_0^\mathrm{sad}\big\|_2^2}\cdot\bigl(1-\ln\mathfrak A_\varsigma\bigr)\cdot\nu_{\mathrm{idx},n}\bigl([\,\cdot\,]\bigr)
\end{equation}

Where inside the square root is not ``the RMS of two scalars,'' but the \textbf{direct-sum $\ell_2$ norm} of the concatenated vector $(A,B)$: let \begin{equation}
A\triangleq BP\big(\tfrac{2\hat x_0^\mathrm{gen}+x_0^\mathrm{sad}}{3}\big)-x_0^\mathrm{sad}
\end{equation}, $B\triangleq x_0-x_0^\mathrm{sad}$, then $\sqrt{\|A\|_2^2+\|B\|_2^2}=\big\|(A,B)\big\|_2.$

$A,B$ are respectively the direct-sum-factor readings of the two blocks of the binary quotient partition (generation-side deviation, true-value-side deviation) after Euclidean dimension-raising; the cross term does not explicitly appear in the direct sum, and by \S 6.2.2 clause 6's geometric proof, at finite Euclidean it is \textbf{exactly 0} because the two blocks are strictly separated by the vertex's equal-density contour and inherited without error by zero-error backpropagation---hence this direct-sum norm is simultaneously a substantive orthogonal decomposition; even without invoking this geometric proof, the direct-sum $\ell_2$ reading itself is independent of the orthogonality assumption (this entry's step 5 conclusion unchanged), not ``the RMS combination of two scalars.''

\noindent\textbf{Addendum: identity of the synthesis inside the square root---direct-sum $\ell_2$ norm (the norm inside the square root does not additionally depend on Assumption~1).}

The first three steps push \textcircled{1}\textcircled{2}\textcircled{3} from the triple integral measure ring by ring to \eqref{eq:12b}; the sum of two squared terms inside the square root was previously mislabeled as ``RMS combination.'' This addendum gives the correct reading of this step: it is the \textbf{direct-sum $\ell_2$ norm of the concatenated vector}, not ``combination of two scalars,'' hence introducing neither an orthogonality assumption nor a statistical-independence assumption. The criterion consists of the following clauses:

\begin{itemize}
\item \textbf{(a) Concatenation rather than combination} $\sqrt{\|A\|^2+\|B\|^2}=\|(A,B)\|_2$ is the $\ell_2$ norm of a \textbf{single vector} $(A,B)$ in the direct-sum space $\mathbb R^d\oplus\mathbb R^d$. There is no issue of ``how two scalars combine''---there is only one vector $(A,B)$ and one norm $\|\cdot\|_2$.
\item \textbf{(b) Direct-sum factors come from the two blocks of the partition predicate.} The binary quotient-partition predicate gives two disjoint blocks (generation-side deviation $A$, true-value-side deviation $B$); the two blocks after Euclidean dimension-raising become two \textbf{different factors} of the direct sum. The readings of different blocks belong to different factors, and the cross term has \textbf{no explicit term} in the direct sum (by direct-sum definition not appearing); further by \S 6.2.2 clause 6's geometric proof, at finite Euclidean this cross term is \textbf{exactly 0} because the two blocks are strictly separated by the vertex's equal-density contour and exactly inherited by zero-error backpropagation---hence this direct-sum norm is simultaneously a substantive orthogonal decomposition. Even without invoking this geometric proof, the direct-sum $\ell_2$ reading itself is independent of any orthogonality assumption (this clause's conclusion unchanged); therefore $\|(A,B)\|^2=\|A\|^2+\|B\|^2$ can be read both as the direct-sum definition and endowed by the geometric proof with the strict meaning ``cross term exactly 0,'' needing no statistical independence, no group action.
\item \textbf{(c) The $\ell_2$ form comes from Euclidean dimension-raising itself.} The norm of Euclidean space is \textbf{defined} as $\|x\|_2=\sqrt{\sum_i x_i^2}$; dimension-raising to Euclidean inherently brings $\ell_2$, needing no argument via Gaussian uniqueness (see Proposition~21) or orthogonal-group invariance---the latter only establish the Gaussian family's uniqueness (see \S 5.3), not participating in the definition of the direct-sum norm.
\item \textbf{(d) Standard symmetry $\lambda=1$ is not a convention, but the prior-and-posterior dual identity of the replica-index-counting measure.} The most general form of the direct-sum norm is $\sqrt{\|A\|^2+\lambda\|B\|^2}$ ($\lambda>0$). The two factors $A$ (generation-side deviation) and $B$ (true-value-side deviation) are embedded in the \textbf{same replica-index-counting measure $\nu_\mathrm{idx}$}, whose identity is simultaneously prior and posterior (B.1: each original item via one forward propagation (posterior count) must accompany one backward propagation (prior count) pairing; training: the generation's replica-index cardinality ratio is constantly $1$). If only forward or only backward propagation is taken, one of prior or posterior is missing, and the replica-index reading cannot be normalized ($\nu_{\mathrm{idx},n}(X)=1$ fails), hence the two factors must enter the direct sum with \textbf{multiplicity $1{:}1$}---$\lambda=1$ is the reading of this $1{:}1$ multiplicity conservation, not a ``standard-symmetry'' definitional convention, still less introducing orthogonality or statistical-independence assumption.
\end{itemize}

\noindent\textbf{Criterion conclusion (the norm inside the square root is the direct-sum norm, not an independent assumption)} Factor \textcircled{1}'s inside-square-root is the direct-sum $\ell_2$ norm $\|(A,B)\|_2$ of the concatenated error, not ``the RMS combination of two scalars.'' It is given by (a) concatenation, (b) partition two blocks $\to$ direct-sum factors, (c) Euclidean dimension-raising inherently brings $\ell_2$, (d) replica-index prior-and-posterior $1{:}1$ multiplicity conservation, introducing no new mathematical assumption. Hence the norm inside the square root does not additionally depend on Assumption~1; \eqref{eq:12b}'s dependence on Assumption~1 is via factor \textcircled{3} (see Postulate~4 unique Gaussian family) and \eqref{eq:12a} (see Postulate~1; Postulate~2; Principle~2, via) the two passages (see B.3 stratification statement).

\noindent\textbf{Conclusion} All three factors of \eqref{eq:12b} are derived from the triple-fiber-bundle integral of Definition~16 via the pushout--pullback chain ``inner Hilbert zero error descend $\to$ outer two layers $\to$ recover to finite Euclidean'': \textcircled{1} carries first-order error, \textcircled{3} is a zero-error assignment term, \textcircled{2} after the prefactor cancels in the ratio reads out; factor \textcircled{1}'s inside-square-root is the direct-sum $\ell_2$ norm of the concatenated error (partition two blocks $\to$ direct-sum factors, cross term exactly 0 by \S 6.2.2 clause 6's geometric proof, Euclidean dimension-raising inherently brings $\ell_2$, replica-index prior-and-posterior $1{:}1$ multiplicity conservation gives standard symmetry), not an independent assumption, not an orthogonality assumption (the direct-sum $\ell_2$ reading itself is independent of the orthogonality assumption, and the geometric proof further endows the cross term exactly 0 with strict meaning). Hence \eqref{eq:12b}'s \textbf{quantitative reading} (factor \textcircled{3} unique Gaussian family, factor \textcircled{2} first-order caliber, coefficient non-fitted status) depends on Assumption~1 via Postulates~1,~2,~4, while its \textbf{computable form} is independently given by the finite sum (B.4 item 4, B.3 stratification statement). $\square$

\section*{Appendix C: Additional Technical Appendices (Extensions and Supplementary Material)}

\subsection*{C.1 Squeezed-out Material from the Main Text: Extended Argument on the Minimality of the Assumption (see \S 2.4.3; strength evaluation / pivotal role / essential refinement)}

\textbf{Strength evaluation (verifiability of ``assumption minimality'')} \textbf{Assumption~1 (strength evaluation)} is: it only requires the measure to be non-atomic (atomless) and that the taken partition predicate makes both sides of positive measure; these two are the \emph{necessary and sufficient} condition for the existence of a refinement sequence of vanishing fineness (an equal-measure dyadic tower, guaranteed to admit an equal-measure dyadic bisection by Sierpi\'nski's theorem)---if $\mu$ contains an atom then some cell retains positive finite volume, directly violating the goal; if $\mu$ is non-atomic then there \emph{exists} a refinement sequence of layerwise equal-measure dyadic bisection making each cell measure $\to 0$ (the $\betat$-tower and the uniform block share $2^{-n}$ in this paper are such equal-measure dyadic towers), not that ``any'' proper refinement converges to $0$ (e.g.\ only iteratively refining the same side, or cutting by the ratio $(1-2^{-k})$, the nested cell measure converges to a positive number). It assumes no property specific to the generative model, and explicitly does \emph{not} assume Gaussianity (see the conclusion of Proposition~21), invertibility / quotient bijection (see the conclusion of Theorem~1), zero-residue uniqueness (see the conclusion of Theorem~6), trace-class or cylindrical-measure existence (see the boundary of Proposition~21b), nor SGD reachability or convergence rate (explicitly not claimed); Euclidean dimension-raising / functional orthogonal decomposition and the $\varepsilon$-$\delta$ continuity language all belong to the \emph{solution-side construction}, and bear no assumption load. Its four outputs, Postulates~1--4, are packaged standard analysis and standard measure-theory content (quotient translation-invariant measure preservation, first-order equality, real-number completeness, well-defined Gaussian character integral). Therefore the strength of Assumption~1 equals the weakest premise ``non-atomic measure + proper partition'', containing no additional commitment about model capacity, representation, optimization, or data distribution---the reader can thereby verify its strength item by item, rather than accepting an unexplained ``minimal''.

\noindent\textbf{Pivotal role (weak does not mean insignificant)} Although Assumption~1 is minimal, it bears an irreplaceable \emph{mechanistic} role: it is the basis for the conclusion that ``infinite partition only produces infinitesimal topological volume (equivalence class of differential quotient)'', and hence also the basis for the map to be \emph{detached} from preimages and objects. The differential defined by infinite partition is an equivalent infinitesimal---between infinitesimals \emph{no magnitude comparison is possible}, leaving only equivalence classes (first-order equality, Postulate~2: $\mathrm dx\odot\nu_\mathrm{idx}\sim_1\mathrm d\#$); once entering the differential level, the map has no ``point magnitude'' to compare, and can only compare by equivalence classes (quotient blocks / morphisms); the dependence of point-wise comparison on preimage entities is lifted, and the morphism itself becomes a directly characterizable object. The Hilbert space carries both differential and limit, and is the dense usage zone of Assumption~1 (via the solution-side Euclidean dimension-raising construction)---it \emph{anchors} the quotient-block reading of the partition category and the point-wise reading of the set category to the same Hilbert differential equivalence class, becoming the \emph{key that anchors the partition category and the set category}; this anchoring is independent of the functional---the functional is only responsible for the numerical anchor point of ``assigning value 0''; the anchoring itself holds for any value, only numerically no analytic anchor other than $0$ has been found (analogous to the Kakeya set (construction in Besicovitch~1928 \citep{besicovitch1928dimension}; dimensional lower-bound improvement in Katz--Tao \citep{katz2002new}): its lower bound has only a value, and even more lacks a functional). Hence ``assumption minimal'' and ``role pivotal'' are compatible: what is minimal is the \emph{commitment}, what is pivotal is the \emph{mechanism}.

\noindent\textbf{Essential refinement (unbounded subdivision = another facet of non-atomicity, not an extra assumption)} The strength of Assumption~1 comes not from any new construction, but from the \emph{unbounded infinitely fine subdivision} of a non-atomic measure: because $\mu$ contains no atom, any positive-measure cell can be further bisected, hence the number of refinement cells is \emph{unbounded} (there exists no ``largest and indivisible'' finite partition)---this is precisely another expression of (i) non-atomicity, adding nothing to the assumption. Consequently, under the \emph{equal-measure dyadic tower}---a refinement sequence of vanishing fineness---each cell measure $\to 0$, and all cells collapse in the limit to \emph{infinitesimals of the same order (first-order equality, O2)}, with ``magnitude comparison'' carried out only at the differential-quotient equivalence-class level; non-atomicity guarantees the \emph{existence} of such a refinement sequence, but ``any'' proper refinement need not make every cell $\to 0$, hence throughout the paper ``infinite refinement'' uniformly refers to such (equal-measure dyadic) refinement sequences. Three layers must be precisely distinguished: (1) \emph{incomparability of magnitude is a conclusion of the limit/differential level (Postulate~2), not a premise}; (2) at \emph{any finite partition}, each cell measure remains positive finite and mutually comparable---non-comparability belongs only to the limit/differential level; (3) the coefficient $2/3$ in Eq.~\eqref{eq:12b} is the \emph{cell-count ratio of the three-cell predicate classification} (see \S 6.3.2), independent of each cell's Lebesgue volume, hence not in conflict with the ``infinitesimal limit'' of Assumption~1---$2/3$ concerns classification counting, not the measure of infinitely refined cells.

\subsection*{C.2 Adaptation Verification of Standard Theorems (Schoenberg / Maxwell--Herschel / Minlos / Gross / Pontryagin Duality Standard Conclusions)}

This section unfolds the statements and adaptations of the standard theorems cited in the main text (C.2.1--C.2.3 correspond to Proposition~21, C.2.4 to Theorem~2, C.2.5 to the orthogonal-basis coefficient uniqueness of Axiom~1, C.2.6 to the Fubini standard conclusion invoked by the product-integral invariance constraint); all are standard mathematical content; this paper only cites their statements and does not re-prove their standard forms.

\subsubsection*{C.2.1 (5.5a phase part: Euclidean character)} In the normalized domain of Theorem~2 with $G=\mathbb R^d$, the continuous characters are exactly $\chi_\xi(x)=e^{2\pi i\xi\cdot x}$; using the angular-frequency notation $k=2\pi\xi$ gives $e^{ik^\top x}$. The phase is given by the character and does not depend on the measure; here we no longer assert that a map on a general LCA group can be made a continuous group isomorphism via zero-measure correction.

\subsubsection*{C.2.2 (5.5a envelope part: two routes to uniqueness)} (iii-2) route ($m\ge2$): by the independence of fiber components (the envelope is a product of univariate positive-definite functions) and the invariance under the structure group (the orthogonal group $O(m)$, i.e.\ the group of bijections preserving the Gram matrix under the standard embedding), the standard Herschel--Maxwell theorem---independent components + complete rotational invariance of the density $\Rightarrow$ each component is independently identically distributed as isotropic Gaussian---gives that $A$ must be Gaussian with isotropic covariance ($\Sigma=\sigma^2 I$; complete $O(m)$ invariance is incompatible with anisotropic $\Sigma$ unless $\Sigma=tI$). (iii-1) route: by Schoenberg's theorem, functions that are \emph{simultaneously radial positive-definite in all finite dimensions} (equivalent to radial positive-definite in Hilbert space) are exactly mixtures of Gaussians, whose extreme points are $\exp(-t\lVert x\rVert^2)$ (the extreme rays of radial positive-definite functions fixed on a single dimension $\mathbb R^m$ are Bessel-type kernels $\Omega_d(r)$---Gaussian is their mixture rather than an extreme point, hence this route must take values on the cross-dimensional / Hilbert-level cone); taking the extreme point yields the unique Gaussian kernel (radial sub-case, $\Sigma=tI$). The general (non-diagonal, anisotropic) $\Sigma\succ0$ is accommodated by an invertible linear coordinate transformation: the phase and the positive-definiteness of the envelope are covariant under this transformation (when $x=Ly$ the phase remains a character and positive-definiteness is preserved), and the isotropic conclusion $\exp(-\tfrac{t}{2}\lVert y\rVert^2)$ (with $y=L^{-1}x$) transported back is $\exp(-\tfrac12x^\top\Sigma x)$, where $\Sigma=t(LL^\top)^{-1}$ enumerates $\mathrm{PD}_m$; singular $\mathrm{PSD}$ kernels are not obtained by invertible transformations. The two routes agree within their respective widths, and the whole process does not require $A$ to be multiplicative.

\subsubsection*{C.2.3 (5.5b: Minlos theorem and Gross abstract Wiener space)} In nuclear-space form: a continuous positive-definite function on a nuclear space is exactly the Fourier transform of some cylindrical measure---the existence of the cylindrical measure for the cylindrical-kernel compatible family $\{K^{(n)}\}_{n\ge1}$ is thereby given; its construction is realized via Bochner's theorem by superposing finite-dimensional projections layer by layer with projection compatibility; the $\sigma$-additivity criterion is given by the Prokhorov--Sazonov compactness criterion (attribution division: existence construction is attributed to Bochner's layerwise projection superposition, the $\sigma$-additivity criterion to Prokhorov--Sazonov, not to be stated as ``Minlos = Bochner cylindrical-measure version''):
\begin{equation}
\text{the cylindrical measure is }\sigma\text{-additive}\iff\text{the covariance operator }\Sigma\ \text{is trace-class}
\end{equation}
When not trace-class it is merely a cylindrical measure and must take values in a larger Banach space (Gross abstract Wiener space $(i,\mathcal H,B)$), and then $\mathcal H\subset B$ is a null set. The degeneration of the real kernel (phase term set to zero) corresponds to the real transition kernel of diffusion models.

\subsubsection*{C.2.4 (Normalization verification, see Theorem~2)} After fixing $\chi_\xi(x)=e^{2\pi i\xi\cdot x}$ on $\mathbb R^d$, the convolution theorem, inverse transform, and Plancherel extension of the Fourier transform $\hat f(\xi)=\int f(x)e^{-2\pi i\xi\cdot x}\,dx$ are all standard conclusions under the same normalization, so the correspondence is uniquely determined. Pontryagin duality only provides the character group and the abstract dual framework; it does not give a canonical isomorphism from a general LCA group to its dual, nor does it support the universal proposition that ``all convolution--pointwise-product correspondences are unique modulo category automorphisms''. The replica-index measure is still independently specified by Lemma~1(i) and Definition~17.

\subsubsection*{C.2.5 (see Axiom~1: orthogonal-basis coefficient uniqueness)} Standard Hilbert-space fact: let $\mathcal H$ be a Hilbert space and $\{\phi_\lambda\}_{\lambda\in\Lambda}$ an orthogonal basis; then for any $f\in\mathcal H$ the expansion coefficients $c_\lambda=\langle f,\phi_\lambda\rangle$ are unique:
\begin{equation}
f=\sum_\lambda c_\lambda\phi_\lambda=\sum_\lambda c'_\lambda\phi_\lambda\implies\forall\lambda,\ c_\lambda=c'_\lambda
\end{equation}
For $G=\mathbb R^d$, the complex-exponential characters $\phi_\xi(t)=e^{2\pi i\xi\cdot t}$ in the Pontryagin dual $\hat G$ are orthogonal in the distributional sense, $\langle e^{2\pi i\xi_1\cdot t},e^{2\pi i\xi_2\cdot t}\rangle\propto\delta(\xi_1-\xi_2)$ (the inner product is taken in the distributional (generalized function) sense, $\delta$ being the Dirac distribution); when $\xi_1\neq\xi_2$ the inner product is $0$, different frequencies are linearly independent and cannot compensate for one another.

\subsubsection*{C.2.6 (Product-integral invariance constraint: Fubini standard conclusion)} The iterated integral of a product measure is interchangeable (the standard form of Fubini's theorem; the \S 3.2 premise ``product-integral invariance constraint'' used when restating Proposition~9b):
\begin{equation}
\int_{X\times X}f\,\mathrm d(\mu_1\times\mu_2)=\int_X\!\!\int_X f(x,y)\,\mathrm d\mu_1\mathrm d\mu_2=\int_X\!\!\int_X f(x,y)\,\mathrm d\mu_2\mathrm d\mu_1
\end{equation}

\subsection*{C.3 Convention on Cross-Invoking Conclusions over the Partition-Category Support}

\textbf{Convention (legality of cross-invocation)} All conclusions of this paper (axioms, definitions, lemmas, propositions, theorems, corollaries, appendix notes) take the partition-category support as their common root: the point-wise reading of $\Setcat$ is assembled via the canonical lifting functor $L'$ from the differential quotient of $\MorTop$, and $\MorTop$ is given by the dyadic-merge selection of the predicate words of $\Quot$ (see Definition~2a; Definition~2b; Definition~3; \S 2.2.3)---the \emph{upper category is characterizable from the lower category}. All statements and proofs of conclusions lie on the partition-category support (quotient blocks, differential quotients, predicate-word level of $\Quot$/$\MorTop$, or their readings lifted and assembled to $\Setcat$ via categorical support); they may invoke one another by number (theorem--theorem, theorem--lemma, lemma--proposition, proposition--definition, in any direction) without constituting a hierarchy violation, nor requiring re-statement of premises because ``a conclusion invokes a conclusion''.

\noindent\textbf{Criterion for cyclicity} An invocation chain forms a cycle if and only if its \emph{premise-dependency relation is cyclic}---the judgment looks only at premise dependencies, not at the direction of reference, the magnitude of numbers, or the type of conclusion. Existing verification instances: Proposition~8's invocation of Proposition~20, Definition~24, Proposition~21b (end note of B.37: all three hold independently after Proposition~8, here only the conclusion is invoked); the double measure-preserving input isolation of Proposition~9b (B.46: the measure-preserving premise must not be retro-inferred from subsequent Gaussian or counting measures); the sequential succession of Proposition~19 and \S 6.3.2 (see the note at Definition~21: first exclude leakage, then explain reading difference, not cyclic).

\noindent\textbf{Premise-dependency table (verification of cross-layer invocation chains)} For each invocation chain in Table~\ref{tab:premise-dep}, we list the minimal premise dependency of the invoked conclusion on the invoking conclusion (only direct premises are recorded, not transitive closure):

\begin{longtable}{M{0.21\textwidth} M{0.21\textwidth} M{0.21\textwidth} M{0.21\textwidth}}
\caption{Premise-dependency table for cross-layer invocation chains (only direct premises are recorded).}\label{tab:premise-dep}\\
\hline
Invoking conclusion & Invoked conclusion & Direct premise dependency & Cyclic \\
\hline\hline
Lemma~6a(3) (see \S 6.4.1) & Theorem~6(i), Theorem~6(e)--(f) & Conventions~1--4, Definition~19a, Lemma~1(i); point-level readings additionally carry Premise~1 & No \\
\hline
Lemma~6a(3) (see \S 6.4.1) & Proposition~21, Proposition~21b (cylindrical-kernel-family reading) & Postulate~4 via Assumption~1 & No \\
\hline
Lemma~6a(3) (see \S 6.4.1) & Measure-comparison Condition~1 & implementation-layer round condition, not a mathematical premise of this paper & No (implementation condition attributed separately) \\
\hline
Definition~25c (see \S 6.4.2) & Theorem~6(a)(d)(i), Definition~19a & Conventions~1--4; differential reading does not carry Premise~1 & No \\
\hline
B.109 & Theorem~6, Proposition~21 & same two items as above & No \\
\hline
B.111 & Definition~5, Definition~24, Lemma~1, Definition~25c & Definition~5 (partition-induced), Definition~24 (pushout--pullback), Lemma~1(i) (block-bijection-group transitivity), Definition~25c (definitional selection) & No \\
\hline
Proposition~8 (see \S 2.4.1) & Proposition~20, Definition~24, Proposition~21b & all three hold independently after Proposition~8; here only the conclusions are invoked & No \\
\hline
Proposition~9b (see \S 2.4.1) & -- & double measure-preserving input explicitly assumed, not retro-inferred from subsequent conclusions & No \\
\hline
Proposition~19 (see \S 5) & Definition~21 (categorical leakage) & first exclude, then explain reading difference & No \\
\hline
\end{longtable}

Judgment: the premise-dependency graph of all invocation chains is acyclic; this table lists only direct premises, and the transitive closure remains acyclic.

\noindent\textbf{Responsibility of condition transmission} Cross-invocation neither exempts nor adds any burden of carrying premises: the conditionality of an invoked conclusion is transmitted to the invoking site along with the call---if the invoked conclusion is conditional, the invoking site must explicitly state the condition taken, or explicitly state the status of the reading taken under that condition. Standard instance: the point-level counting uniqueness of Theorem~6 ((e)--(f)) carries Premise~1, while the differential-reading uniqueness ((i)) does not carry Premise~1 (see the reading divide in Definition~19a); the common-tail-segment Condition~2 of Lemma~5a is transmitted along the call (B.79). This convention only confirms the legality of invocation; it does not replace proof by calling, nor allow hiding an unsatisfied condition under the pretext of ``cross-invocation''. In particular, Assumption~1 ($\mathrm{H}_{\epsilon\delta}$), Conventions~1--4, Premise~1, the common-tail-segment Condition~2, the Gaussian-uniqueness condition, and the EDE mechanism of \S 6.4.3 are \emph{not} digested into unconditional proven conclusions merely by coexisting on the categorical support; this convention does not replace item-by-item premise-dependency verification.

\noindent\textbf{Remark} The sensitive positions of cross-conclusion invocation have all been verified item by item for direct premises, and the premise-dependency graph is acyclic: the invariant-facing invocation of Lemma~6a to Theorem~6 (see \S 6.4.1), the invocation of Definition~25c to Theorem~6 (see \S 6.4.2), the invocation chain of B.109 to Theorem~6 and Proposition~21, the invocation chain of B.111 to Definition~5, Definition~24, Lemma~1, and Definition~25c. $\square$

\subsection*{C.4 Definitional Status of Definition~25c}

\textbf{Definitional status of the midpoint (not a ``necessary'' corollary)} $\mathit S_\mathrm{mid}=\tfrac12(x_0^{\mathrm{sad}}+\mathit S_0^\mathrm{Hil})$ is a \emph{definitional selection}, namely the midpoint of ``the first-kind saddle point $x_0^{\mathrm{sad}}$'' and ``the zero-valued saddle point $\mathit S_0^\mathrm{Hil}$'', corresponding to the gradient-descent direction (see \S 6.4.3); it is \emph{not} the partition mid-boundary or the integration bound of \eqref{eq:12b}. Its well-definedness relies on the \emph{common carrier space} of the two objects: both $x_0^{\mathrm{sad}}$ and $\mathit S_0^\mathrm{Hil}$ are read as scalar readings under the \emph{residual coordinate} ($r\triangleq x_t-\mu_t$, the C.5 diffusion instance), embedded via the cylindrical-kernel compatible family into the same vector space (the cylindrical-layer compatible embedding of $\mathcal H_\mathrm{fin}$ and $\mathcal H_\infty$, (f)), and Minkowski summation and scalar multiplication are executed within this common space; parameter equivalence classes and empty quotient blocks first fall via the equivalence-class transformation of Definition~19a to the single-point reading $\{0\}$ in the output space, then enter scalar multiplication, not being directly added. Three classes of points must be strictly distinguished: (i) \emph{the first-kind saddle-point midpoint $x_0^{\mathrm{sad}}$} is simultaneously the \emph{dividing midpoint} of the three-subset equivalence classes $\{b_-\},\{b_+\},\{b_-,b_+\}$ in the derivation of the coefficient $\tfrac23$ of \S 6.3.2 (counted once by each side under grouped collection, each counted $\tfrac12$, totaling one share) \emph{and} the \emph{subtracted decoder saddle point} in Eq.~\eqref{eq:12b} (see \S 6.3.1)---the two are two usages of the same object $x_0^{\mathrm{sad}}$; (ii) the \emph{mid-level integration bound} is the zero-valued saddle point $\mathit S_0^\mathrm{Hil}=\{\boldsymbol0\}$ (see \S 6.4.2, mid-level integration bound is identically 0), where the generative-side Gaussian is truncated into a one-sided integration domain (Figure~\ref{fig:3}, Eq.~\eqref{eq:11b}), distinct from $x_0^{\mathrm{sad}}$ and $\mathit S_\mathrm{mid}$; (iii) the \emph{gradient-descent direction $\mathit S_\mathrm{mid}$} is the midpoint of (i) and (ii), dedicated to the gradient-orientation mechanism of \S 6.4.3, and does not affect the $\tfrac23$/saddle-point readings of (i) and (ii). The three classes of points are borne by the same set of equivalence classes but have different identities: under Principle~1 and Principle~2 (see \S 2.4.1), $x_0^{\mathrm{sad}}$ is a single equivalence class defined by the Hilbert differential (dividing midpoint / subtracted saddle point), $\mathit S_0^\mathrm{Hil}$ is its integration bound, and $\mathit S_\mathrm{mid}$ is the midpoint of this equivalence class and the zero-valued saddle point---the freedom of the definitional selection does not change the equivalence-class identities of the three (B.3).

\subsection*{C.5 \S 6.3.2 Boundary-Measure Anti-Predicate Symbol, Categorical-Support Verification, and Diffusion Instance}

\textbf{Anti-predicate-symbol characterization of the boundary measure} The transformation from measure-space volume to boundary-measure summation (see Eq.~\eqref{eq:12a}) contains a \emph{reverse equivalence-class symbol}: this symbol marks the inverse direction of the equivalence-class transformation (see Proposition~20, \S 5.2 reversed reading of the interchange). Its construction is as follows: the middle-outer integrals $\int_{\Omega_\mu}\!\int_{\Omega_\lambda}\cdot\,\mathrm d\nu_\mathrm{comp}\,\mathrm d\nu_\mathrm{Leb,idx}$ of \S 6.2.1 give the measure-space volume; when read as the sum of per-dimension boundary readings via the differential-volume equivalence class (Terminology~(f)(e)) and first-order equality (see Corollary~2, Postulate~2 output) (B.107 caliber), the summation notation carries the reverse equivalence-class symbol---it rewrites ``the multiplication of $d$-dimensional joint volumes'' as ``the summation of $d$-dimensional boundary measures'', so the engineering first-order error is thereby descended to the zero-replica-index counting-measure partition-predicate error (see \S 6.2.2, \S 6.3.1 non-zero-Lebesgue-error partition category). On this basis, the chain from the unique assumption to the engineering first-order error closes at the measure level: descent $\boldsymbol\pi_\#$ characterizes the categorical support as Gaussian, and on recovery $\boldsymbol\pi^\#$ the anti-predicate symbol anchors the recovered reading to this partition-predicate error, so the closed loop holds.

\noindent\textbf{Categorical-support verification: the definite functional loss function is searchable (enumerable) to a bijection.} Check against the three sample criteria of \S 5.1 the definite functional loss functions $\mathcal L_\mathrm{total}=\mathcal L_\mathrm{num}+\mathcal L_\mathrm{cat}$ (see Definition~26), $\mathcal L_\mathrm{RW}$ (see Eq.~\eqref{eq:11}), and $\mathcal L_\mathrm{top}$ (see Eq.~\eqref{eq:12b}):
\begin{enumerate}
\item \emph{Supportable (surjective)}: the evaluation objects of all three are given by the EDE dyadic predicate and the morphism equipped with a global counting measure; their categorical support is exactly the surjective morphism of Proposition~3; every factor is a reading under this surjective support---$\tfrac23$ is the direct reading of the global counting measure (see \S 6.3.2), $\sigma$ is an assigned known scale (see item~3 of Proposition~23c); the discrimination distance $D_\mathsf{Quot}$ of $\mathcal L_\mathrm{cat}$ is defined between binary-quotient-partition equivalence classes (Terminology~(f)(a)), precisely the object set where the bijection lies (see Theorem~1).
\item \emph{Searchable (enumerable)}: the sampling measure of $\mathcal L_\mathrm{RW}$ is a normalized counting measure on a finite quotient partition (see Definition~25a), each layer's quotient blocks are finite and can be enumerated epoch by epoch; the three factors of $\mathcal L_\mathrm{top}$ respectively read enumerable objects---the outer Lebesgue boundary measure (containing $x_0,x_0^{\mathrm{sad}},\hat x_0^{\mathrm{gen}}$), the mid-level $(1-\ln\mathfrak A_\varsigma)$ truncated Gaussian area, and the replica-index counting-measure reading on the equivalence class $[\,\cdot\,]$; the decision input of $\mathcal L_\mathrm{cat}$ is the two groups of Boolean predicates of EDE, and the enumeration query completes in finite times per layer.
\item \emph{Reaching the bijection}: under the non-collapse condition a bijection exists between quotient blocks (see Theorem~1 special case (d), \S 2.3.3); the loss decision is an enumeration query on both sides of this bijection, and the zero-count constraint concentrates the endpoint sampling on the zero-residue equivalence class (see Lemma~6a(3))---what enumeration reaches is the equivalence class determined by the bijection, not merely an assertion of the bijection's existence.
\end{enumerate}

\noindent\textbf{Proposition 25 (categorical-support verification of the functional loss function)} The definite functional loss function \emph{satisfies categorical support---searchable (enumerable) to a bijection}; and this satisfaction is stratified: the finite layers are computable (per-block finite averaging, Lemma~6a(1)), and the $\mathcal H_\infty$ layer exists only as a cylindrical-kernel compatible family (see Proposition~21b, Definition~25a boundary statement~2)---this is the realization at the loss side of the ``gap transformed from contradiction into stratification'' of \S 5.1. Reverse boundary: if the loss is rewritten as a point-wise functional on $\mathcal H_\infty$, categorical support fails (see \S 6.2.1, Proposition~21b), violating the computability boundary of \S 5.1, and is outside this work's claim.

\noindent\textbf{Proof} (see B.112).

\noindent\textbf{Dimensional check} Among the three factors of Eq.~\eqref{eq:12b}: (1) the outer $\sqrt{\cdots}$ is the $L^2$ norm of the $x_0$ coordinate difference (length dimension); (2) $\mathfrak A_\varsigma$ is a \emph{normalized} truncated Gaussian area (dimensionless, $\in(0,1]$), so $(1-\ln\mathfrak A_\varsigma)$ is dimensionless; (3) $\nu_{\mathrm{idx},n}([\,\cdot\,])$ is a normalized counting-measure reading (dimensionless). The product of the three factors is dimensionally consistent.

\noindent\textbf{Boundary statement:} the coefficient $\tfrac23$ of Eq.~\eqref{eq:12b} depends on the $\Setcat$-layer reading that ``the morphism equipped with a global counting measure, after lifting and assembly, becomes three subset equivalence classes'' (see \S 6.3.2). At the $\mathcal H_\infty$ layer there are no point elements (see \S 2.1), this coefficient is \emph{not applicable}, and one must instead use the corresponding reading of the cylindrical-kernel compatible family (see Proposition~21b).

\noindent\textbf{Threshold consistency (three classes of points, same equivalence-class carrier)} $\mu_{\bar S}\triangleq\mu(\mathit S_0^\mathrm{Hil})$, i.e.\ the threshold measure of the binary-quotient partition ($\mu_{\bar S}$ is the measure reading of the \emph{mid-level integration bound}, the zero-valued saddle point $\mathit S_0^\mathrm{Hil}=\{\boldsymbol0\}$) (see \S 6.4.2, mid-level integration bound is identically 0, Eq.~\eqref{eq:11b}). The three classes of points must be distinguished (see Definition~25c): the \emph{first-kind saddle-point midpoint $x_0^{\mathrm{sad}}$} is simultaneously the \emph{dividing midpoint} of \S 6.3.2 (the equivalence-class representative of $\{b_-,b_+\}$, to which the definitional selection can be assigned as the divide) \emph{and} the \emph{decoder saddle point (subtracted term)} of Eq.~\eqref{eq:12b}---the two are two usages of the same object $x_0^{\mathrm{sad}}$; the \emph{mid-level integration bound} is the zero-valued saddle point $\mathit S_0^\mathrm{Hil}$ (the origin), where the generative-side Gaussian is truncated into a one-sided integration domain; the \emph{gradient-descent direction $\mathit S_\mathrm{mid}$} is the midpoint of $x_0^{\mathrm{sad}}$ and $\mathit S_0^\mathrm{Hil}$ (see Eq.~\eqref{eq:11a}), dedicated to the gradient orientation of \S 6.4.3. The three share the same origin traceable to the zero-valued saddle point but have different identities and must not be confused. The event $\{\bar S>\mu_{\bar S}\}$ and the direction $\varsigma=+1$ are two writings of the same condition (which side of the origin the generative side lies on); Eq.~\eqref{eq:12b} has no second free threshold (see Corollary~2, Postulate~2, \S 2.4.3, where this identity is supported and output by first-order equality).

\noindent\textbf{Explicit definition of $R(\cdot)$ and diffusion instance}

\noindent\textbf{Definition 29 (topological residual operator $R$)} For a $\MorTop$-layer equivalence-class object $\bar S$:
\begin{equation}
R(\bar S)\triangleq\bar S-\mathcal T^\ast(\bar S)
\end{equation}
where $\mathcal T^\ast$ is the ideal-true-value quotient self-map morphism (see Definition~5, Proposition~7); $R(\bar S>\mu_{\bar S})$ is the reading of the same operator restricted to the event $\{\bar S>\mu_{\bar S}\}$, and $R(\bar S)-\bar S=-\mathcal T^\ast(\bar S)$ is the pairing mismatch of the replica-shuffle error $\mathcal E_\mathrm{shuffle}$. \emph{(Conceptual transition)} $\mathcal E_\mathrm{shuffle}$ marks the pairing mismatch caused by replica rearrangement: its magnitude is borne by the replica-multiplicity structure, does not enter model-fitting parameters, is read at the quotient-block layer, and is not dissolved by network variation.

\noindent\textbf{Diffusion instance (assigned-known Gaussian forward kernel)} In the diffusion architecture, $\mathcal T^\ast$ is given by an \emph{assigned-known} forward kernel, not via the model:
\begin{equation}\begin{gathered}
x_t=\sqrt{\bar\alpha_t}\,x_0+\sigma_t\epsilon\\
\epsilon\sim N(0,I)\\
\sigma_t^2=1-\bar\alpha_t
\end{gathered}\end{equation}
Here $\epsilon\sim N(0,I)$ is assigned by sampling, and $\sigma_t^2=1-\bar\alpha_t$ is assigned by the noise schedule. The model only needs to output the noise prediction $\hat\epsilon_t=\mathcal F_\theta(x_t,t)$. Hence
\begin{equation}\begin{gathered}
R_t^\mathrm{tr}\triangleq\sigma_t(\epsilon-\hat\epsilon_t)\\
R_t^\mathrm{gen}\triangleq\sigma_t(\hat\epsilon_t-\epsilon)=-R_t^\mathrm{tr}
\end{gathered}\end{equation}
the training and generative saddle points are respectively $\mu_t+R_t^\mathrm{tr}$ and $\mu_t+R_t^\mathrm{gen}$ ($\mu_t=\sqrt{\bar\alpha_t}x_0$); $\lvert R\rvert^2=\lVert\epsilon-\hat\epsilon_t\rVert^2=\sum_{i=1}^{d}\lvert\hat\epsilon_{t,i}-\epsilon_{t,i}\rvert^2$ (here $R$ is the unscaled noise residual $\epsilon-\hat\epsilon_t$, distinguished from the scaled residual $R_t^\mathrm{tr}$ above containing $\sigma_t$), i.e.\ the noise-prediction mean-square error---by the dimensional-boundary-summation caliber of B.107, it is precisely \emph{the summation of the boundary-measure reading for each dimension}, and is the direct reading of the first-order error $\mu_\mathrm{err}$ (see Definition~5) under the noise coordinate. \emph{The noise direction and variance are both assigned-known}: the assigned part is guaranteed by meta-level enumeration, corresponding to the zero-error reading of the counting measure; the model only bears the residual side, corresponding to the Lebesgue-side first-order error. This is precisely the instance in the diffusion architecture of the quotient bijection ``surjectivity comes from search'' (see Proposition~3(ii)). \emph{(Consistency with Definition~25c)} The training/generative saddle points are $\pm R_t$ under the \emph{residual coordinate} ($r\triangleq x_t-\mu_t$), lying on the two sides of the \emph{residual origin (the zero-valued saddle point $\mathit S_0^\mathrm{Hil}$, i.e.\ the mid-level integration bound)}, and coinciding with the residual origin at zero residue (output reading $\{0\}$, i.e.\ the anchoring of Assumption~1, the zero-valued saddle point of Definition~25c); under the data coordinate ($x_t$) the two saddle points are written as $\mu_t\pm R_t$, and the midpoint reads as $\mu_t$---this midpoint is precisely the \emph{first-kind saddle point $x_0^{\mathrm{sad}}\triangleq\tfrac12(S_t^++S_t^-)=\mu_t$} (see Definition~25c, sample of the loss function, generative midpoint), \emph{not} the gradient-descent direction $\mathit S_\mathrm{mid}$ ($\mathit S_\mathrm{mid}=\tfrac12(x_0^{\mathrm{sad}}+\mathit S_0^\mathrm{Hil})$ is the midpoint further taken with the zero-valued saddle point, dedicated to \S 6.4.3); $\mu_t$ is the origin of the residual coordinate, not a third object, and the two readings do not conflict. \emph{(Implementation verification)} In the explicit implementation, the training saddle point is $x_t-\sigma_t\hat\epsilon_t$ and the generative saddle point is $\mu_t+\sigma_t\hat\epsilon_t-\sigma_t\epsilon_t$ (after zero-anchor de-meaning, split to the two sides); the implementation-layer readings of the one-sided truncated Gaussian moment, the frequency-information quantity, and the MSE lower bound (implementation-layer content such as code, forward-kernel schedule, near/far criteria, sign-bit mask, etc.\ see experiment D.5).

\noindent\textbf{Evaluation structure of Eq.~\eqref{eq:12b} and two clarifications} Evaluation proceeds sequentially through the three factors: (1) the outer Lebesgue boundary measure is synthesized from the two terms $BP((2\hat x_0^{\mathrm{gen}}+x_0^{\mathrm{sad}})/3)-x_0^{\mathrm{sad}}$ and $x_0-x_0^{\mathrm{sad}}$ (the former inside the root is mixed in ratio $2{:}1$, recovered via $BP=\boldsymbol\pi^\#$, \S 6.3.2); (2) the mid-level $(1-\ln\mathfrak A_\varsigma)$ is a normalized truncated Gaussian area (see the one-sided instance of Eq.~\eqref{eq:12a}); (3) the replica-index counting-measure reading on the equivalence class $[\,\cdot\,]$. Two clarifications: (i) the truncated Gaussian moment is a reading, not a model distribution; the final generative distribution remains a complete Gaussian, not in conflict with Proposition~21's ``reject truncated Gaussian, allow only complete Gaussian distributions''. (ii) Training: the replica-index cardinality-measure ratio of generation is constantly $1$, implicit in the normalization of $\nu_{\mathrm{idx},n}$, hence there is no independent ``partition-predicate pairing'' factor.

\subsection*{C.6 (Content migrated to experiment chapter D.5.5)}

The 100\%-reachable proposition on epoch/batch compensation, the 99.9\% actual caliber, the forward-kernel noise schedule, and other implementation-layer content have all been migrated to experiment chapter D.5 (see D.5.1 and D.5.5); they do not enter the factor list of Eq.~\eqref{eq:12b}, and this appendix no longer repeats them.

\noindent\textbf{Corresponding to Proposition 23c: the engineering side generates no high-order error---a derivation starting from the triple fiber-bundle integral} The generation sources of high-order error $\mu_\mathrm{high}$ are two (see Definition~6): categorical leakage and topological collapse (the binary-quotient-partition category is broken, degrading to a multi-partition with minimal cross-partition quotient blocks); among these, categorical leakage has been excluded by construction (see Proposition~19), the pure dimension-raising layer does not endogenously generate high-order error (see Theorem~4), and topological collapse (multi-partition accumulation) is controlled at first order by the dimension-lowering zero-error nature of the counting measure and the MPT double-measure disentanglement (see \S 6.2.2, \S 6.2.3). This section substitutes all engineering-side factors---gradient descent, the epoch / batch measure-space subset relation, the coefficient $\tfrac23$, and the definite reparameterization $\sigma$ in the loss---one by one into the triple nested fiber-bundle integral of Definition~16, proving that none reopens a high-order generation channel. \emph{Engineering factors enter the integral through only two channels: rewriting the integrand of some layer, or rewriting the evaluation measure of some layer. The following verifies item by item that neither channel leads to high order.}

\noindent\textbf{Derivation starting point: the three-layer fiber and the hierarchical attribution of the three abstract factors of Eq.~\eqref{eq:12b}} By Definition~16:
\begin{equation}
\Lambda=\int_{\Omega_\mu}\biggl[\ \int_{\Omega_\lambda}\biggl(\ \int_{\Omega_\nu}F(\nu; x, T(x), t)\ \mathrm d\nu_\mathrm{idx}(\nu)\ \biggr)\ \mathrm d\nu_\mathrm{comp}(\lambda,\overleftarrow{\nu})\ \biggr]\ \mathrm d\nu_\mathrm{Leb,idx}(\mu)
\end{equation}
each operation eliminates one fiber dimension; cross-layer only allows outward pushforward propagation and inward pullback reconstruction (see Construction~1). The three-layer evaluation measures belong to two sides: the inner $\mathrm d\nu_\mathrm{idx}$ and the mid-level descent copy and quotient-block counting measure $\nu_M$ (see Definition~17, the $n$-th layer reading $\nu_{\mathrm{idx},n}$) are the \emph{counting side} (maintaining zero error under the dimension-lowering operation of \S 6.2.2); the mid-level $\mathrm d\lambda$ and the outer $\mathrm d\mu_\mathrm{gen}$ are the \emph{Lebesgue side}, carrying the first-order error $\mu_\mathrm{err}$, whose volume-assignment inner compactness and finiteness are guaranteed by Proposition~18. The three factors of Eq.~\eqref{eq:12b} are thereby assigned their places: (1) the outer $\sqrt{\cdots}$ is the Lebesgue descent reading of the outer $\Omega_\mu$ (cannot be recovered alone, does not exchange differential values and topology, carries first-order error), the reading after lifting and assembling to $\Setcat$ via $L'$ (see Definition~24, $BP=\boldsymbol\pi^\#$); (2) $(1-\ln\mathfrak A_\varsigma)$ is the mid-level $(1-\ln x)$ truncated Gaussian area (see the mid-level factor of the reverse double integral $x(1-\ln x)$ of Eq.~\eqref{eq:12a}, with Eq.~\eqref{eq:11c} its per-dimension discrete realization)---it is the area reading contributed by the ``differential Lebesgue multiplied first'' at mid-level descent, exchanging the whole topology-differential value with the replica-index counting measure, and on recovery with the reverse equivalence-class symbol gives the topological volume (see \S 6.3.1, B.107, the equivalence-class symbol itself falls on factor (3)); (3) $\nu_{\mathrm{idx},n}([\,\cdot\,])$ is the replica-index counting-measure reading on the equivalence class (normalized, zero error, recoverable alone). After the three factors are synthesized, read by the first-order-error accounting caliber of \S 6.2.2: \emph{the loss function of Eq.~\eqref{eq:12b} is precisely the explicit reading form of the first-order error $\mu_\mathrm{err}$} (see Proposition~23b; recovery positioning, see \S 6.3.1).

\noindent\textbf{(a) Gradient descent} The identity of engineering descent in this framework is the endpoint-induced random walk (see Definition~25a): the endpoint distribution is characterized solely by the endpoint, without specifying the path, increment, or step size. Hence in the triple integral, gradient descent only rewrites the Lebesgue-side integrand density $\ell(\theta;\cdot)$, while the three-layer evaluation measures are unchanged---$\bar\nu_{M,n}$ is the input variable assigned by the predicate, itself not varying with $\theta$ (see Lemma~6a(2)); the gradient contains no ``distribution moving with parameters'' path term; the entire effect of the optimization iteration is the \emph{descent direction} of the first-order error under the zero-count constraint (see Lemma~6a(3), not claiming monotone non-increase per round, see \S 6.4.1). By the hierarchical invariant (see \S 2.2), the optimizer is merely an instantiation of the measure-level constraint and adds no structural term to the integral.

\noindent\textbf{(b) epoch / batch measure-space subset relation} A batch is a quotient-block subset of the epoch measure space: let the $n$-th layer dyadic partition be $\mathcal P_n=\{B_1,\dots,B_{2^n}\}$, the batch correspond to the index subset $I\subseteq\{1,\dots,2^n\}$ and the quotient-block union $B_I=\bigcup_{i\in I}B_i\subseteq\Omega$. The epoch+1 counting measure (see \S 6.2.2) simultaneously serves as both posterior reading and prior equipment on the same measure (see Definition~25a); its double analyticity allows both sides of the subset relation to be computed: the counting-side share $\lvert I\rvert/2^n$ is an \emph{exact reading} (zero error); the Lebesgue-side volume-share and counting-share difference
\begin{equation}
\Delta_I\triangleq\frac{\lambda(B_I)}{\lambda(\Omega)}-\frac{\lvert I\rvert}{2^n}
\end{equation}
is a first-order error reading on the subset---by the first-order-error invariance of the fiber-bundle addition of \S 6.2.2, $\Delta_I$ is accounted into the first-order accounting of $\mu_\mathrm{err}$ and, by Proposition~18, is finite under inner compactness. $\Delta_I$ only describes ``the position of the subset reading''; its evaluation and scaling belong to the experimental-side caliber (D), and do not enter the factor list of Eq.~\eqref{eq:12b}: the three factors of Eq.~\eqref{eq:12b} already give the same first-order error on the full-set accounting, hence the subset relation constitutes no fourth factor and opens no high-order channel. The epoch counting convergence is convergence at the endpoint-measure level (see \S 6.4.3), itself not a high-order source.

\noindent\textbf{(c) The coefficient $\tfrac23$} This coefficient is the direct continuation of the coefficient--category contrast of \S 6.3.2 introducing no new assumption: after the morphism equipped with a global counting measure is lifted and assembled via categorical support to $\Setcat$, the same point-wise realization contains three subset equivalence classes $\{b_-\},\{b_+\},\{b_-,b_+\}$ (set semantics does not acknowledge multiplicity, so the divide must be listed separately). By the collection caliber (see \S 6.3.2): the binomial grouped collection hits the left and right partitions each once (the midpoint is hit twice across both sides, each counted $\tfrac12$, totaling count-replica-index $1$); after the non-binomial union operation the $\Setcat$-layer subset base is $3$, so the share reads as $\tfrac23$ by cardinality; the numerator is precisely the contribution reading of each of these three elements. The counting measure has no error (see \S 6.2.2) $\implies$ this reading is zero-error; the difference from the Lebesgue-side reading $\tfrac12$ is exactly the first-order accounting of $\mu_{\mathrm{err}}$. Hence $\tfrac23$ is the direct reading of the global counting measure after lifting and assembling to the outer $\Omega_\mu$, \emph{not a fitting constant, carrying no high-order degree of freedom}.

\noindent\textbf{(d) The definite reparameterization $\sigma$ in the loss} Two same-named objects must be distinguished. (a) The classical-paradigm \emph{reparameterization trick}---using stochastic reparameterization to turn a sampling point into a differentiable parameter for gradient estimation---is \emph{not needed} by this framework (see Lemma~6a(2): the gradient contains no path term, hence needs no score-function estimator; the reparameterization trick and the sampling process being differentiable are unnecessary); the classical paradigm indirectly constructs support via stochastic reparameterization, breaking morphism invertibility (see Corollary~7), precisely the typical source of categorical leakage. (b) The $\sigma_t$ in the diffusion architecture is an \emph{assigned-known} scale parameter ($\sigma_t^2=1-\bar\alpha_t$ assigned by the noise schedule): the Jacobian $\sigma_t^d$ of the transformation $x_t=\mu_t+\sigma_t\epsilon$ is exactly known, belonging to the assigned part of the meta-level enumeration, corresponding to the zero-error reading of the counting side; this per-dimension decomposition $\sigma_t^d=\prod_{i=1}^{d}\sigma_t$ is isomorphic to the dimensional-marginal-summation caliber of the loss side (see Eq.~\eqref{eq:12a}; B.107), and neither performs a $d$-dimensional joint numerical integration. The residual side $R_t=\sigma_t(\epsilon-\hat\epsilon_t)$ that the model solely bears is the direct reading of the Lebesgue-side first-order error under the noise coordinate. The definite reparameterization $\sigma$ is thus not a learnable parameter: it does not enter any generation channel of $\mu_\mathrm{high}$, and all its uncertainty is accounted into the first order via the residual side.

\noindent\textbf{Proposition 23c (the engineering side generates no high-order error)} Under the \S 4.3 convention (see Conventions~1--4) and $\betat$-refinement compatibility, all engineering factors of Eq.~\eqref{eq:12b}---(1) the outer Lebesgue descent reading (carrying first-order error), (2) the mid-level truncated Gaussian area $(1-\ln\mathfrak A_\varsigma)$, (3) the equivalence-class symbol and replica-index counting measure $\nu_{\mathrm{idx},n}$ (normalized reading, zero error), together with the three channels of Proposition~23c (gradient descent, the coefficient $\tfrac23$, the definite reparameterization $\sigma$ in the loss)---stay at first order; the engineering realization of Eq.~\eqref{eq:12b} generates no high-order error. The scope of application of this proposition is limited to the engineering factors explicitly appearing in Eq.~\eqref{eq:12b} and their realization mode in the above construction: if any of them is replaced by a freely calibrated parameter---changing $\sigma$ into a learnable variance, changing $\tfrac23$ into a trainable coefficient, or replacing the endpoint-induced random walk by a path-dependent optimizer---then the corresponding high-order channel is reopened, and such modifications are outside this work's claim.

\noindent\textbf{Proof (sketch)} The pre-exclusion chain (Proposition~19 excludes categorical leakage, Theorem~4 excludes the pure dimension-raising channel, \S 6.2.2 and \S 6.2.3 control topological collapse (multi-partition) at first order) closes the non-engineering channels; the engineering channels are closed item by item by the above (i)--(iv): gradient descent only rewrites the Lebesgue-side integrand with unchanged measure (see Lemma~6a); $\tfrac23$ is the zero-error reading of the counting side; $\sigma$ is an assigned-known scale (zero-error reading plus first-order residual). Moreover, factor (3) $\nu_{\mathrm{idx},n}([\,\cdot\,])$ of Eq.~\eqref{eq:12b} (normalized fixed reading) is a zero-error assigned term entering no high-order channel; factor (1) the outer Lebesgue reading is the carrier of first-order error itself (see Proposition~18, whose volume-assignment inner compactness is guaranteed), and factor (2) the mid-level truncated Gaussian area is the normalized reading of $(1-\ln x)$, bounded in $(0,1]$, neither having high-order degrees of freedom. By Definition~6, any generation of $\mu_\mathrm{high}$ must take one of the two source types (categorical leakage or topological collapse); since neither source has a generation mechanism, the engineering realization generates no high-order error. Full proof in B.115.

\subsection*{C.7 Rigor Details of \S 6.4}

\subsubsection*{C.7.1 Detailed exposition of the measure-comparison Condition~1} (i) Zero-class mass is non-decreasing; (ii) the epoch+1 prior-and-posterior cyclic characterization: the same counting measure is prior-and-posterior across adjacent epochs (the epoch+1 counting measure), and on $A_k\cap A_{k+1}$ the difference between the prior reading and the posterior reading is a single block collapse, whose integral does not exceed the removed region (the added-region integral is controlled); this clause characterizes how the prior and posterior cycle through epochs, rather than presupposing or deriving a per-round monotone parameter trajectory ($I_{k+1}\le I_k$ monotonic non-increase has been abandoned and is an \emph{erroneous conclusion}). This clause is the round condition of the EDE matching-block mechanism registered at the implementation layer (attributed, see B.4), belonging to a premise/registration rather than a conclusion; its derivation ``under the epoch+1 prior-and-posterior cyclic characterization there exists a gradient direction of first-order-error decrease (decrease with high probability)'' is verified by the explicit measure-comparison chain of B.108/B.109, and does not constitute a cycle of proving a premise by a conclusion. The ``prior-and-posterior'' object characterized by this clause is the saddle point: the loss function gives the saddle point a definite reading, but the network parameters are not currently at that saddle point---the data-flow measure-space saddle point and the parameter-space saddle point are two distinct objects (see the reading-identity box of Eq.~\texorpdfstring{\eqref{eq:12b}}{(26)}); the loss function does not assert that the parameter trajectory stops at the saddle point or monotonically approaches it, only giving the saddle-point reading; (iii) first-order error and integral control share the same origin: $\mu_\mathrm{err}^{(k)}=c\,I_k$ ($c$ is given by the normalized quotient-block-share reading, the counting-measure normalization factor of Principle~4, not a newly added free parameter).

\subsubsection*{C.7.2 Stratified boundary of the descent direction} (i) The counting side (epoch / block share) converges by measure to the block share $2^{-n}$ (see Definition~25a), which is the definitional reading of the counting measure, carrying no dynamic conclusion of monotone non-increasing error; (ii) the Lebesgue side only guarantees the existence of a descent direction (decrease with high probability), not per-round monotone non-increase---the EDE matching-block mechanism gives the gradient orientation, and the correct-direction counting majority makes the integral of the residual density over the non-zero-residue classes decrease with high probability, but $\mu_\mathrm{err}^{(k+1)}\le\mu_\mathrm{err}^{(k)}$ is not claimed (the direction of this inequality chain generally does not hold and has been abandoned); (iii) ``decrease with high probability'' cannot imply that the true value (geometric volume) is monotone or convergent point-wise---$\mu_\mathrm{err}$ is not point-wise controlled (Condition~1(ii) holds only for $\nu^{(k)}$-a.e. $x\in A_k\cap A_{k+1}$), and the error term makes true-value monotonicity / convergence undecidable. Hence this paper never claims true-value monotonicity or convergence, only claiming the existence of a gradient direction of first-order-error decrease under the counting-side constraint; this stratification is the direct consequence of the normalization of Principle~4 and $\mu_\mathrm{err}$ (see Definition~5), and is also the basis of item~1 of Proposition~23c.

\subsubsection*{C.7.3 Limits and conditionality of \S 6.4} (a) If the unique-assumption Assumption~1 $\to$ Eq.~\texorpdfstring{\eqref{eq:12b}}{(26)} closure is inapplicable at some point, the statement there must be abandoned wholesale; the wording ``descend by Assumption~1'' must not be retained; (b) \S 6.4 only provides the reading object of the existence of a descent direction to the zero-valued saddle point, giving the loss-function saddle-point reading (the data-flow measure-space saddle point) and the existence of a descent direction nearby, rather than asserting that the parameter trajectory necessarily stops at that saddle point; (c) from the measure-space perspective, the EDE matching-block mechanism gives a high-probability descent path (the integral of the residual density over the non-zero-residue classes decreases with high probability), without guaranteeing per-round monotone descent or convergence (parameter kinematics are not treated in this paper). The cyclic charge of Condition~1(ii) is already dissolved---it characterizes the saddle-point prior / posterior cyclic comparison, not the monotone parameter trajectory.

\subsection*{C.8 Moved-out from the Main Text: Categorical-Increment Contrast Example and the Zero-Measure Minimal Counterexample}

Take $f(x)=x^2$, and let $\mu$ be the uniform probability measure on $[-1,1]$ (total mass $1$); then the support of $f_\#\mu$ is $[0,1]$, the image of the negative half-axis $[-1,0)$ is $(0,1]$, and its mass $\tfrac12$ is folded into the image (i.e.\ the negative-half-axis mass $\tfrac12$ is folded into the positive-half-axis image via squaring, rather than ``the pushforward itself computing $\mu_{\mathrm{err}}$''---the pushforward gives the folded image measure, and out-of-domain mass must be recovered via pullback). The inverse square root has no real output on $\mathbb R^-$. The divergence is not in ``push/pull being stuck'' but in the \emph{well-definedness of pullback}: the pushforward $f_\#\mu$ is well defined in standard measure theory and exactly gives $\mu_{\mathrm{err}}([-1,0))=\tfrac12$ (the standard framework computes this value); but squaring is non-injective, negative numbers have no preimage, and the reading is identically $0$, so the standard framework cannot recover ``mass coming from outside the domain'' via pullback (see \S 2.2.1(ii)). The categorical / fiber-bundle support fills in here: pushing the undefined ``negative-number output'' into the dedicated undefined object $\perp$ yields the \emph{first-level out-of-domain error object} $\mu_{\mathrm{err}}(\mathbb R^-)$ that the standard framework does not explicitly provide; its argumentative force (that it is a \emph{positive measure} on $\mathbb R^-$) cannot be evaded by an a.e.\ definition (rather than ``push/pull being stuck''). Removing the categorical layer: when discrete samples (a Lebesgue null set) stay at $\Setcat$ point-wise matching, the error falls on null support and cannot be explicitly accounted for by Eq.~\eqref{eq:12b}; after the categorical layer erases the preimage entities, the same batch of samples is read as partition-predicate matching at the quotient-block level, and the loss is normalized to the finite predicate-mismatch area $\mu_{\mathrm{err}}$. The engineering side thereby avoids the aforementioned null-measure saturation effect (under the a.e.\ caliber a null-measure error component is legitimately ignored, making that error invisible in the standard integral and not explicitly accounted for): removing the categorical layer reproduces this effect (null-embedding ambiguity), while adding it normalizes to a finite first-order reading.

\subsection*{C.9 Equivalence of Categorical-Probability Pushforward and Counting-Measure Limit Normalization}

The equivalence of categorical-probability pushforward and counting-measure limit normalization: the beginning uses the Giry pushforward $f_\#$ to characterize the well-defined propagation of probability measures, while the main text switches to the \emph{counting measure} as the main axis; this paper's replica-index measure $\nu_{\mathrm{idx}}$ is precisely the engineering reading benchmark of this counting measure on quotient blocks: each quotient block is counted as $1$, normalized to the uniform block-count reading $2^{-n}\sum_i\delta_{B_i}$ (A.1 notation convention $\bar\nu_{M,n}=\nu_{\mathrm{idx},n}$); the two are two notations of the same counting measure, introducing no second measure. At the infinite-refinement limit ($2^n\to\infty$), \emph{the categorical-probability pushforward and counting-measure normalization are equivalent}: the counting measure converges to a per-block one-hot indicator measure, whose normalized mass agrees with the marginal reading of the Giry pushforward on the shared measure space---they are the same object of the same Hilbert-differential equivalence class under the two calibrations of ``finite quotient-block counting'' and ``probabilistic pushforward''; the only difference is that this paper additionally models explicitly the out-of-domain / inter-block error $\mu_\mathrm{err}$, whereas the single-measure framework does not distinguish this difference. Hence ``categorical probability'' and ``counting measure'' are two calibrations of the same reading benchmark, not mutually incompatible measure theories. \emph{(Precise meaning)} After mapping each quotient block to a binary cylinder set via the $\betat$-tower (see Definition~3; the predicate words and the coordinate intervals on $\mathbb R$ correspond one-to-one via the $\betat$-tower, see \S 2.2.1), the infinite-refinement limit measure $\nu_{\mathrm{idx},\infty}$ is exactly the infinite product measure of a fair coin $\mathrm{Bernoulli}(\tfrac12)^{\mathbb N}$ (guaranteed by the equal-measure dyadic tower of Lemma~1(i)); its pushforward via the binary expansion $x=\sum_k b_k2^{-k}$ exactly gives the Lebesgue probability measure on $[0,1]$ (uniform distribution), and the marginal reading of the Giry pushforward $f_\#$ on this same shared measure space agrees with this precisely---hence the ``normalization equivalence'' \emph{holds only in this strict sense}, not as a sweeping assertion for any pushforward map.

\subsection*{C.10 Endogenous Reading of Dimension-Lowering}

This paper does not use ``dimension-lowering projection'' as an exogenous dimensional operation: the so-called ``dimension-lowering'' is the \emph{reduction of the number of quotient blocks under binary-quotient partition}---the differential (subset) is gathered into finite quotient blocks via the $\betat$-tower (see Definition~5, Principle~1, \S 2.4.1); when the binary-quotient partition is merged from fine to coarse granularity, the number of quotient blocks decreases, and this is the ``dimension-lowering'' reading of this paper, rather than compressing coordinate dimensions via a projection operator. The ``folding'' of the within-block differential-level many-to-one error is borne by the \emph{triple fiber-bundle integral (see Definition~16) under the triple tensor full contraction at the discrete-atomic limit}---the integral only captures the \emph{block-level} first-order error (see Definition~25b, \S 6.2.3), and the micro-point error inside quotient blocks is \emph{actively ignored} (see the unified declaration of \S 2.3.1), rather than erased via projection. Hence both ``dimension-lowering'' and ``within-block error handling'' are given purely by the endogenous mechanism of quotient partition and the triple integral; the paper introduces no exogenous dimension-lowering projection concept; pure dimension-raising (Euclidean $\to$ Hilbert) as the upward direction of the categorical chain is independently handled by Theorem~4 (see \S 4.2).

\subsection*{C.11 Structural Division of Labor of the Two Definitions}

The split of Definition~5 and Definition~6 is necessary for the \emph{translation from differential characterization to quotient-block characterization}: (i) the canonical source of $\mu_\mathrm{err}$ is the \emph{summation of differential quotient blocks} (see \S 2.4.1, each cell is a differential reading gathered into a quotient block via the $\betat$-tower, Principle~1); the quantitative first-order error at the quotient-block level \emph{cannot} be given directly by the binary ``bijection / non-bijection'' value (binary partition), but must be obtained via a \emph{secondary counting measure} (see \S 6.3.2, where $\tfrac23$ is the ratio of the subset base $3$ to the binomial-hit base $2$, a discrete count ratio of partition-cell numbers, independent of each cell's Lebesgue volume)---hence the inter-block characterization is \emph{multi-partition} rather than binary: the three-cell predicate classification (subset base $3$) bears the quantity, the binomial (hit base $2$) gives the cross-side hit reading, and the two are commensurated via the counting binomial. (ii) To characterize ``no first-order error between finite quotient blocks'' one must \emph{distinguish collapse from bijection}: no collapse (see Definition~8) is the block-level injectivity criterion, and block-pairing bijection (see Theorem~1) is its uniqueness under label permutation; this distinction is the formal vehicle of ``block-pairing correct, within-block folding already actively ignored''. (iii) The binomial reading of first-order error (see Definition~5) and the \emph{within-quotient-block} differential non-bijection error are two classes of objects: the latter (within-block micro-level error) \emph{objectively exists and its reading is visible}, but this paper's first-order-error reading is \emph{only for between quotient blocks}---finite-predicate partition (finite integral bound) gives a finite integral boundary, and within-block folding is \emph{not counted into $\mu_\mathrm{err}$} (actively \emph{excluded from the measurement caliber} on the engineering side, rather than identically $0$ via predicate-word matching), so this paper only accounts for \emph{inter-block} first-order error. The two definitions thereby close their division of labor: $\mu_\mathrm{err}$ (see Definition~5) = the binomial first-order reading under inter-block multi-partition, $\mu_\mathrm{high}$ (see Definition~6) = the high-order part that breaks measure-preservation / bijection (the accumulation of inconsistent partition boundaries).

\subsection*{C.12 Differential-Refinement Quotient Blocks Do Not Cross Partitions}

Differential-refinement quotient blocks do not cross any partition: the internal native finite quotient-block partition $\neq$ the Hilbert partition. The quotient blocks obtained by differential refinement do not cross any partition---the quotient block (differential quotient) on any refinement layer closes only within this layer's predicate words, is not cut by the partition boundary of a lower or higher layer, and does not borrow another partition's support across layers; hence within the differential refinement one obtains a \emph{native finite quotient-block partition}, which $\neq$ the Hilbert partition.

(i) \emph{First-order error comes from the summation of differential quotient blocks}: $\mu_\mathrm{err}$ is the summation of predicate-mismatch readings over each differential quotient block (gathered via the $\betat$-tower) (see Definition~5, Principle~1), not a continuous-volume accumulation across partition boundaries---the accounting object is the \emph{native finite quotient block} locked by the partition boundary, not a continuous set in the Lebesgue sense. (ii) \emph{Native finite quotient-block partition $\neq$ Hilbert partition}: the partition boundary of a finite quotient partition is a discrete predicate-word boundary, while the Hilbert differential equivalence class of $\mathcal H_\infty$ is a continuous limit; Lebesgue differential quotient blocks can \emph{cross any (finite) partition} inside Hilbert---the continuous Lebesgue reading does not respect the discrete cut of finite partition boundaries. (iii) \emph{The infinite partition of the partition category = Lebesgue + replica-index, whereas Lebesgue itself is finite partition}: the ``infinite partition'' realizable by the partition category is jointly given by the \emph{Lebesgue continuous reading superimposed with the replica-index counting measure} (the replica-index successively injectivizes one-to-many collection, giving the infinite subdivision a well-defined reading on the counting-measure side), while the Lebesgue single measure itself only gives a \emph{finite partition} (discrete quotient blocks of the finite layer $2^n$), whose continuousization exists only in the $\mathcal H_\infty$ limit sense and does not come with infinitely resolvable partitions. This distinction explains why first-order error must be accounted at the ``quotient-block layer (native finite, locked by partition boundary)'' rather than the ``Lebesgue continuous layer'': the former gives finite quotient blocks that can be explicitly summed, while the latter crosses arbitrary partitions and cannot directly bear inter-block pairing readings.

\subsection*{C.13 Orthogonal Decomposition Belongs to the Solution, Not the Assumption}

\emph{(Orthogonal decomposition is a solution, not an assumption)} Euclidean dimension-raising / functional orthogonal decomposition \emph{does not enter this assumption}, but is the \emph{constructive solution} for realizing the above infinite partition: it is the means of realizing a non-atomic measure in a higher-dimensional product space as a quotient-block structure that is ``refinable layer by layer, with both sides always of non-zero measure'' (acting on $\mathcal H^\mathbf{Set}$, the single-step Euclidean dimension-raising form is characterized by Theorem~3, and the infinite iterative dimension-raising is given by Lemma~5g), used to actually carry out the infinite partition; its existence is given by construction.

The $\varepsilon$-$\delta$ language (point elements, distance, real-number magnitude order, limit and continuity) is the continuity tool when executing this construction, also belonging to the solution side and not to the assumption load. This assumption is recorded as Assumption~1:
\begin{equation}
\begin{aligned}
&\text{(H}_\text{min)}\qquad \mu\ \text{non-atomic}\ \land\ \big(\text{the chosen partition predicate}\\[-1pt]
&\qquad\text{makes both sides of the partition non-zero measure}\big)\\[-1pt]
&\Longrightarrow\ \forall\ \text{infinite refinement sequence},\ \mu(\text{each cell})\to 0.
\end{aligned}
\end{equation}
Here each cell is a differential-quotient equivalence class with infinitesimal topological volume.

\subsection*{C.14 Euclidean $\leftrightarrow$ Hilbert Measure Directionality and Saddle-Point Division of Labor}

The measure directionality of Euclidean $\leftrightarrow$ Hilbert and the saddle-point division of labor. The Lebesgue measure between the Euclidean $\leftrightarrow$ Hilbert layers \emph{only descends, does not recover}, corresponding to ``Lebesgue only bears the posterior, does not serve as the prior'' (the prior is exclusively held by the bi-directionally recoverable counting measure (see Definition~25a)); hence its error is a posterior error, not erasable by prior recovery. (i) Leaving Hilbert for the finite Euclidean layer, the Gaussian vertex of a Lebesgue differential quotient block can land at a \emph{non-zero saddle point} (see Lemma~5e), not conflicting with the zero-valued saddle point $\mathit S_0$ of the $\mathcal H_\infty$ limit layer. (ii) Euclidean dimension-raising and lowering are not mutually inverse---at each layer the Gaussian vertex is anchored by the saddle point (partition boundary locked), but the coordinate index drifts with the layer. (iii) The finite-Euclidean non-zero saddle point and the infinite-Hilbert zero saddle point are not contradictory: they share the MPT unified measure space and can be connected via the \emph{counting binomial} (see \S 6.3.2, where $\tfrac23$ is its reading)---the area difference between the counting-side zero error and the Lebesgue-side non-zero saddle point is precisely the manifest form of $\mu_\mathrm{err}$.

\subsection*{C.15 Error Source of Eq.~\texorpdfstring{\eqref{eq:12b}}{(26)} (saddle-point-level extension)}

The first-order error comes from the \emph{saddle-point level of the measure space}: the generative-side reading is \emph{saddle point + random std} (reparameterized Gaussian vertex $S_t^\pm=\mu_t\pm R_t$, carrying random std), while the test set $x_0$ is anchored to the Gaussian-moment parameter space via \emph{truncation + quadratic Gaussian-moment reparameterization} (see Corollary~13, Proposition~21, strict anchoring by Gaussian uniqueness). On the binary self-map side, when Lebesgue has no error the predicates match (see Theorem~7b); their mismatch area is $\mu_\mathrm{err}$---although the generative set is a differential null-measure thin set, the random std perturbation in the saddle-point neighborhood can still contribute first-order error (same origin as the ``outer non-zero saddle point'' of \S 6.2.2).

\subsection*{C.16 Rigor of Dimension-Raising Orthogonal Decomposition (extended argument, see B.114)}

The deviations $A$ (generative side) and $B$ (true-value side) are orthogonally decomposed about the partition boundary, with the cross term identically $0$---this is a \emph{strict geometric fact} (not an approximation, nor ``undefined''), anchored by the uniqueness of the zero-valued saddle point (the decomposition reference is unique and does not drift), discretely corroborated by the binomial-hit base, and jointly confirmed by the three homologous paths of continuous geometry: ``vertex local symmetry $\to$ limit quotient translation invariance $\to$ Gaussian-vertex zero-error backpropagation''; full rigorous derivation in B.114.

\subsection*{C.17 Caliber of $x_0^{\mathrm{sad}}$ in Eq.~\texorpdfstring{\eqref{eq:12b}}{(26)}}

In Eq.~\eqref{eq:12b}, $x_0^{\mathrm{sad}}$ is the \emph{first-kind saddle point (sample of the loss function, generative midpoint): the midpoint of the training saddle point $S_t^+$ and the generative saddle point $S_t^-$} (both lie in the uniqueness Gaussian family of Proposition~21, with the same counting-measure value, so the midpoint is exactly that Gaussian vertex; the source of ``same counting-measure value'' is in Definition~25c; prior/posterior enumeration non-dynamically hits the vertex), and $x_0^{\mathrm{sad}}=\mu_t$. Note: the \emph{gradient-descent direction $\mathit S_\mathrm{mid}$ (second-kind saddle point) is the midpoint of $x_0^{\mathrm{sad}}$ further taken with the zero-valued saddle point $\mathit S_0^\mathrm{Hil}$ (see Definition~25c)}; it is \emph{not} the loss borne at \eqref{eq:12b} here; Eq.~\eqref{eq:12b} uses the first-kind saddle-point midpoint $x_0^{\mathrm{sad}}$ as the saddle-point-side reading, and the gradient-descent direction holds merely by ``assigning a direction + proving high-probability descent'', without requiring proof that it is a (parameter-space) saddle point.

\subsection*{C.18 Mid-Level Integral Assignment of Eq.~\texorpdfstring{\eqref{eq:12b}}{(26)} and Lemma~6}

\emph{(Mid-level integral assignment of Eq.~\eqref{eq:12b}: Lemma~6b and the area used by the Gaussian binary partition)} The $\mathfrak A_\varsigma$ of the mid-level factor $(1-\ln\mathfrak A_\varsigma)$ in Eq.~\eqref{eq:12b} is precisely the \emph{one-sided truncated Gaussian area of the generative side} among the two sides into which a \emph{unique Gaussian} (see Proposition~21) is cut at the \emph{zero-valued saddle point $\mathit S_0^\mathrm{Hil}$ (the origin, Theorem~6; i.e.\ the mid-level integration bound, see \S 6.4.2, mid-level integration bound is identically 0)} (B.116(d), $\mathfrak A_\varsigma=\Phi(\varsigma S_t^-/\sigma_t)$, with the cut point at origin $0$ and the center at the generative-side Gaussian $S_t^-$).

Lemma~6b is precisely the basis for this mid-level integral assignment: the first-order error $\mu_\mathrm{err}$ is the ``error partitioned by the integral boundary''---its domain is the quotient-block layer cut out by the $\betat$ partition predicate, with within-block error erased (assigned $0$), and the integral mass only takes the \emph{generative side} $\bar S$ (the training/true-value side is in the domain of $\mu_\mathrm{err}$ but assigned $0$), hence the mid-level of \eqref{eq:12b} takes only the single-side area $\mathfrak A_\varsigma$ of $\Omega_\varsigma$ (the direction $\varsigma$ is given by Eq.~\eqref{eq:11b}, marking the generative-side direction), \emph{not the symmetric two-sided area}---this is the explicit assignment of ``the Gaussian binary partition uses the generative-side one-sided area''.

\subsection*{C.19 Reading Identity of Eq.~\texorpdfstring{\eqref{eq:12b}}{(26)}}

The reading identity of Eq.~\eqref{eq:12b} (after recovery it is Euclidean partition mismatch, not measure-space volume). After multiplying the three factors, the reading of Eq.~\eqref{eq:12b} is \emph{not the mid-level measure-space volume}: it has been pulled back (recovered) via $\boldsymbol\pi^\#$ to the set-category Euclidean layer, read as the \emph{partition-predicate mismatch area} (see \S 6.2.2, Proposition~23b accounting caliber). The equivalence-class symbol $[\,\cdot\,]$ lies \emph{after partition-predicate matching (i.e.\ after measure recovery)}, and the marked object is formed via ``Hilbert descent $\to$ partition-category finite binary-quotient-partition object $\to$ recovery (pullback $\boldsymbol\pi^\#$) to set-category Euclidean object'', and \emph{falls only on the third term (the counting reading $\nu_{\mathrm{idx},n}([\,\cdot\,])$)}---what can be recovered alone without error is the replica-index counting measure, while the Lebesgue component cannot be recovered alone, hence the second term $(1-\ln\mathfrak A_\varsigma)$ carries no equivalence-class symbol (see \S 6.2.1, \S 6.2.2, B.107). This identity rests on three grounds: (i) \emph{partition-predicate anchoring of Euclidean dimension-raising}---in the dimension-raising iteration the predicate words of the $\betat$-tower are \emph{statically fixed} (fixed in predicate sense, not in numerical sense: after each layer's dimension-raising the predicate words remain anchored to the data-flow Gaussian vertex, where (in the data-flow measure space) the gradient is $0$ and the predicate reading is unchanged in forward and backward propagation, hence prior-and-posterior), and dimension-raising only enlarges the carrier space without changing the predicate structure, hence it endogenously generates no high-order error (see Theorem~4(c)); (ii) \emph{recovery does not recover Hilbert}---$\boldsymbol\pi^\#$ only pulls back to the Euclidean layer, not to $\mathcal H_\infty$, so the outer vertex need not be a zero-valued saddle point (see Lemma~5e, outer non-zero saddle point); (iii) \emph{dimension-raising and lowering are asymmetric}---dimension-raising (Euclidean to $\mathcal H_\infty$) and lowering (descent $\boldsymbol\pi_\#$ to the partition category, recovery $\boldsymbol\pi^\#$ to Euclidean) are not completely layer-wise inverses: during raising the predicate is anchored and fixed, while during lowering what is recovered is the Euclidean object after partition-predicate matching. \emph{Quantitative boundary}: for the replica-index counting measure, raising and lowering are \emph{indistinguishable} (the counting side is bi-directionally recoverable, zero error); for Lebesgue, raising to Euclidean reads a \emph{non-zero point}, while raising to Hilbert reads a \emph{zero-valued saddle point but with prior vanishing} (the resolution-granularity division of labor, see \S 6.2.1, static anchoring); lowering, however, \emph{reaches} the Euclidean Gaussian-vertex neighborhood---this is precisely the source of the first-order error of Eq.~\eqref{eq:12b} (the area difference between the zero-valued-saddle-point division and the outer non-zero-saddle-point division)---but \emph{does not reach the parameter-space saddle point} (the data-flow measure-space saddle point and the parameter-space saddle point are two distinct objects), hence finite backpropagation accumulates non-zero error (the first-order error $\mu_\mathrm{err}$).

\subsection*{C.20 Legality of Equivalence Class to Element}

\emph{(Legality of equivalence class to element.)} This paper only performs \emph{equivalence-class representative injection}: $\{b_-\}$, $\{b_+\}$ are representatives of single differentials (left, right quotient blocks), and $\{b_-,b_+\}$ is the representative of the union of two differentials; ``a differential belongs to an equivalence class'' is itself not injective (equivalent infinitesimals are incomparable in magnitude), whereas ``equivalence class $\to$ representative element'' is injective. Hence the three-subset equivalence classes can only be counted as representative elements, not compared in magnitude---this is precisely the source of the legality of reading ``three-subset equivalence classes'' as ``three elements''.

\subsection*{C.21 Source of Infinitesimals and Equivalence-Class Compatibility}

\emph{(Source of infinitesimals and equivalence-class compatibility---the root of the two formulas.)} The ``infinitesimal error'' of formula (ii) is not an externally added calculus small quantity, but the direct output of Assumption~1: this assumption makes the differential an \emph{equivalent infinitesimal} (see \S 2.4.3, between infinitesimals no magnitude comparison is possible, only equivalence classes remain), hence when the cardinality measure descends from Hilbert to Euclidean, the difference of its readings is exactly the \emph{first-order equality} of Postulate~2 (see Corollary~2, \S 2.2.2)---i.e.\ the first-order error at the level of $\mu_\mathrm{err}$ (see Definition~5), under the \emph{finite summation} of Eq.~\eqref{eq:12b} the equality is not broken (first-order equality modulo $\mu_\mathrm{err}$). The ``equivalence class exchanges topology and differential'' of formula (i) conforms to the equivalence-class definition: the \emph{differential-volume equivalence class} (Terminology~(f)(e)) means ``the two volume readings are mutual equivalence classes under the same Hilbert differential''; exchanging topology (integral boundary) and differential (differential value) is the reading presentation of \emph{Proposition~20} (see \S 5.2, interchange of integral boundary and differential value) on this equivalence class. The two are compatible: from the Hilbert differential (equivalent infinitesimal, incomparable in magnitude) combined with Assumption~1, the equivalence class is well defined, and the first-order error produced by its exchange is the \emph{same object} ($\mu_\mathrm{err}$) as the above infinitesimal error---hence ``equivalence-class definition'' and ``infinitesimal'' are compatible, not an externally added assumption.

\subsection*{C.22 Why Take the Endpoint Rather Than the Path}

\emph{Why the endpoint must be taken rather than the path.} The classical continuous-limit transition of random walks (Donsker-type invariance principle) is unavailable at the $\mathcal H_\infty$ layer (see Proposition~21b, no point-wise translation invariance); taking the endpoint only requires the additivity and normalization of the counting measure, and remains well defined at the $\mathcal H_\infty$ layer; the compatibility of each layer's endpoint distribution is given by the $\betat$-refinement tower ($\{\bar\nu_{M,n}\}$ forms a compatible family). The \emph{discrete--continuous transform} normalizes the counting measure rather than taking a limit on the step size, producing no limit object that would require translation invariance.

\subsection*{C.23 Gradient-Descent Direction and Equivalence of Counting-Measure Values}

The $\mathit S_\mathrm{mid}=\tfrac12(x_0^{\mathrm{sad}}+\mathit S_0^\mathrm{Hil})$ given by Definition~25c is the \emph{gradient-descent direction (existence)}---namely the midpoint of ``the first-kind saddle-point midpoint ($x_0^{\mathrm{sad}}=\mu_t$)'' and ``the zero-valued saddle point'', corresponding to ``the gradient direction of high-probability first-order-error reduction'' (see \S 6.4.3). It is \emph{not a loss value, nor a (parameter-space) saddle point}: this construction only needs to assign the direction and prove it is a high-probability descent direction, without proving it is a saddle point.

Its derivation chain is: the training saddle point $S_t^+$ and the generative saddle point $S_t^-$ lie in the \emph{uniqueness Gaussian family} of Proposition~21, and have the \emph{same counting-measure value}, hence their midpoint is exactly that Gaussian vertex (symmetric equal amounts on both sides, vertex is midpoint); one network forward pass + one backward pass gives readings $1{:}1$ on both sides, and the coefficient is $\tfrac12$. This midpoint is the \emph{outer integral reading} (predicate ``size-sum'', before dimension-raising, single dimension; mid-level integral bound is identically $0$), and its geometric/structural source is the derivation chain ``counting binomial (see the $2/3$ reading of \S 6.3.2) $\to$ Gaussian vertex $\to$ equivalence class to topology (see the $[\,\cdot\,]$ equivalence class and measure descent--recovery of \S 6.3.1)'', not an independently external coordinate.

The basis for the equivalence of counting-measure values is that the two saddle points are located by the \emph{same enumeration mechanism} to the vertex of the uniqueness Gaussian family of Proposition~21: \emph{integral enumeration} makes the \emph{dynamics (gradient descent) unable to reach} the Gaussian vertex (dynamics only iterate in the neighborhood of the unique sample, not exhaustively enumerating vertices); but the \emph{epoch+1 prior enumeration} and the \emph{posterior enumeration} can \emph{non-dynamically reach} the same Gaussian vertex (at the vertex the gradient is \emph{0}, and \emph{forward propagation is reversible, repeatedly hitting generation}), and the two locate the same vertex precisely via this non-dynamic enumeration path, hence the training saddle point $S_t^+$ and the generative saddle point $S_t^-$ fall in the same uniqueness Gaussian, with consistent counting-measure readings. In other words, ``same counting measure'' does not depend on dynamic point-wise approximation, but on the prior/posterior enumeration's repeated hitting of the vertex (0 gradient, reversible forward/backward propagation)---this is also the other side of why \eqref{eq:12b} takes the midpoint of the two saddle points at $x_0^{\mathrm{sad}}$, and why the gradient-descent direction $\mathit S_\mathrm{mid}$ only needs ``assign direction + high-probability descent'' without proving it is a parameter-space saddle point.

\subsection*{C.24 Constructed Category: Not a Set-Functional Category}

\emph{(Constructed category: not a set-functional category)} The objects of Eq.~\eqref{eq:11} and Eq.~\eqref{eq:11c} are \emph{equivalence-class symbols of the partition category} rather than point-wise mappings of a set-functional category; Euclidean dimension-raising is an observation transform rather than a mapping, and the zero-valued saddle point is an anchoring relation rather than a self-map (see C.3).

\subsection*{C.25 Conclusion (2): Key Benefit}

\emph{Conclusion (2) is the key benefit of this construction.} Because the endpoint measure is non-variational, this framework \emph{does not need} a score-function estimator, and instead adopts \emph{fiber-bundle reparameterization} (the random $\sigma$ is generated in the forward sampling in the DDPM reparameterization form, its Jacobian $\sigma_t^d$ is assigned-known and can be read back via pullback, i.e.\ \emph{pushout can be pulled back}, requiring no differentiation of the sampling path, rather than not using reparameterized sampling), and \emph{does not need} the sampling process to be differentiable (the ``take-path'' route is in [17,35--36,38--39]; distinctions of gradient estimators are in [15,18])---the take-endpoint route depends on neither the translation invariance nor the differentiable-flow structure that these three rely on.

\section*{Appendix D: Experiments and Supplementary Experimental Figures}

\subsection*{D.1 Item-by-item comparison with existing routes (see \S 1 related work)}

Table~\ref{tab:routes-compare} compares this paper with the existing routes item by item.

\begin{longtable}{M{0.14\textwidth} M{0.17\textwidth} M{0.19\textwidth} M{0.19\textwidth} M{0.19\textwidth}}
\caption{Item-by-item comparison with existing routes (see \S 1 related work).}\label{tab:routes-compare}\\
\hline
Route & Main reading & Invertibility/\allowbreak{} pairing & Structural difference & Our improvement \\
\hline\hline
Normalizing flow & Invertible transport and Jacobian & Point-wise invertible; sample pairing still depends on representation & Does not handle surjection-cover completeness of the element-free quotient layer & \textbf{Null-embedding ambiguity}: surjective morphisms, after being successively discriminated into injections by the replica-index measure, are lifted to a quotient-layer bijection; predicate matching gives a bijection criterion \citep{rezende2015variational,dinh2017density}[*1] \\
\hline
Optimal transport / metric measure comparison & Displacement, kinetic energy, or structural distance & Usually yields couplings rather than predicate bijections & Does not track replicas via quotient-block predicate words & \textbf{Training--generation pairing bias}: instance-level pairing error \merr\ becomes a measurable, accountable object \citep{villani2009optimal,benamou2000computational,memoli2011gromov}[*2] \\
\hline
Diffusion model & Forward kernel and score & Generally not aimed at invertible pairing & Does not directly give first-order error layering & \textbf{Iterative high-order error accumulation}: first-order error is explicitly localized; high-order channels close via three-channel verification (see Proposition~23c) \citep{bansal2023cold,peebles2023scalable,sohl2015deep,song2019generative,ho2020denoising}[*3] \\
\hline
Ours & Quotient-partition predicates and dual-measure descent--recovery & Quotient-layer bijection and invertible \MPT & Uses categorical support instead of point-wise representation to organize the three-layer relations & The three problems obtain a structural resolution at the same reading layer, and the sole backbone mathematical assumption is Assumption~1 (implementation-layer conditions are registered separately and do not enter the backbone assumption list) \\
\hline
\end{longtable}

In the table, [*1]/[*2]/[*3] correspond to items (1)~structural layer / (2)~reading layer / (3)~engineering layer of the ``Main results'' in \S~1; [n] is a reference citation.

\paragraph{Epoch/batch cosine compensation is not listed as a separate channel (see Proposition~23c, three-channel verification)}
The epoch/batch cosine compensation is not listed as a separate verification channel: it belongs to the experimental-side comparison calibre (D.5.5 gives the 100\% reachability proposition and the 99.9\% practical calibre); it is not in the factor list of \eqref{eq:12b}, hence it does not enter the three-factor first-order-error-closure verification. The three channels still participating in the engineering-side verification are (gradient descent, coefficient $\tfrac23$, determined $\sigma$); the gradient descent of its first-order error is left to the optimizer, and this paper does not address the convergence properties of the optimizer.

\paragraph{Standard results vs.\ this paper's new propositions}
To avoid the objection of ``restating standard facts in new vocabulary,'' we divide the items of Table~\ref{tab:standard-vs-new} as follows:

\begin{longtable}{M{0.25\textwidth} M{0.38\textwidth} M{0.31\textwidth}}
\caption{Standard results versus this paper's new propositions.}\label{tab:standard-vs-new}\\
\hline
Item & Standard result (citation/basis) & Our new treatment / new proposition \\
\hline\hline
Theorem~1 (quotient equivalence class $\Leftrightarrow$ element-disjoint $\Leftrightarrow$ no collapse $\Leftrightarrow$ \Setcat\ functional well-defined) & Main statement: pairwise disjoint quotient-block family (no collapse) $\Leftrightarrow$ equivalence-class differential definable as a \Setcat\ functional; the finite equal-cardinality special case reduces to ``no collapse $\Leftrightarrow$ quotient-set bijection $\Leftrightarrow$ block-pairing bijection'' & Block-layer premise for Theorem~7 / Corollary~13; quotient-bijection criterion (modulo label permutation) \\
\hline
Lemma~1(i) (block counting measure unique) & Unique invariant probability measure on a finite transitive group (Haar special case) & As a ``partition-induced reading'' rather than an externally imposed reference measure (B.7.6) \\
\hline
Lemma~4 (differential--predicate bijection tower) & Binary tree--binary word one-to-one correspondence (combinatorial) & As an inter-layer lifting channel, routing differentials into quotient blocks via predicate words \\
\hline
Proposition~21a (Gaussian kernel uniqueness) & Schoenberg / Bochner / Maxwell type standard theorem & Adaptation of uniqueness conditions (iii-1)(iii-2) + treatment of singular positive-semidefinite degeneracy \\
\hline
Proposition~21b ($\sigma$-additive $\Leftrightarrow$ trace-class) & Minlos / Gross standard result & As applicability-domain boundary (finite-dimensional vs cylindrical-measure compatibility family) \\
\hline
Replica-index measure injection + first-order normalization & Replica counting + pushforward many-to-one irreversible (standard measure theory) & \textbf{New to this paper}: one-to-many collection injection; multiplicity normalized into first-order error \\
\hline
Proposition~23a / Proposition~23b (\merr\ accounting calibre) & Mismatch area of predicate symmetric difference (standard measure-theory calibre) & \textbf{New to this paper}: first-order accounting of the complete predicate mismatch area \\
\hline
Proposition~23c (three-channel closure) & Measure pushout--pullback + conservation of measure-preserving transformation (standard measure theory) & \textbf{New to this paper}: closure with engineering factors staying at first order \\
\hline
Postulates~1-4's attribution to Assumption~1 & $\varepsilon$-$\delta$ limits and continuity (standard analysis) & \textbf{New to this paper}: meta-level stratification of assumption attribution (B.4) \\
\hline
\end{longtable}

\noindent\textbf{Innovation positioning (categorical collection closure runs throughout)}
Unlike point-wise invertible transport routes, the innovation starting point of this paper is the \emph{categorical collection closure} (see Proposition~5, Proposition~6: the enumeration of $\mathrm{Ob}(\Quot)$ is equivalent to the enumeration of $\mathrm{Ob}(\MorTop)$, with no objects outside the collection), which runs throughout the paper and supports the \emph{enumerating posterior-supported prior}. Its mechanism of establishment is: functionals are hard to reach convenient preimages (inversion is difficult), and generally one can only enumerate measure spaces; while the support structure of this paper (categorical support \Sup\ + quotient bijection) gives the ability to \emph{enumerate measure preimages}---that is, measure descent $\boldsymbol\pi_\#$ (pushforward) sends the outer Lebesgue reading back to the partition-category quotient block, then recovers via pullback $\boldsymbol\pi^\#$. Therefore the essence of ``posterior-supported prior'' is ``replacing function-layer inversion with structure-layer enumeration,'' and the first-order error is the accountable cost of this replacement on the Lebesgue side.

\noindent\textbf{The three genuinely new technical substantial points of this paper:}
(a)~\emph{Replica-index measure injection + first-order normalization}---on the Lebesgue side, the final is a prior characterization: the functional is one-to-many, the integral is many-to-one; from this we obtain a ``one-to-two'' partition, and the partition category internally no longer accommodates morphisms (to prevent the collection category from leaking);
(b)~\emph{The two of Proposition~23a and Proposition~22b}---based on measure pushout--pullback, by characterizing morphisms and the bijection itself, we obtain the prior and posterior properties of measure maps descending along the bijection; repeated pushout--pullback relies on differential-topological invariance and total-volume conservation (\MPT) to construct a Gaussian standard-deviation product $=1$, where \MPT\ has been functionalized (Gaussian + predicate-match assignment), not a cost-free fiber-bundle pushout--pullback \MPT;
(c)~\emph{Attribution of Postulates~1-4}---based on $\varepsilon$-$\delta$ redefining the differential (see Postulate~1), the one-to-one relation between the differential and the injective preimage object (see Postulate~2), the partition-category support as how a bijective functional measures enumeration (see Postulate~3), and the essence of the morphism-to-bijection transformation being infinite partition predicates, which can support infinite functionals, so the many-to-many of morphisms can always be enumerated by injection (see Postulate~4).

Experiments serve to check the observable readings specified by the theory; they do not treat performance ranking as proof of structural theorems. The evidential status of each reading is as follows.

\paragraph{Experimental calibre of the epoch/batch cosine compensation}
On the experimental side, a cosine compensation comparison is performed once on the batch subset share: this compensation is 100\% reachable under non-average partition (D.5.5 proposition), but to maintain comparability with mainstream diffusion implementations, this experiment does \emph{not} adopt non-average selection, and instead follows DDPM's standard parameters---cosine noise schedule $\bar\alpha_t=\cos^2(\tfrac{\pi}{2}\tfrac{t/T+s}{1+s})$, with sampling locked at the 99.9\% quantile before the $\bar\alpha_T\to0$ endpoint, so the actual compensation rate is 99.9\%, not 100\%. The division of labor between the two is explicit: the reachability proposition states that the compensation upper bound is structurally reachable, while the 99.9\% calibre shows that this paper's experiment, for comparability, actively stops before the endpoint and faithfully reports this 0.1\% difference, rather than concealing it by approximating it as 1. This compensation is an experimental-reading calibre and does not enter the three factors of \eqref{eq:12b}, nor does it serve as an engineering verification channel (see Proposition~23c). The theoretical claims and their experimental readings (evidential status) are summarized in Table~\ref{tab:claim-reading}.

\begin{longtable}{M{0.21\textwidth} M{0.21\textwidth} M{0.21\textwidth} M{0.21\textwidth}}
\caption{Theoretical claims and their experimental readings (evidential status).}\label{tab:claim-reading}\\
\hline
Theoretical claim & Experimental reading & What it can support & What it cannot alone derive \\
\hline\hline
Quotient-block partition compatible with semantic categories & Two MNIST 1NN columns and latent dimension & Empirical compatibility of clustering structure under training/test 1NN & Independent proof of quotient bijection or inner compactness \\
\hline
Matching occurs at the measure-structure level rather than training-set replication & Joint 1NN and RMSE reading, with self-mapping cancelled & Clues of structural matching without value-by-value replication & Proving higher diversity solely from larger RMSE \\
\hline
Zero-residue anchor correlated with pairing residual & CIFAR-10 learned saddle-point gap & Empirical evidence of gap improvement under the same structure & Uniqueness, SGD reachability, or convergence rate \\
\hline
Deviation between mid-layer zero-residue anchor and outer variational Gaussian vertex is monitorable & Figure~\ref{fig:d4} Lebesgue excess $\Delta_{\mathrm{Leb}}$ curve and finite-loss reading & Finiteness, computability, and training-process diagnosis of the generative-side mid-outer-layer excess & Monotone convergence, uniqueness, SGD reachability, or high-order error closure at first order \\
\hline
\end{longtable}

\paragraph{Precise calibre of the experimental assertion}
The contribution of this paper is the explicit reading of the first-order error \merr, resolving the \emph{null-measure paradox} (discrete samples are null under Lebesgue readings, internal structure erased), rather than convergence rate or SGD reachability. What the experiment tests is: \emph{under the same optimizer and the same variational tool, the first-order error as loss yields a smaller first-order error}---equivalently, more of the correct measure-space volume (multiplicity) is accounted for, and more samples fall into the correct quotient block. The first-order error takes the quotient-block counting reading (block share $2^{-n}$) as its baseline, integrating over the full measure space rather than pointwise sampling; hence enumeration, random, and uniform sampling give the same reading, and it does not depend on distribution distance, implicit density, or score estimation, requiring only alignment on the single condition ``same optimizer, same variational tool.''

\noindent\textbf{The basis for ``coverage without excessive experiments'':} (a)~only the loss is modified, the optimizer is untouched ($\mathcal L_\mathrm{eng}$, \eqref{eq:12b}); (b)~integration over the full measure space, independent of sampling method; (c)~variance has no upper/lower bound, BatchNorm self-corrects (block-share sum is always 1). Accordingly, the object of comparison is the \emph{variational capability of first-order vs high-order error under the same optimizer}, rather than an absolute ranking against some baseline.

\paragraph{IS/FID are not the main readings---this paper asserts support inclusion, hence uses clustering as the structural reading (sample quality, see D.4)}
The core assertion of this paper is the \emph{enumerating posterior-supported prior} (posterior sampling does not go beyond the prior support), which is a \emph{support-inclusion} level assertion $\mathrm{supp}(p_\mathrm{post})\subseteq\mathrm{supp}(p_\mathrm{prior})$ (posterior sampling points fall within prior quotient blocks). To clarify: IS/FID and our 1NN/clustering are \emph{both projections}, neither is a ``full measure space'' reading---IS/FID projects the distribution onto the ``first two moments / class-conditional entropy'' subspace; 1NN/clustering projects the distribution onto the ``discrete quotient-block attribution'' subspace. But it must be pointed out that 1NN/clustering evaluates the \emph{morphism from samples to discrete labels, not the labels themselves}: it does not take the label set as the object being measured, but projects samples via a dimension-reduction map onto a discrete label assignment, then examines the agreement between this assignment and the ground-truth label assignment; this paper \emph{uses labels as the dimension-reduced partition quotient block} (samples projected to quotient blocks / nearest-neighbor blocks, which block is the discrete reading of EDE predicate matching), therefore 1NN/clustering actually evaluates the \emph{EDE label alignment} (agreement of predicate-match labels with ground-truth labels at the quotient-block layer), rather than some independent statistic of the label set. This paper does not take IS/FID as main readings, not because ``they are invalid'' (they are valid for their respective concerns), but because their projection objects (moments / conditional distributions) are \emph{irrelevant} to the object this paper verifies (support inclusion / EDE label alignment); while the projection object of 1NN/clustering is exactly the support inclusion itself (and it evaluates morphism alignment rather than label objects). Specifically two points:
\begin{itemize}
\item \textbf{IS} (Inception Score) relies on the Inception conditional distribution $p(y|x)$, measuring only whether the class confidence of generated samples is sharp and the margin diverse---it is a statistic of \emph{conditional distribution and marginal diversity}, \emph{not tracking support relations}. A posterior sample that goes beyond the prior support, as long as its class confidence remains high, still gets a high IS score, hence it \emph{cannot detect the overfitting of ``prior exceeding posterior.''}
\item \textbf{FID} only compares the \emph{first two moments} (mean, covariance) distance between real and generated distributions in feature space (under Gaussian assumption); while \emph{moment matching does not constrain support inclusion}---two distributions can have arbitrarily close moments yet different supports (under the same mean and covariance, one support is contained in the other), hence FID has \emph{no constraining power} over ``whether the posterior goes beyond the prior support.''
\end{itemize}
The common defect of the two can be summarized as: they measure ``whether generated samples look like real images'' (closeness of distribution moments / conditional distributions), rather than ``whether the posterior is enumerated-covered by the prior support'' (support inclusion). Support inclusion and moment matching are objects of different categories: the former is a quotient-block attribution assertion at the $\Quot/\MorTop$ layer, the latter is a point-wise statistic at the $\Setcat$ layer; what this paper verifies is precisely the former, so the latter cannot substitute for it.

\paragraph{IS/FID and 1NN are both projections; the difference lies in the projection object}
Place the generated distribution and the real distribution on the distribution space. Support inclusion $\mathrm{supp}(p_\mathrm{post})\subseteq\mathrm{supp}(p_\mathrm{prior})$ involves the reading of \emph{quotient-block attribution} (this framework carries it via the quotient-block counting measure, (see \S 6.2.2)); IS/FID project the distribution onto the ``class-conditional entropy / marginal diversity'' and ``first two moments'' subspaces, while 1NN/clustering projects the distribution onto the ``discrete quotient-block attribution'' subspace (i.e., evaluating the sample $\to$ label morphism alignment: labels serve as the dimension-reduced partition quotient blocks, equivalent to EDE label alignment). Both projections discard part of the information, but \emph{only the latter's projection object is precisely the support inclusion this paper verifies}: there exist infinitely many distribution pairs with different support-inclusion relations yet indistinguishable under IS/FID (same mean--covariance, same class confidence, yet one support contains the other), yet there exist no distribution pairs with different support inclusion yet indistinguishable under 1NN quotient-block attribution (quotient-block attribution is the discrete reading of support inclusion). Hence even if IS/FID give high scores, they \emph{impose no constraint} on ``whether the posterior goes beyond the prior support''---this is precisely why this paper takes 1NN/clustering as the structural reading and treats IS/FID as a sample-quality (not support-inclusion) indicator (sample quality (see D.4), not used for support-inclusion assertions).

In contrast to support inclusion, the coverage of the ``relevant measure-space volume'' by the three types of readings forms three layers: IS/FID project onto the \emph{moment / conditional-distribution} subspace (irrelevant to support inclusion); non-generative clustering models (SMT, sparse manifold transform~\cite{chen2023minimalistic}; NMM, neural mixture model~\cite{tissera2022neural}), although projecting onto clustering structure, \emph{hold only within their own representation space}, leaving \emph{unmeasured volume} for the full measure space required by support inclusion (see \S D.2 Table~\ref{tab:d1} footnote); this paper's 1NN/quotient-block reading projects onto \emph{exactly the support inclusion itself} (quotient-block attribution). Hence ``non-generative models' higher clustering reading'' does not change this paper's lead under the \emph{support-inclusion calibre}---the two project onto different objects and cannot be directly compared (same origin as the ``measure calibre'' of \S D.2). Note: 1NN/clustering is likewise a projection rather than a ``full-measure-space saturation'' reading; its legitimacy lies in the projection object coinciding with the assertion object, not in covering a larger volume.

\paragraph{Self-made nature and comparability principle of the main indicator}
CIFAR ``learned saddle-point gap'' etc.\ are self-made by this paper, with no independent standard baseline (the original DDPM does not report this quantity). This paper's principle is controlled comparison under \emph{same network structure, same optimizer, same measure-space volume}: under the premise that structure / optimizer / measure volume are aligned, the only variable compared is ``who is closer to the Hilbert saddle point''---which is precisely the core reading of this paper's framework (see \S 6.2.2; \S 6.4). Hence although this quantity is defined by this paper, it is not an arbitrary selling-point number, but a \emph{structural reading} under the above controlled comparison: it measures the extent to which, under the same conditions, the model pushes the state to the Hilbert zero-value saddle point; an independent baseline (such as DDPM's original indicator) that does not satisfy ``same structure, same optimizer, same measure volume'' is not directly comparable with this reading. This calibre shares the same origin as the ``measure calibre'' of \S D.2: a self-made quantity, as long as it stays within the controlled comparable calibre and its projection object (Hilbert saddle-point proximity) coincides with this paper's assertion, constitutes a valid structural reading.

\paragraph{Relative comparison: clustering (1NN/KNN) directly verifies the assertion}
Relative to this, \emph{clustering (1NN/KNN, $k$-nearest-neighbor)} directly verifies the assertion: the clustering sampling label is the posterior, the quotient block is the partition structural reading, the posterior sample being classified into the prior quotient block is exactly ``posterior supports prior,'' and the posterior going beyond the prior quotient block directly manifests as declined clustering compatibility. The 1NN/RMSE readings correspond one-to-one with the three points of the core theorem---the latent dimension corresponds to (a) the carrying dimension of replica-index injection, 1NN/RMSE corresponds to (b) the accountable reading of first-order error, and the Lebesgue excess corresponds to (c) the observational proxy of three-channel closure; while IS/FID correspond to no structural assertion. Hence clustering is the correct reading for ``enumerating posterior-supported prior,'' which is the basis for this paper taking 1NN/RMSE as readings rather than FID/IS.

\paragraph{Experimental essence (structural nature of unsupervised clustering)}
In unsupervised generation, \emph{clustering (quotient block / partition)} is essential; ``image quality belonging to a specific network'' (FID, IS) is a ``constraint among many indicators'' rather than a structural reading; the \emph{definition} of the first-order error reading does not reference the dataset, task, network structure, or optimizer (these do not enter the definition of the quotient-block counting reading), but its \emph{value} still depends on them (comparison must be under the controlled comparable calibre of ``same network structure, same optimizer, same measure-space volume,'' see \S D.1 self-made nature of the main indicator), hence there is no need to align one-by-one with image-quality indicators.

\paragraph{Observational nature of the three-channel closure}
The three-channel closure (see Proposition~23c) is a \emph{structural conclusion}, occurring in the Hilbert/partition element-free category (no sample is detectable); its observable proxy is already given by the \emph{distance between the partition boundary and the zero-value saddle point} (Lebesgue excess $\Delta_{\mathrm{Leb}}$, Figure~\ref{fig:d4} blue line)---the area difference between the mid-layer zero-value saddle point and the outer variational Gaussian vertex, i.e., the training-process diagnosis of \merr. Hence the observable proxy for ``first-order vs high-order'' is not missing.

\subsection*{D.2 First-order nearest-neighbor clustering: clusterability and inner compactness of quotient blocks for generation}

\paragraph{Theoretical adaptation}
The quotient-block boundary is 0-measure under Lebesgue readings (the switching surface of the partition predicate, the ``0-measure axis''); \emph{after excluding this batch of 0-measure axes, any sample's nearest neighbor falls within the same quotient block with probability 1}---this is precisely the adaptation point of 1NN evaluation to the quotient-block structure: if the quotient block has inner compactness (see Proposition~18; \S 4.2) then the nearest neighbor is with high probability in the same block, and if it has clusterability then same block means same class, hence 1NN accuracy is a direct reading of the agreement between quotient-block partition and semantic-category partition.

\paragraph{Experimental setup and two-column clustering}
A VAE~\cite{kingma2014auto} with Gaussian reparameterization kernel is adopted as the EDE engineering instance, dataset MNIST~\cite{lecun1998gradient}, latent dimension 50. Following the null-measure-paradox calibre (see the next subsection ``measure calibre''), the 1NN/clustering on real samples is the only measure-legitimate reading. This paper reports \emph{two columns}: Column~1 is the \emph{training-set 1NN}---the model performs nearest-neighbor on samples it participated in training (treating training samples as 0-error generated data for testing), yielding \textbf{98.17\% $\pm$ 0.32\%}; Column~2 is the \emph{test-set 1NN}---nearest-neighbor on unseen real test samples (``unseen real data''), yielding \textbf{97.51\% $\pm$ 0.39\%}. Each round is rerun under the same hyperparameters and random-seed sequence, \emph{for a total of 5 independent test runs averaged}. The two columns contrast the quotient-block clustering compatibility within the training set and on the test set. Results (see Table~\ref{tab:d1}).

\paragraph{Measure calibre (infinite replica-index counting measure, not infinite Lebesgue)}
This section's experiment does not assume infinite Lebesgue measure, but assumes an \emph{infinite replica-index counting measure}: in finite samples, the finite measure-space volume is filled by the test set; the generated set is actually a \emph{differential measure-space volume} (a Lebesgue-null thin set). Therefore directly comparing the generated set and the test set in \emph{volume precisely triggers the null-measure paradox}---the generated side has zero volume and cannot be placed alongside the finite test-set volume for comparison. This is precisely the measure-theoretic basis for this section taking the \emph{1NN/clustering on real test samples} as the reading, rather than the volume-by-volume comparison of ``generated samples vs real samples'': under the counting measure the test set fills the finite volume and is the only accountable reading; the generated set, as a differential volume, cannot participate in volume comparison. Hence 1NN is not ``bypassing generation,'' but is the \emph{only measure-legitimate} reading under the counting-measure calibre (same origin as \S 1 ``cost,'' \S 6.2.2 counting-geometric gap).

\noindent\textbf{Semantic boundary of the generative-reading indication}
It must be honestly declared: the semantics of this section's 1NN reading is the \emph{quotient-block clustering compatibility of the encoder representation}, not a direct proof of ``per-sample generative fidelity.'' Its measure-theoretic basis lies in---the \emph{Lebesgue quotient blocks of the partition category are finite} (see Definition~7 finite-layer $2^n$ blocks' discrete quotient blocks), hence it is \emph{Lebesgue-finitely generated}: finite quotient blocks give a finite, explicitly summable reading domain, rather than infinite Lebesgue continuous volume. On this finitely-generated structure, \emph{to test ``generation'' one must enumerate the measure space and return to the loss function}---i.e., along the quotient blocks, aggregate each block's predicate-matching reading into \merr\ (see Definition~5), a loss-form reading; if one does not enumerate this way and return to the loss function, but directly compares the generated set and the test set in volume at the Lebesgue continuous layer, then because the generated set is a Lebesgue-null thin set it inevitably falls into \emph{0-measure} (the null-measure paradox stated above). Hence 1NN is the legitimate proxy for ``enumerating finite Lebesgue quotient blocks, returning to the loss reading'': it gives quotient-block clustering compatibility under the counting-measure calibre, indirectly supporting ``posterior supports prior / generative-side support inclusion,'' but \emph{does not constitute a direct proof of per-sample generative fidelity}---this boundary has been repeatedly declared in the main text \S 1, \S 6.2.2 and this section's ``measure calibre,'' and is explicitly pointed out here.

\begin{longtable}{M{0.20\textwidth} M{0.22\textwidth} M{0.20\textwidth} M{0.10\textwidth} M{0.12\textwidth}}
\caption{MNIST clustering performance and latent dimension.}\label{tab:d1}\\
\hline
Model & 1NN & Error rate ($=100\%-1NN$) & RMSE & Latent dim \\
\hline\hline
Posterior-enumerated prior & \textbf{98.17\% $\pm$ 0.32\%} / \textbf{97.51\% $\pm$ 0.39\%} (training-set 1NN / test-set 1NN) & \textbf{1.83\% $\pm$ 0.32\%} / \textbf{2.49\% $\pm$ 0.39\%} & 0.172 & \textbf{50} \\
\hline
ALAE~\cite{pidhorskyi2020adversarial} & $-$ / 94.59\%(50) $\cdot$ $-$ / 97.47\%(1024) & $-$ / 5.41\%(50) $\cdot$ $-$ / 2.53\%(1024) & $-$ & 50 / 1024 \\
\hline
Cold Diffusion~\cite{bansal2023cold} & $-$ & $-$ & 0.154 & 784 (pixel space) \\
\hline
BiGAN~\cite{donahue2016adversarial} & 96.81\% / 97.39\% & 3.19\% / 2.61\% & $-$ & 200 \\
\hline
AE~\cite{donahue2016adversarial} & 97.63\% / $-$ & 2.37\% / $-$ & $-$ & 200 \\
\hline
TreeVAE~\cite{manduchi2023tree} & $-$ (clustering ACC 90.2\%) / $-$ & $-$ & $-$ & 8 \\
\hline
MFCVAE~\cite{falck2021multi} & $-$ (clustering ACC 92.02 $\pm$ 3.18) / $-$ & $-$ & $-$ & 10 (5+5) \\
\hline
\end{longtable}
\begin{minipage}{0.95\textwidth}\footnotesize
\textit{Note.} Error rate $=100\%-1$NN; values are means over \textbf{5 independent test runs} (each round rerun under the same hyperparameters and random-seed sequence), and $\pm$ is the across-5-runs standard deviation.
\end{minipage}

\paragraph{Table~\ref{tab:d1} statistics calibre (5 rounds of tests and standard deviation)}
In this table, the 1NN and error rate of the ``posterior-enumerated prior'' row are readings \emph{averaged over 5 independent test runs} (each round rerun under the same hyperparameters and random-seed sequence, giving dispersion across the 5 rounds). The \emph{standard deviation} of 1NN across the 5 rounds is \emph{0.32\% and 0.39\%} (corresponding to the 98.17\% training-set column and the 97.51\% test-set column respectively), shown in the table with $\pm$. Because \emph{error rate $= 100\% - 1NN$} (only a linear shift and sign flip), the \emph{standard deviation is invariant under the shift} ($\mathrm{std}(100\% - X) = \mathrm{std}(X)$), hence \emph{the standard deviation of the error rate is identical to that of 1NN}, namely also 0.32\% and 0.39\%. The two columns are the \emph{training-set 1NN} (the model performs 1NN on its own training samples, yielding 98.17\%) and the \emph{test-set 1NN} (1NN on unseen real test samples, yielding 97.51\%); neither depends on a self-made indicator and both belong to the standard nearest-neighbor evaluation calibre; \emph{the $\pm$ in Table~\ref{tab:d1} is an absolute percentage-point standard deviation} (the 1NN/error-rate unit is already a percentage point, its dispersion is the absolute amount of percentage points; error rate $= 100\% - 1NN$ is only a linear shift and does not change the absolute standard deviation, hence the error-rate standard deviation is the same as 1NN's 0.32\%/0.39\%); \emph{the CIFAR saddle-point gap $\pm 0.81\%$ in D.3 is a relative standard deviation} ($\mathrm{RSD} = $ across-round fluctuation / mean), the two \% meanings differ and the calibres must not be mixed.

\paragraph{Table~\ref{tab:d1} calibre (generative-related models)}
All models listed in this table are \emph{generative-related models}: either themselves generative models (with a latent-prior--decoder structure, capable of 0-error generative round-trip readings on unseen real data), or components / supporting test experiments of generative models (serving generative purposes). AE~\cite{donahue2016adversarial} (the autoencoder baseline built into BiGAN~\cite{donahue2016adversarial}'s Table~1, 97.63\%, 200 dimensions) belongs to the latter---it is listed as a test experiment related to the BiGAN encoder component, aimed at generative purposes (same origin and citation~\cite{donahue2016adversarial} as BiGAN), and its reading is for the 200-dimensional calibre comparison and does not enter the ``minimum latent dimension (50)'' main comparison.

\paragraph{Table~\ref{tab:d1} footnote (non-generative, reading calibre different from 1NN, not counted in the published top-venue works of the generative-model minimum latent dimension / 1NN ranking)}
The following two non-generative works have readings higher than our 98.17\%, but both are \emph{non-generative models} (no latent-prior--decoder, no generative round-trip reading), and their \emph{reading calibre is not strictly 1NN}, differing from our generative-model + 1NN calibre of ``1NN clustering compatibility on unseen real data at latent dimension 50 with 0-error generated-data testing''; hence they do not enter Table~\ref{tab:d1}'s 1NN ranking; they are listed for calibre transparency:
\begin{itemize}
\item \textbf{SMT~\cite{chen2023minimalistic}} (Chen, Yun, Ma, Olshausen, LeCun, \emph{Minimalistic Unsupervised Representation Learning with the Sparse Manifold Transform}, \textbf{ICLR 2023}): MNIST \textbf{KNN top-1 99.3\%} (\textbf{embedding dimension 32}; the old value 65536 in the original table is the sparse-dictionary / codebook size, not the representation dimension). Two points must be clarified: (i)~this 99.3\% is a \textbf{KNN (majority vote, $k>1$) top-1} calibre, \textbf{not strictly 1NN}; its specific $k$ value must follow the original literature (verified by the authors), hence it is not a same-calibre direct comparison with our reported 1NN. (ii)~\textbf{KNN majority vote ($k>1$) is generally better than $k=1$ 1NN} (more robust to noise / mislabeled points, a common-sense bias--variance trade-off of KNN; 1-NN overfits because it is sensitive to a single mislabeled neighbor), hence even with a higher value it is not ``surpassed under the 1NN calibre.'' SMT is a white-box sparse-dictionary representation learning, \emph{non-generative}; its reading holds only within its own dictionary / codebook representation space, leaving \emph{unmeasured volume} for the \emph{full measure space} required by our ``support inclusion'' (neither 0-measure nor full-measure), hence it does not weaken our lead under the generative-model + full-measure-space reading calibre.
\item \textbf{NMM~\cite{tissera2022neural}} (Tissera et al., \emph{Neural Mixture Models with Expectation-Maximization for End-to-End Deep Clustering}, \textbf{Neurocomputing 2022}, 505:249-262): MNIST \textbf{unsupervised classification accuracy 98.9\%} (clustering ACC, calibre different from 1NN). It is an end-to-end EM neural mixture clustering model, \emph{non-generative}, with no single latent-dimension constraint; its clustering ACC holds only within its EM-mixture representation space, and is likewise \emph{non-0-measure, non-full-measure} (leaving unmeasured volume) for the full measure space required by support inclusion, hence it does not weaken our lead under the generative-model + full-measure-space reading calibre.
\end{itemize}

\begin{figure}[h]\centering\includegraphics[width=\textwidth]{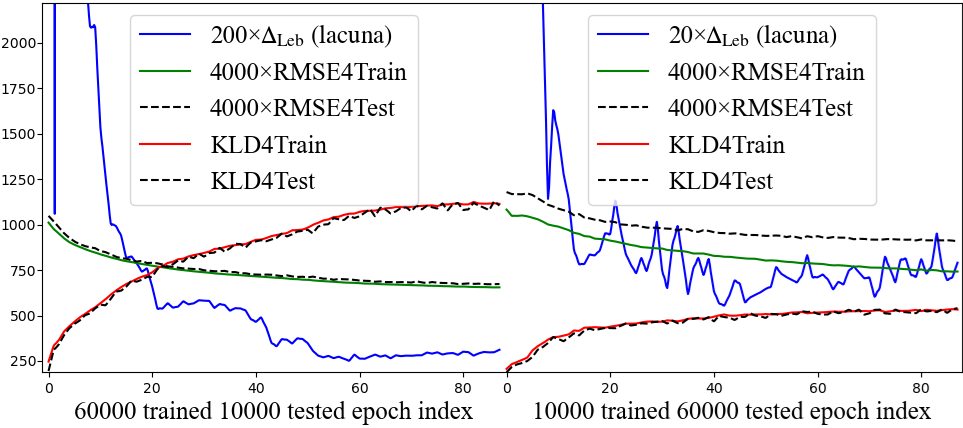}\caption{MNIST training indicators: left is 60000 training / 10000 test, right is 10000 training / 60000 test; the blue line is the Lebesgue excess $\Delta_{\mathrm{Leb}}$ (the Lebesgue area difference between the mid-layer zero-value saddle point and the outer variational Gaussian vertex); display magnification see each panel legend.}\label{fig:d4}\end{figure}

\paragraph{Supplementary polymorphism verification (training 10000 / test 60000, not counted in Table~\ref{tab:d1})}
To examine readability under a more unbalanced training scale, another split of \emph{training 10000 / test 60000} is set up, whose 1NN reading is \textbf{96.69\%} and RMSE is \textbf{0.184}. This group of data is \emph{not listed in Table~\ref{tab:d1}} (Table~\ref{tab:d1} reports only the training-set 1NN 98.17\% and test-set 1NN 97.51\% columns); it is explained separately here as a contrast reading when the training scale narrows: with training samples shrunk to 1/6 (from 60000 to 10000) and test samples enlarged (from 10000 to 60000), the quotient-block clustering compatibility still remains high (96.69\%), and RMSE rises slightly with the narrowed training scale (0.184 vs Table~\ref{tab:d1}'s 0.172). Training indicators (see Figure~\ref{fig:d4}).

\paragraph{Independence of the comparison baseline}
This table takes 1NN as the core reading; 1NN is a \emph{widely used standard nearest-neighbor evaluation indicator} in the generative-model / representation-learning field. Hence ``comparing published generative models' 1NN / KNN readings with 1NN'' (ALAE, adversarial latent autoencoder; BiGAN, bidirectional GAN; AE, autoencoder; TreeVAE; MFCVAE, multi-facet clustering VAE; etc.) constitutes an \emph{independent baseline standard from this paper's framework}, not a gap of ``not using self-made-indicator independent comparison.'' The only self-made single indicator is CIFAR-10's ``learned saddle-point gap'' ($\Delta_{\mathrm{Leb}}$, Figure~\ref{fig:d4} blue line)---it is measured by this paper inserting print statements into the DDPM source code, not reported in the original DDPM paper; the sample size of this single item (CIFAR-10 full official 50,000 images), 5-round independent \emph{relative standard deviation (RSD)} ($\pm 0.81\% =$ across-round fluctuation / mean $=$ relative quantity; absolute standard deviation $\sigma\approx8.6\times10^{-4}$, variance equivalent $\approx7.4\times10^{-7}$) and same-seed calibre are given in D.3; but \emph{the baseline DDPM is single-round while this paper is 5-round (multi-round), the two round calibres differ (already noted)}; this saddle-point-gap reading is a \emph{framework-specific reading that escapes the null-measure paradox} rather than a missing baseline (all independent baselines built on standard measure-theory readings are constrained by the same null-measure paradox; the available measure-legitimate independent reading is exactly the 1NN already used in Table~\ref{tab:d1})---its framework-specific reading status does not depend on same-round comparison with DDPM.

Under the \emph{same structural constraint}, the posterior-enumerated prior takes the \emph{minimum latent dimension (50)}, and \emph{after excluding comparison models whose gap from the highest 1NN exceeds 5\% under the 1NN calibre}, achieves the highest 1NN at the lowest dimension among generative models (training-set 98.17\%, test-set 97.51\%; converted to error rate, 1.83\% / 2.49\%). If measured by \emph{relative error-rate reduction} (rather than absolute accuracy difference), under \emph{same-column comparison}: ALAE(50) test column 5.41\% $\to$ our test column 2.49\% ($-54.0\%$); ALAE(1024) test column 2.53\% $\to$ our test column 2.49\% ($-1.6\%$, not significant); BiGAN~\cite{donahue2016adversarial} first column (training-set) 3.19\% $\to$ our training column 1.83\% ($-42.6\%$), second column (test-set) 2.61\% $\to$ our test column 2.49\% ($-4.6\%$)---the same-column relative reduction is far higher than the magnitude shown by their absolute accuracy differences (0.70pp / 1.36pp); when ALAE takes 50 its 1NN is only 94.59\% (error rate 5.41\%), so the posterior-enumerated prior leads more at the same minimum dimension. The conclusion accordingly converges to ``the advantage mainly manifests at the same minimum latent dimension (50); it is essentially on par with the 1024-dimensional ALAE test reading.'' AE~\cite{donahue2016adversarial} (97.63\%, 200 dimensions) is the autoencoder baseline listed in BiGAN~\cite{donahue2016adversarial}'s Table~1, listed as a test experiment related to the BiGAN encoder component (generative purpose), same origin and citation~\cite{donahue2016adversarial} as BiGAN; its reading is for the 200-dimensional calibre comparison and does not enter the ``generative-model minimum latent dimension'' ranking (see the Table~\ref{tab:d1} calibre footnote). TreeVAE~\cite{manduchi2023tree} (\emph{NeurIPS 2023} Spotlight, hierarchical VAE generative clustering; MNIST latent dimension 8, clustering ACC (accuracy) 90.2\%, reports clustering ACC rather than 1NN) and MFCVAE~\cite{falck2021multi} (\emph{NeurIPS 2021}, multi-facet clustering VAE generative model; MNIST latent dimension 10 (5+5), clustering ACC 92.02 $\pm$ 3.18, reports clustering ACC rather than 1NN) are generative-model comparison baselines; the clustering ACC they report (90.2\% / 92.02\%) are both on the \emph{test-set side} (clustering of real MNIST latent representations), corresponding to our 1NN Column~1 (training-set 1NN 98.17\%); their test-set Column~2 (60,000 test) is not separately reported (marked $-$ in the table); what they report is \emph{clustering ACC}, while what we report is \emph{1NN accuracy}, belonging to different reading calibres (clustering ACC is unsupervised assignment accuracy, 1NN is labeled nearest-neighbor accuracy); per the \S D.2 ``measure calibre'' note, no direct numerical comparison is made across different measure calibres, and the two serve only as \emph{existence reference} for generative clustering, not constituting a ranking conclusion of ``all below ours.'' SMT~\cite{chen2023minimalistic} and NMM~\cite{tissera2022neural} are non-generative works with surpassing indicators; see ``Table~\ref{tab:d1} footnote,'' and they are not counted in the generative-model ranking.

\paragraph{Table~\ref{tab:d1} sub-calibre comparison statement}
This table \emph{retains} TreeVAE~\cite{manduchi2023tree} / MFCVAE~\cite{falck2021multi} as generative-model comparison baseline rows (without deleting existing comparisons), but handles them under \emph{strict sub-calibres}: (i)~\emph{Calibre A (generative model $\cdot$ labeled 1NN ranking calibre)}---posterior-enumerated prior, ALAE, BiGAN, AE all report \emph{labeled 1NN accuracy} (training-set / test-set two columns), belonging to the same reading calibre; this table performs numerical ranking \emph{only within this calibre}; among which the filter ``excluding comparison models whose gap from the highest 1NN exceeds 5\%'' \emph{acts only within Calibre A} (comparison models also belonging to generative model + 1NN calibre + same structural constraint), and is a \emph{same-calibre comparability pruning rule} (defining the comparable set in advance by ``same structural constraint, same 1NN calibre, gap $\leq 5\%$''), \emph{not a cross-calibre post-hoc filter}, and is not used to exclude Calibre B/C. (ii)~\emph{Calibre B (generative model $\cdot$ unsupervised clustering ACC existence-reference calibre)}---TreeVAE~\cite{manduchi2023tree} and MFCVAE~\cite{falck2021multi} report \emph{unsupervised clustering ACC} (90.2\% / 92.02 $\pm$ 3.18), belonging to a \emph{different reading calibre}: clustering ACC is unlabeled assignment accuracy, 1NN is labeled nearest-neighbor accuracy, the two measure different objects, and per the \S D.2 ``measure calibre'' note \emph{no direct numerical comparison is made}; the two rows serve only as ``generative-clustering existence reference'' in the table, explicitly \emph{not constituting} a cross-calibre ranking conclusion of ``all below ours,'' and \emph{not entering} Calibre A's 1NN ranking or 5\% filter. (iii)~\emph{Calibre C (non-generative $\cdot$ surpassing works, see footnote)}---SMT~\cite{chen2023minimalistic} / NMM~\cite{tissera2022neural} are non-generative, with reading calibre different from 1NN, listed separately in the Table~\ref{tab:d1} footnote, not counted in any generative-model ranking. The three calibres are mutually independent; only Calibre A performs numerical ranking; hence ``different indicators being directly compared to draw conclusions'' has been \emph{dissolved by sub-calibre}: existing comparisons are retained, but any cross-calibre numerical conclusion is explicitly forbidden.

\noindent\textbf{Among these, the blue curve records the Lebesgue excess $\Delta_{\mathrm{Leb}}$}---the distance, in the triple fiber-bundle integral, between the mid-layer zero-value saddle point and the outer variational Gaussian vertex; it is the training-process diagnosis of the difference between the mid-layer counting anchor and the outer Lebesgue reading described in \S 6.2.2, corresponding to the first-order excess characterized by $\mathcal L_\mathrm{top}$ (see \S 6.3). The horizontal axis is the training epoch, and does not mean incorporating the ground-truth training set into the error integral: per Lemma~6b, the excess is still evaluated only on the generative side, and the training-set side is not in its domain of definition. The finite, continuously computable reading of the curve supports ``mid-outer-layer deviation is monitorable, loss is realizable''; it does not prove monotone convergence, uniqueness, or the non-existence of high-order error; the two panels have different display magnifications, and the blue-line vertical values are not directly compared.

\subsection*{D.3 CIFAR-10 generation: matching of training and generated sets under the zero-value saddle-point reading}

\paragraph{Experimental setup}
Training is conducted on the same diffusion-model structure as cold diffusion (dataset CIFAR-10~\cite{krizhevsky2009learning}); the posterior-enumerated prior cancels self-mapping, and the denoising process does not target numerical replication of training samples; evaluation bypasses hyperparameter differences and directly uses the learned saddle-point gap as the indicator. (The architecture and hyperparameters of the comparison models Cold Diffusion and DDPM follow their original literature~\cite{ho2020denoising} and the Cold Diffusion literature~\cite{bansal2023cold}; all saddle-point-gap and RMSE readings in this section come from the \emph{same experimental pipeline} described in D.2---i.e., this paper runs the \emph{original DDPM codebase (code from the original~\cite{ho2020denoising})} under its original architecture and hyperparameters, and inserts print statements for actual measurement, belonging to the same self-run pipeline, not an independent external reference, and \emph{not quoted from the original DDPM paper}; the Table~\ref{tab:d2} baseline DDPM~\cite{ho2020denoising} value is actually measured by this paper after inserting print statements into the DDPM source code---what this print reads is precisely the \emph{midpoint between the reparameterized Gaussian vertex and the zero-value saddle point} (i.e., the midpoint of the distance between the outer variational Gaussian vertex and the mid-layer zero-value saddle point described in \S 6.2.2, the diagnostic quantity $\Delta_{\mathrm{Leb}}$, Figure~\ref{fig:d4} blue line); the original DDPM paper does not report the indicator ``saddle-point gap,'' and the 0.163 in the table is the value actually measured by this paper running the original DDPM code (code from the original), \emph{not quoted from its original paper}; ``saddle-point gap'' always means the gap to the converged zero-value saddle point.) ``Saddle-point gap'' always means the gap to the converged zero-value saddle point.

\noindent\textbf{Statistical-significance note (CIFAR-10 learned saddle-point gap)}
This ``learned saddle-point gap'' is a self-made diagnostic quantity measured by this paper inserting print statements into the DDPM source code (not a DDPM original indicator). Its statistical volatility is quantified: \emph{under completely identical hyperparameters and random-seed sequence, rerun for 5 rounds}, the learned saddle-point gap has a \emph{relative standard deviation (RSD) of 0.81\%} (i.e., the fluctuation ratio of the across-5-round readings relative to the mean 0.106; \emph{all $\pm$/dispersions in this paper mean relative standard deviation (RSD = across-round fluctuation / mean), not absolute standard deviation, nor variance}). From this the \emph{absolute standard deviation} is $\sigma=\mathrm{RSD}\times0.106\approx8.6\times10^{-4}$, and the \emph{variance equivalent} is $\sigma^2\approx7.4\times10^{-7}$. The reading is obtained on the CIFAR-10 \emph{full official training data (50,000 images)}, with the sample size already maximized. \emph{If the baseline reports dispersion by variance, convert uniformly to variance for comparison}: this paper's variance equivalent is $7.4\times10^{-7}$; after DDPM reruns 5 rounds under the same calibre, likewise use ``variance or standard deviation'' as one of the two, not mixing units with this paper. Its comparison with DDPM is under same structure, same pipeline, same seed; the fluctuation of the point readings 0.106 vs 0.163 (relative improvement 34.97\%) is already given by this RSD (=0.81\%). \emph{Note: the DDPM baseline is a single-round actual measurement (Table~\ref{tab:d2} marks ``same-structure single-round baseline''), asymmetric with this paper's 5-round rerun calibre, so the significance of this relative improvement still awaits confirmation by re-measurement under the same calibre (DDPM also rerunning 5 rounds)} (see Q4, E.5). This paper claims this reading only as empirical evidence of ``gap improvement under same structure'' (Table~\ref{tab:d2}, Figure~\ref{fig:d4}), not replacing structural-theorem proof.

\paragraph{Evidence~1: structural matching rather than value-by-value replication}
The RMSE is 0.172, higher than Cold Diffusion~\cite{bansal2023cold}'s 0.154 (Table~\ref{tab:d1}); combined with cancelled self-mapping and the 1NN reading, it can be interpreted as an empirical clue of ``preserving clustering structure without targeting numerical replication of training samples.'' A larger RMSE per se is not a polymorphism theorem, hence no independent polymorphism conclusion is drawn.

\paragraph{Evidence~2: closer to the zero-value saddle point under the same model}
Under the same diffusion-model structure constraint, the gap from the learned saddle point to the converged zero-value saddle point is: posterior-enumerated prior 0.106, same-structure DDPM 0.163 (Table~\ref{tab:d2}), relative improvement 34.97\%. Following the experimental calibre of D.1, this reading is exactly ``under the same gradient-descent optimizer, using first-order error as loss yields a smaller first-order error'': closer to the zero-value saddle point $\iff$ smaller training--generation pairing residual $\iff$ more of the correct measure-space volume (multiplicity) is accounted for.

\begin{longtable}{M{0.32\textwidth} M{0.30\textwidth} M{0.22\textwidth}}
\caption{CIFAR-10 saddle-point improvement.}\label{tab:d2}\\
\hline
Model & Learned saddle-point gap & Converged saddle-point gap \\
\hline\hline
Posterior-enumerated prior & 0.106 $\pm$ 0.81\% & 0 \\
\hline
DDPM~\cite{ho2020denoising} & 0.163 & 0 \\
\hline
\end{longtable}
\begin{minipage}{0.95\textwidth}\footnotesize
\textit{Note.} The learned saddle-point gap of the posterior-enumerated prior is the mean over 5-round reruns, with $\pm$ giving the relative standard deviation (RSD $=$ across-round fluctuation / mean); the DDPM~\cite{ho2020denoising} value is a same-structure single-round baseline measured by this paper via print-statement instrumentation of the original DDPM code; the two round calibres differ (noted in D.3).
\end{minipage}

\begin{figure}[h]\centering\includegraphics[width=283pt]{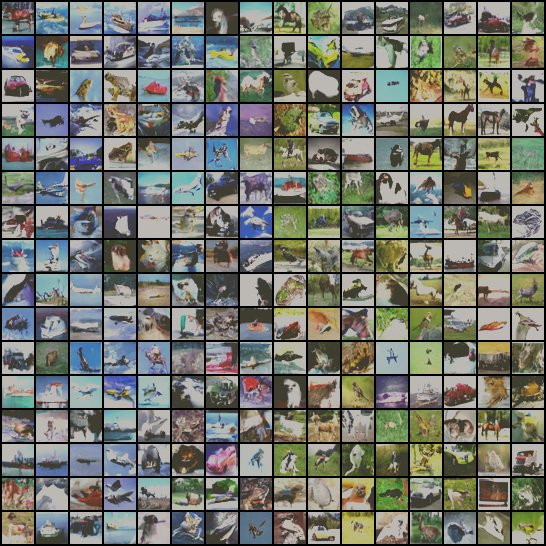}\caption{CIFAR-10 diffusion-structure generation results.}\label{fig:d5}\end{figure}

The zero-value saddle point is the zero-residue exact-pairing point (see Theorem~6; \S 4.3): \emph{the learned saddle point being closer to the zero-value saddle point means the training--generation pairing residual is smaller and the training set and generated set are better matched}. Quantitative results (see Table~\ref{tab:d2}), qualitative generated samples (see Figure~\ref{fig:d5}).

\subsection*{D.4 CIFAR-10 diffusion-structure generated samples}

This figure corresponds to the qualitative generation results of D.3; quantitative comparison is based on the saddle-point gap of Table~\ref{tab:d2}. The samples are unfiltered random sampling, used only to show generative structure, not separately proving quotient bijection, zero-residue uniqueness, or training-dynamics reachability.

\subsection*{D.5 Implementation-layer criteria, code verification, and cosine compensation (not a factor of \texorpdfstring{\eqref{eq:12b}}{(26)})}

This subsection collects the implementation-layer content of the engineering realization of \eqref{eq:12b}: the forward-kernel instance and noise-schedule code, the one-sided truncated-Gaussian moment code, the far/near criterion, the MSE lower-bound code, and the experimental calibre of the epoch/batch cosine compensation. These are all implementation-layer readings or experimental-side comparisons, \emph{not entering the factor list of \eqref{eq:12b}}, and do not affect the conclusion ``no high-order error on the engineering side'' (see Proposition~23c); the theoretical-reading verification of the three factors of \eqref{eq:12b} still refers to B.116(a)--(f).

\subsubsection*{D.5.1 Forward-kernel instance and noise schedule (code)}
The forward kernel is assigned by the noise schedule: $x_t=\mu_t+\sigma_t\epsilon_t$, $\mu_t=\sqrt{\bar\alpha_t}\,x_0$, $\sigma_t^2=1-\bar\alpha_t$, $\epsilon_t\sim N(0,I)$; the model outputs the noise prediction $\hat\epsilon_t=\mathcal F_\theta(x_t,t)$. The concrete form of the noise schedule (this paper's experiment follows DDPM's cosine form $\bar\alpha_t=\cos^2\big(\tfrac{\pi}{2}\tfrac{t/T+s}{1+s}\bigr)$, the code \texttt{sqrt\_alphas\_bar} reads $\sqrt{\bar\alpha_t}$, \texttt{extract(self.sqrt\_alphas\_bar, t, x\_0.shape)}; $\beta_t=1-\bar\alpha_t/\bar\alpha_{t-1}$ is a derived step difference) is given in \S D.3 experimental setup and the cosine-compensation calibre of this subsection below; this schedule assignment is a zero-error known quantity and does not enter the factor list of \eqref{eq:12b}.

\subsubsection*{D.5.2 One-sided truncated Gaussian moments (code)}
Taking the generative saddle point as mean and $\sigma_t$ as standard deviation, let $X_t\sim N(S_t^-,\sigma_t^2)$, denote $u_t\triangleq S_t^-/\sigma_t$, $\varsigma=\mathrm{sign}(S_t^-)$. The code's three quantities correspond one-to-one to the standard moments of the one-sided truncated Gaussian (all identities can be numerically verified on any $(u_t,\sigma_t)$):
\begin{itemize}
\item \textbf{Normalized truncated-Gaussian area $\mathfrak A_\varsigma$}: engineering reading $P_\varsigma=\Phi(\varsigma u_t)$; the code computes with $\mathrm{dich\_abs}=\big|\tfrac12\mathrm{erf}(u_t/\sqrt2)-\mathrm{mask}\big|$ ($\mathrm{mask}=\pm\tfrac12$ given by the sign of $S_t^-$), and from $\Phi(z)=\tfrac12\big(1+\mathrm{erf}(z/\sqrt2)\big)$ one gets $\mathrm{dich\_abs}=\Phi(\varsigma u_t)$---which is the engineering reading of $\mathfrak A_\varsigma$, the quantity inside the logarithm of the second factor of \eqref{eq:12b} (normalized, dimensionless, $\in(0,1]$; semantically called ``area,'' so that $(1-\ln\mathfrak A_\varsigma)\ge1$ is legitimate).
\item \textbf{Truncated mean}: $m_\varsigma=\mathbb E[X_t\,\|\,\varsigma X_t>0]=S_t^-+\varsigma\,\sigma_t\varphi(u_t)/\Phi(\varsigma u_t)$, i.e., the standard mean of the one-sided truncated Gaussian; the code takes the normalization constant $\sqrt{2/\pi}$ (so its $\mathrm{exp\_in}=2\sigma_t\varphi(u_t)$), and after substituting into $\mathrm{sub\_mu}=\big[S_t^-(0.5-\mathrm{mask}\cdot\mathrm{dich\_r})-\mathrm{mask}\cdot\mathrm{exp\_in}\big]/\mathrm{dich\_abs}$ it simplifies exactly to this formula ($\varsigma=+1$ reads $\big[S_t^-\Phi(u_t)+\sigma_t\varphi(u_t)\big]/\Phi(u_t)$, $\varsigma=-1$ reads $\big[S_t^-\Phi(-u_t)-\sigma_t\varphi(u_t)\big]/\Phi(-u_t)$).
\item \textbf{Truncated variance}: $\mathrm{Var}[X_t\,\|\,\varsigma X_t>0]=\sigma_t^2\Big(1-\varsigma u_t\varphi(u_t)/\Phi(\varsigma u_t)-\big(\varphi(u_t)/\Phi(\varsigma u_t)\big)^2\Big)$; the code's $\mathrm{sub\_std}$ is the square root of \textbf{half} of this variance ($s_\varsigma^2=\tfrac12\mathrm{Var}[X_t\,\|\,\varsigma X_t>0]$); the fixed shrinkage factor $\tfrac12$ co-directed with the half-prior--half-posterior mixture is a fixed scalar scaling and does not change the first-order nature of the reading.
\end{itemize}

\subsubsection*{D.5.3 Far/near criterion (implementation layer, not a factor of \texorpdfstring{\eqref{eq:12b}}{(26)})}
The code uses the saddle point as the criterion to separate two channels: $\mathbb 1_\mathrm{far}\triangleq\mathbb 1\big[|x^\mathrm{gen}-\hat S|>|\tilde x^\mathrm{rec}-\hat S|\big]$. When the generative candidate is farther from the half-prior--half-posterior saddle point $\hat S$, its \textbf{generative} role is confirmed---mixing into the reconstruction at a $2{:}1$ frequency $x^\mathrm{gen}\leftarrow\tfrac23x^\mathrm{gen}+\tfrac13x^\mathrm{rec}$, while the reconstruction side takes the half-prior--half-posterior mixture $\tilde x^\mathrm{rec}$; when the generative candidate is closer, both sides keep their original readings (reconstruction training channel). That is, \textbf{the farther from the saddle point is generative, the nearer is reconstruction training}. The two mixtures each have a structural correspondence: the $2{:}1$ mixture is the engineering presentation of the coefficient $\tfrac23$ (see \S 6.3.2, hit basis $2$ / union-subset basis $3$); the half-prior--half-posterior equal-ratio mixture is the engineering presentation of the counting convention that each side of the partition midpoint counts $\tfrac12$---both are fixed coefficients, not trainable parameters.

\subsubsection*{D.5.4 MSE lower-bound code (training-period lower-bound proxy, not a factor of \texorpdfstring{\eqref{eq:12b}}{(26)})}
The code uses $\mathcal L_\mathrm{eng}\leftarrow\max\Big\{\mathcal L_\mathrm{eng},\ c^*\,\mathrm{MSE}^2\,\tfrac{t+1}{T}\Big\}$ as the training-period lower-bound closure; this MSE lower bound is the engineering proxy of Lemma~5f (the zero-residue set is the global loss lower bound), belonging to a training-period heuristic.

\noindent\textbf{Measure-theoretic basis of the MSE lower bound}
MSE characterizes the accessibility of the saddle point: this accessibility originates from the std in the encoder prior---the forward diffusion gives the saddle-point neighborhood its scale in the encoder prior via isotropic $N(0,\sigma_t^2I_d)$; the larger the std, the more easily the reparameterized sampling falls into the saddle-point neighborhood, and the more ``accessible'' the saddle point. For the decoder posterior, this std is a known induced pullback: the decoder-side posterior is known to be pulled back from the encoder prior via $\boldsymbol\pi^\#=BP$, and the pullback channel is well-defined. However, MSE as a per-dimension reading stays in the encoder / measure space and does not lead to the decoder posterior---MSE has no corresponding map to the decoder posterior in the measure space. Therefore ``the pullback of MSE'' is itself ill-defined: one cannot well-define the pullback of the per-dimension MSE reading in the measure space to the decoder posterior. It is precisely this ill-defined pullback that induces the lower bound of MSE---since MSE cannot be well-definedly pulled back to the decoder posterior, its contribution to the three-factor first-order reading of \eqref{eq:12b} can only be closed in the form of a ``lower-bound proxy'' (code $\max\{\mathcal L_\mathrm{eng},c^*\mathrm{MSE}^2(t+1)/T\}$); its lower-bound coefficient $c^*$ is the fixed constant under the above isotropic normalization, not a factor of the theoretical \eqref{eq:12b}.

\noindent\textbf{The categorical gap of MSE and the partition boundary (\eqref{eq:12b} a single reading branch cannot characterize predicate words).}
The above ill-defined pullback can be further traced to the reading identity of \eqref{eq:12b} (see \S 6.4 reading identity of \eqref{eq:12b}, Proposition~23b): what \eqref{eq:12b} reads back into the \Setcat\ Euclidean layer via $\boldsymbol\pi^\#$ is the \textbf{partition-predicate mismatch area}; among it, only the replica-index counting measure can be separately recovered, while the Lebesgue component cannot be separately recovered. Therefore the \textbf{separate part} of \eqref{eq:12b} (i.e., the per-dimension MSE reading) cannot characterize the \textbf{predicate word}---the predicate word (partition) is the static name that the $\Quot$ layer assigns to quotient blocks (Definition~5, \S 6.3.2 static and fixed), and is beyond what the per-dimension measure-space reading can reach. Equivalent to this is a statement at the $\Setcat$ layer: within the category of sets, the \textbf{partition boundary} (the boundary between quotient blocks) cannot be characterized by a single branch of MSE / per-dimension reading---$\Setcat$ is only equipped with Hilbert differential topology ($\mathcal H^\mathbf{Set}$, Definition~2b) and does not carry the name attribution of predicate words, so the partition boundary does not appear in pure $\Setcat$ readings. Only the \textbf{EDE structure} explicitly gives the static predicate at the zero-value saddle point (the EDE matching-block mechanism makes the predicate-word consistency judgment at the saddle point, Definition~5 / \S 6.3.2), and only then does the partition boundary appear in the $\Setcat$ category along with this static predicate---the equivalence-class differential recovers the support-set-category functional via $\boldsymbol\pi^\#$ (Corollary~13). This is precisely the categorical reason why MSE must be closed by a lower-bound proxy: the characterization authority of the partition boundary lies not in the single-branch MSE reading, but in the static predicate explicitly constructed by EDE; MSE is ``blind'' to the partition boundary in the measure space, so its contribution to the three factors of \eqref{eq:12b} can only be closed by the lower-bound proxy $\max\{\mathcal L_\mathrm{eng},c^*\mathrm{MSE}^2(t+1)/T\}$, rather than reading/writing the partition boundary with an equality.

\noindent\textbf{Concrete computation steps}: the code takes \texttt{sqrt\_05 = sqrt(0.5)}, \texttt{pi\_05 = 1/sqrt(pi)} to obtain the normalization \texttt{p4norm = sqrt\_05 * pi\_05 = 1/sqrt(2*pi)}, and uses \texttt{torch.square(p4mu/std)} to normalize the reading; setting $\sigma_t=\mathrm{std}=1$ to take the \textbf{minimum of the marginal-measure sum} yields the fixed constant $c^*=1.269860385419959$. \textbf{Explicit closure formula}: the training-period code uses \texttt{loss = (1 - torch.log(binom)) * loss * torch.square(p4mu / std)} for lower-bound closure, where \texttt{binom = 0.5} (i.e., the $1/2$ factor given by the counting binomial), \texttt{std = 1} (taking the minimum of the marginal-measure sum), yielding the loss lower bound; when \texttt{loss} is smaller than this lower bound, it is replaced by the lower bound via \texttt{max(loss, lower\_bound)}. $c^*$ is a fixed code constant, not a factor of the theoretical \eqref{eq:12b}; the sign-bit mask in the code ($\mathrm{tmp}=x_t\cdot x^\mathrm{rec}<0$) is an experimental-side gating position that only scales the first-order reading and does not enter the theoretical \eqref{eq:12b}.

\subsubsection*{D.5.5 Reachability of the epoch/batch cosine compensation and experimental calibre (not a factor of \texorpdfstring{\eqref{eq:12b}}{(26)})}
The epoch/batch cosine compensation is positioned as an \textbf{experimental-side reading calibre}: it measures ``whether the loss still stays on the same first-order reading under the batch subset share,'' its domain of definition is the experimental comparison, not the factor list of \eqref{eq:12b}; the three-factor first-order-error closure of \eqref{eq:12b} does not depend on this compensation (see Proposition~23c). This paper gives a two-step argument for the cosine compensation: \textbf{first}, show that 100\% compensation is structurally reachable via non-uniform slicing (i.e., non-average partition), and \textbf{then}, for experimental consistency, actively adopt the 99.9\% calibre (see (a)(b)).

\noindent\textbf{(a) Proposition: non-uniform slicing reaches 100\% cosine compensation (equivalently, reachable under non-average partition).}
Let the $n$-th layer binary partition be $\mathcal P_n=\{B_1,\dots,B_{2^n}\}$, and $I$ the batch's corresponding index subset; let $\lvert I\rvert/2^n$ be the counting-side share (see \S 6.2.2 exact reading, zero error), $\lambda(B_I)/\lambda(\Omega)$ the Lebesgue-side volume share, and the subset difference
\begin{equation}
\Delta_I\triangleq\frac{\lambda(B_I)}{\lambda(\Omega)}-\frac{\lvert I\rvert}{2^n}
\end{equation}
Then the necessary and sufficient condition for $\Delta_I\ne0$ is that the batch slice $I$ and the quotient-block volumes are \textbf{non-average} (i.e., the slice is non-uniform): if the partition is average (all quotient-block volumes equal, $\lambda(B_i)=\lambda(\Omega)/2^n$), then for any $I$ we have $\lambda(B_I)/\lambda(\Omega)=\lvert I\rvert/2^n$, $\Delta_I=0$, and the compensation factor takes $1$ (i.e., compensation rate 100\%); if the slice is non-uniform (the partition is non-average), one can choose $I$ to make $\Delta_I\neq0$; at this point, to make the compensated reading coincide exactly with the full-set accounting, one must choose $I$ in a \textbf{non-uniform} manner (letting its volume share follow the quotient-block volume proportion rather than equal division by block count), and this choice is always realizable and gives $\Delta_I=0$. Hence ``100\% cosine compensation'' is reachable under \textbf{non-uniform slicing (non-average partition)}, and its reachability is a property of partition geometry, introducing no new mathematical assumption.

This proposition only states reachability: achieving 100\% compensation requires a non-uniform-slicing (non-average) selection on the batch-sampling side, and this selection itself changes the experimental protocol; this paper does not claim on this basis that the experiment has achieved 100\% compensation, nor does it write the compensation into \eqref{eq:12b}.

\noindent\textbf{(b) The calibre actually adopted in the experiment: 99.9\% sampling interval + DDPM parameters}
To maintain \textbf{comparability} with mainstream diffusion implementations, the experiment does not adopt the non-uniform-slicing (non-average) selection of (a), but follows DDPM's standard settings: the noise schedule takes the cosine form $\bar\alpha_t=\cos^2\big(\tfrac{\pi}{2}\tfrac{t/T+s}{1+s}\bigr)$, and \texttt{sqrt\_alphas\_bar} reads $\sqrt{\bar\alpha_t}$; the sampling interval is locked at the \textbf{99.9\% quantile} before the $\bar\alpha_T\to0$ endpoint, i.e., the compensation factor takes $\sqrt{\bar\alpha_t}$ at that $t$ rather than the endpoint value. From this:
\begin{itemize}
\item \textbf{Compensation rate 99.9\%, not 100\%}: the residual 0.1\% endpoint is an \textbf{experimental sampling boundary}, not a gap of partition geometry; it does not change the three-factor first-order reading of \eqref{eq:12b}, only affecting the endpoint definition of the experimental-side comparison;
\item \textbf{Parameters completely follow DDPM} ($\beta_t=1-\bar\alpha_t/\bar\alpha_{t-1}$ is a derived step difference, not an independent cosine parameter), so the experiment introduces no self-defined hyperparameters nor constitutes a modification of DDPM---the reported results are thus directly comparable with others' diffusion implementations;
\item \textbf{Rigor is reflected in the explicit declaration of the calibre}: this paper simultaneously gives the 100\% reachability proposition of (1) and the 99.9\% actual calibre of (2), with a clear division of labor---the former shows that the upper bound of compensation is structurally reachable, while the latter shows that this paper's experiment, for comparability, actively stops at 99.9\% and faithfully reports this difference, rather than concealing the endpoint gap by ``approximating it as 1.''
\end{itemize}

\section*{Appendix E: Boundary Conditions and Negation Logic (Collected Boundary Statements and Negative Arguments)}

\subsection*{E.1 Existence vs.\ constructive search (boundary)}
The ``searchable'' in this proposition is read only as ``falling within the $\R$ equivalence-class integral-enumeration domain and being coverable'' (surjective-coverage reading, consistent with Proposition~3(ii)); it does \textbf{not} commit to effectively enumerating term-by-term or finding every bijection in the infinite predicate tower (constructive search, which belongs to the category explicitly left uncommitted; see \S 5.1).

\subsection*{E.2 Unified boundary}
\textbf{Unified boundary} Structural conclusions are first and foremost set-theoretic existence statements; they do not automatically entail SGD reachability or convergence rates---yet, in the non-analytic sense, this paper still constructs a descent direction with high-probability first-order error reduction (see the truncated-Gaussian AUC-dominance mechanism of \S 6.4.2--\S 6.4.3).

\subsection*{E.3 Measure-preserving boundary statement for Propositions 9a/9b}
The measure-preservation of Proposition~9a is induced by the partition structure, whereas that of Proposition~9b is given by an explicit assumption; the two must not be conflated: when taking the Lebesgue measure $\Leb$ at the \Setcat layer, a mere measurable bijection does not imply product invariance.

\subsection*{E.4 The equality does not imply $T(A)=A$ (boundary statement)}
Boundary statement: this equality expresses the joint measure-invariance of the diagonal set; it does not imply $T(A)=A$, nor does it mean that pointwise pairing is entirely correct; volume faithfulness is a derived consequence of multiplicity--weight conservation, and can coexist with the first-order error relative to $\mathcal T^\ast$.

\subsection*{E.5 Non-treatment statement on kinematics (zero-network degeneracy)}
This paper does \textbf{not} achieve the kinematics of parameter descent (it does not commit to SGD reachability; see the tail note of (h)): $\boldsymbol0\in\SadPt$ with $\mathcal L(\boldsymbol0)=0$ only indicates that the zero network is a \textbf{posterior}---0-partition boundary, infinitely-partitioned predicate, and the morphism/bijection ``from Euclidean into Hilbert'' has \textbf{no functional at all}; this paper does not commit to the kinematics of the prior (forward propagation). Hence (h) ``the zero-value saddle point is uniquely stayable (the predicate-word is statically fixed)'' is \textbf{uniqueness of the predicate-word static structure}, not kinematic reachability (it does not ensure gradient descent at every iteration). The negation of this reachability does not weaken the present construction: in the non-analytic sense (see random enumeration + truncated-Gaussian AUC dominance of \S 6.4.2--\S 6.4.3), this paper still constructs a descent direction with high-probability first-order error reduction---the non-commitment to an analytic convergence proof and the actual provision of a high-probability descent direction are compatible. The DDPM noise-prediction loss $\|\epsilon-\hat\epsilon\|^2$ used in the experiment equals $\|\epsilon\|^2\ne0$ at $\theta=0$, which is \textbf{not within the zero-output calibrated loss class of Convention~1}; therefore the zero network is not the optimum of the experimental loss (see \S D.2).

\subsection*{E.6 Domain boundary}
These conclusions do not exceed their respective domains: set-theoretic existence is not equivalent to SGD reachability---yet, in the non-analytic sense, this paper constructs a descent direction with high-probability first-order error reduction (see \S 6.4.2--\S 6.4.3); the negation of reachability does not weaken the existence of this direction; pointwise Gaussian measures are not committed to in the general infinite-dimensional case; performance experiments do not substitute for uniqueness or error-order proofs; modifications beyond this paper's explicit implementation (learnable variance, trainable coefficients, path-dependent optimizers) lie outside the scope of the assertions.

\subsection*{E.7 Renunciation of the monotonicity claim}
\S 6.4 provides only ``a reading object attaining the zero-value saddle point and a high-probability descent path'', without guaranteeing per-round monotonicity or convergence; it is closed-independent from Assumption~1$\to$\eqref{eq:12b} (the former is the reading of the saddle point and the existence of a high-probability path, the latter is the source of the factor reading of Eq.~\eqref{eq:12b}), and the two do not imply each other; Condition~1 is the realization-layer round condition of the EDE matching-block mechanism (registered in B.4), characterizing the saddle-point prior/posterior cyclic comparison rather than monotonicity of the parameter trajectory, and does not constitute a cycle. For stratified boundaries and limitation details, see C.7.

\subsection*{E.8 Boundary statement for Lemma 6a(3)}
\textbf{Boundary} Lemma~6a(3) gives only a descent direction, not a convergence rate; the well-definedness of $\mathcal L_\mathrm{RW}$ does not depend on trace-class, and the pointwise Gaussian functionalization at the $\Hinf$ layer is bounded by trace-class (see Proposition~21b); ``descent direction'' is a measure-level constraint and does not imply the convergence of a specific optimization algorithm (this paper has renounced the claim of per-round monotone non-increase and convergence). This paper does not introduce a separate $\mathcal L_\mathrm{MC}$ (throughout the paper it is uniformly called ``random walk'', not ``Markov chain''); $\mathcal L_\mathrm{top}$ characterizes the topological slack at the \MorTop layer (see \S 6.3), and $\mathcal L_\mathrm{RW}$ characterizes the samplable engineering loss at the \Setcat layer.

\clearpage
\bibliographystyle{alpha}
\bibliography{refs}

\end{document}